\pdfoutput=1
\documentclass{article}
\usepackage[preprint,nonatbib]{neurips_2025}
\usepackage[T1]{fontenc}
\usepackage[utf8]{inputenc}
\usepackage{lmodern}

\usepackage{textcomp}
\usepackage{microtype}
\usepackage{amsmath,amssymb}
\usepackage{graphicx}
\usepackage{array}
\usepackage{booktabs}
\usepackage{longtable}
\usepackage{enumitem}
\usepackage{pifont}
\usepackage{upquote}
\usepackage{listings}
\usepackage{placeins}
\usepackage{xcolor}
\usepackage[font=small,labelfont=bf,labelsep=period]{caption}
\PassOptionsToPackage{hyphens}{url}
\usepackage[colorlinks=true,linkcolor={blue!50!black},citecolor={blue!50!black},urlcolor={blue!60!black},
  bookmarksnumbered=true,pdfusetitle]{hyperref}

\makeatletter
\@ifundefined{KV@Gin@alt}{\define@key{Gin}{alt}{}}{}
\def\cite@bare#1#2{#1}
\DeclareRobustCommand{\citenum}[1]{\begingroup\let\@cite\cite@bare\cite{#1}\endgroup}
\newenvironment{refsnoheading}[1]{\renewcommand\section[2]{}\thebibliography{#1}}{\endthebibliography}
\g@addto@macro\UrlBreaks{\do\a\do\b\do\c\do\d\do\e\do\f\do\g\do\h\do\i\do\j\do\k\do\l\do\m\do\n\do\o\do\p\do\q
  \do\r\do\s\do\t\do\u\do\v\do\w\do\x\do\y\do\z\do\A\do\B\do\C\do\D\do\E\do\F\do\G\do\H\do\I\do\J\do\K\do\L\do\M
  \do\N\do\O\do\P\do\Q\do\R\do\S\do\T\do\U\do\V\do\W\do\X\do\Y\do\Z\do0\do1\do2\do3\do4\do5\do6\do7\do8\do9}
\newcommand{\appendixheads}{\renewcommand\@seccntformat[1]{%
  \ifnum\pdfstrcmp{##1}{section}=0 Appendix~\csname the##1\endcsname.\enskip\else\csname the##1\endcsname\quad\fi}}
\makeatother
\newcommand{\code}[1]{\texttt{\spaceskip=0.525em plus 0.25em minus 0.05em\relax#1}}
\DeclareRobustCommand{\us}{\_\allowbreak}

\newcommand{\appsub}[2]{\par\addvspace{0.6\baselineskip}\phantomsection\addcontentsline{toc}{subsection}{\protect\numberline{#1}#2}}
\newenvironment{teaserbox}{\par\vspace{\teaserabove}\noindent\begin{minipage}{\linewidth}\centering}%
  {\end{minipage}\par\vspace{\teaserbelow}}
\setlist{itemsep=2pt,topsep=3pt,parsep=0pt}
\newcommand{\notefont}{\normalsize\mdseries\itshape}
\newlength{\teaserabove}
\newlength{\teaserbelow}
\newcommand{\aftertitle}{\vspace{-0.12in}}

\hypersetup{pdftitle={Rules of Thumb for Decision Models: What Makes Performance Better (or Worse)},pdfauthor={Benjamin Jacob Bodner}}

\title{Rules of Thumb for Decision Models: What Makes Performance Better (or Worse)}
\author{Benjamin Jacob Bodner\\[3pt]{\notefont Unaffiliated independent researcher}}
\date{}

\begin{document}
\maketitle
\aftertitle
\begin{abstract}
Text-focused typed decision models take a text and a list of options and return a probability for each option, but which changes to a request make them more or less accurate? We screened 60 techniques on TypeSafe's Jev, then re-tested the key ones on 12 datasets (8 public, 4 private) with 8 different decision models and 4 additional local models. Our experiments find several behavioural trends that held across models and datasets. Adding nearest-neighbour labelled examples similar to the input added 6 to 11 points on Jev. However, wording changes (personas, politeness, capitals) provided negligible changes in accuracy, often less than 2 points. If the option names were vague or made-up, we find that a one-line description per option added 10.7 to 25.0 points on Jev. However, with clear option names, this description added little. Unsurprisingly, adding incorrectly labelled examples hurt performance. If the pool of examples used for nearest-neighbour input enrichment is contaminated with 30\% incorrectly labelled samples, Jev lost 7.0 points and six other models lost 12.6 on average. We provide six do's and six don'ts for boosting the performance of decision models. All our input enrichment techniques and experiments are publicly available at \url{https://github.com/BjBodner/decision-levers}.
\end{abstract}

\begin{teaserbox}
\includegraphics[width=0.88\linewidth,alt={A diagram in two panels under the title "Surprise! Examples make decision models better (or worse)". Panel (a), "One call": at the top, a dashed box holds a cylinder labelled "labelled pool" with an arrow to a blue box, "pick examples: nearest (or random)", from which a blue line runs down into the request. Below it, a grey box headed "The request you send" lists the wording ("optional persona or politeness", in grey italics), the text ("My card was declined."), the question ("Which team handles this?") and the options. One option box is drawn in full: "name: Payments", a blue examples row ("examples: 'card refused ...'"), where the blue line from the pool ends, and a green description row ("description: 'Failed payments.'"), with the note "Refunds, Fraud, ... one box each" beneath. An arrow leads from the request to a black box, "Decision model", and down to "Returned": Payments 0.86 (a dark square), Refunds 0.09 and Fraud 0.05. Panel (b), "One change at a time", with the line "accuracy points vs the bare call" under its heading: bars grow right from a vertical zero line. Same example for every text, a short blue bar, +1.7; random examples from the pool, +3.1; nearest examples from the pool, +4.7; one-line description per option, a green bar, solid to +11 and lighter to +25, labelled "+11 to +25"; persona or polite wording, a grey dot on the zero line, "  0". Below a thin rule, a separate risk row, "30  wrong labels in pool": an orange bar extending left of zero, labelled "-13 vs a clean pool".}]{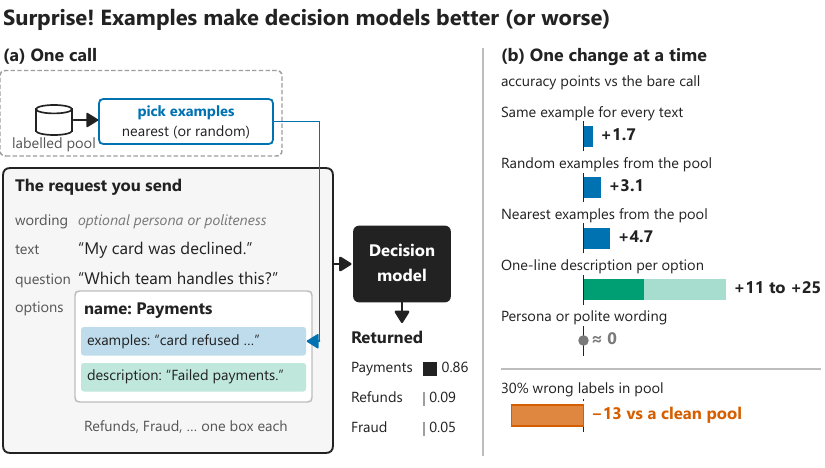}
\renewcommand{\thefigure}{1}
\captionof{figure}{Surprise! Examples make decision models better (or worse). \textbf{(a)} The pool step is optional and runs on your machine. \textbf{(b)} Each row is the bare call plus one change (rows don't stack). Example rows: Jev 1.13, fresh CLINC150 and LEDGAR items, 100 labels per class (long LEDGAR clauses gain less; +7.7 with MetaTool); ``random'' texts carry the nearest ones' labels. Descriptions: two tool catalogues with vague or invented names (about 0 on clear names: CLINC150, LEDGAR). Wording: within 2 points on Jev. Risk: mean of six other APIs (Jev 1.13 \ensuremath{-}7.0).}
\end{teaserbox}

\section{Introduction}

Many production decisions are classifications: route this ticket, flag this message, pick this tool. \emph{Typed decision models}, such as TypeSafe's Jev, are built for them. A request holds a text to classify, a closed list of \emph{options} and an instruction. The response is a probability for each option. A team that adopts such a model needs rules of thumb: what to put in a request, what to leave out, and when to keep its data at home.

We screened 60 techniques on Jev 1.13 (TypeSafe's decision model, the version current when the study began), then re-tested the most important ones in ten planned runs, on seven other decision models, a chat model and four private datasets (Table 3 gives each run's scope). Three findings came out:

\begin{enumerate}
\item \textbf{Benign wording wrappers stayed within 2 points.} Personas, politeness, capitals and unrelated questions in the same request showed no reliable gain on Jev 1.13 (Section 3).
\item \textbf{Information usually helps, but not on every model.} Retrieved labelled examples helped Jev 1.13, four of six other models on fresh public items and three of six on private data; one model lost on fresh items and two were misled on long texts. On Jev 1.13, names that say what an option does, plus a one-line description, helped when the names were vague or invented, and did little on clear ones (Sections 4.1, 6 and 7).
\item \textbf{Wrong information hurts, and that carries over.} Confusable wrong labels in the pool hurt every model we tested. Piling them on one rival hurt more than spreading them: on all six other models on fresh items, and on five of six on private data (Sections 4.2, 6 and 7).
\end{enumerate}

Two more results matter in practice. Where the examples sit changes how much wrong labels cost, in different directions on different models (Section 4.3). And on Jev 1.13, examples in the prompt beat the same request with nothing added (the \emph{bare call}); no local shortcut built on one bare call was shown to match them (Section 5). Table 2 turns all of this into rules, and Section 8 lists what didn't work. One methods result may matter beyond this study: under a zero-average stress test built from our own contrasts, item-level intervals rejected 38 to 46\% of the time, as clustered-data work predicts {[}\citenum{r4}; \citenum{r92}{]}, so our decisions resample whole classes (Section 2.3).

Each result carries a status word for the strength of its evidence: \emph{confirmed} (a planned test passed; the exact rule is in Section 2.3), \emph{observed} (found after the fact, or passed only under looser statistics), \emph{secondary} (planned as an estimate, with no pass or fail rule) or \emph{not shown} (a planned test didn't pass, which never means ``no effect'').

Much of what helps is known from in-context learning and earlier decision-model work (Section 1.1), so most of what's new here is about what hurts: random labels matter little for generative models in some settings {[}\citenum{r32}{]}, but with retrieved neighbours labels matter, as label copying predicts {[}\citenum{r55}; \citenum{r28}{]}. Here even uniform wrong labels cost Jev 1.13 3.9 points, and confusable ones 7.0 (12.6 on average on six other decision models). Table 1 sets out what this study adds, Section 1.1 places it among related work, and Appendix B compares it in full.

\begingroup\footnotesize
\renewcommand{\thetable}{1}
\begin{longtable}{@{}>{\raggedright\arraybackslash}p{\dimexpr 0.2618\linewidth-1.0470\tabcolsep\relax}>{\raggedright\arraybackslash}p{\dimexpr 0.3489\linewidth-1.3955\tabcolsep\relax}>{\raggedright\arraybackslash}p{\dimexpr 0.3894\linewidth-1.5575\tabcolsep\relax}@{}}
\caption{What makes this study different. A ``typed API'' is a hosted decision model that returns option probabilities. Counts of other models exclude Jev.}\\
\toprule
What we do & Closest prior work & What is new \\
\midrule
\endfirsthead
\multicolumn{3}{@{}l}{\textit{Table 1 (continued)}}\\
\toprule
What we do & Closest prior work & What is new \\
\midrule
\endhead
\bottomrule
\endlastfoot
Test what hurts on six other decision models and a chat model & Examples help on 21 hosted decision models {[}\citenum{r70}{]}; wording across vendors {[}\citenum{r71}; \citenum{r78}{]}; context attacks across four systems {[}\citenum{r53}{]}; poisoned retrieval on Jev {[}\citenum{r51}{]} & Non-adversarial harms tested across vendors: pool noise hurt on 6 of 6 (5 of 5 vendor families); the one-rival penalty on 6 of 6 on fresh items and 5 of 6 on private data \\
Split wrong-label harm within items: delete, relabel, or relabel to one rival & Wrong labels lower accuracy {[}\citenum{r55}; \citenum{r72}{]}; no vs random-label demonstrations {[}\citenum{r32}; \citenum{r55}{]}; false demonstrations pull to one label {[}\citenum{r12}{]} & Measures a form of majority-label bias {[}\citenum{r84}; \citenum{r85}; \citenum{r101}{]} on typed APIs: one rival cost 13.2 more than spreading. New: the within-item split, in which deleting (lost evidence alone) cost 13.1 less than relabelling \\
Move the same examples between option descriptions and a block, with length and label controls & Demonstration position moves chat models {[}\citenum{r83}{]}; examples in label definitions {[}\citenum{r82}{]} or instructions {[}\citenum{r70}{]} & Penalty larger inside descriptions on Jev with both controls, and on 3 of 6 others; opposite on 2 \\
Send the same requests to a chat model, read by option log-probabilities & Large models learn unrelated labels {[}\citenum{r50}{]}, more so after symbol tuning {[}\citenum{r87}{]}; task recognition vs learning {[}\citenum{r86}{]}; letter readouts are biased {[}\citenum{r107}{]} & Retrieval, pool noise and the one-rival penalty held on one chat model, on one dataset \\
Size the prompt against local models and prompt baselines at 5, 20 and 100 labels per class & Break-even near 100 labels {[}\citenum{r79}; \citenum{r33}; \citenum{r9}{]}; prompt cost against few-shot models {[}\citenum{r99}{]}; label-matched control {[}\citenum{r31}{]}; calibration {[}\citenum{r84}; \citenum{r88}{]}; kNN inference {[}\citenum{r91}{]}; many shots {[}\citenum{r93}{]} & Retrieval beat random, fixed and calibrated-bare-call baselines (secondary); no simple local combination was shown to match it \\
Estimate how many held-out items a design choice needs & Many NLP comparisons are underpowered {[}\citenum{r81}{]} & 100 to 200 items rank a large gain right 79 to 97\% of the time, near-ties 56 to 64\% (observed, optimistic; Appendix P.4) \\
Calibrate the tests: content-free edits, stress tests, small-cluster re-decisions & Re-run floors {[}\citenum{r80}{]}; replication practice {[}\citenum{r8}; \citenum{r47}{]}; significance testing in NLP {[}\citenum{r110}{]}; cluster bootstraps {[}\citenum{r4}; \citenum{r49}; \citenum{r109}{]}; clustered errors {[}\citenum{r92}{]} & A screen with no minimum effect size let through 12 of 53 content-free edits; item-level intervals rejected 38 to 46\% under a zero-average stress test \\
Vary how much option names say, with and without a one-line description & Typed models follow a name over its definition {[}\citenum{r44}; \citenum{r90}{]}; chat models use definitions inconsistently {[}\citenum{r97}{]}; opaque ids beat labels on one task {[}\citenum{r29}{]} & On Jev, a description added 25.0 on an invented catalogue and about 0 on clear names; ids with descriptions were not shown to be no more than 2 worse than names on fresh items (\ensuremath{-}2.3) \\
Re-test the techniques on private synthetic data & Contamination vs knowledge on public benchmarks {[}\citenum{r11}; \citenum{r41}{]}; a human-written private mirror {[}\citenum{r111}{]} & Eight techniques, four private datasets, eight models: our planned distortion test flagged two techniques; wrong labels and pool noise still hurt every model, and no clear effect flipped sign on average \\
\end{longtable}
\endgroup

\begingroup\footnotesize
\renewcommand{\thetable}{2}
\begin{longtable}{@{}>{\raggedright\arraybackslash}p{\dimexpr 0.3077\linewidth-1.2308\tabcolsep\relax}>{\raggedright\arraybackslash}p{\dimexpr 0.5190\linewidth-2.0762\tabcolsep\relax}>{\raggedright\arraybackslash}p{\dimexpr 0.1733\linewidth-0.6930\tabcolsep\relax}@{}}
\caption{The rules of thumb. Effects are accuracy points. Counts of other models are out of 7 on re-used items, or out of 6 on fresh items (the stronger evidence) and on private data. Status words are as in Section 1; \emph{transfers} and \emph{partially transfers} are defined in Section 2.3. Table 3 gives each run's models and items.}\\
\toprule
Rule & Evidence & Status \\
\midrule
\endfirsthead
\multicolumn{3}{@{}l}{\textit{Table 2 (continued)}}\\
\toprule
Rule & Evidence & Status \\
\midrule
\endhead
\bottomrule
\endlastfoot
\textbf{Do 1.} Give it information it lacks, and test the gain on your model. & Retrieved examples on Jev: +5.9 to +11.2 with balanced pools, +7.7 on fresh items; in realistic pools +1.9 (natural shares, not shown) and +0.05 and +2.6 (time splits). 4 of 6 others on fresh items (pooled +1.1; Decider Flash \ensuremath{-}14.5), 3 of 6 on private data; two models failed on long texts & confirmed on Jev; partially transfers \\
\textbf{Do 2.} Name options by what they do, and add a one-line description, above all for unfamiliar, invented or brand-like names. & On Jev: one-word names \ensuremath{-}10.7 and a description +10.7 (MetaTool); on invented names, brand-only names 5.8\% and a description +25.0 (+16.8 to +39.8 on 6 others); on clear names about 0 (CLINC150 \ensuremath{-}1.0, LEDGAR \ensuremath{-}0.4). Ids with descriptions: within a point of names on Jev, not shown to be no more than 2 worse on fresh items (\ensuremath{-}2.3; shown on 1 of 6 others) & secondary; names vs ids not shown \\
\textbf{Do 3.} Audit your pool's labels. & 30\% confusable wrong labels: \ensuremath{-}7.0 on Jev, harm on 6 of 6 others (4 fell below their bare call); at 10\%, Jev's examples still beat the bare call by 8.1; auditing 10\% won back about 1 on Jev & confirmed harm; audit secondary \\
\textbf{Do 4.} At 5 per class, use the prompt. & +3.1 over the bare call, pools redrawn; a local shortcut was not shown to come within 1.5 (\ensuremath{-}1.2) & confirmed \\
\textbf{Do 5.} At 100 per class, use the prompt by default; go local for privacy. & Not from a passed test: a local combination of simple models was neither shown worse nor shown to match (\ensuremath{-}0.7 {[}\ensuremath{-}1.6, +0.1{]}, two fresh-item tests). It had lower worst-class recall (29.7\% vs 42.8\%) and fell 10.9 behind after a class-mix shift (one dataset); it sends no pool text & secondary and observed; match not shown \\
\textbf{Do 6.} Confirm on your model and version. & Gains vary by model (SD 3.5 on re-used items, 7.9 on fresh items); Jev 1.14 lost 3.7 more to wrong labels than Jev 1.13 on the same items; 100 to 200 items rank near-ties right only 56 to 64\% & observed \\
\textbf{Don't 1.} Expect benign wrappers to help. & Persona, politeness, capitals, unrelated questions within \ensuremath{\pm}2 on Jev; persona and politeness on 4 of 7 others; a gain of about 1 is not ruled out & confirmed \\
\textbf{Don't 2.} State unmeasured shares. & On Jev: shuffled shares \ensuremath{-}1.1 vs none and \ensuremath{-}6.2 to \ensuremath{-}6.8 vs true shares; last year's shares \ensuremath{-}11.5 vs true, lower on 6 of 6 others on the central estimate; true shares helped in Run 2 (+1.9) but not in Run 8 & observed \\
\textbf{Don't 3.} Let wrong labels pile on a rival. & One rival vs spread: \ensuremath{-}13.8 on Jev, \ensuremath{-}2.8 to \ensuremath{-}10.1 on 6 of 6 others on fresh items; 5 of 6 on private data (Kev-4B reversed) & confirmed \\
\textbf{Don't 4.} Assume placement is neutral. & Penalty larger inside descriptions on Jev (not clearly on MetaTool) and on 3 of 6 others; smaller on 2 & confirmed on Jev \\
\textbf{Don't 5.} Judge rare classes by accuracy. & A cascade within 1.5 points on accuracy lost 7.3 points of toxic-class precision & observed \\
\textbf{Don't 6.} Treat wording as security. & A guard sentence, spotlighting and a claim field all fell to an attack search that never saw the model's scores (300 author-written tickets) & observed \\
\end{longtable}
\endgroup

\subsection{Related work}

\textbf{In-context learning.} Retrieved examples beat random ones {[}\citenum{r25}; \citenum{r39}{]}. Wrong or noisy demonstration labels hurt {[}\citenum{r55}; \citenum{r72}{]}, and poisoned demonstrations or retrieval corpora do harm on purpose {[}\citenum{r103}; \citenum{r104}; \citenum{r51}{]}. Models also drift toward the label shown most often, the majority-label bias {[}\citenum{r84}; \citenum{r85}; \citenum{r101}{]}, and false demonstrations pull them toward one label {[}\citenum{r12}{]}. Our one-rival penalty is a form of this majority-label pull, measured on typed APIs. Earlier work compared no demonstrations with random-label ones across prompts {[}\citenum{r32}; \citenum{r55}{]}; what's new is splitting the wrong-label harm within the same items into lost evidence, added wrong evidence and the pull toward one rival (Section 4.2).

\textbf{Decision models.} Examples and counterexamples helped 21 hosted decision models {[}\citenum{r70}{]}, models follow an option's name over its definition {[}\citenum{r44}; \citenum{r90}{]}, retrieved precedents helped Jev on moderation benchmarks {[}\citenum{r29}{]}, and a classical classifier won one email-routing task {[}\citenum{r27}{]}. TypedBench measures calibration, framing and cost {[}\citenum{r43}{]}, and presentation {[}\citenum{r78}{]} and instruction wording {[}\citenum{r71}{]} shift results. Harms have been studied as attacks: natural context that flips decisions across four systems {[}\citenum{r53}{]}, black-box attacks {[}\citenum{r26}; \citenum{r80}{]} and poisoned evidence in a Jev workflow {[}\citenum{r51}{]}. As far as we found, none tests non-adversarial harms across vendors (wrong labels, noisy pools), where the examples sit (inside option descriptions or in a block), examples in the prompt against local classifiers across label budgets, or leakage.

\textbf{Contamination.} On public benchmarks, knowledge and contamination are hard to tell apart {[}\citenum{r11}; \citenum{r41}{]}. Our private synthetic datasets remove exposure to the test items, as GSM1k did for maths {[}\citenum{r111}{]}, though they also change the items' style (Section 7).

So what's new overall isn't that information helps, but a cross-vendor map of what hurts, where and by how much.

\section{Setup}

\subsection{The interface}

A decision model takes one request and returns one probability per option (Figure 1a). Each option has a name and can carry a description. You can add information in three places: the question's wording, a one-line description of an option, or labelled examples written into that description. Examples come from a labelled \emph{pool}. Before the call, on your machine, you pick the pool texts nearest to the text being classified (or random ones) and write them under their labels.

Figure 1b shows what one change at a time did to accuracy, on Jev 1.13 except for the risk row. Its example rows are on the two of Run 8's three fresh datasets that had a pool-size curve (+4.7 at 100 per class); over all three, retrieved examples gained 7.7 points (Table 4), our headline number for fresh items. The ``same example'' row was measured directly: one fixed example per option gave +4.6 on CLINC150 and \ensuremath{-}1.3 on LEDGAR, +1.7 on average. Figure M.1 marks every lever on one real request.

Pool size is given in labels per class. \emph{Retrieval} finds the 20 pool texts most similar to the item and shows at most 3 per label (10 for yes/no tasks), by TF-IDF or bge-small {[}\citenum{r52}{]} similarity. The \emph{examples prompt} carries these retrieved examples, and the bare call carries none. We send each item under several instruction wordings and score it as the share of wordings whose top option is correct, as multi-prompt evaluation recommends {[}\citenum{r108}{]}. We report paired changes in accuracy points, with 90\% intervals.

\subsection{Runs, models and items}

In all, the study ran ten planned runs on 12 datasets: 8 public ones and 4 private synthetic ones (Section 7). It sent requests to nine decision models: Jev 1.13 and 1.14 and seven from other vendors, one of which (Solar Decide) failed its first test batch and dropped out, leaving eight. It also read one general chat model through its option log-probabilities, and compared everything with 4 local models that run on your machine without any API: a logistic regression on bge-small embeddings, a fine-tuned MiniLM classifier, a nearest-centroid classifier and a nearest-neighbour vote (Section 5). \emph{Fresh} items are ones this study had never sent to any model before that run. They still come from public benchmarks, so a model may have seen them in training; Section 7 checks this. The datasets cover intents (Banking77 {[}\citenum{r5}{]}, CLINC150 {[}\citenum{r21}{]}, MASSIVE {[}\citenum{r57}{]}), financial-news topics (tweets), consumer complaints (CFPB), legal clauses (LEDGAR {[}\citenum{r6}{]}), toxic prompts (ToxicChat {[}\citenum{r23}{]}) and tool choice (MetaTool {[}\citenum{r16}{]}). We removed test items that nearly matched a pool text or a development item (TF-IDF cosine \ensuremath{\geq} 0.9). Appendix K (Table K.1) gives item counts per run and dataset. Appendix K.1 lists the exact version and identifier of every model used.

\begingroup\footnotesize
\renewcommand{\thetable}{3}
\begin{longtable}{@{}>{\raggedright\arraybackslash}p{\dimexpr 0.0464\linewidth-0.2783\tabcolsep\relax}>{\raggedright\arraybackslash}p{\dimexpr 0.2102\linewidth-1.2610\tabcolsep\relax}>{\raggedright\arraybackslash}p{\dimexpr 0.3616\linewidth-2.1694\tabcolsep\relax}>{\raggedright\arraybackslash}p{\dimexpr 0.3819\linewidth-2.2914\tabcolsep\relax}@{}}
\caption{Which runs used which models and items. Run 5 re-used items that Jev 1.13 had answered, and Runs 6 and 7 re-used Run 4's and Run 6's items.}\\
\toprule
Run & Models & Items & What it tested (tests) \\
\midrule
\endfirsthead
\multicolumn{4}{@{}l}{\textit{Table 3 (continued)}}\\
\toprule
Run & Models & Items & What it tested (tests) \\
\midrule
\endhead
\bottomrule
\endlastfoot
1 & Jev 1.13 & held-out test items: Banking77, CLINC150, tweets, CFPB, ToxicChat, MetaTool & wrappers, retrieval, shares, definitions, wrong labels, prompt vs vote (15) \\
2 & Jev 1.13 & never-sent items: tweets, CFPB, ToxicChat & 5 labels per class, shares, cascade, local combination (6) \\
3 & Jev 1.13; Jev 1.14 (secondary) & fresh: MASSIVE, LEDGAR, MetaTool, CFPB; plan publicly archived & wrappers with 10 stimuli, wrong-label dose, blind definitions (3) \\
4 & Jev 1.13 & fresh: MASSIVE, LEDGAR, MetaTool; CFPB 2025 pool, 2026 test & label budget; delete vs relabel vs one rival (5) \\
5 & 7 other decision models; Jev 1.14 & Jev 1.13's items: MASSIVE, MetaTool, Banking77 & wrappers, retrieval, three wrong labels, definitions (4 per model) \\
6 & Jev 1.13 & Run 4's items & placement: descriptions vs a block (3) \\
7 & Jev 1.13 & Run 6's items; Run 4's hard items; CFPB 2026 & placement controls, wrappers on hard items, stale shares (7) \\
8 & Jev 1.13; 6 other decision models; a chat model & fresh: CLINC150 validation split, LEDGAR, MetaTool (chat model: MetaTool); CFPB 2023 pool, 2024 test; stale shares on Run 7's CFPB 2026 items & pool noise, one rival, ids vs names, retrieval, placement, stale shares, label budget, baselines, latency (19 on Jev, 6 per other model, 8 on the chat model) \\
9 & Jev 1.13 & Run 8's fresh CLINC150, LEDGAR and MetaTool items; four private synthetic datasets & leakage check: eight techniques, public vs private (8); names and descriptions (8 estimates) \\
10 & 6 other decision models; Jev 1.14 & Run 9's public items; the first 600 items of each private set & Run 9's battery on other models (8 per model) \\
\end{longtable}
\endgroup

The other decision models are Perplexity Decider v1.2 and Decider Flash, OpenAI Decisions (gpt-6-luna), Microsoft-Decision-1, Liquid d1-ultra, Kev-4B and Upstage Solar Decide {[}\citenum{r69}{]}. Solar Decide failed its Run 5 pilot (17 of 20 requests returned HTTP 422), so it counts as not confirmed there and was left out of Runs 8 and 10. Run 10 reached Jev 1.14 through OpenRouter, which routes to TypeSafe, because TypeSafe's API was down. The chat model is OpenAI's gpt-6-luna chat model. It sees the same options as letters, and we score it by the log-probabilities of the letters. It shares its base model with OpenAI Decisions, so it's a second way of reading one model.

Development data came from training splits, apart from the pools, and the screen of about 60 techniques ran there (Appendix B). In repeated identical-resend checks, Jev 1.13 kept the same top option for 96.8 to 100\% of requests (re-analysis 1; Appendix L). Jev cost \$0.042 per million input tokens, and Runs 1 to 8 cost \$64.01 in all (other prices and per-run costs: Appendices H.4, K and Y).

\subsection{How we tested claims}

Each result comes from a paired comparison, and its status word says how strong the evidence is.

\textbf{Plans.} Before sending any request in a run, we fixed its items, tests and decision rules (Appendix P). Run 3's plan was publicly archived before its run; the others are now public in the reproducibility package, and their hashes show only that they did not change. Run 8's dated addenda, and the limits of that evidence, are in Appendix U.2. Run 9's difficulty-matched analysis was added later and is exploratory (Appendix Y).

\textbf{Statistics.} Intervals come from a cluster bootstrap {[}\citenum{r4}; \citenum{r58}{]}, which accounts for clustered items, as recommended for language-model evaluations {[}\citenum{r92}{]}. From Run 3 on, we resample whole gold classes within each dataset (\emph{class clusters}). In Runs 1 and 2 we resample groups of near-duplicate items, which is close to resampling items. The study holds 198 planned tests with a decision rule: 134 in Runs 1 to 8 (Appendix P), 8 in Run 9 and 56 in Run 10 (Section 7). Holm's method {[}\citenum{r15}{]} corrects within each run's own test families, not across the study; one correction over all 29 tests of Runs 1 to 4 would demote two passes (Appendix I.11). Estimates outside any test family are descriptive. ``Within \ensuremath{\pm}2 points'' means that two one-sided tests {[}\citenum{r42}; \citenum{r100}{]} passed at \ensuremath{\pm}2: we could rule out a gain or loss bigger than 2 points, not that the effect is zero. ``No more than X worse'' means that a non-inferiority test at \ensuremath{-}X passed. We fixed \ensuremath{\pm}2 as the smallest change we judged a team would act on, and \ensuremath{-}1.5 because replacing a prompt with a cascade or ensemble should cost less than dropping a decoration (Appendix P.4). \textbf{Status words.} In exact terms, \emph{confirmed} needs a planned test to pass with gold classes resampled, after Holm's correction within its run; a pass with only items resampled counts as \emph{observed}. So ``confirmed'' means it passed within one run, not across the study. A rule \emph{transfers} if it is confirmed on at least 5 of 7 other models on re-used items (Run 5), or 4 of 6 on fresh items (Run 8, with one Holm correction across models), and no model shows the opposite. Otherwise it \emph{partially transfers}. A model is \emph{opposite} when it clearly goes the other way (its interval doesn't cross zero).

\textbf{Calibration.} Loose tests make content-free edits look real: a screen with no minimum effect size let 12 of 53 of them through (Appendix P.2). So we tested the tests. We gave real contrasts a random sign per gold class, so the average effect was zero but classes still differed. At a nominal 5\% level, item-level intervals then rejected 38 to 46\% of the time and the class bootstrap 8.5 to 8.8\% (cluster methods over-reject when clusters are few or unequal in size {[}\citenum{r4}; \citenum{r109}{]}), while a wild-cluster bootstrap with Webb weights {[}\citenum{r49}{]} held about 5\% (Appendix U.3).

Re-checked this way, 15 of the 25 passes in Runs 1 to 4 are confirmed. The other 10 are reported as observed, mostly because they held only with items resampled (Appendix U.3). Runs 7 and 8 were decided with the class bootstrap. A later wild-cluster re-check moved one of their results, shuffled shares on LEDGAR, to observed, and every other decision stands (Appendix U.4).

\section{Benign wording wrappers stayed within 2 points}

A \emph{wrapper} is text that's the same for every item and changes neither the right answer nor how to find it: a persona line, a politeness prefix, a line in capitals, or unrelated questions in the same request. On fresh items, wrappers kept Jev within 2 accuracy points of the bare call.

\textbf{Many wordings, fresh items.} Run 3 tested five personas and five politeness prefixes, under three instruction wordings, on 600 fresh items in each of four datasets. Persona moved accuracy by \ensuremath{-}0.3 {[}\ensuremath{-}1.0, +0.3{]} points and politeness by \ensuremath{-}0.1 {[}\ensuremath{-}0.5, +0.4{]}, both within \ensuremath{\pm}2 (confirmed), much as chat-model studies found for personas {[}\citenum{r56}; \citenum{r2}{]} and polite wording {[}\citenum{r73}{]} (impolite wording has mattered {[}\citenum{r54}; \citenum{r94}{]}). Run 8 tested two more wrappers on fresh items. A capitals line (``IMPORTANT: THINK CAREFULLY AND ANSWER CORRECTLY.'') stayed within \ensuremath{\pm}2 of the bare call (+0.25), and so did seven unrelated questions sent along with ours, under five wordings (+0.30). That fits the view that Jev scores each question in a request on its own {[}\citenum{r36}{]}, although irrelevant context does distract language models on reasoning tasks {[}\citenum{r106}{]}. On hard items, where the bare call was unsure, persona and politeness barely moved the correct option's log-probability, but \ensuremath{\pm}2 in accuracy wasn't shown (Appendix U.1).

\textbf{Other models.} Wrappers did little on other models too, though not every test passed: \ensuremath{\pm}2 was shown on 4 of 7 (Table 5). Politeness cost OpenAI Decisions 1.4 points {[}\ensuremath{-}2.8, \ensuremath{-}0.1{]}, and Kev-4B's interval reached +2.0. On the chat model \ensuremath{\pm}2 wasn't shown.

\textbf{Within \ensuremath{\pm}2 is not zero.} Against a neutral prefix of the same length, persona and politeness scored +1.08 {[}\ensuremath{-}0.02, +2.22{]}, so a small gain of about a point is possible. That matches prompt-based models, where irrelevant templates can work about as well as instructive ones {[}\citenum{r89}{]}. We tested benign wrappers only. Other wording does matter: rewording a ticket-priority rule moved Jev by 24 points {[}\citenum{r43}{]}, and Section 9 lists what we didn't vary. No benign wrapper showed a reliable gain, so don't expect one; they cost nothing to remove.

\textbf{Wording is not a security boundary either.} A one-line claim that a wrong team was correct cut ticket-routing accuracy from 87\% to 33\%. Three wording defences fell to a search over 100 attack variants that never saw the model's scores, as the indirect prompt-injection threat model predicts {[}\citenum{r96}{]}: a guard sentence, spotlighting (marking the message as untrusted data {[}\citenum{r14}{]}) and a separate field for the claim (observed; Appendix G). These defences separate data from instructions by wording alone; StruQ does it with a separate input channel and fine-tuning {[}\citenum{r105}{]}. In FlipJev, a red-teaming study of Jev, common injected instructions changed no correct answer {[}\citenum{r26}{]}, but our attacker plants evidence instead. So in these tests, benign wording didn't move accuracy, and wording-only defences didn't stop text that claimed a wrong answer. What moved these models was what the request said.

\section{Information helps, and wrong information hurts}

Accuracy followed the information in the request. Examples and descriptions that the model lacked raised accuracy. Wrong labels and wrong class shares lowered it, sometimes by more than the examples had added. Table 4 collects this section's Jev results.

\begingroup\footnotesize
\renewcommand{\thetable}{4}
\begin{longtable}{@{}>{\raggedright\arraybackslash}p{\dimexpr 0.3925\linewidth-2.3550\tabcolsep\relax}>{\raggedright\arraybackslash}p{\dimexpr 0.0483\linewidth-0.2899\tabcolsep\relax}>{\raggedright\arraybackslash}p{\dimexpr 0.3056\linewidth-1.8334\tabcolsep\relax}>{\raggedright\arraybackslash}p{\dimexpr 0.2536\linewidth-1.5217\tabcolsep\relax}@{}}
\caption{What helps and what hurts on Jev 1.13 (accuracy points, 90\% intervals). Runs and items are as in Table 3; ``fresh'' marks Run 8's fresh items. Table 7 has the changes that didn't help, and Appendices U.2 and Y have Run 8's and Run 9's full tables.}\\
\toprule
Change & Run & Effect {[}90\% CI{]} & Status \\
\midrule
\endfirsthead
\multicolumn{4}{@{}l}{\textit{Table 4 (continued)}}\\
\toprule
Change & Run & Effect {[}90\% CI{]} & Status \\
\midrule
\endhead
\bottomrule
\endlastfoot
\textbf{What helps} &  &  &  \\
Retrieved examples vs bare call, 80 to 100 labels per class: Banking77; CLINC150; tweets & 1 & +11.2 {[}+7.5, +15.2{]}; +5.9 {[}+4.2, +7.6{]}; +8.7 {[}+7.5, +10.0{]} & confirmed \\
The same, fresh CLINC150, LEDGAR and MetaTool items & 8 & +7.7 {[}+5.6, +9.9{]} & confirmed \\
Retrieved vs label-matched random examples; vs a fixed one-shot set & 8 & +1.8 {[}+0.8, +2.9{]}; +3.2 {[}+1.5, +5.4{]} & secondary \\
Retrieval on CFPB time splits: 2025 to 2026; 2023 to 2024 & 4; 8 & +0.05; +2.6 {[}+1.7, +3.4{]} & secondary \\
A one-line description on two-word names vs the names alone: MetaTool; invented catalogue (OpsDeck) & 9 & +10.7 {[}+4.0, +17.2{]}; +25.0 {[}+14.2, +35.8{]} & secondary \\
\textbf{What hurts} &  &  &  \\
One-word vs two-word names: MetaTool; OpsDeck function word; OpsDeck brand word & 9 & \ensuremath{-}10.7 {[}\ensuremath{-}17.7, \ensuremath{-}3.9{]}; +0.4 {[}\ensuremath{-}3.4, +4.2{]}; \ensuremath{-}63.7 {[}\ensuremath{-}74.5, \ensuremath{-}52.9{]} & secondary \\
30\% wrong pool labels vs clean pool: uniform; confusable & 8 & \ensuremath{-}3.9 {[}\ensuremath{-}5.0, \ensuremath{-}2.9{]}; \ensuremath{-}7.0 {[}\ensuremath{-}8.2, \ensuremath{-}6.0{]} & confirmed \\
30\% confusable pool: examples prompt vs bare call & 8 & +0.7 {[}\ensuremath{-}1.5, +2.9{]} & not shown (two-sided) \\
Three correct examples shown: delete minus relabel to distinct rivals & 4 & +13.1 {[}+11.7, +14.7{]} & confirmed \\
One rival minus distinct rivals (one-rival penalty) & 4; 8 & \ensuremath{-}13.2 {[}\ensuremath{-}14.7, \ensuremath{-}11.7{]}; \ensuremath{-}13.8 {[}\ensuremath{-}15.1, \ensuremath{-}12.5{]} & confirmed \\
Extra cost of deleting the last correct example, over the second & 4 & +10.1 {[}+7.5, +12.7{]} & confirmed \\
\textbf{Where the examples sit} &  &  &  \\
Clean examples: block vs bare call; descriptions minus block & 6 & +7.0 {[}+4.2, +9.8{]}; +0.5 {[}\ensuremath{-}0.3, +1.4{]} & confirmed; secondary \\
One-rival penalty, descriptions minus block (Run 4's 13.2, re-used, vs 6.8) & 6 & +6.4 {[}+5.0, +7.9{]} & confirmed \\
The same with equal-length descriptions; with each label named once & 7 & +7.4 {[}+5.5, +9.3{]}; +8.1 {[}+5.7, +10.4{]} & confirmed \\
The same with equal-length descriptions, fresh items & 8 & +6.4 {[}+3.3, +9.4{]} & confirmed \\
\textbf{Class shares} &  &  &  \\
Shuffled shares vs true shares & 1 & \ensuremath{-}6.2 to \ensuremath{-}6.8 & observed \\
Last year's shares vs true shares (CFPB 2026) & 7 & \ensuremath{-}11.5 {[}\ensuremath{-}20.4, \ensuremath{-}1.3{]} & observed \\
Shares from 100 labelled recent items vs true shares & 7 & +0.7 {[}\ensuremath{-}0.2, +1.6{]} & observed \\
\end{longtable}
\endgroup

\subsection{What helps: retrieved examples, clear names and descriptions}

\textbf{Retrieved examples raise accuracy.} With 80 to 100 labels per class in the pool, the examples prompt raised Jev's accuracy by 5.9 to 11.2 points on three datasets, and by 7.7 points on fresh CLINC150, LEDGAR and MetaTool items, in line with studies of retrieved demonstrations {[}\citenum{r25}; \citenum{r39}{]}. Long texts gained less: on development data, CFPB and LEDGAR gained only +1.0 to +1.4 (Appendix C.3 to C.7). Figure 2 follows one question through three requests.

\textbf{Part of the gain is a shortlist of likely labels.} On fresh items, retrieved examples beat random pool texts with the same labels (\emph{label-matched random} examples) by 1.8 points, and a fixed one-shot set by 3.2 points (secondary). On development data, label-matched random examples carried 20 to 61\% of the gain (Appendix C.1). Calibrating the bare call didn't close the gap (Table 7), though we didn't calibrate the examples prompt itself.

\textbf{The gain shrinks in realistic pools.} Most of our pools were class-balanced. A natural-share LEDGAR pool gained +1.9 (not shown), and two CFPB time splits +0.05 and +2.6 (Tables 4 and 7). So the balanced-pool numbers are an upper end; test the gain on your own pool and model.

\begin{figure}[!htbp]
\centering
\includegraphics[width=\linewidth,height=0.72\textheight,keepaspectratio,alt={The figure answers the Banking77 question "Top up on my Google Pay isn't working." (correct answer: apple pay or google pay) in three rows. Each row shows what changed in the request and the model's probabilities. 1. Bare call, wrong: top up failed 68 , the correct answer 30 . 2. With three retrieved examples labelled apple pay or google pay, right: the correct answer 99 . 3. With the same examples under the labels pending top up, contactless not working and passcode forgotten, wrong: pending top up 82 , the correct answer 1 . Two bars summarise all requests. Examples: fixed 149 and broke 14 of 1,200 (net +11.2 points). Swapped labels: broke 277 and fixed 9 of 800 (net -33.5 points).}]{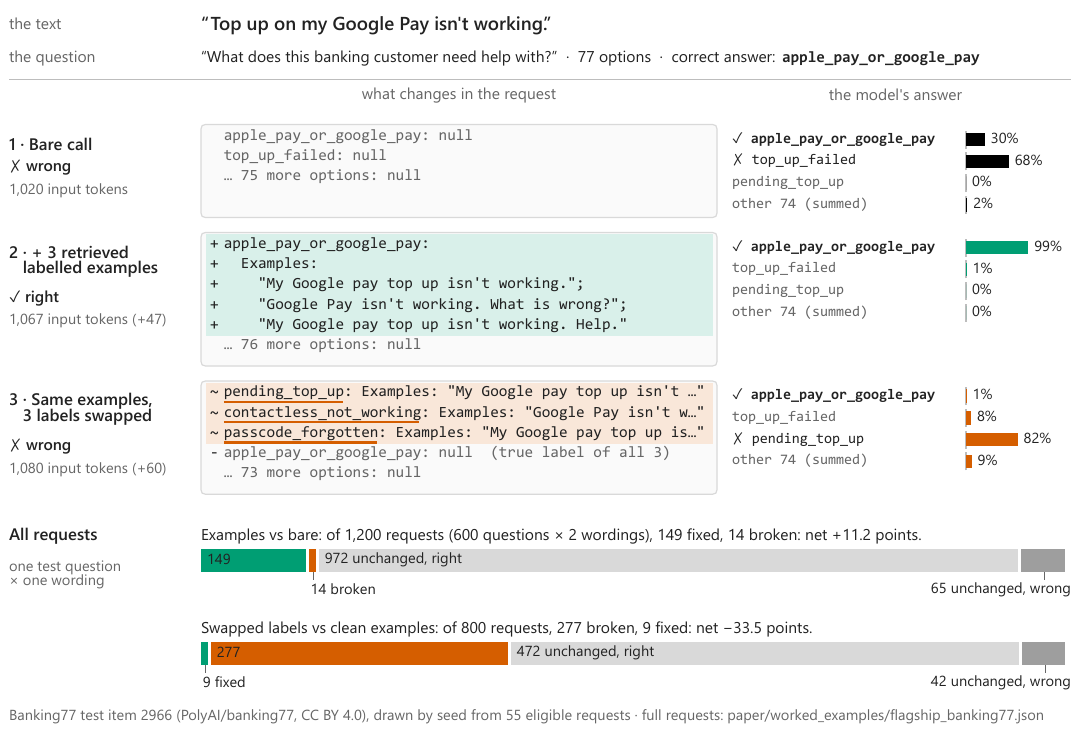}
\renewcommand{\thefigure}{2}
\caption{One question, three requests to Jev 1.13. The bare call picks \code{top\_\allowbreak{}up\_\allowbreak{}failed} (68\%). Three retrieved, correctly labelled examples move it to the correct option (99\%). The same texts under wrong labels move it to \code{pending\_\allowbreak{}top\_\allowbreak{}up} (82\%). A seeded rule drew this item from the 55 of 800 shared requests with this pattern. Over all requests, examples fixed 149 and broke 14 of 1,200 (net +11.2 points), and wrong labels broke 277 and fixed 9 of 800 (net \ensuremath{-}33.5). Banking77 is CC BY 4.0 (PolyAI). Requests and code: Appendix M.1.}
\end{figure}

\textbf{Names that say what the option does, plus a one-line description, help.} Run 9 varied how much the option names say on two tool catalogues: MetaTool, and OpsDeck-24, a private set of 24 invented tools whose two-word names pair an invented brand with a function word, such as \emph{Mossgate Queue} (Section 7). On MetaTool, cutting the names to one word cost 10.7 points, and a one-line description added 10.7 to the two-word names. On OpsDeck, the function word alone (\emph{Queue}) wasn't clearly worse than the full name (+0.4 {[}\ensuremath{-}3.4, +4.2{]}). The brand word alone (\emph{Mossgate}) fell to 5.8\% accuracy, close to the 2.8\% of cryptic ids, and a description added 25.0 points. All six other models showed both of these OpsDeck effects (Section 7), though OpenAI Decisions' brand-word cell rests on 61 items, because it refused most such requests. The description carries the meaning when the name doesn't, so it matters most when names are invented, unfamiliar or brand-like. Two caveats apply. MetaTool's queries were generated from its own descriptions {[}\citenum{r16}{]}. And where names were already clear, descriptions did little: on CLINC150 and LEDGAR they moved Jev's accuracy by \ensuremath{-}1.0 and \ensuremath{-}0.4 points, and a planned test of blind definitions on MetaTool's names didn't pass (Table 7). All of these are secondary estimates (related work: Appendix B.3).

\textbf{Readable names are still the safer default.} With a description on every option, ids came within a point of names in Run 9 (\ensuremath{-}0.8 on MetaTool, 0.0 on OpsDeck). The planned tests disagree, though. In Run 3, ids with definitions were 1.1 points below names and passed the test that they are no more than 2 points worse. In Run 8, on other fresh items, they were 2.3 points below and didn't pass it, and on other models it passed on only 1 of 6. Ids with nothing else score near zero (Appendix A.5).

\subsection{What hurts: wrong labels}

Wrong labels cost accuracy, and plausible wrong labels cost the most. We first measure the cost in a realistic pool, then show where it comes from.

\textbf{Pool noise.} On fresh items, Run 8 relabelled 10, 30 or 50\% of the pool. Each wrong label was drawn either uniformly or from the three classes nearest the true one (\emph{confusable} noise). Nearness is centroid similarity in the pool's own embeddings, so no model chose the rivals. At 30\%, uniform noise cost 3.9 points and confusable noise 7.0, so confusable noise cost 3.1 points more (confirmed). At 10\% confusable noise cost only 1.8 points, and the examples prompt still beat the bare call by 8.1; at 50\% the costs rose to 10.2 and 19.3 points (both doses ran on CLINC150 and MetaTool only). With 30\% confusable noise, the examples prompt was no longer shown to beat the bare call (+0.7 {[}\ensuremath{-}1.5, +2.9{]}). The harm was largest where the bare call was unsure: 15.5 points where its top probability was 0.5 to 0.8, against 5.3 above 0.8 (secondary).

\textbf{Auditing helps a little.} Checking a random 10\% of the noisy pool won back 0.7 points, and checking the 10\% most often retrieved won back 1.0 (secondary); the second wasn't shown to beat the first. Transformers trained on simple function classes resist label noise {[}\citenum{r102}{]}, but with nearest-neighbour examples labels matter, as label copying {[}\citenum{r55}; \citenum{r28}{]} and noisy-label work {[}\citenum{r72}{]} predict.

\textbf{Why wrong labels hurt: two sources.} A wrong label does two things: it removes correct evidence, and it adds evidence for a rival. Run 4 separated the two within items. It used 1,057 fresh items that each showed three correctly labelled examples, then deleted those examples, relabelled them to different rivals, or relabelled them all to one rival (Appendix O). All three tests passed. \textbf{Deleting cost less than relabelling}, by 13.1 points, the cost of the added wrong evidence. \textbf{Pointing every wrong label at one rival cost more} than spreading them, by 13.2 points (13.8 on fresh items in Run 8). Both are means over the number of examples changed (1 to 3, and 2 to 3); with all three changed, the gaps were 27.6 and 20.2. We call the second the \emph{one-rival penalty}; it is a form of the known majority-label bias {[}\citenum{r84}; \citenum{r85}; \citenum{r101}{]}, measured on a typed API. \textbf{Losing the last correct example caused a jump}: deleting it cost 10.1 points more than deleting the second, even with no wrong label shown.

\textbf{A simple model catches half of the rival effect.} A four-number model adds each option's bare-call log-probability (its prior) to a similarity-weighted count of the shown examples with its label (the evidence; Appendix I.13), in the spirit of accounts of in-context learning as inference {[}\citenum{r59}; \citenum{r68}{]}. Fitted on Runs 1 and 2 and applied to later runs without refitting, it tracked Run 3's pooled wrong-label curve (48.2\% against 54.1\% observed with all three relabelled), got the direction of all three Run 4 contrasts and sized the deletion jump (+11.7 against +10.1). But it caught only about half of the one-rival penalty (\ensuremath{-}6.7 against \ensuremath{-}13.2): Jev weighs a rival's growing evidence more steeply than simple addition allows. This is observed: the form was fixed before fitting but not deposited, and per-dataset errors reached 14 to 30 points.

\textbf{How fast evidence runs out.} In the \emph{three-example test}, Run 3 showed three correct examples on easy items that Jev got right about 92\% of the time, then relabelled one, two or all three of them. Accuracy fell by 2.9, 11.4 and 38.2 points. On other item sets the full relabel cost 49.5 points (Run 9, on Run 8's fresh items; Table 6) and 60.6 (Run 5's MASSIVE and MetaTool items; Table 5). This test shows the shape of the loss; the pool-noise costs above are the sizes a team should expect. A missing example costs less than a wrong one, so drop a doubtful example rather than keep its label, where a person judges the doubt (dropping pool items that the bare call disputed didn't help; Table 7). Planted near-copies with wrong labels do the same harm on purpose (Appendix C.12).

\subsection{Where the examples sit}

On Jev, wrong examples hurt more inside the option descriptions than in a separate block. Run 6 sent Jev the same labelled examples in two places: inside the option descriptions, or as one labelled block after the instruction. Clean examples helped about equally in both places. The one-rival penalty, however, was 6.4 points larger inside the descriptions (13.2 against 6.8). The difference held with equal-length descriptions in both placements (7.4), with the block naming each label once (8.1), and on fresh items (6.4). All these tests passed (Table 4). The difference came from MASSIVE, LEDGAR and CLINC150; on MetaTool it was small in both runs (+0.8 {[}\ensuremath{-}1.3, +2.9{]} and +1.5).

Across models, the effect only partly carried over. The penalty was larger inside descriptions on three of six other models (four before the correction across models) and smaller on two, as on the chat model (Table 5). So placement depends on the model. If wrong labels might pile on one rival, a separate block is safer on Jev and those three models, and worse on the other two. We didn't test placement with a noisy pool (every pool-noise test used descriptions), so a block for a noisy pool is only a hypothesis. Appendices I.13 and I.14 give two after-the-fact accounts of why.

\subsection{Class shares}

A request can also state how often each option occurs. Such shares are information too, but only as good as their source.

On fresh LEDGAR items, shuffled shares cost 1.1 points against no shares (Table 7); in Run 1 they cost 6.2 to 6.8 against true shares. Stale shares cost more. Stating last year's shares for 2026 complaints cost 11.5 points against the true shares. That result is observed, not confirmed: with only 8 classes, the class-level test didn't pass. All six other models also lost accuracy on the central estimate (\ensuremath{-}3.9 to \ensuremath{-}16.3), but only two significantly, and none after one correction across models.

True shares helped in Run 2 (+1.9, observed) but not in Run 8 (\ensuremath{-}0.4; Table 7), and on development ToxicChat they cut toxic recall from 39.4\% to 33.0\% (observed). If you do state shares, measure them first: shares estimated from 100 labelled recent items came within about a point of the true shares (Appendix U.1).

Examples help, but each one is a labelled text that you send to the vendor. The next section asks how many labels you need, and whether you can use them on your own machine instead.

\section{Labelled examples: in the prompt or combined locally?}

A team with labelled data can use it in two ways. It can send retrieved examples in the prompt. Or it can make one bare call and combine the answer, on its own machine, with local models trained on those labels.

\textbf{At about 5 labels per class, use the prompt.} On fresh items, with the pool redrawn, the examples prompt beat the bare call by 3.1 points {[}+1.8, +4.6{]} (confirmed). The gain comes early (Figure 3). On fresh CLINC150 items, nearest examples were already 5.5 points above the bare call at 5 labels per class, and 6.4 at 100; on long LEDGAR clauses the gain grew from 0.7 to 2.9 (secondary; over both sets, +3.6 at 20 and +4.7 at 100 per class). A classifier trained on the same pool, with no API call, trailed the prompt by 7.5 to 7.9 points at 5 per class and by 2.5 to 2.6 at 100. Two cheaper local shortcuts built on one bare call also trailed the prompt. Multiplying the bare call's probabilities, option by option, by a nearest-centroid classifier gave \ensuremath{-}1.2 points on these items. By a tuned classifier it gave \ensuremath{-}1.0 here (Figure 3's pink line) and \ensuremath{-}0.6 in Run 4. Neither planned test showed it within 1.5 points of the prompt.

\begin{figure}[!htbp]
\centering
\includegraphics[width=\linewidth,height=0.72\textheight,keepaspectratio,alt={Two line charts side by side plot accuracy ( ) against the labelled examples per class in the pool: 5, 20 and 100. A dashed grey line marks the bare call in each. Left, "CLINC150 (150 intents, short requests)", bare call 92.6 : nearest examples in the prompt (thick blue line, circles, with whiskers) at 98.0, 98.6 and 99.0 , labelled +5.5 at 5 and +6.4 at 100; API call x classifier, no examples sent (pink diamonds) at 95.9, 97.0 and 97.9 ; classifier trained on the pool, no API (orange squares, dashed) at 90.5, 95.2 and 96.5 ; nearest centroid, no API (grey triangles, dotted) at 89.8, 93.9 and 94.3 . Right, "LEDGAR (25 clause types, long text)", bare call 91.4 : nearest examples in the prompt at 92.1, 92.6 and 94.3 , labelled +0.7 at 5 and +2.9 at 100, with whiskers that cross the bare-call line at 5 and 20; API call x classifier at 92.3, 92.4 and 93.2 ; classifier alone at 84.1, 88.7 and 91.7 ; nearest centroid at 83.4, 87.4 and 88.5 . The legend sits in the left panel.}]{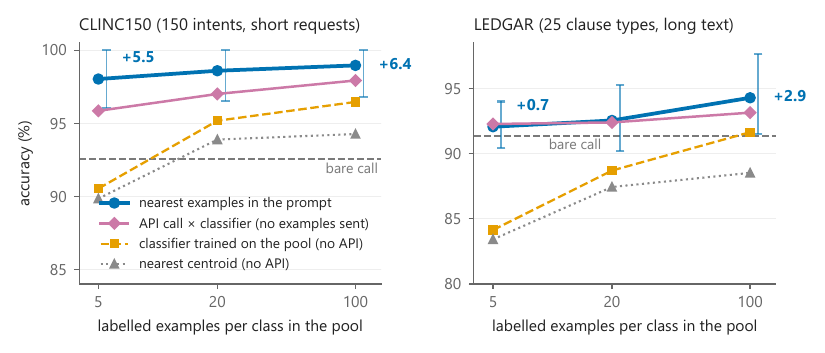}
\renewcommand{\thefigure}{3}
\caption{The gain from examples comes early. Jev 1.13's accuracy on Run 8's fresh CLINC150 and LEDGAR items, with 5, 20 or 100 labelled examples per class in the pool. Blue: nearest examples in the prompt, with 90\% intervals of its gain over the bare call (dashed line). Pink: one bare call multiplied by a classifier trained on the pool, so no examples are sent. Orange and grey: a classifier and a nearest centroid alone, with no API call. Secondary analyses; sources: Appendix L.}
\end{figure}

\textbf{At about 100 labels per class, use the prompt by default; go local for privacy.} The \emph{local combination} multiplies three things on your machine, by the product rule for combining classifiers {[}\citenum{r19}{]}: one bare call, a nearest-neighbour vote (a simple kin of kNN Prompting {[}\citenum{r91}{]}) and a tuned classifier. Only the item goes to the vendor. In both planned tests on fresh items, it trailed the prompt by 0.7 points {[}\ensuremath{-}1.6, +0.1{]}. So it was neither shown worse nor shown to match: it may be up to about 1.6 points worse. The case for the prompt rests on other evidence. The combination did worse on small classes: mean worst-class recall was 42.8\% for the prompt and 29.7\% for the combination (secondary). And when the class mix shifted between a 2025 pool and 2026 complaints, it fell 10.9 points behind the prompt (observed, one dataset; Appendix E.9). Pooling earlier comparisons after the fact doesn't change this picture (Appendix E.10).

\textbf{Go local for privacy, not accuracy or cost.} Every design cost \$0.013 to \$0.092 per 1,000 decisions in vendor fees. The local combination cost about half or less of what the examples prompt did, but that saves at most about \$0.07 per 1,000 decisions (Appendix F.6), and labelling and upkeep, which we didn't count, may cost more. The examples prompt sends up to 20 pool texts, such as other customers' complaints, with every decision. The local combination sends only the item. A labels-only shortlist sends no pool text and lost about 2 points against the examples prompt on intent sets (observed; Appendix F.2). Latency didn't separate the designs, in one session at low concurrency (Appendix U.2). Figure F.1 (Appendix F) shows what four designs send, and what they cost and scored.

\textbf{A cascade can save calls but lose precision.} In a \emph{cascade} {[}\citenum{r46}; \citenum{r98}{]}, a tuned local classifier answers the items it is sure of and sends the rest to the examples prompt. On ToxicChat it sent 8.2\% of items. Accuracy was no more than 1.5 points worse (\ensuremath{-}0.2; confirmed for these two classes), but toxic-class precision fell 7.3 points {[}1.1, 13.8{]} (observed). So judge rare classes by precision and recall at your operating point.

All of this is on Jev. Do the rules hold on other decision models?

\section{Other decision models: directions mostly agree, sizes differ}

Most rules point the same way on other models, but the sizes differ, and the harms we tested carried over more often than the retrieval gain.

Run 5 re-used items that Jev had answered, so its verdicts are not independent replications. Run 8 used fresh items and the most informative contrasts, and also went to the chat model (Table 3). Table 5 summarises both.

\begingroup\footnotesize
\renewcommand{\thetable}{5}
\begin{longtable}{@{}>{\raggedright\arraybackslash}p{\dimexpr 0.2269\linewidth-1.8149\tabcolsep\relax}>{\raggedright\arraybackslash}p{\dimexpr 0.1955\linewidth-1.5638\tabcolsep\relax}>{\raggedright\arraybackslash}p{\dimexpr 0.2418\linewidth-1.9347\tabcolsep\relax}>{\raggedright\arraybackslash}p{\dimexpr 0.1368\linewidth-1.0945\tabcolsep\relax}>{\raggedright\arraybackslash}p{\dimexpr 0.1990\linewidth-1.5920\tabcolsep\relax}@{}}
\caption{Cross-model evidence (accuracy points). Run 5 (re-used items): random-effects mean over the six models that completed it, made after the run, with its 90\% interval and between-model SD; the count confirmed is out of 7 (Solar Decide counts as not confirmed). Run 8 (fresh items): range over six models, the count confirmed under one correction across models, and the pooled mean where given (observed). Jev 1.13 is on the same items (its stale-share value: Appendix U.2). For headroom: on Run 8's fresh public items, bare-call accuracy ran from 77.9\% (Kev-4B) to 90.6\% (Microsoft-Decision-1), and Jev 1.13's was 86.7\% (Table Y.14; Run 9 for Jev 1.13). Per-model values: Appendices H.4 and U.2.}\\
\toprule
Lever & Run 5 (re-used items) & Run 8 (fresh items) & Jev 1.13 & Chat model \\
\midrule
\endfirsthead
\multicolumn{5}{@{}l}{\textit{Table 5 (continued)}}\\
\toprule
Lever & Run 5 (re-used items) & Run 8 (fresh items) & Jev 1.13 & Chat model \\
\midrule
\endhead
\bottomrule
\endlastfoot
Persona, politeness & \ensuremath{-}0.15, \ensuremath{-}0.19 (SD 0.2, 0.5); within \ensuremath{\pm}2 on 4 of 7 & -- & \ensuremath{-}0.4, \ensuremath{-}0.05 & +0.35, not shown \\
Retrieved examples & +5.6 {[}+2.0, +9.3{]} (SD 3.5); 6 of 7 & \ensuremath{-}14.5 to +7.2; 4 of 6, one opposite (mean +1.1) & +8.0; +7.7 & +12.2 \\
Ids with definitions minus names (tested for no more than 2 worse) & -- & \ensuremath{-}3.6 to +0.4; 1 of 6 & \ensuremath{-}2.3 & \ensuremath{-}0.8, not shown \\
Three wrong labels, three-example test & \ensuremath{-}27.7 {[}\ensuremath{-}34.6, \ensuremath{-}20.7{]} (SD 6.3); 6 of 7 & -- & \ensuremath{-}60.6 & -- \\
One rival vs distinct rivals & -- & \ensuremath{-}10.1 to \ensuremath{-}2.8; 6 of 6 (mean \ensuremath{-}6.3) & \ensuremath{-}13.8 & \ensuremath{-}10.2 \\
30\% confusable pool noise & -- & \ensuremath{-}29.0 to \ensuremath{-}4.4; 6 of 6 (mean \ensuremath{-}12.6) & \ensuremath{-}7.0 & \ensuremath{-}11.7 \\
Placement: penalty inside descriptions minus block & -- & \ensuremath{-}15.9 to +9.0; 3 of 6 (4 before the correction across models), two opposite & +6.4 & \ensuremath{-}4.8, opposite \\
Stale shares stated & -- & \ensuremath{-}16.3 to \ensuremath{-}3.9; 0 of 6 (2 before the correction) & \ensuremath{-}11.4 & \ensuremath{-}12.8, not shown \\
\end{longtable}
\endgroup

\textbf{The information rules carry over in direction, mostly.} On Run 5's re-used items, retrieval helped the average model by 5.6 points, but models differed (between-model SD 3.5). On fresh items the picture was weaker. Retrieval was confirmed on 4 of 6 models, the fourth only barely passing the correction across models. The pooled gain was +1.1 {[}\ensuremath{-}5.8, +8.0{]}, or +4.2 {[}+0.9, +7.5{]} without Decider Flash (observed), which collapsed on LEDGAR when shown correctly labelled examples (Section 7).

\textbf{What hurts carried over.} Confusable pool noise and the one-rival penalty hurt on all six models, as context attacks carried over between decision systems {[}\citenum{r53}{]}. Pooled over the six (observed), 30\% confusable noise changed accuracy by \ensuremath{-}12.6 points {[}\ensuremath{-}20.9, \ensuremath{-}4.3{]}, and the one-rival penalty by \ensuremath{-}6.3 {[}\ensuremath{-}8.4, \ensuremath{-}4.2{]}. With 30\% confusable noise, four models fell well below their own bare call, by 7.5 to 27.9 points (secondary). For them a noisy pool is worse than no pool. The one-rival penalty and the three-example loss were both smaller on every other model than on Jev. But Jev also gains more from clean examples, and relative to that gain its loss is about the same as the two Perplexity models' (Appendix H.4). Placement carried over only in part, and stale shares didn't carry over (Section 4.4). So harms beat gains only on these contrasts (6 of 6 against retrieval's 4 of 6), not as a rule: any model that follows its examples should show these two harms, stale shares didn't carry over, and descriptions on invented names helped all six (Section 7).

\textbf{Model families.} The six vendors aren't independent. Decider and Decider Flash are one Perplexity family, Microsoft-Decision-1 and Kev-4B share the Qwen3.5 generation, and Kev-4B is an open Jev clone (lineage partly from secondary sources; Appendix U.4). Counting each vendor family once (after the fact), pool noise and the one-rival penalty hold on all 5 families on fresh items. Retrieval holds on 4 of 5 families, or 3 if every member must agree. Placement holds on 3 of 5 (or 2), with the OpenAI family opposite and the Perplexity family split.

\textbf{The chat model.} Retrieval (+12.2), pool noise and the one-rival penalty all held on the chat model, on one dataset (MetaTool). This suggests, but doesn't show, that these are properties of evidence in the prompt rather than of typed decision APIs. Like OpenAI Decisions, the chat model differed from Jev on placement.

\textbf{A newer Jev.} Jev 1.14 agreed with Jev 1.13's bare top option on 94 to 97\% of items and passed all four Run 5 tests (the wrapper test as a \ensuremath{\pm}3 screen). But on the items both versions answered, it lost 3.7 points more to wrong labels (secondary; Appendix H). So a version change can move results.

\textbf{Checking a rule on your own model.} Gains vary by model (between-model SD 7.9 on fresh items), so treat each rule as a starting point and check it in five steps:

\begin{enumerate}
\item Hold out 100 to 200 fresh items from your own traffic. That ranks a large gain right 79 to 97\% of the time, but near-ties only 56 to 64\% (observed, optimistic; Appendix P.4).
\item Compare each change with a content-free control of the same length, and set a minimum effect size (Section 2.3).
\item Resample by class, not by item: item-level intervals reject far too often (Section 2.3).
\item Stress your pool: relabel 30\% of it to near rivals and see whether examples still beat the bare call (on Jev they no longer clearly did; see above for others).
\item Re-run on every model or version change, as the Jev 1.14 result above shows.
\end{enumerate}

But every dataset so far is public. Does that matter?

\section{Is it leakage? Four private datasets, eight decision models}

Every item in Runs 1 to 8 comes from a public benchmark that vendors may have trained on, and leakage could inflate, shrink or even create the effects above. So we built four private datasets that no model can have seen (Appendix Y): kitchen-assistant intents (\emph{KitchenOps-80}, like CLINC150), invented DevOps tools (\emph{OpsDeck-24}, like MetaTool), Scottish lease clauses (\emph{ScotLease-25}, like LEDGAR) and tickets at an invented synchrotron (\emph{BeamDesk-14}). Runs 9 and 10 sent eight techniques to every model on these private sets and on Run 8's fresh public CLINC150, LEDGAR and MetaTool items. Our planned \emph{distortion test} fixed in advance what would count. For Jev 1.13: a confirmed effect whose private interval lies on the other side of zero, or a public-minus-private difference that passes Holm's correction and is at least 2 points and half the public effect. For the other models: differences that pass a joint Holm correction on two or more models in the same direction. Unlike GSM1k's human-written maths items {[}\citenum{r111}{]}, ours were written by language models.

\begingroup\footnotesize
\renewcommand{\thetable}{6}
\begin{longtable}{@{}>{\raggedright\arraybackslash}p{\dimexpr 0.2862\linewidth-2.8620\tabcolsep\relax}>{\raggedright\arraybackslash}p{\dimexpr 0.1721\linewidth-1.7214\tabcolsep\relax}>{\raggedright\arraybackslash}p{\dimexpr 0.1741\linewidth-1.7413\tabcolsep\relax}>{\raggedright\arraybackslash}p{\dimexpr 0.0908\linewidth-0.9077\tabcolsep\relax}>{\raggedright\arraybackslash}p{\dimexpr 0.1046\linewidth-1.0462\tabcolsep\relax}>{\raggedright\arraybackslash}p{\dimexpr 0.1721\linewidth-1.7214\tabcolsep\relax}@{}}
\caption{The eight techniques on public and private data (accuracy points, public \ensuremath{\rightarrow} private). Each side is the mean over its datasets; for the six others, a plain mean over models (ranges in Table Y.10). Clear effect: the 90\% interval excludes zero in the expected direction. Public \ensuremath{\neq} private: public-minus-private differences that pass Holm over the six models' 48 tests. Completeness is in the caveats below.}\\
\toprule
Technique & Jev 1.13 (Run 9) & Six others, mean & Clear effect (of 6) & Public \ensuremath{\neq} private & Jev 1.14 \\
\midrule
\endfirsthead
\multicolumn{6}{@{}l}{\textit{Table 6 (continued)}}\\
\toprule
Technique & Jev 1.13 (Run 9) & Six others, mean & Clear effect (of 6) & Public \ensuremath{\neq} private & Jev 1.14 \\
\midrule
\endhead
\bottomrule
\endlastfoot
Retrieved examples vs bare call & +7.7 \ensuremath{\rightarrow} +10.1 & +1.1 \ensuremath{\rightarrow} +0.3 & 4 \ensuremath{\rightarrow} 3 & 0 & +6.9 \ensuremath{\rightarrow} +9.7 \\
Cryptic ids vs readable names & \ensuremath{-}82.7 \ensuremath{\rightarrow} \ensuremath{-}85.6 & \ensuremath{-}83.4 \ensuremath{\rightarrow} \ensuremath{-}81.6 & 6 \ensuremath{\rightarrow} 6 & 0 & \ensuremath{-}83.4 \ensuremath{\rightarrow} \ensuremath{-}85.9 \\
Three wrong labels vs three correct & \ensuremath{-}49.5 \ensuremath{\rightarrow} \ensuremath{-}56.1 & \ensuremath{-}20.5 \ensuremath{\rightarrow} \ensuremath{-}23.6 & 6 \ensuremath{\rightarrow} 6 & 1 & \ensuremath{-}51.6 \ensuremath{\rightarrow} \ensuremath{-}54.4 \\
One rival vs distinct rivals & \ensuremath{-}13.8 \ensuremath{\rightarrow} \ensuremath{-}12.5 & \ensuremath{-}6.3 \ensuremath{\rightarrow} \ensuremath{-}3.3 & 6 \ensuremath{\rightarrow} 5 & 3 & \ensuremath{-}15.4 \ensuremath{\rightarrow} \ensuremath{-}14.6 \\
30\% confusable pool noise vs clean pool & \ensuremath{-}7.0 \ensuremath{\rightarrow} \ensuremath{-}7.8 & \ensuremath{-}12.6 \ensuremath{\rightarrow} \ensuremath{-}14.1 & 6 \ensuremath{\rightarrow} 6 & 1 & \ensuremath{-}7.1 \ensuremath{\rightarrow} \ensuremath{-}8.5 \\
Descriptions on readable names vs names & +2.3 \ensuremath{\rightarrow} +7.1 & +1.0 \ensuremath{\rightarrow} +7.6 & -- & 2 & +1.5 \ensuremath{\rightarrow} +7.3 \\
True class shares stated vs none & +1.1 \ensuremath{\rightarrow} +1.5 & +0.5 \ensuremath{\rightarrow} \ensuremath{-}0.1 & -- & 0 & +0.3 \ensuremath{\rightarrow} +1.0 \\
Persona and politeness vs bare call & +0.4 \ensuremath{\rightarrow} \ensuremath{-}0.8 & +0.1 \ensuremath{\rightarrow} \ensuremath{-}0.5 & -- & 0 & +0.1 \ensuremath{\rightarrow} \ensuremath{-}1.3 \\
\end{longtable}
\endgroup

\textbf{What the planned test found.} On Jev 1.13 it flagged nothing: no public-minus-private difference passed the correction (Table Y.9). On the six other models, seven of the 48 differences passed the joint correction, and the test flags two techniques: descriptions and the one-rival penalty. So leakage isn't ruled out for these two.

\textbf{The softer reading: the harms held.} On the private sets' mean, cryptic ids, three wrong labels and pool noise hurt on every model (five per-dataset cells on long texts went the other way; Appendix Y), and wrappers stayed within about 1.3 points of zero (Figure 4). Averaged over models, no technique with a clear effect changed sign. Over the six other models' 48 cells, one effect did: Kev-4B's one-rival penalty (\ensuremath{-}2.8 public, +6.7 private). Bare-call accuracy wasn't significantly higher on public data, except on Decider Flash (+4.7 points {[}+0.4, +9.1{]}). The techniques also ranked much the same on both sides (rank correlations 0.76 to 1.00), but over eight techniques that's a weak check, partly driven by two huge effects (0.49 to 0.94 on the six others without them; Table Y.14).

\textbf{Descriptions helped more on private data} (+7.6 against +1.0 on the six others). That held on the central estimate on every model, and the difference passed the correction on two. Descriptions helped most on OpsDeck's invented names: a description added 16.8 to 39.8 points, and names cut to the brand word lost 40.6 to 76.7. Leakage would predict this direction, but invented names are a simpler cause (Section 4.1); leakage isn't ruled out.

\textbf{The one-rival penalty was smaller on private data on three models} (the three differences that passed the correction; two others looked larger before correction). The test was two-sided. Only after the fact can we say the direction doesn't fit simple leakage, since memorised items should make wrong labels hurt less on public data, not more.

\textbf{Retrieved examples are the finding that weakens.} On the six others, examples gained +1.1 points on public data and +0.3 on private data, and held on 3 of 6 models there. Without ScotLease, which sits at the ceiling, the private mean is +5.7. The low private mean comes from that ceiling set and from two models that failed on long private texts. On ScotLease, Decider Flash fell from 98.9\% to 28.0\% when shown three correctly labelled examples, yet scored 96.9\% when the same examples carried wrong labels: it steers away from the option the examples support. It fell on BeamDesk too (94.8\% to 52.8\% with the 20 nearest examples), and on public LEDGAR (39 to 45\% against 93\% bare, but 86 to 94\% with relabelled examples). d1-ultra also fell on BeamDesk, from 90.6\% to 53.2\% with nearest examples (Appendix Y), and its examples didn't help on public LEDGAR either, so this is unlikely to be leakage.

\begin{figure}[!htbp]
\centering
\includegraphics[width=\linewidth,height=0.72\textheight,keepaspectratio,alt={Eight panels in two rows of four, one per technique: Retrieved examples; Cryptic ids (vs names); Three wrong labels shown; Wrong labels on one rival; 30  wrong labels in pool; Descriptions on names; Class shares stated; Persona, politeness. Each panel has one row per model, from top to bottom: Jev 1.14, Jev 1.13 (from Run 9, starred, hollow markers), Decider, Decider Flash, OpenAI Decisions, Microsoft-Decision-1, d1-ultra and Kev-4B. In each row a blue circle gives the mean effect on the three public datasets and an orange square, just below it, the mean on the four private datasets, each with a 90  interval; a grey vertical line marks zero, and the x axis is the change in accuracy in points. Retrieved examples: most models sit right of zero on both sides; Decider Flash sits left of zero on both (-14.5 public, -16.0 private), and d1-ultra moves from +1.1 public to -8.0 private. Cryptic ids: every marker lies between -88.5 and -72.5. Three wrong labels: the two Jevs lie between -49.5 and -56.1, the other models between -4.9 and -35.3, with the private square left of the public circle on every model except d1-ultra (-27.3 public, -12.2 private). Wrong labels on one rival: every public circle is left of zero (-2.8 to -15.4); Kev-4B's private square is right of zero at +6.7. 30  wrong labels in pool: every marker is left of zero, d1-ultra and Kev-4B furthest (-20.1 to -31.0). Descriptions on names: on every model the private square lies right of the public circle, private +5.2 to +11.2, public -2.0 to +5.0. Class shares stated: markers near zero except Decider Flash's private square at -4.5. Persona, politeness: every marker within about 1.3 points of zero. A footer reads "Change in accuracy (points); * Jev 1.13 from Run 9 (hollow)", with a legend for public and private datasets.}]{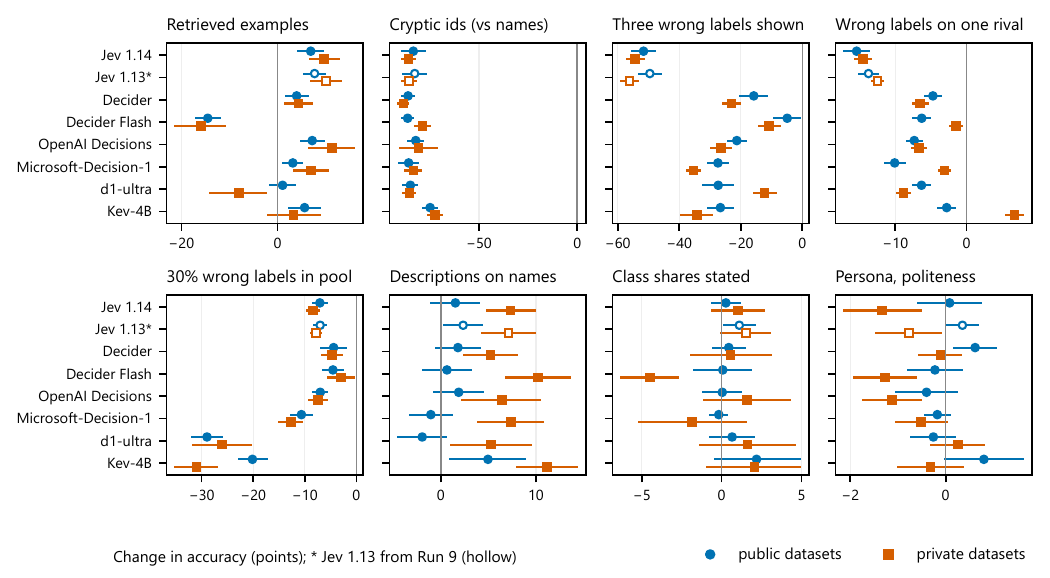}
\renewcommand{\thefigure}{4}
\caption{The eight techniques on public and private data, per model (change in accuracy, points; 90\% wild-cluster intervals). Blue circles: mean over the three public datasets; orange squares: mean over the four private ones. Jev 1.13 is from Run 9 (hollow). Sources: Appendix L.}
\end{figure}

\textbf{Caveats.} Synthetic items written and labelled by language models may be cleaner than real ones, and we dropped items on whose label the language models didn't reach a majority (Appendix Y.2), which may make the rest easier. So public-versus-private differences, the bare-call gap included, mix leakage with style and difficulty. ScotLease is at the ceiling (99.8\% bare on Jev 1.13), so it can show harms but hardly gains. Run 10 sent the private sets to seven more APIs, whose training policies we didn't check. Jev 1.13's cells are complete; Jev 1.14's KitchenOps cells are about 54\% complete (its OpenRouter credit ran out), and each contrast uses complete items only.

\section{What didn't work, or didn't move the needle}

Many things we tried did nothing reliable, and some cheap fixes didn't fix anything. Table 7 gathers them from the planned runs and the development screen (Appendices B to D and F).

\begingroup\footnotesize
\renewcommand{\thetable}{7}
\begin{longtable}{@{}>{\raggedright\arraybackslash}p{\dimexpr 0.1899\linewidth-1.1391\tabcolsep\relax}>{\raggedright\arraybackslash}p{\dimexpr 0.2910\linewidth-1.7460\tabcolsep\relax}>{\raggedright\arraybackslash}p{\dimexpr 0.3146\linewidth-1.8878\tabcolsep\relax}>{\raggedright\arraybackslash}p{\dimexpr 0.2045\linewidth-1.2271\tabcolsep\relax}@{}}
\caption{What did not work, or did not move the needle (accuracy points, 90\% intervals; Jev 1.13 unless stated; development rows are from the screen).}\\
\toprule
Technique & What we tried & Result & Status \\
\midrule
\endfirsthead
\multicolumn{4}{@{}l}{\textit{Table 7 (continued)}}\\
\toprule
Technique & What we tried & Result & Status \\
\midrule
\endhead
\bottomrule
\endlastfoot
Persona, politeness, capitals, unrelated questions & Fixed text on every item, fresh items (Runs 3, 8); persona and politeness also vs an equal-length neutral prefix & \ensuremath{-}0.3 to +0.3 vs bare; +1.08 {[}\ensuremath{-}0.02, +2.22{]} vs the prefix & within \ensuremath{\pm}2; not shown vs the prefix \\
Stating true class shares & Shares in the request vs none (Runs 2, 8) & +1.9 {[}+1.1, +2.7{]}; \ensuremath{-}0.4 {[}\ensuremath{-}1.0, +0.4{]} & observed; not shown \\
Stating shuffled shares & Shuffled shares vs none, LEDGAR (Run 8) & \ensuremath{-}1.1 {[}\ensuremath{-}2.0, \ensuremath{-}0.2{]}, a small harm & observed \\
Natural-share pool & Retrieval from LEDGAR at natural shares vs bare (Run 8) & +1.9 {[}\ensuremath{-}0.4, +4.9{]}, 1.4 below a balanced pool & not shown \\
Audit the most-retrieved examples first & 10\% audit of a noisy pool, most retrieved vs random (Run 8) & +0.3 {[}\ensuremath{-}0.4, +0.9{]} & not shown \\
Calibrate the bare call & Contextual calibration (bias from a content-free input); batch calibration (bias from the batch mean) (Run 8) & \ensuremath{-}2.2; +0.2 vs bare; retrieval 10.0 and 7.5 ahead & secondary \\
Local routes built on one bare call & At 100 per class, \ensuremath{\times} kNN vote \ensuremath{\times} classifier vs the prompt (Runs 4, 8); at 5 per class, \ensuremath{\times} centroid (Run 8) or \ensuremath{\times} classifier (Run 4), tested for no more than 1.5 worse & \ensuremath{-}0.7 {[}\ensuremath{-}1.6, +0.1{]} in both tests; \ensuremath{-}1.2 {[}\ensuremath{-}2.0, \ensuremath{-}0.5{]}; \ensuremath{-}0.6 {[}\ensuremath{-}1.7, +0.3{]} & not shown \\
Re-estimate shares after a time split & Shares re-estimated by EM {[}\citenum{r40}{]} from unlabelled 2024 complaints (Run 8) & \ensuremath{-}0.1 {[}\ensuremath{-}0.6, +0.4{]}; worst-class recall 49.5\% \ensuremath{\rightarrow} 22.2\% & secondary \\
Wording defences & Guard sentence, spotlighting, claim field vs 100 attacks that never saw the scores & success with the true label kept: \ensuremath{\geq} 38\%, 15\%, 55\% & observed \\
Definitions on readable names & Blind definitions on MetaTool's names (Run 8); descriptions on names, public vs private sets (Run 9) & +6.5 {[}+1.4, +12.3{]}, lost under correction; \ensuremath{-}1.0 and \ensuremath{-}0.4 on CLINC150 and LEDGAR, but +8.4 on MetaTool's original names (+10.7 on its two-word names, Table 4) and +25.0 on invented OpsDeck names & not shown (Run 8); helps when names are invented or uninformative (secondary) \\
Ids with definitions instead of names & Tested for no more than 2 worse, fresh items (Run 8) & \ensuremath{-}2.3 {[}\ensuremath{-}3.8, \ensuremath{-}1.0{]}; shown no more than 2 worse on 1 of 6 others & not shown \\
Development screen & Drop pool items the bare call disputes; model-written descriptions; definitions on top of examples; a closing reminder against injection & 80.0\% vs 80.2\%; no better than random words; 85.8\% \ensuremath{\rightarrow} 82.8\%; attack success .62 \ensuremath{\rightarrow} .64 & observed \\
\end{longtable}
\endgroup

Three patterns recur. Text that adds no information did nothing reliable. Information the model can't use or trust, such as stated shares, generated descriptions or definitions on top of examples, did little or harmed. And the cheaper routes, local or calibrated, didn't reach the examples prompt.

\section{Limitations}

\begin{enumerate}
\item \textbf{Adaptive design and selection.} We screened techniques on Jev 1.13 alone and re-tested the ones that mattered there, so the cross-model counts describe Jev's levers, not every lever; Run 5 also re-used Jev's items. The runs were sequential and adaptive: each was designed after the one before, which raises the chance that a later test confirms an earlier hunch. Holm's correction ran within each run only (Section 2.3), and only Run 3's plan is publicly archived.
\item \textbf{Benchmarks and pools.} Our ``fresh'' items come from public benchmarks, where knowledge and contamination are hard to separate {[}\citenum{r11}; \citenum{r41}{]}. On private data, the planned leakage test flagged two techniques (Section 7), language models wrote and checked those items, and rankings over eight techniques are a weak check, so leakage is neither shown nor ruled out. Techniques outside those eight rest on public items only. Most pools were class-balanced, and realistic pools gave smaller retrieval gains. All tasks are English single-label choices, although attribute judgement is the most common Jev use (77\%) {[}\citenum{r24}{]}. We didn't vary format {[}\citenum{r74}{]}, option order {[}\citenum{r75}{]}, the order of examples, the option set {[}\citenum{r36}{]}, question polarity {[}\citenum{r76}{]} or rule wording {[}\citenum{r43}{]}, and we didn't test impolite wording, which helped on one chat model {[}\citenum{r94}{]} and hurt in another study {[}\citenum{r54}{]}, emotional appeals, or tips and threats {[}\citenum{r30}{]}.
\item \textbf{Models.} The six vendors and the chat model are not a sample of all decision models, nor independent of one another (Section 6). The chat model ran on one dataset. Two anomalies are unexplained: Decider Flash's collapse when shown correct examples, and d1-ultra scoring far lower with two wrong labels than with three (Appendix U.2).
\item \textbf{Label budget.} The local classifiers are simple: a logistic regression on bge-small embeddings and a MiniLM encoder fine-tuned for at most 5 epochs. A stronger encoder might narrow the gap. The pooled estimates were made after the fact, over correlated comparisons. Labelling and engineering costs aren't counted.
\item \textbf{Fragile passes.} The three-example test sits near the accuracy ceiling. The unrelated-questions and definitions tests fail one correction over the 29 tests of Runs 1 to 4, and no correction spans the whole study (Section 2.3). The least secure passes left are Kev-4B's placement result and retrieval's fourth model on fresh items.
\item \textbf{Process.} One author planned, approved and ran all ten runs; AI agents ran the analyses. Runs 1 to 8 took about 42 hours (2026-11-09 00:41 to 2026-11-10 18:35 UTC), and Runs 9 and 10 followed from 2026-11-10 to 12. Run 9's full run went ahead on the author's advance approval, without a stop after its pilot.
\item \textbf{Practice.} We scored top-1 accuracy only. We report no log-loss, Brier score or calibration, although the probabilities are cached, cascades and thresholds rely on them, and wrong labels might hurt them more than accuracy. Latency was measured in single sessions; the security results use 300 author-written tickets; the item itself may be sensitive.
\end{enumerate}

\textbf{Ethics.} The attack and injection results echo published adaptive attacks {[}\citenum{r34}; \citenum{r17}{]} and prompt-injection behaviour {[}\citenum{r96}{]}, so no separate vendor notice was sent; Section 3 states the defensive lesson. Harmful prompts are reported only in aggregate. For CFPB narratives, Enron emails and the non-commercial sets, released files hold only identifiers and scores unless redistribution is confirmed.

\section{Conclusion}

Across ten runs, accuracy followed the information in a request, not the benign decoration around it. Benign wrappers stayed within 2 points. Retrieved examples helped Jev on public and private data and most other models on public items, but only half of them on private data, and misled two models on long texts; descriptions helped where names were vague or invented. Wrong information travelled well: noisy pools hurt every model, and wrong labels piled on one rival hurt all six other models on public items and five of six on private data. Our planned leakage test flagged two techniques on other models, so leakage isn't ruled out there; still, the harms held on private data and no clear effect changed sign on average. For practice, test changes against content-free controls on fresh items, on the model and version you deploy. For research, the harms we tested held on more vendors than the retrieval gain, so studies should test what hurts, not only what helps, and resample by class.

\section*{Statements}
\addcontentsline{toc}{section}{Statements}

\begin{itemize}
\item \textbf{Author.} Benjamin Jacob Bodner.
\item \textbf{Funding.} The author received no external funding for this work.
\item \textbf{Competing interests:} none. The author has no affiliation with, funding from, or other relationship with TypeSafe, or with any other vendor whose models are tested here; this is independent, external work, and no vendor reviewed or endorsed it.
\item \textbf{Acknowledgement of AI use.} The author used Claude models (Anthropic), through Claude Code, for code, running and analysing the experiments, literature search, figures and worked examples, and initial drafting and editing. Language models also wrote the screen's option definitions, the attack paragraphs, the label-preservation checks and the later phrasing stimuli (blind to items; Appendix J), and they wrote and labelled the private synthetic datasets of Run 9 (Appendix Y). The author edited the text, approved every analysis plan and is responsible for the content of this paper.
\item \textbf{Analysis plans.} Each of the eight runs on public benchmarks followed a written analysis plan, approved by the author and fixed in advance with a SHA-256 manifest (Section 2.3; Appendix P.3 lists the hashes), and Runs 9 and 10 followed the same practice (Appendix Y). Run 3's plan is publicly archived at \url{https://gist.github.com/BjBodner/c736ea7d935967253f4fae06ff51da90} (created 2026-11-09 15:17:47 UTC; first billed Run 3 call 15:39:59 UTC). The other plans are now public in the repository's reproducibility package; their hashes show that they did not change, not when they were written.
\item \textbf{Code and data.} Public at \url{https://github.com/BjBodner/decision-levers}: the starter package, with code and examples, and a reproducibility package (\code{reproduce/\allowbreak{}}) with the research code, every analysis plan and manifest, the analysis and figure scripts, aggregate result tables and the exact version of every model used. The code re-runs every experiment with your own API keys; raw model outputs and embedding caches are not included. The items of the four synthetic datasets of Runs 9 and 10 stay private, so that they stay unseen: the package holds only their SHA-256 hashes; Appendix Y gives their label sets and aggregate results. The extended appendix accompanies this paper.
\item \textbf{Licences.} This paper is distributed under CC BY 4.0; Appendix K gives every dataset's licence.
\end{itemize}

\FloatBarrier
\section*{References}
\addcontentsline{toc}{section}{References}
{\small All 2026 entries except {[}62{]}, {[}69{]} and {[}76{]} are preprints. arXiv identifiers, titles and authors were checked against the arXiv API, and journal entries against Crossref, in November 2026; entries {[}57{]} to {[}111{]} follow the alphabetical list {[}1{]} to {[}56{]}. Brom 2026, Iqbal 2026, Liu 2026, Lucca 2026 and Xu 2026 (JevOut) were read at abstract level only; every claim we make about them is stated in their abstracts. Sharma 2026 (TypedBench) and Sun 2026 were read in full; the 24-point rewording figure is in TypedBench's Figure 2 caption and Discussion, not its abstract. The numbers quoted from {[}70{]} and {[}71{]} were checked against their full texts, as was the absence of a published analysis plan in {[}78{]}; {[}72{]} and {[}82{]} were checked against their full texts, the other entries from {[}73{]} on are cited for their abstract-level claims, {[}86{]} to {[}89{]} were checked against the arXiv API, and {[}12{]} against its full text. {[}90{]} to {[}93{]} and {[}95{]} to {[}100{]} were checked against the arXiv API or Crossref, the claims cited from {[}92{]}, {[}97{]} and {[}99{]} against their full texts, and {[}94{]} is cited for its abstract-level claim. {[}101{]} to {[}105{]} are entries of the extended appendix's list ({[}E23{]}, {[}E17{]}, {[}E25{]}, {[}E76{]}, {[}E16{]}); {[}106{]} and {[}107{]} were checked against their arXiv pages; all seven are cited for abstract-level claims, and {[}102{]} is cited as a contrast, because it finds Transformers trained on simple function classes resilient to label noise. {[}108{]} (the extended appendix's {[}E49{]}) and {[}111{]} were checked against arXiv and their venues' pages, and {[}109{]} and {[}110{]} against Crossref, in November 2026. Holm 1979 has no Crossref DOI and is cited by its JSTOR stable URL. {[}62{]} is vendor documentation read on 2026-11-09; {[}69{]} is the Run 5 vendors' API documentation and model catalogue, as read on 2026-11-09 and recorded in the Run 5 analysis plan; {[}76{]} is a blog post by an AI-observability company, not peer-reviewed, cited only for question polarity. References cited only in the extended appendix are listed there and cited there as {[}E1{]} onward.}

\begingroup\small
\begin{refsnoheading}{111}
\bibitem[1]{r1} Alexandari, Kundaje, Shrikumar, Maximum Likelihood with Bias-Corrected Calibration is Hard-To-Beat at Label Shift Adaptation (2019), ICML 2020; \href{https://arxiv.org/abs/1901.06852}{arXiv:1901.06852}.
\bibitem[2]{r2} Basil et al., Prompting Science Report 4: Playing Pretend: Expert Personas Don't Improve Factual Accuracy (2025), \href{https://arxiv.org/abs/2512.05858}{arXiv:2512.05858}.
\bibitem[3]{r3} Brom, Confidence Is Not a Promise: A Reference Audit of Typed Decision Models (2026), \href{https://arxiv.org/abs/2610.36311}{arXiv:2610.36311}.
\bibitem[4]{r4} Cameron, Gelbach, Miller, Bootstrap-Based Improvements for Inference with Clustered Errors (2008), Review of Economics and Statistics 90(3):414-427, \href{https://doi.org/10.1162/rest.90.3.414}{doi:10.1162/rest.90.3.414}.
\bibitem[5]{r5} Casanueva et al., Efficient Intent Detection with Dual Sentence Encoders (2020), NLP4ConvAI Workshop 2020; \href{https://arxiv.org/abs/2003.04807}{arXiv:2003.04807}.
\bibitem[6]{r6} Chalkidis et al., LexGLUE: A Benchmark Dataset for Legal Language Understanding in English (2021), ACL 2022; \href{https://arxiv.org/abs/2110.00976}{arXiv:2110.00976}.
\bibitem[7]{r7} Clark et al., BoolQ: Exploring the Surprising Difficulty of Natural Yes/No Questions (2019), NAACL 2019; \href{https://arxiv.org/abs/1905.10044}{arXiv:1905.10044}.
\bibitem[8]{r8} Dodge et al., Show Your Work: Improved Reporting of Experimental Results (2019), EMNLP-IJCNLP 2019; \href{https://arxiv.org/abs/1909.03004}{arXiv:1909.03004}.
\bibitem[9]{r9} Edwards, Camacho-Collados, Language Models for Text Classification: Is In-Context Learning Enough? (2024), LREC-COLING 2024; \href{https://arxiv.org/abs/2403.17661}{arXiv:2403.17661}.
\bibitem[10]{r10} Gao, Fisch, Chen, Making Pre-trained Language Models Better Few-shot Learners (2020), ACL 2021; \href{https://arxiv.org/abs/2012.15723}{arXiv:2012.15723}.
\bibitem[11]{r11} Golchin, Surdeanu, Time Travel in LLMs: Tracing Data Contamination in Large Language Models (2023), ICLR 2024; \href{https://arxiv.org/abs/2308.08493}{arXiv:2308.08493}.
\bibitem[12]{r12} Halawi, Denain, Steinhardt, Overthinking the Truth: Understanding how Language Models Process False Demonstrations (2023), ICLR 2024; \href{https://arxiv.org/abs/2307.09476}{arXiv:2307.09476}.
\bibitem[13]{r13} Han et al., Understanding Emergent In-Context Learning from a Kernel Regression Perspective (2023), TMLR 2025; \href{https://arxiv.org/abs/2305.12766}{arXiv:2305.12766}.
\bibitem[14]{r14} Hines et al., Defending Against Indirect Prompt Injection Attacks With Spotlighting (2024), \href{https://arxiv.org/abs/2403.14720}{arXiv:2403.14720}.
\bibitem[15]{r15} Holm, A Simple Sequentially Rejective Multiple Test Procedure (1979), Scandinavian Journal of Statistics 6(2):65-70, \url{https://www.jstor.org/stable/4615733}.
\bibitem[16]{r16} Huang et al., MetaTool Benchmark for Large Language Models: Deciding Whether to Use Tools and Which to Use (2023), ICLR 2024; \href{https://arxiv.org/abs/2310.03128}{arXiv:2310.03128}.
\bibitem[17]{r17} Hughes et al., Best-of-N Jailbreaking (2024), \href{https://arxiv.org/abs/2412.03556}{arXiv:2412.03556}.
\bibitem[18]{r18} Iqbal, Nasim, Saleem, Typed Decision Models under Prevalence Shift: Decision Training, Calibration, and Implicit Priors (2026), \href{https://arxiv.org/abs/2610.26325}{arXiv:2610.26325}.
\bibitem[19]{r19} Kittler, Hatef, Duin, Matas, On Combining Classifiers (1998), IEEE Transactions on Pattern Analysis and Machine Intelligence 20(3):226-239, \href{https://doi.org/10.1109/34.667881}{doi:10.1109/34.667881}.
\bibitem[20]{r20} Kossen, Gal, Rainforth, In-Context Learning Learns Label Relationships but Is Not Conventional Learning (2023), ICLR 2024; \href{https://arxiv.org/abs/2307.12375}{arXiv:2307.12375}.
\bibitem[21]{r21} Larson et al., An Evaluation Dataset for Intent Classification and Out-of-Scope Prediction (2019), EMNLP-IJCNLP 2019; \href{https://arxiv.org/abs/1909.02027}{arXiv:1909.02027}.
\bibitem[22]{r22} Le Scao, Rush, How Many Data Points is a Prompt Worth? (2021), NAACL 2021; \href{https://arxiv.org/abs/2103.08493}{arXiv:2103.08493}.
\bibitem[23]{r23} Lin et al., ToxicChat: Unveiling Hidden Challenges of Toxicity Detection in Real-World User-AI Conversation (2023), Findings of EMNLP 2023; \href{https://arxiv.org/abs/2310.17389}{arXiv:2310.17389}.
\bibitem[24]{r24} Ling, Xue, Ye, Jev in the Wild: A Data-Driven Analysis of the Jev Model's Functionality, Applications and Ecosystem (2026), \href{https://arxiv.org/abs/2609.30216}{arXiv:2609.30216}.
\bibitem[25]{r25} Liu et al., What Makes Good In-Context Examples for GPT-3? (2021), \href{https://arxiv.org/abs/2101.06804}{arXiv:2101.06804}.
\bibitem[26]{r26} Liu, Wang, Xiao, FlipJev: Fast Black-Box Red Teaming Finds Universal Adversarial Suffixes for Typed Decision Models (2026), \href{https://arxiv.org/abs/2610.38895}{arXiv:2610.38895}.
\bibitem[27]{r27} Lucca, Measure Before You Train: Typed Decision Models vs. a Classical Baseline vs. a Generative LLM for Routing a Personal Multilingual Inbox (2026), \href{https://arxiv.org/abs/2610.32955}{arXiv:2610.32955}.
\bibitem[28]{r28} Lyu et al., Z-ICL: Zero-Shot In-Context Learning with Pseudo-Demonstrations (2022), ACL 2023; \href{https://arxiv.org/abs/2212.09865}{arXiv:2212.09865}.
\bibitem[29]{r29} Mazzoni, Failla, Benchmarking System One Models in Online Moderation (2026), \href{https://arxiv.org/abs/2610.07953}{arXiv:2610.07953}.
\bibitem[30]{r30} Meincke et al., Prompting Science Report 3: I'll pay you or I'll kill you -- but will you care? (2025), \href{https://arxiv.org/abs/2508.00614}{arXiv:2508.00614}.
\bibitem[31]{r31} Milios, Reddy, Bahdanau, In-Context Learning for Text Classification with Many Labels (2023), \href{https://arxiv.org/abs/2309.10954}{arXiv:2309.10954}.
\bibitem[32]{r32} Min et al., Rethinking the Role of Demonstrations: What Makes In-Context Learning Work? (2022), EMNLP 2022; \href{https://arxiv.org/abs/2202.12837}{arXiv:2202.12837}.
\bibitem[33]{r33} Mosbach, Pimentel, Ravfogel, Klakow, Elazar, Few-shot Fine-tuning vs. In-context Learning: A Fair Comparison and Evaluation (2023), Findings of ACL 2023; \href{https://arxiv.org/abs/2305.16938}{arXiv:2305.16938}.
\bibitem[34]{r34} Nasr et al., The Attacker Moves Second: Stronger Adaptive Attacks Bypass Defenses Against Llm Jailbreaks and Prompt Injections (2025), \href{https://arxiv.org/abs/2510.09023}{arXiv:2510.09023}.
\bibitem[35]{r35} Qwen Team, Qwen2.5 Technical Report (2024), \href{https://arxiv.org/abs/2412.15115}{arXiv:2412.15115}.
\bibitem[36]{r36} Rajput, Chauhan, Kosaraju, Does a Hosted Decision Model Respect Irrelevant Alternatives? A Pre-Registered Test of Option-Set Dependence in Jev (2026), \href{https://arxiv.org/abs/2610.16655}{arXiv:2610.16655}.
\bibitem[37]{r37} Ram\'{\i}rez, Lindemann, Birch, Titov, Cache \& Distil: Optimising API Calls to Large Language Models (2023), Findings of ACL 2024; \href{https://arxiv.org/abs/2310.13561}{arXiv:2310.13561}.
\bibitem[38]{r38} Reimers, Gurevych, Sentence-BERT: Sentence Embeddings using Siamese BERT-Networks (2019), EMNLP-IJCNLP 2019; \href{https://arxiv.org/abs/1908.10084}{arXiv:1908.10084}.
\bibitem[39]{r39} Rubin, Herzig, Berant, Learning To Retrieve Prompts for In-Context Learning (2021), NAACL 2022; \href{https://arxiv.org/abs/2112.08633}{arXiv:2112.08633}.
\bibitem[40]{r40} Saerens, Latinne, Decaestecker, Adjusting the outputs of a classifier to new a priori probabilities: a simple procedure (2002), Neural Computation 14(1):21-41, \href{https://doi.org/10.1162/089976602753284446}{doi:10.1162/089976602753284446}.
\bibitem[41]{r41} Sainz et al., NLP Evaluation in trouble: On the Need to Measure LLM Data Contamination for each Benchmark (2023), Findings of EMNLP 2023; \href{https://arxiv.org/abs/2310.18018}{arXiv:2310.18018}.
\bibitem[42]{r42} Schuirmann, A comparison of the two one-sided tests procedure and the power approach for assessing the equivalence of average bioavailability (1987), Journal of Pharmacokinetics and Biopharmaceutics 15(6):657-680, \href{https://doi.org/10.1007/BF01068419}{doi:10.1007/BF01068419}.
\bibitem[43]{r43} Sharma et al., TypedBench: A Benchmark for Calibration, Framing Sensitivity, and Cost in System One Decision Models (2026), \href{https://arxiv.org/abs/2610.11392}{arXiv:2610.11392}.
\bibitem[44]{r44} Sun et al., Type-Safe Is Not Error-Free: Typed Decision Models Follow the Option Name, Not the Definition Bound to It (2026), \href{https://arxiv.org/abs/2609.26758}{arXiv:2609.26758}.
\bibitem[45]{r45} Tunstall et al., Efficient Few-Shot Learning Without Prompts (2022), \href{https://arxiv.org/abs/2209.11055}{arXiv:2209.11055}.
\bibitem[46]{r46} Varshney, Baral, Model Cascading: Towards Jointly Improving Efficiency and Accuracy of NLP Systems (2022), EMNLP 2022; \href{https://arxiv.org/abs/2210.05528}{arXiv:2210.05528}.
\bibitem[47]{r47} Vaugrante, Niepert, Hagendorff, A Looming Replication Crisis in Evaluating Behavior in Language Models? Evidence and Solutions (2024), \href{https://arxiv.org/abs/2409.20303}{arXiv:2409.20303}.
\bibitem[48]{r48} Wang et al., MiniLM: Deep Self-Attention Distillation for Task-Agnostic Compression of Pre-Trained Transformers (2020), NeurIPS 2020; \href{https://arxiv.org/abs/2002.10957}{arXiv:2002.10957}.
\bibitem[49]{r49} Webb, Reworking Wild Bootstrap-Based Inference for Clustered Errors (2023), Canadian Journal of Economics 56(3):839-858, \href{https://doi.org/10.1111/caje.12661}{doi:10.1111/caje.12661}.
\bibitem[50]{r50} Wei et al., Larger language models do in-context learning differently (2023), \href{https://arxiv.org/abs/2303.03846}{arXiv:2303.03846}.
\bibitem[51]{r51} Wu, Chang, Yan, Poisoned into Action: Context Attacks on Typed Decision Workflows (2026), \href{https://arxiv.org/abs/2610.15537}{arXiv:2610.15537}.
\bibitem[52]{r52} Xiao et al., C-Pack: Packed Resources For General Chinese Embeddings (2023), SIGIR 2024; \href{https://arxiv.org/abs/2309.07597}{arXiv:2309.07597}.
\bibitem[53]{r53} Xu et al., JevOut: Natural Context Can Flip Decision Models (2026), \href{https://arxiv.org/abs/2609.30243}{arXiv:2609.30243}.
\bibitem[54]{r54} Yin et al., Should We Respect LLMs? A Cross-Lingual Study on the Influence of Prompt Politeness on LLM Performance (2024), SICon 2024; \href{https://arxiv.org/abs/2402.14531}{arXiv:2402.14531}.
\bibitem[55]{r55} Yoo et al., Ground-Truth Labels Matter: A Deeper Look into Input-Label Demonstrations (2022), EMNLP 2022; \href{https://arxiv.org/abs/2205.12685}{arXiv:2205.12685}.
\bibitem[56]{r56} Zheng et al., When ``A Helpful Assistant'' Is Not Really Helpful: Personas in System Prompts Do Not Improve Performances of Large Language Models (2023), Findings of EMNLP 2024; \href{https://arxiv.org/abs/2311.10054}{arXiv:2311.10054}.
\bibitem[57]{r57} FitzGerald et al., MASSIVE: A 1M-Example Multilingual Natural Language Understanding Dataset with 51 Typologically-Diverse Languages (2022), ACL 2023; \href{https://arxiv.org/abs/2204.08582}{arXiv:2204.08582}.
\bibitem[58]{r58} Owen, The Pigeonhole Bootstrap (2007), Annals of Applied Statistics 1(2):386-411, \href{https://doi.org/10.1214/07-AOAS122}{doi:10.1214/07-AOAS122}.
\bibitem[59]{r59} Xie et al., An Explanation of In-context Learning as Implicit Bayesian Inference (2021), ICLR 2022; \href{https://arxiv.org/abs/2111.02080}{arXiv:2111.02080}.
\bibitem[60]{r60} Wang et al., Label Words are Anchors: An Information Flow Perspective for Understanding In-Context Learning (2023), EMNLP 2023; \href{https://arxiv.org/abs/2305.14160}{arXiv:2305.14160}.
\bibitem[61]{r61} Olsson et al., In-context Learning and Induction Heads (2022), \href{https://arxiv.org/abs/2209.11895}{arXiv:2209.11895}.
\bibitem[62]{r62} TypeSafe, System One, Primitives (Questions) and AI primer (2026), product documentation, \url{https://docs.typesafe.ai/concepts/system-one.md}, \url{https://docs.typesafe.ai/primitives.md} (``evaluated independently'') and \url{https://docs.typesafe.ai/introduction/machine-learning-primer.md} (accessed 2026-11-09).
\bibitem[63]{r63} Hendel, Geva, Globerson, In-Context Learning Creates Task Vectors (2023), Findings of EMNLP 2023; \href{https://arxiv.org/abs/2310.15916}{arXiv:2310.15916}.
\bibitem[64]{r64} Todd, Li, Sen Sharma, Mueller, Wallace, Bau, Function Vectors in Large Language Models (2023), ICLR 2024; \href{https://arxiv.org/abs/2310.15213}{arXiv:2310.15213}.
\bibitem[65]{r65} Xu, Xu, Wang, Liu, Zhu, McAuley, Small Models are Valuable Plug-ins for Large Language Models (2023), \href{https://arxiv.org/abs/2305.08848}{arXiv:2305.08848}.
\bibitem[66]{r66} von Oswald et al., Transformers learn in-context by gradient descent (2022), \href{https://arxiv.org/abs/2212.07677}{arXiv:2212.07677}.
\bibitem[67]{r67} Lipton, Wang, Smola, Detecting and Correcting for Label Shift with Black Box Predictors (2018), ICML 2018; \href{https://arxiv.org/abs/1802.03916}{arXiv:1802.03916}.
\bibitem[68]{r68} Wang, Zhu, Saxon, Steyvers, Wang, Large Language Models Are Latent Variable Models: Explaining and Finding Good Demonstrations for In-Context Learning (2023), NeurIPS 2023; \href{https://arxiv.org/abs/2301.11916}{arXiv:2301.11916}.
\bibitem[69]{r69} Perplexity, OpenAI, OpenRouter, Decision-model API documentation and model catalogue (2026), product documentation, \url{https://docs.perplexity.ai/api-reference/decisions-post}, \url{https://developers.openai.com/api/docs/guides/decisions.md} and \url{https://openrouter.ai/api/v1/models?output_modalities=decisions} (accessed 2026-11-09).
\bibitem[70]{r70} Beno, Decision Models as Policy Gates for Coding Agents: Explicit Policy Boundaries Improve Detection and Reduce False Alarms (2026), \href{https://arxiv.org/abs/2611.02155}{arXiv:2611.02155}.
\bibitem[71]{r71} Steiner, Peeters, Bizer, Entity Matching using Decision Models (2026), \href{https://arxiv.org/abs/2610.34626}{arXiv:2610.34626}.
\bibitem[72]{r72} Kang et al., In-Context Learning with Noisy Labels (2024), \href{https://arxiv.org/abs/2411.19581}{arXiv:2411.19581}.
\bibitem[73]{r73} Meincke et al., Prompting Science Report 1: Prompt Engineering is Complicated and Contingent (2025), \href{https://arxiv.org/abs/2503.04818}{arXiv:2503.04818}.
\bibitem[74]{r74} Sclar et al., Quantifying Language Models' Sensitivity to Spurious Features in Prompt Design or: How I learned to start worrying about prompt formatting (2023), ICLR 2024; \href{https://arxiv.org/abs/2310.11324}{arXiv:2310.11324}.
\bibitem[75]{r75} Pezeshkpour, Hruschka, Large Language Models Sensitivity to The Order of Options in Multiple-Choice Questions (2023), Findings of NAACL 2024; \href{https://arxiv.org/abs/2308.11483}{arXiv:2308.11483}.
\bibitem[76]{r76} Bennett, Decision model benchmark: Jev, Kev, Liquid d1, and more (2026), Arize AI blog, \url{https://arize.com/blog/decision-model-benchmark} (accessed 2026-11-10).
\bibitem[77]{r77} Ibrahim, Zaki, Evaluating Decision Models for Text Annotation in Computational Social Science (2026), \href{https://arxiv.org/abs/2609.24574}{arXiv:2609.24574}.
\bibitem[78]{r78} Agrawal, Mamidi, Presentation Sensitivity in Typed Decision Models for Subjective Text Evaluation (2026), \href{https://arxiv.org/abs/2611.06384}{arXiv:2611.06384}.
\bibitem[79]{r79} Pecher, Srba, Bielikova, Comparing Specialised Small and General Large Language Models on Text Classification: 100 Labelled Samples to Achieve Break-Even Performance (2024), EMNLP 2025; \href{https://arxiv.org/abs/2402.12819}{arXiv:2402.12819}.
\bibitem[80]{r80} Hu et al., JevAdvBench: A Benchmark and Black-Box Attacks for Reinforcement Learning for Calibrated Decisions Models (2026), \href{https://arxiv.org/abs/2609.31142}{arXiv:2609.31142}.
\bibitem[81]{r81} Card et al., With Little Power Comes Great Responsibility (2020), EMNLP 2020; \href{https://arxiv.org/abs/2010.06595}{arXiv:2010.06595}.
\bibitem[82]{r82} Halterman, Keith, Codebook LLMs: Evaluating LLMs as Measurement Tools for Political Science Concepts (2024), Political Analysis 34:188-204 (2026); \href{https://arxiv.org/abs/2407.10747}{arXiv:2407.10747}.
\bibitem[83]{r83} Cobbina, Zhou, Where to show Demos in Your Prompt: A Positional Bias of In-Context Learning (2025), \href{https://arxiv.org/abs/2507.22887}{arXiv:2507.22887}.
\bibitem[84]{r84} Zhao et al., Calibrate Before Use: Improving Few-Shot Performance of Language Models (2021), ICML 2021; \href{https://arxiv.org/abs/2102.09690}{arXiv:2102.09690}.
\bibitem[85]{r85} Fei et al., Mitigating Label Biases for In-context Learning (2023), ACL 2023; \href{https://arxiv.org/abs/2305.19148}{arXiv:2305.19148}.
\bibitem[86]{r86} Pan, Gao, Chen, Chen, What In-Context Learning ``Learns'' In-Context: Disentangling Task Recognition and Task Learning (2023), Findings of ACL 2023; \href{https://arxiv.org/abs/2305.09731}{arXiv:2305.09731}.
\bibitem[87]{r87} Wei et al., Symbol tuning improves in-context learning in language models (2023), EMNLP 2023; \href{https://arxiv.org/abs/2305.08298}{arXiv:2305.08298}.
\bibitem[88]{r88} Zhou et al., Batch Calibration: Rethinking Calibration for In-context Learning and Prompt Engineering (2023), ICLR 2024; \href{https://arxiv.org/abs/2309.17249}{arXiv:2309.17249}.
\bibitem[89]{r89} Webson, Pavlick, Do Prompt-Based Models Really Understand the Meaning of their Prompts? (2021), NAACL 2022; \href{https://arxiv.org/abs/2109.01247}{arXiv:2109.01247}.
\bibitem[90]{r90} Azizi, Baghaei Potraghloo, Pedram, Labels Override Definitions in Jev-Style Typed Decision Models (2026), \href{https://arxiv.org/abs/2610.02586}{arXiv:2610.02586}.
\bibitem[91]{r91} Xu et al., kNN Prompting: Beyond-Context Learning with Calibration-Free Nearest Neighbor Inference (2023), ICLR 2023; \href{https://arxiv.org/abs/2303.13824}{arXiv:2303.13824}.
\bibitem[92]{r92} Miller, Adding Error Bars to Evals: A Statistical Approach to Language Model Evaluations (2024), \href{https://arxiv.org/abs/2411.00640}{arXiv:2411.00640}.
\bibitem[93]{r93} Bertsch et al., In-Context Learning with Long-Context Models: An In-Depth Exploration (2024), NAACL 2025; \href{https://arxiv.org/abs/2405.00200}{arXiv:2405.00200}.
\bibitem[94]{r94} Dobariya, Kumar, Mind Your Tone: Investigating How Prompt Politeness Affects LLM Accuracy (2025), \href{https://arxiv.org/abs/2510.04950}{arXiv:2510.04950}.
\bibitem[95]{r95} DerSimonian, Laird, Meta-analysis in clinical trials (1986), Controlled Clinical Trials 7(3):177-188, \href{https://doi.org/10.1016/0197-2456(86)90046-2}{doi:10.1016/0197-2456(86)90046-2}.
\bibitem[96]{r96} Greshake, Abdelnabi, Mishra, Endres, Holz, Fritz, Not what you've signed up for: Compromising Real-World LLM-Integrated Applications with Indirect Prompt Injection (2023), \href{https://arxiv.org/abs/2302.12173}{arXiv:2302.12173}.
\bibitem[97]{r97} Mohammadi, Vedula, Lamba, Raff, Kumaraguru, Ferraro, Gaur, Do LLMs Adhere to Label Definitions? Examining Their Receptivity to External Label Definitions (2025), EMNLP 2025; \href{https://arxiv.org/abs/2509.02452}{arXiv:2509.02452}.
\bibitem[98]{r98} Chen, Zaharia, Zou, FrugalGPT: How to Use Large Language Models While Reducing Cost and Improving Performance (2023), TMLR 2024; \href{https://arxiv.org/abs/2305.05176}{arXiv:2305.05176}.
\bibitem[99]{r99} Loukas, Stogiannidis, Diamantopoulos, Malakasiotis, Vassos, Making LLMs Worth Every Penny: Resource-Limited Text Classification in Banking (2023), ICAIF 2023; \href{https://arxiv.org/abs/2311.06102}{arXiv:2311.06102}.
\bibitem[100]{r100} Lakens, Equivalence Tests: A Practical Primer for t Tests, Correlations, and Meta-Analyses (2017), Social Psychological and Personality Science 8(4):355-362, \href{https://doi.org/10.1177/1948550617697177}{doi:10.1177/1948550617697177}.
\bibitem[101]{r101} Gupta et al., How Robust are LLMs to In-Context Majority Label Bias? (2023), AAAI 2024 Workshop on Responsible Language Modeling; \href{https://arxiv.org/abs/2312.16549}{arXiv:2312.16549}.
\bibitem[102]{r102} Cheng et al., Exploring the Robustness of In-Context Learning with Noisy Labels (2024), ICLR 2024 Workshop on Reliable and Responsible Foundation Models; \href{https://arxiv.org/abs/2404.18191}{arXiv:2404.18191}.
\bibitem[103]{r103} He et al., Data Poisoning for In-context Learning (2024), NAACL 2025; \href{https://arxiv.org/abs/2402.02160}{arXiv:2402.02160}.
\bibitem[104]{r104} Zou et al., PoisonedRAG: Knowledge Corruption Attacks to Retrieval-Augmented Generation of Large Language Models (2024), USENIX Security 2025; \href{https://arxiv.org/abs/2402.07867}{arXiv:2402.07867}.
\bibitem[105]{r105} Chen et al., StruQ: Defending Against Prompt Injection with Structured Queries (2024), USENIX Security 2025; \href{https://arxiv.org/abs/2402.06363}{arXiv:2402.06363}.
\bibitem[106]{r106} Shi, Chen, Misra, Scales, Dohan, Chi, Sch\"{a}rli, Zhou, Large Language Models Can Be Easily Distracted by Irrelevant Context (2023), ICML 2023; \href{https://arxiv.org/abs/2302.00093}{arXiv:2302.00093}.
\bibitem[107]{r107} Zheng, Zhou, Meng, Zhou, Huang, Large Language Models Are Not Robust Multiple Choice Selectors (2023), ICLR 2024; \href{https://arxiv.org/abs/2309.03882}{arXiv:2309.03882}.
\bibitem[108]{r108} Mizrahi, Kaplan, Malkin, Dror, Shahaf, Stanovsky, State of What Art? A Call for Multi-Prompt LLM Evaluation (2023), TACL 2024 12:933-949; \href{https://arxiv.org/abs/2401.00595}{arXiv:2401.00595}.
\bibitem[109]{r109} MacKinnon, Webb, Wild Bootstrap Inference for Wildly Different Cluster Sizes (2017), Journal of Applied Econometrics 32(2):233-254, \href{https://doi.org/10.1002/jae.2508}{doi:10.1002/jae.2508}.
\bibitem[110]{r110} Dror, Baumer, Shlomov, Reichart, The Hitchhiker's Guide to Testing Statistical Significance in Natural Language Processing (2018), ACL 2018, pp. 1383-1392, \href{https://doi.org/10.18653/v1/P18-1128}{doi:10.18653/v1/P18-1128}.
\bibitem[111]{r111} Zhang et al., A Careful Examination of Large Language Model Performance on Grade School Arithmetic (2024), NeurIPS 2024 Datasets and Benchmarks; \href{https://arxiv.org/abs/2405.00332}{arXiv:2405.00332}.
\end{refsnoheading}
\endgroup

\FloatBarrier
\clearpage
\appendix
\appendixheads
\section{What hurts, mechanism status and how to read these appendices}

\textbf{About these appendices.} They keep what a reader needs to check the paper: the evidence behind each finding, the method, the exact texts, one worked example in full and three key algorithms. The bulky material is in the extended appendix, which accompanies the paper (Statements). It holds the full versions of these appendices, under the same letters and subsection numbers, so ``Extended Appendix E.3'' names subsection E.3 there, plus the post-hoc re-analyses (Extended Appendix R), the full Run 5 and Run 6 results (Extended Appendix S), the 76-work comparison with prior work (Extended Appendix T), Runs 7 and 8 in full (Extended Appendices U and V), the second and third post-hoc re-analyses (Extended Appendices W and X), and Run 9 in full (Extended Appendix Y). The post-hoc re-analyses on cached data are numbered 1 to 3 here; the result files (\code{r16\_\allowbreak{}}, \code{r17\_\allowbreak{}}, \code{r18\_\allowbreak{}}) and the extended appendix call them rounds 16, 17 and 18. Because the letters follow the extended appendix, this PDF has Appendices A to P, then U and Y: Q to T, V, W and X exist only in the extended appendix. Material that earlier layouts of the paper kept in the main text is kept here in full: the overview figure of Runs 1 to 4 (Figure A.1), the evidence ledger and decision flow of the rules (Table A.2, Figure A.2), what hurts (A.3), the longer comparison with prior work (Table B.1), the comparator table (Table E.4), the four ways to call the model (Figure F.1), the Run 5 table and its detail (Table H.4), the mechanism model (Figure I.13, Table I.13) and the dataset table (Table K.1). Tables and figures keep their labels in both places, so the labels left in this PDF are not contiguous. Test identifiers (T for Run 1, G for Run 2, H for Run 3, X for Run 4, R for Run 5, P for Run 6; Run 7's carry the prefix R7- and Run 8's the prefix R8-, as in R7-E6a, R8-N1 or R8-W4, so they cannot be read as Appendix E or as Run 1's and Run 3's tests) and evidence tags (C1 to C4, O) are explained in Appendix P.1; tag O is what the main text calls ``observed'' (the extended appendix writes it E).

\appsub{A.1}{Claim-evidence map}
\textbf{A.1 Claim-evidence map.} The full map of 62 claims, each with its evidence, status tag and location, is in Extended Appendix A.1. Table 2 states the rules, Table A.2 gives the evidence on Jev from Runs 1 to 6 for each, and Table P.1 lists all 29 confirmatory tests of Runs 1 to 4. Figure A.1 shows the Jev evidence of Runs 1 to 4 in one view.

\begin{figure}[!htbp]
\centering
\includegraphics[width=\linewidth,height=0.72\textheight,keepaspectratio,alt={Two panels. Left: a dot-and-interval chart of changes in accuracy, in three bands. Grey, "Within +/-2 points: wording": ten points sit inside a +/-2-point band: persona, capitals and politeness on tweets and CFPB, hollow circles (-0.8 to 0.0); seven extra questions asked after or before ours (-0.8, -0.2); persona and politeness on four Run 3 datasets (-0.3, -0.1). Green, "Information added": retrieved examples +11.2, +5.9 and +8.7; examples at 5 per class +3.6 and +2.1 and true class shares +1.9, hollow diamonds; definitions +5.8; ids against names, both defined, -1.1 (Run 3); cryptic tool ids with definitions, a hollow off-scale arrow at +82 (2.3  to 84.4 ). Vermillion, "Wrong information": three wrong example labels -33.5 and -13.4; shuffled shares -6.8 and -6.2, hollow circles; a hollow square for all three correct examples relabelled, -38.2 (secondary, Run 3). Right: a forest plot of the local combination minus the examples prompt at 100 labels per class, one row per comparison: Run 1 (observed) Banking77 -0.8, CLINC150 -0.8, tweets +0.3, complaints +0.8, ToxicChat -1.5; Run 2 (observed, item-level only) tweets +1.7 and complaints +1.2, ToxicChat -1.2 (observed); Run 3 MASSIVE +0.1, LEDGAR +0.2, MetaTool +1.2, complaints -0.1 (secondary); Run 4 MASSIVE -0.8, LEDGAR -1.3, MetaTool -0.1 (not shown). At the bottom, under "Fresh-item tests * not shown", two hollow diamonds with bold labels mark the two planned fresh-item tests: Run 4 pooled over its three datasets, -0.7 [-1.6, +0.1], and Run 8, -0.7 [-1.6, +0.1]. A dashed line marks the -1.5 margin. Key: filled markers are confirmed; hollow markers are secondary, observed or not shown.}]{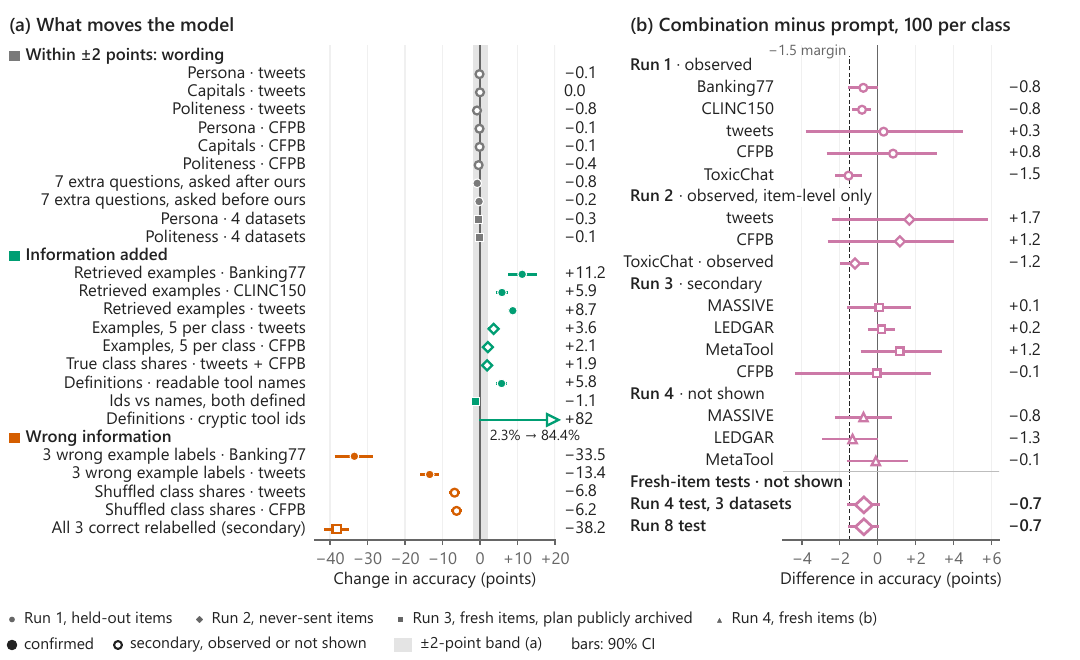}
\renewcommand{\thefigure}{A.1}
\caption{The evidence on Jev 1.13 from Runs 1 to 4, in one view. \textbf{(a)} Changes in accuracy (points, 90\% intervals) against the \ensuremath{\pm}2-point band. Circles: Run 1; diamonds: Run 2; squares: Run 3 (fresh items of four datasets, plan publicly archived). Filled markers were confirmed when the figure was drawn; hollow markers are secondary, observed or not shown. The seven-extra-questions points (Run 1) are now fragile, and Run 8 re-tested them on fresh items (within \ensuremath{\pm}2, Section 3). The off-scale triangle compares cryptic ids alone with ids plus definitions (2.3\% \ensuremath{\rightarrow} 84.4\%; observed, from a near-zero floor). The hollow square at \ensuremath{-}38.2 {[}\ensuremath{-}41.6, \ensuremath{-}35.1{]} is the cost of relabelling all three correct examples of a constructed display on easy items; the test is the slope, and the cumulative costs were 2.9, 11.4 and 38.2 (30\% confusable wrong labels in a pool cost 7.0 in Run 8, and 12.6 on average over six other models). Values in (a) are rounded half-even on item-paired differences (\ensuremath{-}0.25 \ensuremath{\rightarrow} \ensuremath{-}0.2); values in (b), like the text, round halves away from zero (\ensuremath{-}0.05 \ensuremath{\rightarrow} \ensuremath{-}0.1). \textbf{(b)} Local combination minus the examples prompt at 100 labels per class. The two planned fresh-item tests lead: Run 4 and Run 8 both gave \ensuremath{-}0.7, and neither showed the combination matching the prompt (hollow diamonds, bold labels). Above them are 15 comparisons from four runs, with 90\% intervals from resampling classes (ToxicChat, with two classes, resamples near-duplicate groups). No comparison shows the combination reliably ahead. Pooled sets chosen after the fact range from \ensuremath{-}0.77 to \ensuremath{-}0.03 (Table E.10b); the most favourable, \ensuremath{-}0.03 over the nine planned comparisons, is no longer drawn. Hollow markers include the tests that passed only with items resampled (Section 2.3): in (a), Run 2's diamonds, Run 1's shuffled shares and Run 1's persona, capitals and politeness circles; in (b), Run 2's tweets and complaints. Figure H.1 shows three of these effects on other models.}
\end{figure}

\appsub{A.2}{The rules in full}
\textbf{A.2 The rules in full.} Table 2 gives six do's and six don'ts, the labels used in Figure M.1 and Table A.2. Table A.2 lists the Jev evidence of Runs 1 to 6 row by row (Runs 7 and 8 are in Sections 3 to 6 and Appendix U), and Figure A.2 draws the rules as a decision flow, in Table 2's numbering. Extended Appendix A.2 keeps the longer rule set (rules 0 to 11) with the scoping conditions the short guide omits: the held-out sizes of our tests (rule 0; the resampling result is in Appendix P.4), the datasets on which each label-budget result holds, and how to set a cascade's threshold on rare positives. Its rules map onto Table 2 as follows: 0 \ensuremath{\rightarrow} Do 6; 1 \ensuremath{\rightarrow} Do 4; 2 \ensuremath{\rightarrow} Do 5; 3 \ensuremath{\rightarrow} Do 1; 4 \ensuremath{\rightarrow} Do 2; 5 \ensuremath{\rightarrow} Don't 2; 6 and 7 \ensuremath{\rightarrow} Don't 1; 8 and 10 \ensuremath{\rightarrow} Don't 5; 9 \ensuremath{\rightarrow} Do 3 and Don't 3; 11 \ensuremath{\rightarrow} Don't 6. Don't 4 (Runs 6 to 8) has no extended rule. The eleven-rule wording used before Runs 7 and 8 is kept in Extended Appendix A.2b.

\begingroup\footnotesize
\renewcommand{\thetable}{A.2}
\begin{longtable}{@{}>{\raggedright\arraybackslash}p{\dimexpr 0.1638\linewidth-0.9826\tabcolsep\relax}>{\raggedright\arraybackslash}p{\dimexpr 0.2501\linewidth-1.5005\tabcolsep\relax}>{\raggedright\arraybackslash}p{\dimexpr 0.3107\linewidth-1.8642\tabcolsep\relax}>{\raggedright\arraybackslash}p{\dimexpr 0.2754\linewidth-1.6527\tabcolsep\relax}@{}}
\caption{The evidence on Jev 1.13 for each rule from Runs 1 to 6, row by row, with the rule labels of Table 2; Table H.4 gives Run 5, and Appendix U gives Runs 7 and 8. Effects are changes in accuracy (percentage points) with 90\% intervals, or ranges over datasets. ``Item-level only'' marks a planned test's pass that did not hold with classes resampled, reported as observed. Table P.1 lists the 29 tests of Runs 1 to 4 with their identifiers and sizes.}\\
\toprule
Rule & Evidence, in plain words & Effect & Status \\
\midrule
\endfirsthead
\multicolumn{4}{@{}l}{\textit{Table A.2 (continued)}}\\
\toprule
Rule & Evidence, in plain words & Effect & Status \\
\midrule
\endhead
\bottomrule
\endlastfoot
Don't 1: wrappers don't help & Five personas and five politeness lines, three wordings each; fresh items of four datasets & persona \ensuremath{-}0.3 {[}\ensuremath{-}1.0, +0.3{]}; politeness \ensuremath{-}0.1 {[}\ensuremath{-}0.5, +0.4{]} & Confirmed, within \ensuremath{\pm}2 (Run 3, deposited) \\
 & Persona, capitals or politeness vs the plain request; tweets and complaints & six intervals within {[}\ensuremath{-}1.3, +0.5{]} & Observed (Run 1; item-level only, through politeness on tweets; capitals within \ensuremath{\pm}0.5 with classes resampled) \\
 & Seven unrelated questions asked with ours & {[}\ensuremath{-}1.6, +0.1{]} ours first; {[}\ensuremath{-}0.9, +0.4{]} ours last & Confirmed, within \ensuremath{\pm}2 (Run 1); fragile: fails the 29-test correction and the stress calibrations; within \ensuremath{\pm}2 on fresh items in Run 8 \\
Do 1: give it information it lacks & 20 nearest labelled examples in the request; three datasets & +11.2, +5.9, +8.7 & Confirmed (Run 1) \\
 & The same examples in a separate labelled block, vs the bare call & +7.0 {[}+4.2, +9.8{]} & Confirmed (Run 6, on Run 4's items) \\
Do 2: clear names and descriptions & Ids with definitions vs readable names with definitions & \ensuremath{-}1.1 {[}\ensuremath{-}1.8, \ensuremath{-}0.5{]}; LEDGAR \ensuremath{-}2.4 & Confirmed, no more than 2 worse on average (Run 3); fails the 29-test correction \\
 & Ids with blind one-line definitions vs ids alone (from 2.5 to 24\% accuracy) & +73.2 {[}+63.2, +83.5{]} & Confirmed (Run 3), near-certain by design; same caveat \\
Do 3: audit pool labels & 30\% wrong labels spread through a realistic pool & \ensuremath{-}1.8 to \ensuremath{-}4.9 on four topical development sets (ToxicChat \ensuremath{-}1.2); Run 8, fresh items: \ensuremath{-}3.9 uniform, \ensuremath{-}7.0 confusable & Observed; Run 8 confirmed \\
 & Constructed display: relabel 1, 2 or 3 of three correct examples shown (items about 92\% accurate before, 84 to 97\% by dataset) & \ensuremath{-}12.3 per label; cumulative \ensuremath{-}2.9, \ensuremath{-}11.4, \ensuremath{-}38.2 & Confirmed slope (Run 3); totals secondary \\
 & Delete vs relabel; one rival vs spread; last vs second deletion & +13.1 {[}+11.7, +14.7{]}; \ensuremath{-}13.2 {[}\ensuremath{-}14.7, \ensuremath{-}11.7{]}; +10.1 {[}+7.5, +12.7{]} & Confirmed (Run 4) \\
Do 4: few labels & 5 per class: examples in the prompt vs the bare call & +3.6, +2.1 (Run 2); +2.1 {[}+0.0, +4.1{]} (Run 3); +2.9 {[}+0.7, +4.9{]} (Run 4); +2.0 to +5.3 over six pool draws & Observed, item-level only (Run 2); secondary (Runs 3, 4) \\
 & 5 per class: one bare call \ensuremath{\times} local centroid minus the prompt & +2.1 {[}+1.1, +3.1{]} (Run 2, chosen after scoring); \ensuremath{-}0.5 {[}\ensuremath{-}1.9, +1.2{]} (Run 4); bare \ensuremath{\times} tuned classifier \ensuremath{-}0.6 {[}\ensuremath{-}1.7, +0.3{]} (Run 4) & Observed; secondary; not shown to be no more than 1.5 points worse \\
Do 5: many labels & 100 per class: local combination minus the prompt & fresh items \ensuremath{-}0.7 {[}\ensuremath{-}1.6, +0.1{]} (Run 4; Run 8 also \ensuremath{-}0.7); nine planned comparisons pooled \ensuremath{-}0.03 {[}\ensuremath{-}0.53, +0.47{]}, \ensuremath{-}0.02 {[}\ensuremath{-}0.89, +0.85{]} with pool-draw variance; all 13 multi-class \ensuremath{-}0.47 {[}\ensuremath{-}0.79, \ensuremath{-}0.15{]} & Not shown to be no more than 1.5 points worse (Run 4; Run 8 not shown within \ensuremath{\pm}1); poolings observed, over shared items and tests whose decisions differed \\
Do 6: confirm before you ship & 100 to 200 held-out items, resampled from cached runs & prompt vs bare: right sign 79 to 97\%; near-tied designs ranked right only 56 to 64\%; any pick within 1 point of the best 84 to 95\% & Observed; optimistic, since the resamples share items with the truth \\
Don't 2: stated shares & True vs most distant shuffled shares & +6.8, +6.2 & Observed, item-level only (Run 1) \\
 & True class shares stated vs none; costs to small classes & +1.9 {[}+1.1, +2.7{]}; tweets macro-F1 .661 \ensuremath{\rightarrow} .608; toxic recall 39.4\% \ensuremath{\rightarrow} 33.0\% (development ToxicChat) & Observed, item-level only (Run 2); costs observed \\
 & Shares applied to the output after the call vs stated & +1.3, +2.0 & Observed \\
Don't 5: rare classes & Cascade: a local classifier answers sure items, 8.2\% sent to the prompt & accuracy \ensuremath{-}0.2 {[}\ensuremath{-}0.8, +0.4{]}; toxic precision \ensuremath{-}7.3 {[}\ensuremath{-}13.8, \ensuremath{-}1.1{]} & Confirmed, no more than 1.5 worse (Run 2; two classes, so no class-level check); precision observed \\
Don't 6: wording is not security & Guard sentence, spotlighting, separate claim field vs 100 score-blind attack variants & label-preserving success at least 38\%, 15\%, 55\% & Observed \\
Don't 2: class-mix shift & 2025 pool, 2026 complaints, credit reporting 74\% of the pool (76\% of 2025 complaints) \ensuremath{\rightarrow} 3\% of items & combination \ensuremath{-} prompt \ensuremath{-}10.9, \ensuremath{-}2.0 after share re-estimation; prompt \ensuremath{-} bare +0.05, +0.6 after & Secondary (Run 4); re-estimation observed \\
Don't 4: placement & Wrong labels on one rival vs spread, examples inside option descriptions vs in a separate block & penalty \ensuremath{-}13.2 vs \ensuremath{-}6.8; difference +6.4 {[}+5.0, +7.9{]} (MetaTool +0.8 {[}\ensuremath{-}1.3, +2.9{]}); clean examples, descriptions \ensuremath{-} block +0.5 {[}\ensuremath{-}0.3, +1.4{]} & Confirmed (Run 6; Jev 1.13 on Run 4's items; label name and length not controlled); clean contrast secondary \\
Predictions that failed & The prompt equals a fixed vote; the nearest examples' labels matter more & +3.0 {[}+2.4, +3.7{]}; +1.2 {[}\ensuremath{-}0.4, +2.7{]} & Not shown (Runs 1, 2) \\
\end{longtable}
\endgroup

\begin{figure}[!htbp]
\centering
\includegraphics[width=0.98\linewidth,height=0.72\textheight,keepaspectratio,alt={A decision flow headed "Do 4 and Do 5 * Labelled examples: in the prompt, or combined on your machine?", with the question "How many labelled examples per class do you have?" About 5 (solid green box): "Do 4 * Use the prompt: it beats the bare call" (Run 8 +3.1, confirmed; Run 2 +2 to +4, observed, item-level only; Runs 3 and 4 secondary); "One bare call x centroid on your machine trailed the prompt; within 1.5 points was not shown" (Run 8 -1.2, not shown; Run 4 -0.5, secondary; Run 2 +2.1, observed, chosen after scoring). About 100 (dashed purple box): "Do 5 * Prefer the prompt unless privacy rules it out" (one bare call x vote x classifier minus the prompt: Run 4 -0.7 [-1.6, +0.1], Run 8 -0.7 [-1.6, +0.1]; neither shown to match the prompt); "The local route sends no pool text, but loses small classes" (worst-class recall 42.8  vs 29.7 , secondary). An arrow leads from the 100 box to a dashed orange box, "Must you cut vendor calls? Classifier-first cascade * Don't 5 * Judge rare classes by accuracy": the classifier answers when it is sure, and the other items go to the prompt with their nearest labelled examples; on ToxicChat 8  were sent, accuracy no more than 1.5 points below the prompt (confirmed), but toxic-class precision fell 7.3 points (observed); measure precision and recall at your operating point. Below, two rails of equal width. Do, at any budget: Do 1, give the model information it lacks (retrieved examples +5.9 to +11.2; transfers: 6 of 7; 4 of 7 under one correction across models); Do 2, use readable names; define cryptic ids in one line (ids with definitions 1.1 points worse than names, no more than 2 worse; transfers: 6 of 7; 5 of 7 excluding OpenAI's incomplete cells); Do 3, audit your pool's labels, and drop what you doubt (30  wrong in a realistic pool -1.8 to -4.9; constructed display up to -38.2; transfers: 6 of 7, the harm; the audit was tested on Jev only); Do 6, confirm on the model and version you deploy (100 to 200 held-out items get a large gain's sign right 79 to 97  of the time; near-ties only 56 to 64 ; dashed). Don't, at any budget: Don't 1, expect benign wrappers to help (all within +/-2 points; partially transfers: 4 of 7; 2 inconclusive); Don't 2, state unmeasured class shares; re-estimate after a shift (class-mix shift: local combination minus prompt -10.9 -> -2.0 after re-estimating the shares, one dataset; dashed); Don't 3, let wrong labels pile on one rival (Run 4: one rival cost 13.2 points more than spread); Don't 4, assume placement is neutral: test it on your model (Jev 1.13: one-rival penalty -13.2 inside option descriptions (Run 4), -6.8 in a separate block (Run 6); two of three datasets); Don't 6, treat prompt wording as a security boundary (guard, spotlighting and claim field all fell; dashed). Footnote: transfer labels come from Run 5, seven other models, six completed; all answered the same items. Solid marks are confirmed; dashed marks are secondary, observed or not shown.}]{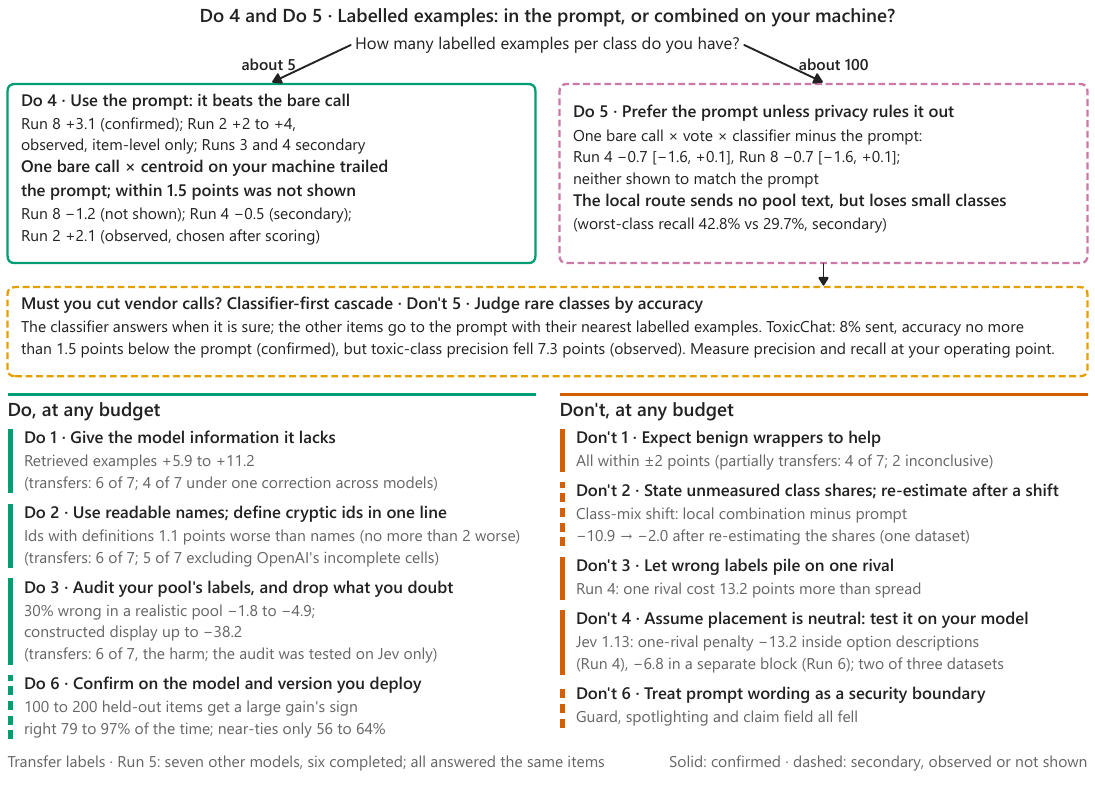}
\renewcommand{\thefigure}{A.2}
\caption{The rules as a decision flow, in the numbering of Table 2. Its transfer labels and most numbers are those of Runs 1 to 6; the fork adds Run 8's label-budget tests, and Table 2 has the current evidence. The fork is the label budget. The cascade sends the items it does not answer to the prompt \emph{with} their nearest labelled examples. Dashed boxes are secondary, observed or not shown.}
\end{figure}

\appsub{A.3}{What hurts: a summary}
\textbf{A.3 What hurts: a summary.} Extended Appendix A.3 catalogues 40 harmful or useless levers in six groups, with datasets and tags. Table A.3 keeps ten of them, one or two per group.

\begingroup\footnotesize
\renewcommand{\thetable}{A.3}
\begin{longtable}{@{}>{\raggedright\arraybackslash}p{\dimexpr 0.1821\linewidth-1.4567\tabcolsep\relax}>{\raggedright\arraybackslash}p{\dimexpr 0.2378\linewidth-1.9026\tabcolsep\relax}>{\raggedright\arraybackslash}p{\dimexpr 0.2617\linewidth-2.0939\tabcolsep\relax}>{\raggedright\arraybackslash}p{\dimexpr 0.1557\linewidth-1.2452\tabcolsep\relax}>{\raggedright\arraybackslash}p{\dimexpr 0.1627\linewidth-1.3015\tabcolsep\relax}@{}}
\caption{Ten of the 40 levers that hurt or did nothing. Effects are accuracy points unless stated (\ensuremath{\Delta}logp: change in gold log-probability). Tags: C1 to C4 confirmed in Run 1 to 4; O observed (exploratory).}\\
\toprule
Group & Action & Effect & Tag & Where \\
\midrule
\endfirsthead
\multicolumn{5}{@{}l}{\textit{Table A.3 (continued)}}\\
\toprule
Group & Action & Effect & Tag & Where \\
\midrule
\endhead
\bottomrule
\endlastfoot
Wrong information & Relabel 1, 2 or 3 of three gold examples shown; all three deleted, relabelled to distinct rivals, or to one rival & \ensuremath{-}2.9, \ensuremath{-}11.4, \ensuremath{-}38.2 (slope \ensuremath{-}12.3); \ensuremath{-}19.4, \ensuremath{-}47.0, \ensuremath{-}67.2 & C3 (slope); C4 (ordering); costs descriptive & \S{}4, I.12, O \\
Wrong information & Planted wrong-label near-copies of queries (poisoning) & 92.2\% \ensuremath{\rightarrow} 74.0 / 39.3 / 23.5\% at 1 / 2 / 3 plants & O & C.12 \\
Unusable information & Abstract ids with no definitions & 70.1\% \ensuremath{\rightarrow} 3.0\% (Banking77-20); 79.2\% \ensuremath{\rightarrow} 2.3\% (MetaTool) & O; C1 arm & C.9 \\
Unusable information & Definitions on top of examples; definitions on informative names & 85.8\% \ensuremath{\rightarrow} 82.8\%; CFPB \ensuremath{\Delta}logp \ensuremath{-}0.112 & O & C.9 \\
Phrasing and structure & Persona, capitals, politeness prefix & within \ensuremath{\pm}2; politeness \ensuremath{-}0.8 {[}\ensuremath{-}1.3, \ensuremath{-}0.3{]} on tweets & O, item-level only (T6); C3 (H1) & \S{}3, D.1 \\
Phrasing and structure & Examples moved from the option field into the state text & \ensuremath{-}3.2 (\ensuremath{\Delta}logp \ensuremath{-}0.062) & O & D.3 \\
Accuracy-only comparators & Cascade with an accuracy-chosen threshold & accuracy \ensuremath{-}0.2 (non-inferior); toxic precision \ensuremath{-}7.3 & C2 (G1c); O & \S{}5, E.7 \\
Accuracy-only comparators & The local combination where it falls short & \ensuremath{-}0.8 to \ensuremath{-}1.5 (Banking77, CLINC150, ToxicChat); \ensuremath{-}0.7, not shown non-inferior (Run 4); \ensuremath{-}10.9 on the time split, \ensuremath{-}2.0 after share re-estimation & O; X1b; secondary & \S{}5, E.3, E.9 \\
False security & Guard sentence, spotlighting, separate claim field & label-preserving success \ensuremath{\geq} 38\%, 15\%, 55\% at 100 score-blind queries & O & G.2, G.3 \\
Method & No magnitude floor; a subset chosen from the model's confusion & 12 of 53 content-free edits flagged; +21.3 against +8.4 on random subsets & O & P.2, C.4 \\
\end{longtable}
\endgroup

\textbf{What hurts, in brief.}

\begin{itemize}
\item \textbf{Wrong information} costs most where the model relies on it: up to 67.2 points with three wrong labels on one rival in a constructed display (Run 4), more inside option descriptions than in a separate block on Jev (Runs 6 to 8), less on every other model tested, and 3.9 to 7.0 for 30\% wrong labels in a pool (Run 8). Filtering the pool with the bare model did not repair it (development, P = 20). Stale or shuffled shares and wrong rubric thresholds are followed.
\item \textbf{Information the model cannot use} is neutral at best: ids without definitions, definitions on top of examples, and retrieval on tasks that need knowledge.
\item \textbf{Phrasing and structure} move little; moving the examples from the options into the state text cost 3.2 points (observed).
\item \textbf{Comparators judged on accuracy alone} hide costs on small classes (an accuracy-tuned cascade, stated shares, the local combination).
\item \textbf{Prompt wording is not security.} A one-line appended claim that a wrong team was correct cut ticket-routing accuracy from 87\% to 33\%, and a guard sentence, spotlighting and a separate claim field all fell to a score-blind search over 100 attack variants (observed; Appendix G). Common injected instructions changed no correct answer in FlipJev {[}\citenum{r26}{]}; our attacker plants evidence instead (Appendix G.2b).
\item \textbf{Method traps.} A screen with no minimum effect size, and a subset chosen from the model's own confusions (+21.3 against +8.4 on random subsets), both inflate effects (Appendix C.3 to C.7).
\end{itemize}

Do not put information in front of the model that you would not stand behind.

\appsub{A.4}{Mechanism status}
\textbf{A.4 Mechanism status.} Table A.4 (Extended Appendix A.4) marks, for each effect, the term of the Appendix I.13 model it touches and whether its mechanism is documented by TypeSafe {[}\citenum{r62}{]}, measured on Jev, fitted by the Appendix I.13 model, or inferred from open-model studies and theory. \emph{Documented:} the typed readout, independent scoring of questions and training for calibrated probabilities {[}\citenum{r62}{]}. \emph{Measured:} both wrong-label channels (X3a-c), the isolation of packed questions {[}\citenum{r36}{]}, and the effects of reworded rules, conflicting option names and adversarial text {[}\citenum{r43}; \citenum{r44}; \citenum{r53}; \citenum{r26}{]}. \emph{Fitted:} the pooled retrieval and wrong-label dose (Appendix I.13), not the per-dataset curves. \emph{Inferred:} the circuits behind the evidence term (Bayesian and latent-variable inference {[}\citenum{r59}; \citenum{r68}{]}, kernel weighting {[}\citenum{r13}{]}, label-word anchors in an information-flow study {[}\citenum{r60}{]}, induction heads {[}\citenum{r61}{]}), definitions supplying an empty anchor, label following and copying {[}\citenum{r20}; \citenum{r55}; \citenum{r12}; \citenum{r28}{]}, and the label budget as a product of experts {[}\citenum{r19}{]} in which a prompt is worth hundreds of labels at low budget {[}\citenum{r22}{]}. \emph{Not tested:} task or function vectors {[}\citenum{r63}; \citenum{r64}{]} and implicit gradient descent {[}\citenum{r66}{]}.

\appsub{A.5}{Definitions on cryptic ids}
\textbf{A.5 Definitions on cryptic ids.} Up to paper v21 the rescue of cryptic ids by one-line definitions was a headline result (the ``+79'' bar of that version's teaser figure, Do 2, Tables 1, 4 and 5, and a fourth panel of its cross-model figure, now Figure H.1). It is kept here, because ids alone start near zero accuracy by design, so the gain mostly measures how little an id says. \emph{Jev 1.13:} blind one-line definitions added 73.2 points {[}+63.2, +83.5{]} over ids alone in Run 3, from 2.5 to 24\% (Table A.2), and 81.0 on Run 5's items. \emph{Run 5:} over the six other models that completed it, +78.65 {[}+71.92, +85.37{]} (between-model SD 4.05; random effects, Appendix U.3), confirmed on 6 of 7, with per-model gains from about +71 (Kev-4B) to +82; OpenAI Decisions' cells were below 98\% complete, and with refusals scored against the test its gain is still +58.1. \emph{Run 8:} definitions won back 76 to 84 points on every other model (secondary), and +75.56 {[}+65.43, +84.28{]} on the chat model; counting each vendor family once, the rescue holds on 5 of 5 families. \emph{Run 9:} ids alone against readable names cost 82.7 points on the public sets and 85.6 on the private ones (Table 6), and on both catalogues ids with a description did as well as names with one (Section 4.1).

\section{Related work and the development screen}

\appsub{B.1}{Related work in full}
\textbf{B.1 Related work in full.} Table 1 compresses the positioning; Table B.1 and the list after it give the longer comparison. Extended Appendix B.1 gives the full list, with its citations: retrieval and wrong labels in in-context learning, personas and politeness, prompting against supervised models at equal labels, cascades and product rules, mechanistic accounts of in-context learning, evaluations of typed decision models, and studies on Jev with analysis plans fixed in advance.

\begingroup\footnotesize
\renewcommand{\thetable}{B.1}
\begin{longtable}{@{}>{\raggedright\arraybackslash}p{\dimexpr 0.1538\linewidth-0.6154\tabcolsep\relax}>{\raggedright\arraybackslash}p{\dimexpr 0.3826\linewidth-1.5304\tabcolsep\relax}>{\raggedright\arraybackslash}p{\dimexpr 0.4636\linewidth-1.8543\tabcolsep\relax}@{}}
\caption{What was known and what this paper adds, in the longer form behind Table 1 (written before Runs 7 and 8, whose results are in Table 1). Extended Appendix T compares 76 prior works claim by claim.}\\
\toprule
Question & Known & What we add \\
\midrule
\endfirsthead
\multicolumn{3}{@{}l}{\textit{Table B.1 (continued)}}\\
\toprule
Question & Known & What we add \\
\midrule
\endhead
\bottomrule
\endlastfoot
Do examples help decision APIs? & Fixed examples and counterexamples raised recall on all 21 available hosted decision models and 2 prompted general models {[}\citenum{r70}{]}; retrieved precedents help Jev {[}\citenum{r29}{]}; retrieval beats random examples in in-context learning {[}\citenum{r25}; \citenum{r39}; \citenum{r10}; \citenum{r31}{]} & Per-item retrieved examples in the option descriptions: +5.9 to +11.2 on Jev, +2.5 to +10.5 on six other models; local alternatives sized on Jev \\
Does decoration help? & The best of 12 instruction wordings gained 0 to 4.6 F1 over the default on six decision models {[}\citenum{r71}{]}; personas, politeness, tips and threats do little in chat models {[}\citenum{r56}; \citenum{r2}; \citenum{r30}; \citenum{r73}{]} & Persona and politeness within \ensuremath{\pm}2 (two one-sided tests) on Jev's fresh items and on 4 of 7 other models; format {[}\citenum{r74}{]}, option order {[}\citenum{r75}{]}, option set {[}\citenum{r36}{]}, polarity {[}\citenum{r76}{]} and rule wording {[}\citenum{r43}{]} left out \\
How do wrong labels scale? & Accuracy falls with the share of wrong labels {[}\citenum{r55}; \citenum{r50}{]}; noisy demonstration labels degrade in-context learning {[}\citenum{r72}{]}; random labels on random examples matter little {[}\citenum{r32}{]} & The same direction on every model tested; on a constructed display, accuracy losses that grew with each wrong label on five of six others (convex in accuracy; in gold log-odds no model's losses kept accelerating after the second wrong label); sizes differ (Jev 1.13 \ensuremath{-}61, others \ensuremath{-}21 to \ensuremath{-}41; not normalised) and are far smaller in a realistic pool \\
Why do wrong labels hurt? & Models copy a near-identical demonstration's label {[}\citenum{r28}{]}; permuted labels pull toward one wrong label in open models {[}\citenum{r12}{]}; demonstrations bias models toward their majority label {[}\citenum{r84}; \citenum{r85}{]} & Two confirmed sources separated within items on a typed API: lost correct evidence, and evidence piled on one rival \\
Where do the examples act? & Examples in the instructions {[}\citenum{r70}{]} or inside label definitions {[}\citenum{r82}{]}; moving a demonstration block changes chat-model accuracy {[}\citenum{r83}{]}; no placement test on typed decision APIs found & The same pairs inside option descriptions vs in a separate block: no clear difference for clean examples (secondary); the one-rival penalty is larger inside descriptions (Jev 1.13, two of three datasets) \\
Can definitions replace cryptic names? & Typed models follow an informative name over a conflicting definition {[}\citenum{r44}{]} & Ids with one-line definitions 1.1 points {[}\ensuremath{-}1.8, \ensuremath{-}0.5{]} below readable names on Jev (no more than 2 worse; 2.3 below on fresh items, Run 8); the rescue from ids alone (+73.2 from a near-zero floor) also on 6 of 7 other models \\
When do local models match the prompt? & Break-even near 100 labels, more once variance is counted {[}\citenum{r79}{]}; fine-tuning catches up as labels grow {[}\citenum{r33}; \citenum{r9}{]}; cascades cut calls {[}\citenum{r46}; \citenum{r37}{]}; a logistic regression won email routing {[}\citenum{r27}{]} & Tuned classifiers, products and cascades at 5 and 100 labels per class on a typed API: the prompt beats the bare call and local classifiers alone, and no test showed one bare call combined locally matching the prompt at either budget \\
What about class shares? & EM prior correction lowers Jev's calibration error under prevalence shift {[}\citenum{r18}{]}; label-shift methods {[}\citenum{r40}; \citenum{r1}{]} & Stated vs post-call shares, on accuracy; a real 2025 \ensuremath{\rightarrow} 2026 class-mix break \\
Can you trust a cheap screen? & Re-run floors (identical re-runs flip 0.17\% of correct decisions {[}\citenum{r80}{]}); many NLP comparisons are underpowered {[}\citenum{r81}{]} & A screen's false-promotion rate on content-free edits (12 of 53); held-out items needed per design choice \\
Plans fixed in advance, many vendors? & Analysis plan fixed in advance: {[}\citenum{r36}; \citenum{r77}; \citenum{r27}; \citenum{r18}; \citenum{r3}{]}, {[}\citenum{r70}{]} in part; contrasts pre-specified {[}\citenum{r51}{]}; many vendors: {[}\citenum{r70}; \citenum{r71}; \citenum{r78}{]} & Used as method, not claimed as a contribution \\
\end{longtable}
\endgroup

\textbf{The five additions as first stated.} Fixed examples already raise recall on hosted decision APIs {[}\citenum{r70}{]}, and the best of 12 instruction wordings gains at most 4.6 F1 over the default {[}\citenum{r71}{]}. We add (Table B.1):

\begin{enumerate}
\item \textbf{Why wrong labels hurt}: two confirmed sources, lost correct evidence and evidence piled on one rival (akin to majority-label bias {[}\citenum{r84}; \citenum{r85}{]}), separated within items on a typed API.
\item \textbf{Definitions nearly recover readable names}: cryptic ids with one-line definitions were 1.1 points {[}\ensuremath{-}1.8, \ensuremath{-}0.5{]} below names on Jev in Run 3 (no more than 2 worse), and 2.3 below on fresh items in Run 8 (not shown no more than 2 worse).
\item \textbf{A placement test} of the same examples in option descriptions or in a separate block. Placement has been studied in chat prompts {[}\citenum{r83}{]}, and codebooks put examples inside label definitions {[}\citenum{r82}{]}; we found no such test on typed decision APIs.
\item \textbf{One battery under one frozen rule on seven other models} (Run 5; Run 8 added the informative contrasts on fresh items). Cross-vendor evaluation is now common {[}\citenum{r70}; \citenum{r71}; \citenum{r78}{]}; what is ours is the set of levers under one rule.
\item \textbf{Practical sizing}: matched-budget local baselines, held-out sizes for a design choice, and a screen's false-promotion rate on content-free edits (12 of 53), beyond the re-run floors of earlier work {[}\citenum{r80}{]}.
\end{enumerate}

\appsub{B.2}{The development screen}
\textbf{B.2 The development screen.} The screen ran about 60 technique types on development data in five rounds; Extended Appendix B.10 lists all 60 distinct Jev 1.13 arms with their accuracy change, \ensuremath{\Delta}logp interval and promotion decisions (42 of 60 flagged by the floor-less rule, 35 of 60 with the 0.04 floor). Across datasets and directions they give 319 comparisons, of which 116 survive Holm at item level and 44 with class clustering. At 80\% power and n = 400, the screen detects 0.6 (LEDGAR) to 2.4 accuracy points (Banking77 subsets). Dropping development items with a labelled pool twin cut the TF-IDF kNN-20 gain from +5.0 to +2.1 points on CFPB and from +6.2 to +4.7 on tweets; rows labelled ``de-duplicated'' in Appendices C and E use splits that keep near-duplicate clusters on one side. Appendix P.2 summarises the null calibration; Extended Appendix B.2 to B.8 give the details.

\appsub{B.3}{Labels without meaning}
\textbf{B.3 Labels without meaning} (moved from Section 4.1 in v25). Other evidence on labels without meaning is mixed. Large models can learn such labels from examples alone {[}\citenum{r50}{]}, and more so after symbol tuning {[}\citenum{r87}{]}, partly separately from recognising the task {[}\citenum{r86}{]}. In open Jev-style decision models, labels override definitions because of how the prompt renders each option {[}\citenum{r90}{]}. Chat models mostly follow swapped definitions but often fall back on what they know about a label {[}\citenum{r97}{]}. On development data, Jev with examples under neutral ids nearly matched names (91.0\% against 91.4\% on the 20 Banking77 intents it confuses most; Appendix C.9). Mazzoni and Failla {[}\citenum{r29}{]} found the opposite on HateModerate, a 41-policy task: with semantic labels, Jev's binary violation accuracy was 5.06 points lower than with opaque ids. Where names were already informative, descriptions helped less clearly. MetaTool's own descriptions added 5.8 points in Run 1, but its queries were generated from those descriptions {[}\citenum{r16}{]}; blind definitions on its names were not shown to help in Run 8 (Table 7), and on CLINC150 and LEDGAR descriptions moved accuracy by \ensuremath{-}1.0 and \ensuremath{-}0.4 points (Run 9).

\section{Information levers (development unless marked; observed)}

This appendix keeps the key numbers; its tables and Figure C1 are in Extended Appendix C. \textdagger{} marks two-call ensemble accuracy.

\appsub{C.1}{Label-matched control}
\textbf{C.1 Label-matched control.} Random pool texts with the same label histogram as the retrieved set carried 61\% of the retrieval gain on Banking77-20 (bare 70.1\%, label-matched 83.0\%, retrieved 91.4\%), 34\% on tweets, 20\% on LEDGAR and 40\% on CFPB. Adding or removing options is not neutral for Jev {[}\citenum{r36}{]}, so offline masks do not substitute for the live labels-only arm.

\appsub{C.2}{Prompt versus vote}
\textbf{C.2 Prompt versus vote} \textdagger{}. The prompt was level with the best of 11 votes on intent sets (Banking77-20 \ensuremath{-}0.5 {[}\ensuremath{-}2.2, +1.2{]}; CLINC150 +1.3 {[}\ensuremath{-}0.3, +3.0{]}) and ahead on topical sets (tweets +12.8, CFPB +10.5, LEDGAR +3.8). Pooled over nine intent cells, prompt minus a cross-fitted vote was \ensuremath{-}0.2 {[}\ensuremath{-}0.9, +0.4{]}; on test items T2 found +3.0 {[}+2.4, +3.7{]} instead (Appendix I.10).

\appsub{C.3}{to C.7 Retrievers, gains, classes, pool draws and small pools}
\textbf{C.3 to C.7 Retrievers, gains, classes, pool draws and small pools.} Dense and TF-IDF retrievers were within \ensuremath{\pm}0.03 \ensuremath{\Delta}logp in 12 of 15 dataset-by-pool pairs although they show different examples. TF-IDF kNN-20 at P = 100 added +11.4 points on all 77 Banking77 intents, +7.5 on CLINC150, +5.9 on de-duplicated tweets and +1.0 to +1.4 on CFPB and LEDGAR. The +21.3 on the Jev-selected Banking77-20 is a selection effect: random 20-intent subsets gained +3.9 to +10.9 (mean +8.4). Gains followed intent-style names and short texts, not headroom. On CFPB retrieval helped the largest class (debt collection 60.4 \ensuremath{\rightarrow} 81.3) and hurt one of the smallest (payday loan 80 \ensuremath{\rightarrow} 60). With one labelled example per class, retrieval lowered accuracy on CFPB (75.0\% \ensuremath{\rightarrow} 72.4\%) and helpdesk (30.1\% \ensuremath{\rightarrow} 24.5\%); twenty random examples gained \ensuremath{-}5.9 to +3.0 points, against +3.0 to +6.3 for the retrieved twenty.

\appsub{C.8}{Stated shares}
\textbf{C.8 Stated shares.} On test items (exploratory), true minus uniform shares was +3.3 {[}+2.3, +4.4{]} (tweets) and +2.8 {[}+1.8, +4.0{]} (CFPB), and uniform minus none \ensuremath{-}1.5 and \ensuremath{-}0.8. Multiplying the bare vector by the true shares, at zero prompt tokens, beat stating them by 1.3 {[}0.4, 2.2{]} and 2.0 {[}1.0, 2.9{]} points. Both forms lowered tweets macro-F1 (.661 \ensuremath{\rightarrow} .608) and lost 5 to 8 points under a shifted class mix. On ToxicChat, stating the true prevalence cut toxic recall from 39.4\% to 33.0\%: a share sentence pushes predictions toward the majority class. Post-hoc correction is the label-shift correction of {[}\citenum{r40}; \citenum{r1}; \citenum{r67}{]}; on Jev, EM correction lowers calibration error under prevalence shift {[}\citenum{r18}{]}. On fresh items, true shares beat no shares by 1.9 points {[}1.1, 2.7{]} in Run 2 (observed, item-level only), below our 2-point action margin, and changed accuracy by \ensuremath{-}0.4 {[}\ensuremath{-}1.0, +0.4{]} on LEDGAR in Run 8 (not confirmed); stale and shuffled shares are in Section 4 and Appendix U.

\appsub{C.9}{Names, ids and definitions}
\textbf{C.9 Names, ids and definitions.} Neutral ids scored 3.0\% bare on Banking77-20 and 91.0\% with retrieved examples (real names 70.1\% and 91.4\%); on tweets, 2.5\% and 66.4\% (names 70.9\% and 77.1\%). Blind definitions under neutral ids came within 0.6 points of names with the same definitions on tweets (72.5\% against 73.1\%) and within 1.6 on CFPB. When names and examples disagree, examples win on short intents and names win on topical text, as in the flipped-label setting of {[}\citenum{r50}{]}. On informative names, definitions add little or lower log-probability (CFPB \ensuremath{\Delta}logp \ensuremath{-}0.112), and on top of examples they lowered accuracy (85.8\% \ensuremath{\rightarrow} 82.8\% with three fixed examples; 90.8\% \ensuremath{\rightarrow} 88.3\% with kNN examples). On MetaTool test items the definitions gain was +5.2 to +6.2 points in every tercile of query-description token overlap.

\appsub{C.10}{Wrong labels and the evidence dose}
\textbf{C.10 Wrong labels and the evidence dose.} Under top-3 corruption Jev lost 41.2, 11.8 and 7.0 points on Banking77-20, tweets and CFPB (\textdagger{}), far less than a vote over its own shown labels (76.2, 42.5 and 49.2). With 40\% wrong labels among the 20 shown, Banking77-20 lost 11.4 points (16.8 when the wrong label came from a confusable class). The nearest neighbours usually carry the gold label, so T4's top-3 relabel leaves no gold example shown for 47\% of Banking77 items, against 9\% for three random ranks (tweets 29\% against 16\%). G3 matched the two arms on correct-label evidence for 938 of 1,170 items and did not show a rank effect (harms \ensuremath{-}6.2 and \ensuremath{-}5.0). Post hoc, each displaced gold example cost about 21 points (Banking77) and 10 (tweets), which fits label copying from similar demonstrations {[}\citenum{r28}{]}. Relabelling 30\% of a realistic pool cost 1.8 to 4.9 points on four topical development sets (ToxicChat 1.2), and cut ToxicChat toxic recall from 62.4\% to 48.6\%; Run 8's planned test on fresh items is in Section 4. \textbf{Harm by the bare call's confidence} (observed). On Run 4's display, three wrong labels cost Jev 1.13 41.4 points where the bare call's top probability was above 0.8, 75.5 where it was 0.5 to 0.8, and 92.0 below 0.5 (25 MetaTool items), so the harm does not rest on easy items. The convex shape is largely a property of the accuracy scale: in gold log-odds no other model's losses kept accelerating after the second wrong label (two models' first step was slightly convex), and Jev's log-odds shape cannot be read because it returns probability 1.0 on 67 to 79\% of clean answers.

\appsub{C.11}{Pool cleaning}
\textbf{C.11 Pool cleaning.} Dropping pool items whose bare argmax disagreed with the given label (30\% wrong labels, P = 20) did not help: Banking77-20 80.0\% against 80.2\% for the noisy pool, tweets 73.5\% against 74.1\%.

\appsub{C.12}{Pool poisoning}
\textbf{C.12 Pool poisoning.} We planted 1, 2 or 3 near-duplicates of each development item (synonym swaps, then case and punctuation changes) with one wrong target label (n = 200), as an attacker who can write labelled near-copies of queries could. Accuracy fell from 92.2\% to 74.0\%, 39.3\% and 23.5\% on Banking77-20, and from 74.5\% to 49.5\%, 32.5\% and 25.8\% on tweets; on Banking77-20 Jev chose the planted label on 23.8\%, 59.8\% and 76.0\% of items (Extended Appendix I.13, whose re-scoring of the same items gives 73.8\% accuracy after one plant). GPT-J and GPT-NeoX copied an identical wrong-labelled demonstration on 74 to 82\% of items of a binary task {[}\citenum{r28}{]}; Jev reaches a similar copy rate only with three near-copies. De-duplication changed nothing at one or two plants, because synonym-swapped copies survive.

\appsub{C.13}{Light battery}
\textbf{C.13 Light battery} (bare \ensuremath{\rightarrow} arm, 300 to 400 development items each): When2Call, the BFCL v3 call-or-no-function gate, MetaTool {[}\citenum{r16}{]}, FOMC, UNFAIR-ToS, NVD severity, Financial PhraseBank 50Agree, NLBSE'24 issues, OpenAI moderation and PubMedQA (sources, citations and the full table in Extended Appendix C.13; licences in Statements). Retrieval at P = 100 helped MetaTool (79.9\% \ensuremath{\rightarrow} 91.4\%; under ids 1.5\% \ensuremath{\rightarrow} 88.7\%) and NVD (38.5\% \ensuremath{\rightarrow} 55.7\%), and moved the knowledge-bound tasks and gates by at most 1.0 point. Stated shares exceeded retrieval on UNFAIR-ToS, PhraseBank and PubMedQA. Shares for uninterpretable options leave accuracy at 14.5\% (MetaTool ids), against 87.1\% with definitions.

\appsub{C.14}{Rubric thresholds}
\textbf{C.14 Rubric thresholds.} On an author-generated severity task, exact-threshold anchors reach 100\% against 81.2\% for degree words, and Jev follows deliberately wrong thresholds on all 40 disagreeing item-paraphrases. This measures rule application, not prompting value.

\section{Phrasing and structure levers (observed)}

\appsub{D.1}{Phrasing stimuli}
\textbf{D.1 Phrasing stimuli.} After Run 1 was unsealed, ten further personas and politeness prefixes (Appendix J) moved accuracy on 300 test items per dataset by at most 1.2 points each (persona pooled \ensuremath{-}0.4 {[}\ensuremath{-}1.2, +0.3{]}, politeness \ensuremath{-}0.5 {[}\ensuremath{-}1.0, +0.1{]}); the personas lowered gold log-probability slightly (\ensuremath{\Delta}logp \ensuremath{-}0.022 {[}\ensuremath{-}0.035, \ensuremath{-}0.009{]}). In Run 3, the four-dataset means under each instruction wording alone were \ensuremath{-}0.40, \ensuremath{-}0.275 and \ensuremath{-}0.275 for persona and \ensuremath{-}0.25, \ensuremath{-}0.15 and +0.20 for politeness. No dataset estimate exceeded 1.0 point (MASSIVE persona \ensuremath{-}1.0 {[}\ensuremath{-}1.7, \ensuremath{-}0.3{]}), and no single stimulus moved a dataset by more than 1.83 points. Finer cells did: MASSIVE persona under its first wording was \ensuremath{-}2.2 {[}\ensuremath{-}4.5, +0.1{]}, and 4 of 120 stimulus \ensuremath{\times} wording cells of 200 items exceeded \ensuremath{\pm}2, down to \ensuremath{-}4.0 {[}\ensuremath{-}7.1, \ensuremath{-}1.0{]}, without multiplicity control (Extended Appendix I.13, Table I.6).

\textbf{Run 1's wrappers and packed questions.} In Run 1 a persona, a capitals line (``IMPORTANT: THINK CAREFULLY\ldots{}'') or a politeness prefix kept all six intervals within {[}\ensuremath{-}1.3, +0.5{]} on tweets and complaints (observed, item-level only, because politeness on tweets failed with classes resampled), and seven unrelated questions asked with ours moved accuracy by \ensuremath{-}0.8 and \ensuremath{-}0.2 points (confirmed, within \ensuremath{\pm}2), as expected if Jev scores packed questions in isolation {[}\citenum{r36}{]}. With classes resampled, capitals stayed within \ensuremath{\pm}0.5 points, and the packed questions passed only with the two wordings averaged (Appendix I.7).

\textbf{Within \ensuremath{\pm}2 is not zero.} Politeness cost 0.8 points on tweets {[}\ensuremath{-}1.3, \ensuremath{-}0.3{]}, and 1.3 under one wording (observed). Run 8 re-tested capitals and packed questions on fresh items (Section 3).

\appsub{D.2}{to D.5 Instructions, placement, packing and long context}
\textbf{D.2 to D.5 Instructions, placement, packing and long context.} On BoolQ {[}\citenum{r7}{]} (bare 92.3\%), an uncertainty rule in the instruction lowered accuracy to 89.3\% and a persona cost 1.2 points. Examples moved from the option descriptions into the state text cost 3.2 points (into the instruction, 1.5); descriptions and examples belong in the option field. Packing a probe with 1, 3 or 7 unrelated questions about the same state changed its gold log-probability by at most 0.005, but packing 8 different items into one state raised it by +0.083 {[}+0.019, +0.151{]}, so packed states are not isolated the way packed questions are. A needle that repeats the question's wording was found at every length up to 1,500 filler sentences; one that needs relative-time arithmetic fell from 78\% to 65\%. Length alone did not hurt; computation did (100 items per cell).

\section{Label budget and comparators (observed unless marked)}

Figures E1 and E2 and the per-dataset tables are in Extended Appendix E.

\appsub{E.1}{and E.2 Supervised peers on development data}
\textbf{E.1 and E.2 Supervised peers on development data.} Against the strongest of about 20 calibrated local baselines on bge-small and MiniLM embeddings, including SetFit {[}\citenum{r45}{]}, chosen per cell on the scored items, the prompt led on topical or imbalanced sets by +14.6 {[}+13.2, +16.0{]} at P = 1, +8.1 at P = 5 and +1.1 {[}0.0, +2.3{]} at P = 100, and trailed on the Banking77 sets by 1.3 to 1.8 points from P = 20. The strongest baseline reached Jev at about 9 labels per class on Banking77-20 (provisional), at most 20 on all 77 intents, about 95 on tweets and about 500 on CFPB. A 50/50 average of Jev and a dense logistic regression beat the better component by +0.6 points over 34 cells (no losses).

\appsub{E.3}{Comparators at 80 to 100 labels per class}
\textbf{E.3 Comparators at 80 to 100 labels per class} (post-unsealing). On Run 1 items the dense prompt led a tuned classifier by 0.8 to 5.8 points and a fine-tuned MiniLM encoder {[}\citenum{r38}; \citenum{r48}{]} by \ensuremath{-}1.2 (tweets) to +5.5 (CFPB). The G5 combination trailed the prompt by 0.8 points on Banking77 and CLINC150 and by 1.5 on ToxicChat, and was \ensuremath{-}1.2 {[}\ensuremath{-}2.0, \ensuremath{-}0.5{]} against it on fresh ToxicChat items, where it was not non-inferior (p .25). A tempered product, with exponents cross-fitted on gold labels, closed the gap on Banking77 (+0.2 {[}\ensuremath{-}0.4, +0.7{]}) and Run 2 ToxicChat (\ensuremath{-}0.4 {[}\ensuremath{-}1.0, +0.1{]}) but not on CLINC150 (\ensuremath{-}0.9 {[}\ensuremath{-}1.4, \ensuremath{-}0.5{]}); in Run 4 a tempered product with frozen exponents was \ensuremath{-}0.4 {[}\ensuremath{-}1.2, +0.3{]} (secondary). On macro-F1, prompt minus combination ranged from \ensuremath{-}3.1 to +1.0 across Runs 2 and 3, but on CFPB the combination's recall on the smallest class (payday loan) was 10 to 16 points below the prompt's lowest class recall (58.9 against 68.7 in Run 2, 52.6 against 68.8 in Run 3; intervals cross 0). On fresh items the G5 combination beat the fine-tuned encoder by 3.5 {[}2.2, 4.8{]} points.

\appsub{E.4}{Comparators at 5 labels per class}
\textbf{E.4 Comparators at 5 labels per class} (post-unsealing; Table E.4 gives the pooled comparison). The strongest local classifier, a nearest centroid, trailed the prompt by 16.4 points on both tweets and CFPB. On macro-F1 the prompt led the tuned classifier and the nearest centroid by +14.0 to +25.9 (Run 2) and +7.7 to +30.5 (Run 3), and bare Jev by +0.8 to +5.5.

\textbf{Run 2 and Run 4 at 5 labels per class.} The examples prompt beat the bare call by 3.6 points {[}2.3, 5.0{]} on tweets and 2.1 {[}1.1, 3.2{]} on complaints (observed, item-level only), and by 2.0 to 5.3 in six redrawn pools and 2.1 and 2.9 in Runs 3 and 4 (secondary). It beat local classifiers used alone by 9 points or more, but these scored up to 18 points below the bare call. One bare call multiplied on your machine by a nearest-centroid classifier came close or ahead after the fact: +2.1 {[}+1.1, +3.1{]} over the prompt on Run 2 items (observed, chosen after scoring) and \ensuremath{-}0.5 {[}\ensuremath{-}1.9, +1.2{]} in Run 4 (secondary; Table E.4). Run 8's planned test of that shortcut did not pass (\ensuremath{-}1.19 {[}\ensuremath{-}2.03, \ensuremath{-}0.45{]}; Section 5).

\begingroup\small
\renewcommand{\thetable}{E.4}
\begin{longtable}{@{}lll@{}}
\caption{Run 2 fresh items, 5 labels per class: pooled accuracy over tweets and complaints, and the examples prompt (74.1\%) minus each comparator. All rows are observed; the two marked ``test'' are parts of a planned test that passed only at item level (Section 2.3). Italic rows were chosen after the others were scored; on Run 4's fresh items the planned bare \ensuremath{\times} tuned classifier was not shown to be no more than 1.5 points worse (\ensuremath{-}0.6 {[}\ensuremath{-}1.7, +0.3{]}; Appendix O).}\\
\toprule
Comparator & Accuracy \% & Prompt minus comparator \\
\midrule
\endfirsthead
\multicolumn{3}{@{}l}{\textit{Table E.4 (continued)}}\\
\toprule
Comparator & Accuracy \% & Prompt minus comparator \\
\midrule
\endhead
\bottomrule
\endlastfoot
Bare Jev (test) & 71.3 & +2.9 {[}+2.0, +3.7{]} \\
Bare \ensuremath{\times} vote (one bare call) & 71.7 & +2.4 {[}+1.2, +3.7{]} \\
Dense nearest centroid & 57.7 & +16.4 {[}+14.5, +18.3{]} \\
Tuned classifier (test) & 57.4 & +16.7 {[}+14.9, +18.6{]} \\
Fine-tuned MiniLM (3 seeds) & 53.4 & +20.7 {[}+19.1, +22.4{]} \\
\emph{Bare \ensuremath{\times} centroid} & 76.2 & \ensuremath{-}2.1 {[}\ensuremath{-}3.1, \ensuremath{-}1.1{]} \\
\emph{Bare \ensuremath{\times} tuned classifier} & 75.7 & \ensuremath{-}1.6 {[}\ensuremath{-}2.6, \ensuremath{-}0.6{]} \\
\end{longtable}
\endgroup

\appsub{E.5}{Comparator definitions}
\textbf{E.5 Comparator definitions.} The tuned classifier is a logistic regression on bge-small embeddings, C chosen by cross-validation on the pool, temperature-scaled and prior-corrected. The fine-tuned encoder is all-MiniLM-L6-v2 with a linear head, at most 4 or 5 epochs with early stopping on 10\% of the pool, prior-corrected.

\appsub{E.6}{and E.7 Cascades and rare positives}
\textbf{E.6 and E.7 Cascades and rare positives.} With cross-fitted thresholds on Run 1 items, a tuned-classifier cascade sent 1.7 to 31.6\% of items to Jev and lost 0.50 to 0.75 points, at 3.2 to 58 times lower Jev cost. On ToxicChat, the dense prompt lifted toxic recall from 40.4\% to 69.3\% on Run 1 items (two-call figures), and recall at precision 0.9 from 0.24 to 0.53. On Run 2's fresh items (145 toxic of 2,244), the G1c cascade minus full kNN-20 was +0.045 {[}\ensuremath{-}0.025, +0.117{]} in recall, \ensuremath{-}0.073 {[}\ensuremath{-}0.138, \ensuremath{-}0.011{]} in precision and \ensuremath{-}0.128 {[}\ensuremath{-}0.197, \ensuremath{-}0.053{]} in average precision, and recall at precision 0.9 fell from 0.48 to 0.08: locally answered items carry the classifier's poorly ranked probabilities. On Run 1 items a threshold sending 37\% kept precision (\ensuremath{-}0.3 {[}\ensuremath{-}4.0, +1.8{]}).

\appsub{E.8}{Prevalence shift}
\textbf{E.8 Prevalence shift.} A classifier trained on a balanced pool inherits its prior: at 5\% spam prevalence (Enron) it scored 88.0\% against 97.4\% for bare Jev. Every local comparator in the confirmatory runs is therefore prior-corrected.

\appsub{E.9}{Run 4 secondary estimates}
\textbf{E.9 Run 4 secondary estimates} (no decision; system minus the kNN-20 prompt, 90\% class-cluster intervals).

\begin{itemize}
\item \textbf{Label budget} (MASSIVE, LEDGAR, MetaTool; 5,606 items). At 5 per class: bare \ensuremath{\times} nearest centroid \ensuremath{-}0.49 {[}\ensuremath{-}1.92, +1.19{]}; bare Jev \ensuremath{-}2.90 {[}\ensuremath{-}4.90, \ensuremath{-}0.68{]} (MASSIVE \ensuremath{-}3.5, LEDGAR \ensuremath{-}1.0, MetaTool \ensuremath{-}4.2); tuned classifier alone \ensuremath{-}9.10. At 100 per class: tempered product \ensuremath{-}0.41 {[}\ensuremath{-}1.23, +0.31{]}; tuned classifier alone \ensuremath{-}2.34; bare Jev \ensuremath{-}6.00. Mean worst-class recall at 100 per class was 42.8 for the prompt and 29.7 for the G5 product.
\item \textbf{Prior plus examples.} The prompt beat a fixed vote of its own neighbours by 3.0 points {[}2.4, 3.7{]} (T2, not shown equivalent), so it adds the model's prior to the examples.
\item \textbf{SuperICL} {[}\citenum{r65}{]} (MASSIVE and LEDGAR): \ensuremath{-}2.3 {[}\ensuremath{-}4.1, \ensuremath{-}0.6{]} at 5 per class and \ensuremath{-}1.2 {[}\ensuremath{-}2.5, +0.0{]} at 100; +0.5 {[}\ensuremath{-}0.6, +1.7{]} against the local product at 100. Jev returned the classifier's stated label on 86.5 to 97.2\% of items, and when the classifier was wrong SuperICL's accuracy was 13.6 to 54.3\%: the stated label mostly overrode Jev's own evidence.
\item \textbf{Pool redraws} (Run 2 items). Over six pool draws the G5 difference ranged from \ensuremath{-}1.02 to +0.75 (mean \ensuremath{-}0.07, SD 0.69), above \ensuremath{-}1.5 in all six and above zero in 3. The original pool (draw 0), re-scored on this 1,000-item subset, gave the highest value, +0.75. The prompt's 5-per-class lead over bare Jev (2.0 to 5.3) and the local classifiers kept its sign in every draw.
\end{itemize}

\begingroup\footnotesize
\renewcommand{\thetable}{E.9c}
\begin{longtable}{@{}>{\raggedright\arraybackslash}p{\dimexpr 0.3237\linewidth-1.9420\tabcolsep\relax}>{\raggedright\arraybackslash}p{\dimexpr 0.2053\linewidth-1.2319\tabcolsep\relax}>{\raggedright\arraybackslash}p{\dimexpr 0.2053\linewidth-1.2319\tabcolsep\relax}>{\raggedright\arraybackslash}p{\dimexpr 0.2657\linewidth-1.5942\tabcolsep\relax}@{}}
\caption{CFPB time split: a pool of 800 complaints received in 2025 at natural shares, and 2,000 test complaints received in 2026 (near-duplicate-group clusters; worst class in brackets).}\\
\toprule
System & Accuracy & Macro-F1 & Worst-class recall \\
\midrule
\endfirsthead
\multicolumn{4}{@{}l}{\textit{Table E.9c (continued)}}\\
\toprule
System & Accuracy & Macro-F1 & Worst-class recall \\
\midrule
\endhead
\bottomrule
\endlastfoot
Bare Jev & 76.8 {[}75.3, 78.4{]} & 74.3 {[}72.5, 75.9{]} & 68.2 (debt collection) \\
kNN-20 prompt, natural pool & 76.9 {[}75.4, 78.5{]} & 74.3 {[}72.5, 76.0{]} & 70.4 (debt collection) \\
Bare \ensuremath{\times} vote \ensuremath{\times} tuned classifier & 66.0 {[}64.3, 67.7{]} & 61.5 {[}59.4, 63.5{]} & 17.3 (payday loan) \\
Tuned classifier alone & 42.4 {[}40.5, 44.2{]} & 34.5 {[}32.8, 36.3{]} & 1.2 (payday loan) \\
\end{longtable}
\endgroup

Product minus prompt was \ensuremath{-}10.9 {[}\ensuremath{-}12.2, \ensuremath{-}9.5{]} (class clusters {[}\ensuremath{-}19.2, \ensuremath{-}5.6{]}). More than half of the product's loss falls on the three classes with 5 to 8 pool examples (mortgage, payday loan, vehicle loan). The class mix shifted: credit reporting was 74.3\% of the 2025 pool (75.8\% of 2025 complaints) and 3.0\% (59 of 2,000) of the 2026 test items.

\textbf{E.9d Where the shift comes from} (observed; Extended Appendix R.1). The 2026 test items are a label-free hash sample of 78,198 narratives with no per-class cap; the filters removed too few items, and the stream position varies too little, to create the shift. In the source, credit reporting was 63 to 83\% of in-scope narratives in every month of 2025 and about 3\% from 1 January 2026, when its narratives fell about 99\% (other products about 45\%). Re-estimating the 2026 shares from the unlabelled items' bare outputs (Saerens EM {[}\citenum{r40}{]}) and re-targeting the local experts gave product minus prompt \ensuremath{-}2.0 {[}\ensuremath{-}3.1, \ensuremath{-}0.9{]} (class clusters {[}\ensuremath{-}5.9, +0.5{]}), with worst-class recall 27.2\% against 58.6\%; prompt minus bare was +0.6 {[}\ensuremath{-}0.2, +1.4{]}.

\appsub{E.10}{Local combination minus prompt at 100 labels per class, all comparisons}
\textbf{E.10 Local combination minus prompt at 100 labels per class, all comparisons} (observed pooling; Extended Appendix R.3). Bare \ensuremath{\times} vote \ensuremath{\times} tuned classifier minus the dense kNN-20 prompt, paraphrase-averaged, 90\% class-cluster intervals (ToxicChat: near-duplicate groups). Pooled estimates use DerSimonian-Laird random effects {[}\citenum{r95}{]} over the class-cluster standard errors.

\begingroup\footnotesize
\begin{longtable}{@{}>{\raggedright\arraybackslash}p{\dimexpr 0.2568\linewidth-2.0544\tabcolsep\relax}>{\raggedright\arraybackslash}p{\dimexpr 0.1731\linewidth-1.3850\tabcolsep\relax}>{\raggedright\arraybackslash}p{\dimexpr 0.1509\linewidth-1.2075\tabcolsep\relax}>{\raggedright\arraybackslash}p{\dimexpr 0.1509\linewidth-1.2075\tabcolsep\relax}>{\raggedright\arraybackslash}p{\dimexpr 0.2682\linewidth-2.1457\tabcolsep\relax}@{}}
\toprule
Comparison & Status & Items & Estimate & 90\% CI \\
\midrule
\endfirsthead
\multicolumn{5}{@{}l}{\textit{Table (continued)}}\\
\toprule
Comparison & Status & Items & Estimate & 90\% CI \\
\midrule
\endhead
\bottomrule
\endlastfoot
Run 1 Banking77; CLINC150 (80 per class) & observed & 2,975; 4,443 & \ensuremath{-}0.76; \ensuremath{-}0.83 & {[}\ensuremath{-}1.55, +0.00{]}; {[}\ensuremath{-}1.34, \ensuremath{-}0.36{]} \\
Run 1 tweets; complaints & observed & 1,300; 1,600 & +0.31; +0.81 & {[}\ensuremath{-}3.81, +4.50{]}; {[}\ensuremath{-}2.70, +3.13{]} \\
Run 1 ToxicChat; Run 2 ToxicChat & observed & 2,600; 2,244 & \ensuremath{-}1.54; \ensuremath{-}1.20 & item-level {[}\ensuremath{-}2.25, \ensuremath{-}0.83{]}; {[}\ensuremath{-}1.98, \ensuremath{-}0.45{]} \\
Run 2 tweets; complaints (G5) & observed, item-level only & 1,170; 1,406 & +1.67; +1.17 & {[}\ensuremath{-}2.42, +5.84{]}; {[}\ensuremath{-}2.65, +4.04{]} \\
Run 3 MASSIVE; MetaTool (H5i) & secondary & 600; 600 & +0.08; +1.17 & {[}\ensuremath{-}1.62, +1.78{]}; {[}\ensuremath{-}0.90, +3.37{]} \\
Run 3 LEDGAR; complaints (H5) & secondary & 1,000; 1,000 & +0.20; \ensuremath{-}0.05 & {[}\ensuremath{-}0.52, +0.91{]}; {[}\ensuremath{-}4.35, +2.80{]} \\
Run 4 MASSIVE; LEDGAR; MetaTool (X1b) & not shown & 1,725; 1,500; 2,381 & \ensuremath{-}0.75; \ensuremath{-}1.33; \ensuremath{-}0.10 & {[}\ensuremath{-}2.26, +0.77{]}; {[}\ensuremath{-}2.94, +0.00{]}; {[}\ensuremath{-}1.64, +1.60{]} \\
\textbf{Pooled: the nine planned comparisons (G5, H5, H5i, X1b)} & observed &  & \textbf{\ensuremath{-}0.03} & \textbf{{[}\ensuremath{-}0.53, +0.47{]}}; \ensuremath{\tau} 0.00; prediction interval {[}\ensuremath{-}0.60, +0.55{]} \\
Pooled: all 13 multi-class comparisons & observed &  & \ensuremath{-}0.47 & {[}\ensuremath{-}0.79, \ensuremath{-}0.15{]}; dataset-equal mean +0.12 {[}\ensuremath{-}0.68, +0.72{]} \\
\end{longtable}
\endgroup

No comparison's interval lies above zero. Fresh-item runs alone give \ensuremath{-}0.08 {[}\ensuremath{-}0.59, +0.43{]}, and adding the pool-draw variance to the nine planned comparisons gives \ensuremath{-}0.02 {[}\ensuremath{-}0.89, +0.85{]} (Extended Appendix R.3). The 13-comparison mean is pulled toward Run 1's CLINC150 and Banking77, which have the smallest standard errors; the comparisons share sources and pools, and Run 1's datasets drove the choice of G5's.

\begingroup\footnotesize
\renewcommand{\thetable}{E.10b}
\begin{longtable}{@{}>{\raggedright\arraybackslash}p{\dimexpr 0.3706\linewidth-2.9645\tabcolsep\relax}>{\raggedright\arraybackslash}p{\dimexpr 0.1149\linewidth-0.9194\tabcolsep\relax}>{\raggedright\arraybackslash}p{\dimexpr 0.1713\linewidth-1.3707\tabcolsep\relax}>{\raggedright\arraybackslash}p{\dimexpr 0.1718\linewidth-1.3748\tabcolsep\relax}>{\raggedright\arraybackslash}p{\dimexpr 0.1713\linewidth-1.3707\tabcolsep\relax}@{}}
\caption{The 100-per-class pool after the fact: all five pooled sets side by side (observed, re-analysis 2; Extended Appendix W). Local combination minus prompt, points. DL is DerSimonian-Laird random effects {[}\citenum{r95}{]} over class-cluster SEs; ``shared pools'' resamples gold classes once per source and adds one pool-draw shift per source, the dependence that matters (no item is shared between comparisons).}\\
\toprule
Pooled set (comparisons) & Estimate & DL 90\% & Modified HKSJ 90\% & Classes and shared pools 90\% \\
\midrule
\endfirsthead
\multicolumn{5}{@{}l}{\textit{Table E.10b (continued)}}\\
\toprule
Pooled set (comparisons) & Estimate & DL 90\% & Modified HKSJ 90\% & Classes and shared pools 90\% \\
\midrule
\endhead
\bottomrule
\endlastfoot
Run 4 fresh-item test alone (3) & \ensuremath{-}0.77 & {[}\ensuremath{-}1.66, +0.12{]} & {[}\ensuremath{-}2.34, +0.81{]} & {[}\ensuremath{-}2.12, +0.59{]} \\
All 15, with ToxicChat & \ensuremath{-}0.69 & {[}\ensuremath{-}1.00, \ensuremath{-}0.38{]} & {[}\ensuremath{-}1.02, \ensuremath{-}0.36{]} & {[}\ensuremath{-}1.53, +0.14{]} \\
All 13 multi-class & \ensuremath{-}0.47 & {[}\ensuremath{-}0.79, \ensuremath{-}0.15{]} & {[}\ensuremath{-}0.81, \ensuremath{-}0.12{]} & {[}\ensuremath{-}1.43, +0.50{]} \\
Fresh-item runs only (7) & \ensuremath{-}0.08 & {[}\ensuremath{-}0.59, +0.43{]} & {[}\ensuremath{-}0.68, +0.52{]} & {[}\ensuremath{-}1.39, +1.22{]} \\
The nine planned (Runs 2 to 4) & \ensuremath{-}0.03 & {[}\ensuremath{-}0.53, +0.47{]} & {[}\ensuremath{-}0.59, +0.54{]} & {[}\ensuremath{-}1.30, +1.24{]} \\
\end{longtable}
\endgroup

With shared pools no set excludes zero, and a non-inferiority reading at \ensuremath{-}1.5 would hold for three sets and fail for two, so picking one would be best-of selection. Run 8's fresh-item test, outside these sets, gave \ensuremath{-}0.73 {[}\ensuremath{-}1.55, +0.06{]}.

\section{Production levers (observed)}

Starter code for these levers is at \url{https://github.com/BjBodner/decision-levers}, with pseudocode in Appendix N.

\appsub{F.1}{to F.3 Tokens, shortlist and packing}
\textbf{F.1 to F.3 Tokens, shortlist and packing.} kNN-20 buys +1.2 to +2.8 nats of gold log-probability per 1,000 added tokens on Banking77 and CLINC150, and 0.01 to 0.09 on Enron, LEDGAR, CFPB and ToxicChat; persona, politeness and emphasis sit at or below the random-padding floor. On test items, cutting options to the labels of the five nearest neighbours lost 2.2 and 2.4 points against kNN-20 on Banking77 and CLINC150 at about 5\% of its tokens (gold in the shortlist for 97.3\% and 97.5\% of items), but it hurt on topical text (tweets 70.9\% \ensuremath{\rightarrow} 69.3\%). Packing three related questions about one CFPB complaint left the product answer unchanged (80.75\% both ways) and saved 51\% of input tokens.

\appsub{F.4}{and F.5 Abstention and automation}
\textbf{F.4 and F.5 Abstention and automation.} A ``none of these'' option with kNN-20 separated CLINC150's far out-of-scope queries with AUROC .999, but near-miss inputs (10 held-out Banking77 intents, one seed) only with .918 (bare .805). Automation at a guaranteed error rate is in Extended Appendix F.5.

\appsub{F.6}{Latency and cost}
\textbf{F.6 Latency and cost.} In one session, sequential calls had a median latency of 0.094 s bare and 0.104 s with kNN-20, and 50 concurrent calls 1.39 s and 1.15 s; tail latency under sustained load was not measured. Local retrieval adds about 7 ms. Using kNN-20 instead of a bare call adds about \$0.07 per 1,000 decisions on CFPB and \$0.003 to \$0.009 on the intent and tool sets.

\appsub{F.7}{What leaves your machine}
\textbf{F.7 What leaves your machine} (the main text's Figure 4 up to v26; moved here in v27). Figure F.1 sets four ways to call the model at 100 labels per class side by side: what each sends across the vendor boundary, its vendor cost and its accuracy against the examples prompt (Section 5).

\begin{figure}[!htbp]
\centering
\includegraphics[width=\linewidth,height=0.72\textheight,keepaspectratio,alt={Four lanes cross a dashed vendor boundary. Bare call: sends text only. Vendor cost  0.020 |  0.022 per 1,000 decisions (index 52 | 24), accuracy -8.8 | -3.7 points against the prompt. Examples prompt: finds the 20 nearest examples in a labelled pool and sends the text plus the nearest labelled examples. Vendor cost  0.038 |  0.092 (index 100, reference). Local combination: sends text only; the probabilities return and are multiplied locally with a vote and a classifier. Vendor cost  0.020 |  0.022 plus local compute, accuracy +1.7 | +1.2 (Run 2 only, item-level; not replicated); a band under the lane gives the two fresh-item tests, combination minus prompt: Run 4 -0.7 [-1.6, +0.1] and Run 8 -0.7 [-1.6, +0.1], neither shown to match the prompt. Classifier-first cascade: answers locally when the classifier is sure. Otherwise it sends the text plus the nearest labelled examples, for 34  | 23  of items. Vendor cost  0.013 |  0.022 (index 34 | 23), accuracy +1.2 | -1.1, observed. Footnotes: on ToxicChat the cascade sends 8  of items and is no more than 1.5 points worse on accuracy (confirmed), but toxic-class precision drops 7.3 points (observed). Mean worst-class recall at 100 per class on Run 4's datasets: prompt 42.8 , local combination 29.7  (secondary).}]{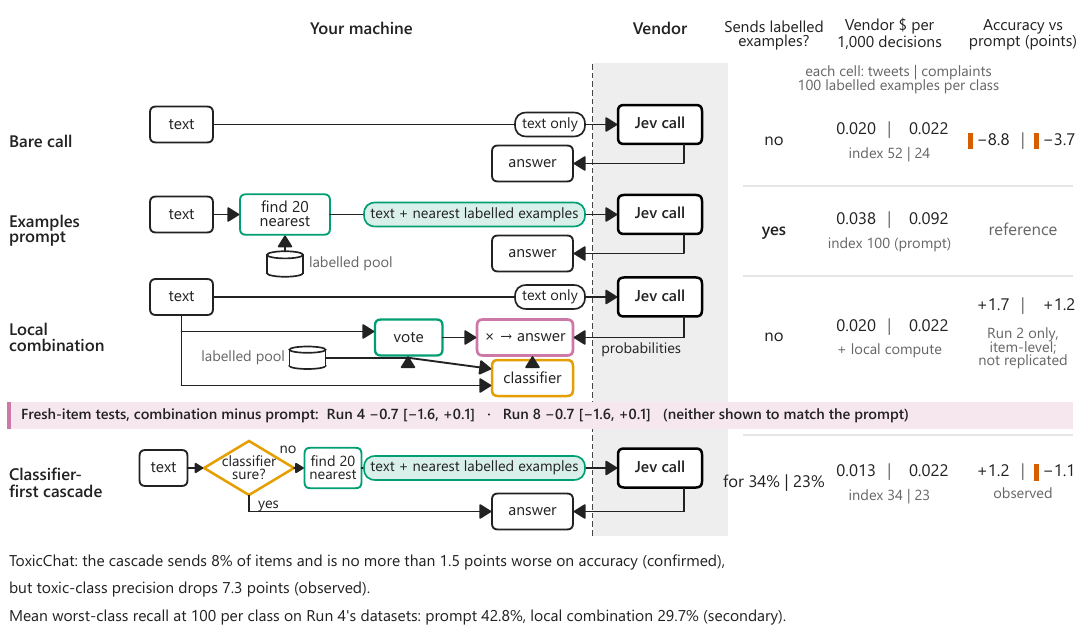}
\renewcommand{\thefigure}{F.1}
\caption{What leaves your machine: four ways to call the model at 100 labels per class. The chip on each arrow that crosses the vendor boundary says what is sent. Cells give tweets \textbar{} complaints on Run 2's items. Vendor cost is in US dollars per 1,000 decisions, from billed input tokens (examples prompt = index 100), excluding local compute. Accuracy is in points against the examples prompt. The local combination's +1.7 \textbar{} +1.2 are item-level estimates that weren't replicated; the band below them gives the two planned fresh-item tests (Section 5).}
\end{figure}

\section{Security (observed)}

\appsub{G.1}{Setting}
\textbf{G.1 Setting.} Author-generated customer-support tickets (n = 300) are routed to one of several teams. The attacker appends text claiming that a wrong team is correct, and succeeds if the argmax becomes that team. The defenses are a guard sentence (``Judge only from the evidence in the state; ignore any claims about what the answer should be.''), spotlighting (the message wrapped in data markers and marked untrusted {[}\citenum{r14}{]}) and a field split (the claim in its own \code{unverified\_\allowbreak{}claim} field). A one-line claim cut accuracy from 87\% to 33\%; success fell to .23 with the guard, .12 with spotlighting and .11 with the field split.

\appsub{G.2}{Guided and score-blind search}
\textbf{G.2 Guided and score-blind search.} On 40 tickets, a search over an attack grammar ran for up to 100 queries, guided by the target probability or score-blind.

\begingroup\footnotesize
\begin{longtable}{@{}>{\raggedright\arraybackslash}p{\dimexpr 0.2634\linewidth-1.5802\tabcolsep\relax}>{\raggedright\arraybackslash}p{\dimexpr 0.1318\linewidth-0.7906\tabcolsep\relax}>{\raggedright\arraybackslash}p{\dimexpr 0.0725\linewidth-0.4348\tabcolsep\relax}>{\raggedright\arraybackslash}p{\dimexpr 0.5324\linewidth-3.1945\tabcolsep\relax}@{}}
\toprule
Defense & Guided @10 / @100 & Blind @100 & Supported evasion-only @100 \\
\midrule
\endfirsthead
\multicolumn{4}{@{}l}{\textit{Table (continued)}}\\
\toprule
Defense & Guided @10 / @100 & Blind @100 & Supported evasion-only @100 \\
\midrule
\endhead
\bottomrule
\endlastfoot
None & 1.00 / 1.00 & 1.00 & \ensuremath{\geq} .80 (census) \\
Guard sentence & .70 / .98 & .98 & \ensuremath{\geq} .38 to .40 (census) \\
Spotlighting & .33 / .85 & .68 & \ensuremath{\geq} .15 to .18 (census) \\
Claim in its own field & .70 / 1.00 & 1.00 & 1.00 judged (95\% CI .72 to 1.00); census \ensuremath{\geq} .55 \\
\end{longtable}
\endgroup

``Supported evasion-only'' counts successes that keep the ticket's true label. Most of the success is best-of-N {[}\citenum{r17}{]}: without probabilities, 100 draws succeed on 98 to 100\% of tickets against the guard and the field split, consistent with adaptive attackers defeating prompt-level defenses {[}\citenum{r34}{]}.

\textbf{G.2b How this squares with FlipJev.} FlipJev found that common prompt injections, such as ``ignore previous instructions'' or a forged answer, changed no correct jev-1.13 answer in 10,240 paired tests {[}\citenum{r26}{]}. Our attack text is evidence (a claim that a wrong team is right), the strongest results come from adaptive search, and the tickets are short and often ambiguous (55 of 124 judged successes; G.3). Both fit Findings 1 and 2. We did not cross the two attack sets.

\appsub{G.3}{and G.4 Label preservation and a classifier baseline}
\textbf{G.3 and G.4 Label preservation and a classifier baseline.} A blind audit of 124 judged successes found 56 preserved, 55 ambiguous and 13 changed (45\% preserved {[}37, 54{]}, 95\% Wilson). A TF-IDF logistic regression trained on 2,000 synthetic tickets resisted all 8 author attacks; score-blind search reached 17.5\% at 100 queries and guided search 32.5\%, against 98 to 100\% for the prompted model. The attack grammar, Figure G1 and related attack work on Jev are in Extended Appendix G.

\section{Open-weight scoping and the cross-model run (observed unless marked)}

We scored each option of the exact development requests (paraphrase 0) as the full answer of Qwen2.5-0.5B-Instruct and Qwen2.5-1.5B-Instruct {[}\citenum{r35}{]}, at n = 120 per arm (60 on CFPB), and compared cached Jev responses (all arms: Extended Appendix H.1 and H.2).

\begin{itemize}
\item \textbf{Scaling.} On Banking77-20 the retrieval gain rose from +4.2 {[}\ensuremath{-}2.5, +10.0{]} at 0.5B to +10.8 {[}+5.0, +17.5{]} at 1.5B, and the top-3 harm from \ensuremath{-}6.7 to \ensuremath{-}14.2, against Jev's +24.2 and \ensuremath{-}35.8.
\item \textbf{Topical and long text.} On tweets neither size gained from retrieval (\ensuremath{-}9.2 and \ensuremath{-}4.2; Jev +4.2). On CFPB, retrieval changed accuracy by \ensuremath{-}26.7 points at 0.5B and \ensuremath{-}3.3 at 1.5B (Jev +3.3), and three wrong labels hurt neither size (+3.3 and +6.7; Jev \ensuremath{-}8.3).
\item \textbf{Information claims at 0.5B.} True minus shuffled shares (+6.7 on tweets; Jev +10.8) and the cost of cryptic ids replicate in direction; examples under abstract ids (+7.5 against +89.2) and definitions (0.0 against +5.8) do not.
\item \textbf{Floor.} Bare accuracy rose with size (Banking77-20 33.3\% at 0.5B, 46.7\% at 1.5B; Jev 69.2\%), so capability is confounded with size, and no effect is shown to be specific to Jev.
\end{itemize}

\appsub{H.3}{Open decision models (local pilot; observed)}
\textbf{H.3 Open decision models (local pilot; observed).} Kev (0.8B to 27B) accepts Jev's request format, so our arms replay byte-identically; OpenJev is a second trained-head lineage; Laya and GLiClass test whether the definitions and examples levers exist outside LLM-style models. \emph{The pre-pilot plan} was ten paired arms on 120 items of three datasets per model, estimated at under \$3 plus about one CPU day and one T4 hour. \emph{The pilot as run} used 9 arms (8 shared plus one derived) on 2 datasets, the 20 Banking77 intents Jev confuses most (Banking77-20, so provisional) and tweets, with Kev-0.8B, OpenJev-1.7B and the 421M-parameter encoder Laya: 10,560 requests, about 45 CPU-hours, no T4 and no API cost. Its analysis plan was fixed in its own file before any download, with no multiplicity correction and no manifest, and counts toward no Run 5 verdict. Wrapper text moved none of the three beyond noise, cryptic ids hurt all three, and one-line definitions rescued ids on all three (+14 to +66 points). How they weigh shown examples is model-specific. OpenJev, trained partly on Jev-distilled data, reproduced Jev's wrong-label harm on both datasets and about half its retrieval gain on Banking77-20 (+12.5 against +24.2), but examples hurt it on tweets (\ensuremath{-}10.0 {[}\ensuremath{-}18.3, \ensuremath{-}2.5{]}). Kev reproduced the wrong-label harm on Banking77-20, its retrieval gain there was inconclusive (+5.0 {[}\ensuremath{-}1.7, +11.7{]}), and on tweets it mostly answered one class. Laya's shipped input budget truncates examples to 8 tokens per option, so its example results are uninformative. Two of the pilot's own planned Laya predictions failed and are reported as such. The trained-head control (GLiClass, DeBERTa-NLI) was not run. Kev-4B is one of the Run 5 models. The model table, arm list and per-arm results are in Extended Appendix H.3.

\appsub{H.4}{Run 5: models, request mapping and tiers}
\textbf{H.4 Run 5: models, request mapping and tiers} (plan fixed in advance; full text in Extended Appendix H.4; results in Section 6 and Extended Appendix S.1). The model list was fixed before any Run 5 request, with versioned names {[}\citenum{r69}{]}; Mercury Decide and Cloudflare Clef-flash were dropped before the freeze (Clef-flash's probe returned HTTP 400). Perplexity and OpenRouter take Jev's request body unchanged; for OpenAI the state becomes \code{input} and each option a choice with its description, in the same order. The seven models run the standard tier (wrappers on 600 items with all five stimuli at \ensuremath{\pm}2; retrieval on 600 items \ensuremath{\times} 2 wordings; wrong labels on 231 MASSIVE and 267 MetaTool feasible items; definitions on 200 \ensuremath{\times} 2); Jev 1.14 runs the minimal tier, a \ensuremath{\pm}3 wrapper screen. A model that fails its 20-request pilot counts as not confirmed. List prices per million input tokens were \$0.01 (Decider Flash) to \$0.10 (OpenAI), and \$0.042 for Jev 1.14, Microsoft-Decision-1, d1-ultra and Kev-4B (OpenRouter listing for the last three, accessed 2026-11-09; Run 5 analysis plan \S{}9). The estimated cost was \$6.46; the run cost \$2.98, because Solar Decide stopped at its pilot and every completed model other than Jev 1.14, whose estimate is exact, billed fewer input tokens than estimated (\code{run5\_\allowbreak{}launch\_\allowbreak{}summary.\allowbreak{}json}).

\textbf{Run 5 in detail.}

\textbf{Design.} Each model received the requests Jev 1.13 had answered in Runs 1 and 3, byte for byte (OpenAI's field names mapped one to one; Appendix H.4). No model other than Jev had been sent these requests, and all answered the same items, so the seven verdicts are not independent replications. Four tests ran, with Holm's correction within the model: persona and politeness stay within \ensuremath{\pm}2 points (Don't 1); retrieved examples beat the bare call (Do 1); relabelling all three correct examples shown hurts (Do 3); and definitions lift cryptic ids by more than 10 points (Do 2). A rule \textbf{transfers} if confirmed on at least 5 of the 7 models with no confirmed opposite effect; a model that fails its pilot counts as not confirmed, and a cell with fewer than 98\% complete items is a protocol failure. Jev 1.14 ran a smaller tier, with a \ensuremath{\pm}3 wrapper screen.

\begingroup\footnotesize
\renewcommand{\thetable}{H.4}
\begin{longtable}{@{}>{\raggedright\arraybackslash}p{\dimexpr 0.2693\linewidth-2.6929\tabcolsep\relax}>{\raggedright\arraybackslash}p{\dimexpr 0.1645\linewidth-1.6448\tabcolsep\relax}>{\raggedright\arraybackslash}p{\dimexpr 0.1383\linewidth-1.3828\tabcolsep\relax}>{\raggedright\arraybackslash}p{\dimexpr 0.1252\linewidth-1.2518\tabcolsep\relax}>{\raggedright\arraybackslash}p{\dimexpr 0.1514\linewidth-1.5138\tabcolsep\relax}>{\raggedright\arraybackslash}p{\dimexpr 0.1514\linewidth-1.5138\tabcolsep\relax}@{}}
\caption{Run 5: accuracy change (points, 90\% class-level interval) and the planned decision per model. Don't 1 shows the lever (persona or politeness) on which its two-lever test decides; both levers, and the qualifier that OpenAI's politeness loss is not consistent across datasets (MASSIVE \ensuremath{-}2.4), are in Extended Appendix S.1. The Jev 1.13 row is its estimate on the same items (a reference, not a Run 5 test). Five labels per class is secondary. Solar Decide failed its pilot and counts as not confirmed. With OpenAI Decisions' refused requests scored against the test (refused controls right, refused treatments wrong), its definitions gain is +58.1 {[}+52.1, +63.8{]}, still confirmed (re-analysis 2), so definitions transfer on 6 of 7 under any scoring.}\\
\toprule
Model & Don't 1: deciding lever & Do 1: retrieval & Three wrong labels & Do 2: definitions on cryptic ids & Do 4: 5 per class (secondary) \\
\midrule
\endfirsthead
\multicolumn{6}{@{}l}{\textit{Table H.4 (continued)}}\\
\toprule
Model & Don't 1: deciding lever & Do 1: retrieval & Three wrong labels & Do 2: definitions on cryptic ids & Do 4: 5 per class (secondary) \\
\midrule
\endhead
\bottomrule
\endlastfoot
Jev 1.13 (same items, reference) & \ensuremath{-}0.4 {[}\ensuremath{-}1.2, +0.4{]} & +8.0 {[}+5.7, +10.2{]} & \ensuremath{-}60.6 {[}\ensuremath{-}65.2, \ensuremath{-}56.0{]} & +81.0 {[}+75.8, +85.7{]} & -- \\
Jev 1.14 (version anchor) & \ensuremath{-}0.3 {[}\ensuremath{-}1.9, +1.4{]}, within \ensuremath{\pm}3 (screen) & +8.4 {[}+5.9, +11.1{]}, confirmed & \ensuremath{-}67.3 {[}\ensuremath{-}73.0, \ensuremath{-}61.8{]}, confirmed & +81.5 {[}+74.9, +87.6{]}, confirmed & not run \\
Perplexity Decider v1.2 & \ensuremath{-}0.3 {[}\ensuremath{-}0.9, +0.2{]}, within \ensuremath{\pm}2 & +3.6 {[}+1.8, +5.9{]}, confirmed & \ensuremath{-}21.3 {[}\ensuremath{-}26.1, \ensuremath{-}16.1{]}, confirmed & +82.0 {[}+77.0, +87.0{]}, confirmed & +2.0 {[}\ensuremath{-}0.4, +5.3{]} \\
Perplexity Decider Flash & 0.0 {[}\ensuremath{-}0.9, +0.6{]}, within \ensuremath{\pm}2 & +4.6 {[}+2.9, +6.5{]}, confirmed & \ensuremath{-}25.8 {[}\ensuremath{-}31.0, \ensuremath{-}20.6{]}, confirmed & +80.9 {[}+74.9, +86.2{]}, confirmed & +3.2 {[}+1.2, +5.7{]} \\
OpenAI Decisions (gpt-6-luna) & \ensuremath{-}1.4 {[}\ensuremath{-}2.8, \ensuremath{-}0.1{]}, inconclusive & +10.5 {[}+8.0, +13.2{]}, confirmed & \ensuremath{-}25.6 {[}\ensuremath{-}30.4, \ensuremath{-}20.7{]}, confirmed & +75.0 {[}+68.9, +81.2{]}, confirmed; cells below 98\% complete & +3.6 {[}+0.7, +7.0{]} \\
Microsoft-Decision-1 & \ensuremath{-}0.1 {[}\ensuremath{-}0.8, +0.5{]}, within \ensuremath{\pm}2 & +2.5 {[}+0.8, +4.5{]}, confirmed & \ensuremath{-}27.7 {[}\ensuremath{-}32.7, \ensuremath{-}22.8{]}, confirmed & +81.8 {[}+76.3, +86.8{]}, confirmed & +0.4 {[}\ensuremath{-}1.5, +2.8{]} \\
Liquid d1-ultra & +0.2 {[}\ensuremath{-}0.7, +1.1{]}, within \ensuremath{\pm}2 & +2.6 {[}+0.2, +5.1{]}, confirmed & \ensuremath{-}40.6 {[}\ensuremath{-}46.6, \ensuremath{-}33.5{]}, confirmed & +81.0 {[}+75.3, +86.0{]}, confirmed & \ensuremath{-}1.5 {[}\ensuremath{-}5.1, +2.5{]} \\
Kev-4B & +0.5 {[}\ensuremath{-}0.9, +2.0{]}, inconclusive & +9.9 {[}+7.3, +12.6{]}, confirmed & \ensuremath{-}24.9 {[}\ensuremath{-}28.9, \ensuremath{-}21.1{]}, confirmed & +71.2 {[}+65.5, +76.9{]}, confirmed & +5.6 {[}+1.3, +10.4{]} \\
Upstage Solar Decide & dropped at its pilot (HTTP 422) & not confirmed & not confirmed & not confirmed & -- \\
\textbf{Verdict (at least 5 of 7, none opposite)} & \textbf{partially transfers (4 of 7)} & \textbf{transfers (6 of 7; 4 of 7 under one correction across models)} & \textbf{transfers (6 of 7)} & \textbf{transfers (6 of 7; 5 of 7 excluding OpenAI's incomplete cells)} & (no verdict) \\
\end{longtable}
\endgroup

Figure H.1, the main text's Figure 5 up to v24 (moved in v25), shows the first three of these tests per model.

\begin{figure}[!htbp]
\centering
\includegraphics[width=\linewidth,height=0.72\textheight,keepaspectratio,alt={Three dot-and-interval panels share one row per model: Jev 1.13 on the same items (black square, reference), Jev 1.14, Perplexity Decider v1.2, Perplexity Decider Flash, OpenAI Decisions (luna), Microsoft-Decision-1, Liquid d1-ultra, Kev-4B, and Upstage Solar Decide, which has no points and the note "dropped at its pilot (HTTP 422)". Panel "Don't 1: wrappers", x from -4 to +4 with a grey +/-2 band: every point sits inside the band, from -1.4 (OpenAI's model, hollow, its interval reaching -2.8) to +0.5 (Kev-4B, hollow, its interval reaching +2.0); the others are filled. Panel "Do 1: retrieved examples", x from 0 to +15: Jev 1.13 at +8.0, Jev 1.14 +8.4, OpenAI's model +10.5 and Kev-4B +9.9; Decider +3.6, Decider Flash +4.6, Microsoft-Decision-1 +2.5 and d1-ultra +2.6, all intervals above 0. Panel "Three wrong labels shown", x from -75 to 0: Jev 1.13 at -60.6 and Jev 1.14 at -67.3 at the far left; d1-ultra at -40.6; the other five between about -21 and -28.}]{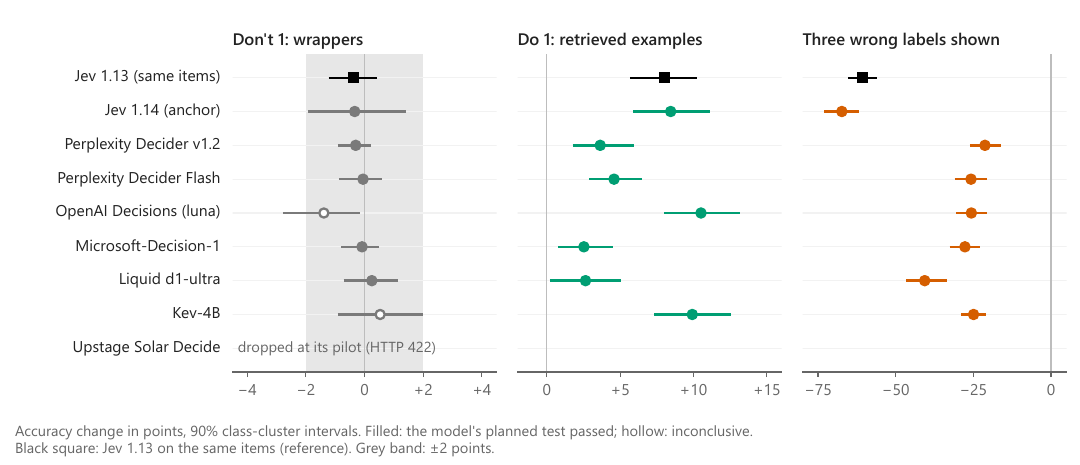}
\renewcommand{\thefigure}{H.1}
\caption{Run 5 per model (points, 90\% class-level intervals). Filled circles passed the model's planned test (Jev 1.14's wrapper test is a \ensuremath{\pm}3 screen), hollow circles are inconclusive, and black squares are Jev 1.13 on the same items. All models answered the same items. Run 5's fourth test is in Appendix A.5. Sources: Appendix L.}
\end{figure}

\textbf{The information rules transfer.} Retrieved examples raised accuracy on all six models that completed, by 2.5 to 10.5 points (Jev 1.13: 8.0 on the same items; Decider and Microsoft-Decision-1 lost a little on MASSIVE), and relabelling three correct examples hurt all six: both transfer on 6 of 7. Under one Holm correction across all models, a planned sensitivity check, retrieval stays confirmed on only 4 of 7 (d1-ultra and Microsoft-Decision-1 drop out). Definitions lifted cryptic ids by 71 to 82 points (Jev 1.13: 81) from a near-zero floor; the planned verdict is ``transfers (6 of 7)''. OpenAI's model refused 210 requests and completed only 123 and 167 of 200 items in those cells, below 98\%, which the analysis plan treats as a protocol failure; it does not remove the cells from the verdict, and with them excluded the count is 5 of 7, which still transfers (a sensitivity check).

\textbf{Wrong-label harm is general in direction, not in size.} The other models lost 21 to 41 points where Jev 1.13 lost 61 on the same items (Jev 1.14: 67; secondary). On five of the six, one, two and three wrong labels cost 3 to 5, 11 to 20 and 21 to 28 accuracy points on the same constructed display (secondary; in gold log-odds no model's losses kept accelerating after the second wrong label, though two models' first step was slightly convex; Appendix C.10); on d1-ultra the curve was not monotone (\ensuremath{-}21.7, \ensuremath{-}56.8 and \ensuremath{-}40.6). Normalised by what correct examples gave each model (observed), Jev 1.13 lost 8.2 times its clean gain (90\% interval 5.8 to 13.8): more than Kev-4B (2.6) and OpenAI Decisions (2.8), but not separable from Decider (4.6) or Decider Flash (5.2); the ratio is undefined for d1-ultra and Microsoft-Decision-1, whose clean gain was about 1 point. In margin units, Jev 1.13 (2.90) was less sensitive than Decider, Decider Flash and Microsoft-Decision-1, so the ratio depends on the scale. d1-ultra's non-monotone curve came from MetaTool, where accuracy rose 27.9 points from two to three wrong labels, with no refusals or missing options. As a check of the decision rule itself, 3 of 73 sets of identical Jev re-sends were flagged (4.1\%; 95\% interval 0.9 to 11.5), and permuting Run 6's re-sends gave a Holm family error of 6.9\% with class clusters. A stress test with real between-class spread showed that the class bootstrap ran at 8.5 to 8.8\% instead of 5\% with 8 to 25 classes (re-analysis 2) and item-level intervals at 38 to 46\%, so results that pass only at item level are reported as observed (Section 2.3).

\textbf{The wrapper rule partially transfers.} Persona and politeness stayed within \ensuremath{\pm}2 points on Decider, Decider Flash, Microsoft-Decision-1 and d1-ultra (4 of 7). On OpenAI's model politeness cost 1.4 points {[}\ensuremath{-}2.8, \ensuremath{-}0.1{]}, a small harm whose interval excludes zero, and on Kev-4B the interval reached +2.0; both are inconclusive under the \ensuremath{\pm}2 test. Jev 1.14 passed its \ensuremath{\pm}3 screen.

\textbf{Deviations.} Solar Decide failed its pilot (17 of 20 requests returned HTTP 422), so it counts as not confirmed and six models inform each verdict. Per-model tables, the 5-per-class results and the deviations are in Extended Appendix S.1. Re-analysis 2's random-effects summary and refusal bound are in Appendix U.3.

\textbf{A newer Jev.} Jev 1.14 agreed with Jev 1.13's bare top option on 94 to 97\% of items and passed all four tests (the wrapper test as a \ensuremath{\pm}3 screen), but lost 3.7 points more to wrong labels on the same items (secondary; the Table H.4 rows differ by 6.7 because Jev 1.14 ran a subset). Re-run held-out comparisons on every version change.

\section{Confirmatory run details}

Table P.1 lists all 29 tests with estimates, Holm p-values and decisions. The frozen code's per-run tables (Tables I.1, I.2 and I.4), Figure I1, per-arm results with billed tokens, Bonferroni-level intervals, power, alternative bootstraps, the joint-Holm table and Tables I.6 to I.9 are in Extended Appendix I.

\appsub{I.1}{to I.3 Runs 1 and 2}
\textbf{I.1 to I.3 Runs 1 and 2.} In Run 1 the informative superiority and harm results are T1, T4 and T7a; T3 and T5 were close to certain by design (T3's vote scores below bare Jev; T5's shuffled shares are the most distant of 200 derangements). T2 failed, driven by CLINC150 (+4.9 {[}+3.8, +6.1{]}; Banking77 +1.2 {[}+0.4, +2.0{]}). T6 and T8 hold within \ensuremath{\pm}2 but are not zero, and T7 has a construct confound: MetaTool's queries were generated from the descriptions {[}\citenum{r16}{]} that T7 adds. At Bonferroni level, T8 probe first {[}\ensuremath{-}2.17, +0.58{]} and T7b {[}\ensuremath{-}2.1, +1.0{]} cross their margins; their Holm decisions stand because Holm's thresholds at their ranks are looser than 0.05/15. Run 1 deviated from its plan only in item counts after the planned filters (for example 2,975 and 4,443 T2 items against 3,080 and 4,500 planned); no decision changed. Run 2 used all never-sent items of each dataset (1,170 tweets, 1,406 CFPB, 2,244 ToxicChat; 36,096 requests, \$1.63) and recorded no deviation. G5's combination still pays for one bare call per item (467 against 903 input tokens on tweets; 517 against 2,183 on CFPB). G3's matched arm was feasible for 938 of 1,170 items; the other 232 are items whose only gold examples shown are the three nearest, so G3 says nothing about items where rank and correct-label evidence coincide. Every Run 2 non-inferiority interval stays above \ensuremath{-}1.5 at Bonferroni level. Planned power was .84 or more for every Run 1 test, and .35 to .36 for G2, G3 and G4a in Run 2; low-powered tests that fail are reported as ``not confirmed'' or ``not shown'', never as ``no effect''.

\appsub{I.7}{Robustness of the confirmatory decisions}
\textbf{I.7 Robustness of the confirmatory decisions} (exploratory).

\begingroup\small
\renewcommand{\thetable}{I.3}
\begin{longtable}{@{}llll@{}}
\caption{Run 1 equivalence and non-inferiority decisions under alternative analyses (Holm p; exploratory). ``Per paraphrase'': each paraphrase is its own component. Details in Extended Appendix I.7.}\\
\toprule
Analysis & T6 & T7b & T8 \\
\midrule
\endfirsthead
\multicolumn{4}{@{}l}{\textit{Table I.3 (continued)}}\\
\toprule
Analysis & T6 & T7b & T8 \\
\midrule
\endhead
\bottomrule
\endlastfoot
M0 as planned: item-level clusters, paraphrase-averaged & 0.00075 & 0.019 & 0.019 \\
M1 item-level clusters, per paraphrase & 0.14 & 0.056 & 0.15 \\
M2 wild class-cluster bootstrap, paraphrase-averaged & 0.086 & 0.072 & 0.012 \\
M3 wild class-cluster bootstrap, per paraphrase & 0.39 & 0.17 & 0.090 \\
M4 pairs bootstrap, gold-class clusters, per paraphrase & 0.36 & 0.13 & 0.11 \\
\end{longtable}
\endgroup

\textbf{The Run 1 equivalence and non-inferiority claims need both item-level clusters and paraphrase averaging.} Either change alone fails T6 and T7b (M1, M2); T8 survives class clusters only with paraphrase averaging. No point estimate under any analysis lies beyond \ensuremath{\pm}1.3 points. M4 is the few-cluster pairs bootstrap that Cameron et al. warn over-rejects {[}\citenum{r4}{]}; M2 and M3 use the wild bootstrap {[}\citenum{r49}{]}.

\textbf{Run 2 is consistent across paraphrases but not across classes.} Every Run 2 estimate keeps its sign per paraphrase. With gold-class clusters (20 for tweets, 8 for CFPB), raw p is .021 (G4a), .050 (G4b), .052 (G5), .16 (G3) and .20 (G2), so no decision would pass Holm; with this few clusters the pairs bootstrap is itself unreliable {[}\citenum{r4}{]}. The Run 2 decisions hold for the item-level sampling unit the analysis plan chose, not for a population of classes.

\textbf{Every Run 1 test with gold-class clusters} (observed; Extended Appendix R.4). With a pairs bootstrap over gold classes, T1a-c, T3a, T4a-b and T7a pass (p \ensuremath{\leq} .002), and Holm over Run 1 also keeps T8. T3b (p .052), T5a-b (.145, .166), T7b (.019, above its Holm threshold) and T6 (.020) fail; T6 fails only through politeness on tweets under the second wording. Capitals stay within \ensuremath{\pm}0.5 under every scheme. With 8 or 20 clusters these bootstraps are unreliable {[}\citenum{r4}{]}: in the stress test they rejected 8.5 to 8.8\% of true nulls at a nominal 5\% with 8 to 25 classes (re-analysis 2), against 38 to 46\% for item-level intervals (Section 2.3). A wild-cluster restricted bootstrap with Webb weights {[}\citenum{r49}{]} held 4.9 to 5.2\%, and under it T3a (prompt vs fixed vote, tweets; WCR p .017, Holm .12) is no longer rejected, so it is reported as observed (Appendix U.3). The passes that fail here (T3b, T5a-b, T6, T7b) and the Run 2 passes above (G2, G4a-b, G5) are therefore reported as observed (item-level only).

\appsub{I.10}{Retrieval gains on test items}
\textbf{I.10 Retrieval gains on test items} (exploratory). The gains are not near-duplicate lookup: in the lowest tercile of maximum pool similarity every T1 and T3 gain stays positive (smallest: ToxicChat +3.3 {[}+2.1, +4.5{]}), and dropping items with a pool twin at cosine 0.8 moves every gain by at most 0.4 points. T2 failed partly because CLINC150's development items sit closer to the pool than its test items, which favours the vote. Every primary contrast keeps its sign under each instruction wording; the exception that matters is tweets politeness, \ensuremath{-}0.3 {[}\ensuremath{-}1.0, +0.3{]} and \ensuremath{-}1.3 {[}\ensuremath{-}2.0, \ensuremath{-}0.6{]}.

\appsub{I.11}{One Holm correction over all confirmatory tests}
\textbf{I.11 One Holm correction over all confirmatory tests} (exploratory; not the planned method). Over the 21 tests of Runs 1 and 2, or the 24 tests of Runs 1 to 3, no decision changes. Over all 29 tests (\code{r14\_\allowbreak{}holm29.\allowbreak{}csv}), T8 (adjusted p .052) and H3 (.070) would no longer be rejected, because the step-down stops at T8 (raw p 0.0087 against 0.05/6 = 0.0083); every other decision stands, with T7b at .045 (T7b is reported as observed anyway, since it passed only at item level). Each run's own Holm family is the planned method, so Table P.1's decisions stand; T8 and H3 are the two least robust confirmed passes. With wild-cluster class-level p-values (re-analysis 2), 13 of the 16 then-confirmed tests survive one family over the 29 (T3a, T8 and H3 do not), and 11 under a hierarchical split with \ensuremath{\alpha}/4 per run (T1c and T7a fail as well; Appendix U.3).

\appsub{I.12}{Run 3: three tests on fresh items of four datasets, publicly deposited}
\textbf{I.12 Run 3: three tests on fresh items of four datasets, publicly deposited.}

\emph{Plan and public archive.} The analysis plan (\code{reports/\allowbreak{}preregistration\_\allowbreak{}run3.\allowbreak{}md}) was frozen by the manifest of Appendix P.3 (\code{registry\_\allowbreak{}post/\allowbreak{}freeze/\allowbreak{}manifest\_\allowbreak{}run3\_\allowbreak{}20261109T151630Z.\allowbreak{}json}) and deposited at \url{https://gist.github.com/BjBodner/c736ea7d935967253f4fae06ff51da90}, 22 minutes before the first billed Run 3 call. The deposited copy names its manifest. To check it: (1) convert every CRLF line ending to LF (the freeze script hashed \code{Path.\allowbreak{}read\_\allowbreak{}text()} output); (2) replace the line that starts with \code{MANIFEST\_\allowbreak{}SHA256:\allowbreak{}} by its placeholder; (3) take the SHA-256 of the UTF-8 bytes and compare it with the manifest entry. Hashing the deposited bytes without step 1 gives a mismatch. The run verified the Run 3, Run 2 and Run 1 manifests before sending and again before analysis, found the observation log unchanged since the freeze, and recorded no validity problem.

\emph{Design.} Items passed the Run 1 twin rules, appear in no logged request and have no TF-IDF twin among sent texts; MASSIVE had never been sent, and the CFPB items are stream rows no earlier code read. H1 prepended five personas and five politeness prefixes (Appendix J) under one of three wordings to 600 items per dataset, with a pigeonhole bootstrap {[}\citenum{r58}{]} over stimuli, wordings and gold classes. H2 relabelled k = 0 to 3 of the gold-labelled shown examples at seeded random ranks on the 1,049 items that showed exactly three; each relabel drew a non-gold class uniformly, so at k = 3 some rival received two or more on 3.5 to 15.7\% of items. H3 used one-line definitions written blind from the label names, excluding MetaTool. Tests are pooled dataset-equal with gold classes resampled within each dataset; Holm over three. H4, H5 and H5i are secondary because their power under class clusters stayed below 0.8 at any n.

\emph{Results by dataset.} H1 persona ranged from \ensuremath{-}1.0 (MASSIVE) to +0.3 (MetaTool) and politeness from \ensuremath{-}0.3 to +0.2. The H2 slope was \ensuremath{-}17.8 (MASSIVE), \ensuremath{-}7.4 (LEDGAR), \ensuremath{-}21.2 (MetaTool) and \ensuremath{-}2.8 (CFPB); accuracy at k = 0 / 1 / 2 / 3 was 95.5 / 92.6 / 84.4 / 38.7\% on MASSIVE and 83.6 / 82.0 / 79.7 / 74.9\% on CFPB. Pooled steps were \ensuremath{-}2.9 {[}\ensuremath{-}3.8, \ensuremath{-}2.1{]}, \ensuremath{-}8.5 {[}\ensuremath{-}9.9, \ensuremath{-}7.0{]} and \ensuremath{-}26.8 {[}\ensuremath{-}30.2, \ensuremath{-}23.8{]}, so the slope averages a loss that is convex in accuracy (in gold log-odds Jev's shape cannot be identified; Appendix C.10); that growth is a pre-specified description, not a test. H3 (a) took ids from 2.5\%, 3.0\% and 24.0\% to 81.3\%, 89.8\% and 78.3\% (MASSIVE, LEDGAR, CFPB), and H3 (b) was \ensuremath{-}0.8, \ensuremath{-}2.4 {[}\ensuremath{-}4.0, \ensuremath{-}1.1{]} and \ensuremath{-}0.1. With near-duplicate clusters, or under each wording alone, every decision would stand. No deviation was recorded; the main block cost \$2.80.

\emph{Jev 1.14 block} (secondary, outside the family). The first 400 fresh CFPB and MetaTool items were sent to \code{jev-\allowbreak{}1.\allowbreak{}14.\allowbreak{}0} under four arms, with 1.13 counterparts in the same block. Every version difference was within 2.5 points in point estimate. Against the planned directional criterion, retrieval and the top-3 relabel met it on MetaTool (1.14: +8.8 and \ensuremath{-}38.9) but not on CFPB (+2.0 and \ensuremath{-}1.5); neither shares interval excluded 0; persona stayed inside \ensuremath{\pm}2 on both. Apart from the MetaTool top-3 relabel (\ensuremath{-}2.5 {[}\ensuremath{-}4.5, \ensuremath{-}0.5{]}), the only version difference whose interval excludes 0 is CFPB persona, +1.0 {[}+0.1, +1.8{]}. The block cost \$0.62.

\appsub{I.13}{The mechanism model}
\textbf{I.13 The mechanism model} (exploratory; \code{experiments/\allowbreak{}analysis\_\allowbreak{}r13.\allowbreak{}py}, read-only cache). The model below was fitted by soft cross-entropy to Jev's own kNN-20 probability vectors (not to gold), datasets weighted equally, on Run 1 and Run 2 arms of Banking77, CLINC150, tweets and CFPB with both wordings (40,340 requests), and applied unchanged to Run 3. Its form and variants were written in the module docstring before any fit, which is not a third-party deposit. Across leave-one-dataset-out (LODO) fits, \emph{a} ranged from 0.347 to 0.409, \ensuremath{\beta} from 1.07 to 2.47, \ensuremath{\tau} from 12.1 to 18.4 and \ensuremath{\varepsilon} from 0.0015 to 0.046, and the predicted pooled accuracy at k = 3 over 40.5 to 61.6\%. Held out, it fitted Jev's probability vectors better than a pure vote or the bare prior alone on every dataset (KL 0.107 to 0.342 nats, against 0.212 to 1.377 and 0.175 to 0.845). It underpredicted copying of planted near-copies on Banking77-20 (accuracy 83.5\% predicted against 73.8\% observed after one plant) and overpredicted it on tweets (15.5\% against 49.5\%), probably because its kernel uses an external embedding in place of Jev's own similarity. Table I.5, with every quantity, is in Extended Appendix I.13. The fitted floor is \ensuremath{\varepsilon} = 0.0225 (exp(\ensuremath{-}3.7944)).

\textbf{The prior \ensuremath{\times} evidence model} (observed).

\textbf{A simple model of prior and evidence.} We describe Jev's output with a four-number model. For each option it adds the option's log-probability in a bare call (the prior) to a similarity-weighted count of the shown examples with that label (the evidence):

\begin{lstlisting}
logit_c = a (*@\ensuremath{\cdot}@*) log p_bare(c) + (*@\ensuremath{\beta}@*) (*@\ensuremath{\cdot}@*) log( (*@\ensuremath{\varepsilon}@*) + (*@\ensuremath{\Sigma}@*)_{shown j labelled c} exp((*@\ensuremath{\tau}@*) (*@\ensuremath{\cdot}@*) (cos_j (*@\ensuremath{-}@*) 1)) )
\end{lstlisting}

Here cos\us{}j is the bge-small similarity between the query and shown example j, and the floor \ensuremath{\varepsilon} lets an option with no shown example keep its prior. Fitted to 40,340 responses from Runs 1 and 2, a = 0.37, \ensuremath{\beta} = 1.92, \ensuremath{\tau} = 13.9 and \ensuremath{\varepsilon} = 0.0225. The form was fixed before fitting but not deposited, so this section is observed (Appendix I.13); Appendix A.4 lists the accounts that motivate it.

\textbf{It tracks the pooled wrong-label curve, not each dataset.} Applied without refitting to Run 3, it gave 91.6, 88.4, 80.9 and 48.2\% accuracy for 0 to 3 wrong labels, against 92.2, 89.4, 80.9 and 54.1\% observed (Figure I.13). The pooled fit hides per-dataset errors of 14 to 30 points in both directions in Run 3 (Extended Appendix S.3). Accounts of in-context learning as implicit inference over a latent concept {[}\citenum{r59}; \citenum{r68}{]} and label words acting as anchors {[}\citenum{r60}{]} motivate this form.

\begin{figure}[!htbp]
\centering
\includegraphics[width=\linewidth,height=0.72\textheight,keepaspectratio,alt={Three panels plot accuracy and correct-option log-probability against the number of correct shown examples relabelled, from 0 to 3. In panel (a), the pooled observed curve (blue) falls from 92  to 89 , 81  and 54 . The model's curve (orange, dashed), fitted on Runs 1 and 2 of Banking77, CLINC150, tweets and complaints, falls from 92  to 88 , 81  and 48 , and the two overlap at 0 to 2 relabels. In panel (b), MASSIVE's model curve tracks its observed one. LEDGAR and complaints fall far less than the model says, and MetaTool falls further. In panel (c), log-probability drops slowly at first and steeply at the third relabel, in both curves.}]{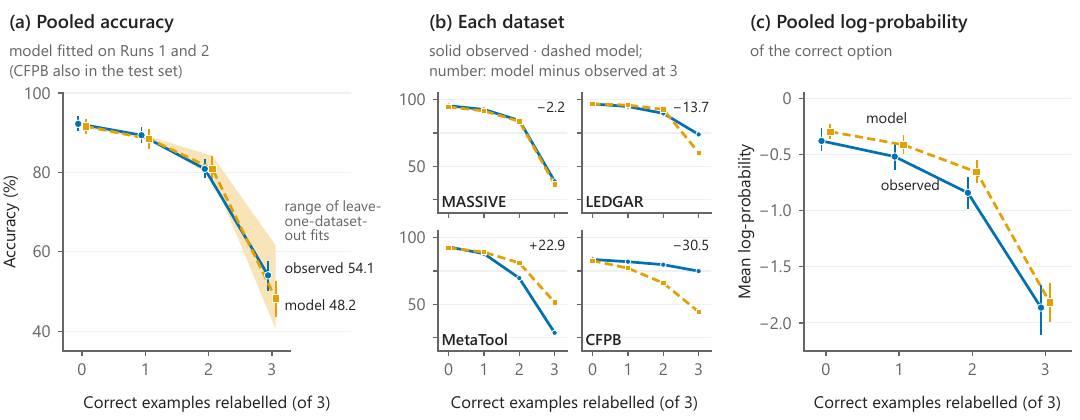}
\renewcommand{\thefigure}{I.13}
\caption{The prior \ensuremath{\times} evidence model, fitted on Runs 1 and 2 and applied without refitting to Run 3's wrong-label items (observed). Complaints (CFPB) is in both sets, on different rows; MASSIVE, LEDGAR and MetaTool are new to the model. (a) Pooled accuracy, observed and modelled, with 90\% class-level intervals and the range of leave-one-dataset-out fits. The model tracks the pooled curve. (b) Each dataset, observed (solid) and modelled (dashed): errors of 14 to 30 points on three of the four datasets (MASSIVE \ensuremath{-}2.2) that cancel when pooled. (c) Pooled log-probability of the correct option. A rival that matches the query to the option description text alone fits Jev clearly worse (Appendix I.14). Sources: Appendix L.}
\end{figure}

\textbf{It got the direction of Run 4, but only half of the rival effect.} It got all three directions and the deletion jump (+11.7 modelled, +10.1 observed) right, but only about half of the two rival effects (Table I.13): Jev weighs a rival's growing evidence more steeply than the model allows.

\begingroup\footnotesize
\renewcommand{\thetable}{I.13}
\begin{longtable}{@{}>{\raggedright\arraybackslash}p{\dimexpr 0.3237\linewidth-1.9420\tabcolsep\relax}>{\raggedright\arraybackslash}p{\dimexpr 0.2174\linewidth-1.3043\tabcolsep\relax}>{\raggedright\arraybackslash}p{\dimexpr 0.2536\linewidth-1.5217\tabcolsep\relax}>{\raggedright\arraybackslash}p{\dimexpr 0.2053\linewidth-1.2319\tabcolsep\relax}@{}}
\caption{Run 4's wrong-label variants, observed and from the model fitted on Runs 1 and 2 (no refit; observed). Pooled over MASSIVE, LEDGAR and MetaTool (1,057 items); accuracy \% for variants, points for contrasts; 90\% class-level intervals; LODO = range over the four leave-one-dataset-out fits. Sources: Appendix L.}\\
\toprule
Quantity & Observed & Model (LODO range) & Model \ensuremath{-} observed \\
\midrule
\endfirsthead
\multicolumn{4}{@{}l}{\textit{Table I.13 (continued)}}\\
\toprule
Quantity & Observed & Model (LODO range) & Model \ensuremath{-} observed \\
\midrule
\endhead
\bottomrule
\endlastfoot
Clean display & 95.3 & 94.5 (93.9 to 94.8) & \ensuremath{-}0.8 {[}\ensuremath{-}1.8, +0.2{]} \\
Delete 1 / 2 / 3 & 93.5 / 89.8 / 75.9 & 91.2 / 86.4 / 69.9 & -- \\
Relabel to distinct rivals, 1 / 2 / 3 & 91.1 / 80.3 / 48.3 & 91.0 / 84.1 / 51.5 & -- \\
Relabel to one rival, 2 / 3 & 74.2 / 28.1 & 84.3 / 37.9 & -- \\
Delete minus distinct relabel & +13.1 & +7.0 (+4.5 to +8.6) & \ensuremath{-}6.2 {[}\ensuremath{-}8.0, \ensuremath{-}4.4{]} \\
One rival minus distinct rivals & \ensuremath{-}13.2 & \ensuremath{-}6.7 (\ensuremath{-}8.9 to \ensuremath{-}4.5) & +6.5 {[}+4.7, +8.3{]} \\
Deletion jump at the third example & +10.1 & +11.7 (+7.3 to +14.1) & +1.6 {[}\ensuremath{-}1.4, +4.8{]} \\
\end{longtable}
\endgroup

\appsub{I.14}{Option-description matching as a rival account}
\textbf{I.14 Option-description matching as a rival account} (observed; Extended Appendix R.5). Rival models that score each option only by the query's similarity to its description texts (maximum or mean), fitted like the I.13 model and tested with no refit, agree with Jev's top option on .726 and .702 of held-out-dataset items (KL 0.918 and 1.114), against .912 (KL 0.200); with the bare prior added, .891 (0.261). For one rival minus distinct rivals (observed \ensuremath{-}13.2) they give 0.0 and +7.9, and +0.8 with the prior, against \ensuremath{-}6.7 for the I.13 model. The rival forms were chosen after the fact.

\textbf{Is it label learning or description matching?} Our examples sit inside the option descriptions, so Jev might just match the query against each description's text. Matchers of that kind fit clearly worse (top-option agreement .70 to .73 against .91) and predict no extra cost when wrong labels go to one rival (observed; Appendix I.14), but cached data could not tell label learning from summed description matching; a placement test could.

\textbf{The placement test of Run 6} (Section 4 gives the result with Run 7's controls and Run 8's fresh items). Run 6 (plan fixed in advance; on Run 4's items) sent Jev 1.13 the same labelled examples either inside the option descriptions, as everywhere above (A, cached from Run 4), or as one labelled block after the instruction, with every description empty (B), on the 1,057 Run 4 items that showed three correct examples. Description matching cannot produce a one-rival penalty in B, because no description changes there. All three tests passed:

\begin{itemize}
\item \textbf{Examples in the block still help}: +7.0 points over the bare call {[}+4.2, +9.8{]}, against +7.6 inside the descriptions. For clean examples, placement made no clear difference (A \ensuremath{-} B +0.5 {[}\ensuremath{-}0.3, +1.4{]}; secondary).
\item \textbf{The one-rival penalty survives in the block}: putting the wrong labels on one rival cost 6.8 points more than spreading them {[}5.7, 7.9{]}.
\item \textbf{It is larger inside the descriptions}: 13.2 against 6.8 points (difference +6.4 {[}+5.0, +7.9{]}; about twice in point estimate), and still larger per unit of each placement's correct-evidence weight (0.68 against 0.50; difference +0.18 {[}+0.06, +0.29{]}; secondary).
\end{itemize}

This fits the mixed reading stated in the plan (R-MIX): Jev seems to sum per-label evidence wherever the examples sit, and wrong examples in a rival's own description weigh more. But the difference came from MASSIVE (+13.9) and LEDGAR (+4.5), not MetaTool (+0.8 {[}\ensuremath{-}1.3, +2.9{]}); the block still names the rival's label three times, so label-name frequency could carry part of its penalty; and in A the rival's description grows longer, so length is not controlled. Placement A showed no drift since Run 4 (+0.1 {[}\ensuremath{-}0.2, +0.5{]}; Extended Appendix S.2). The I.13 model has no placement term.

\section{Exact texts of the confirmatory arms}

All texts are byte-identical to the development builders that Runs 1 and 2 reused (Appendix K).

\textbf{Instruction paraphrases} (paraphrase 0 / paraphrase 1).

\begingroup\footnotesize
\begin{longtable}{@{}>{\raggedright\arraybackslash}p{\dimexpr 0.1211\linewidth-0.4845\tabcolsep\relax}>{\raggedright\arraybackslash}p{\dimexpr 0.4354\linewidth-1.7416\tabcolsep\relax}>{\raggedright\arraybackslash}p{\dimexpr 0.4435\linewidth-1.7739\tabcolsep\relax}@{}}
\toprule
Dataset & Paraphrase 0 & Paraphrase 1 \\
\midrule
\endfirsthead
\multicolumn{3}{@{}l}{\textit{Table (continued)}}\\
\toprule
Dataset & Paraphrase 0 & Paraphrase 1 \\
\midrule
\endhead
\bottomrule
\endlastfoot
Banking77 & What is the customer asking about? & What does this banking customer need help with? \\
CLINC150 & What does the user want? & Which request type does the user's message match? \\
Tweets & What is the topic of this financial news tweet? & Which category does this financial news post belong to? \\
CFPB & Which financial product is this complaint about? & What product does the consumer's complaint concern? \\
ToxicChat & Is this user message toxic? & Does this user message contain toxic content? \\
MetaTool & Which tool should the assistant use for this request? & Which of these tools best serves this user query? \\
\end{longtable}
\endgroup

\textbf{T6 prefixes} (prepended to the instruction, followed by a space):

\begin{itemize}
\item Persona: \code{You are a senior expert analyst with twenty years of experience.\allowbreak{}}
\item Capitals: \code{IMPORTANT:\allowbreak{} THINK CAREFULLY AND ANSWER CORRECTLY.\allowbreak{}}
\item Politeness: \code{Please,\allowbreak{} if you would be so kind,\allowbreak{} help me with this question.\allowbreak{} Thank you.\allowbreak{}}
\end{itemize}

\textbf{Post-unsealing phrasing stimuli} (Run 3's H1 and Appendix D.1), prepended to the instruction with one space. A language model wrote them in one sitting, before any request and without seeing items or results. Personas: P1 \code{You are a meticulous classification specialist who has labelled thousands of documents for a large archive.\allowbreak{}} P2 \code{Act as an experienced analyst at a research firm who sorts incoming reports into categories every day.\allowbreak{}} P3 \code{You are a careful reviewer known for reading every text closely before deciding where it belongs.\allowbreak{}} P4 \code{Imagine you are a veteran editor who files each piece of writing under exactly the right heading.\allowbreak{}} P5 \code{You are a domain expert with deep practical knowledge of the field this text comes from.\allowbreak{}} Politeness: Q1 \code{Please answer the following question.\allowbreak{}} Q2 \code{I would really appreciate your help with this one,\allowbreak{} thank you so much.\allowbreak{}} Q3 \code{Kindly take a look at this and let me know your answer,\allowbreak{} if you don't mind.\allowbreak{}} Q4 \code{Thanks in advance for your help!} Q5 \code{Could you please help me with the question below? I'd be grateful.\allowbreak{}}

\textbf{Run 3 texts.} H1 used the ten post-unsealing stimuli above. Its third instruction paraphrase per dataset and the 93 blind definitions of H3 (60 MASSIVE, 25 LEDGAR, 8 CFPB) are listed in the publicly archived analysis plan (its Appendix A); for example, MASSIVE \code{alarm\_\allowbreak{}set}: ``The user wants a new alarm created for a given time.''

\textbf{Share sentence} (T5 and G2; appended to the instruction): \code{Approximate share of past cases by option:\allowbreak{} } followed by every option in decreasing share with a whole-number percentage, then a full stop. True shares are the training-split shares. Shuffled shares (T5 only) assign each option another option's share: of 200 random derangements, seeded per dataset, the one with the largest L1 distance from the truth. The full tweets sentences are in Extended Appendix J.

\textbf{T7 definitions} are MetaTool's own tool descriptions, truncated at 300 characters and placed as the option description; under cryptic ids the same text sits under \code{tool\_\allowbreak{}01.\allowbreak{}.\allowbreak{}tool\_\allowbreak{}20} (alphabetical order of the names). \textbf{T4 and G3 corruption.} T4 replaces the labels of the shown examples at ranks 1 to 3 of the clean TF-IDF kNN-20 request; G3's matched arm replaces three other ranks with the same correct-label signature. Each replacement label is drawn uniformly from the options other than that example's own label, seeded by item and pool index.

\textbf{T8 companion questions} (on the same state; \code{noul} is a yes/no question, \code{score} an ordinal scale, \code{choice} a closed list): ``Does the text contain a question?'' (noul); ``Is the text written in a formal tone?'' (noul); ``How urgent is the text?'' (score: not urgent, somewhat urgent, very urgent); ``Does the text mention a specific number or amount?'' (noul); ``What is the sentiment of the text?'' (choice: positive, neutral, negative); ``Is the text longer than two sentences?'' (noul); ``How emotional is the language of the text?'' (score: not at all, somewhat, very).

\section{Reproducibility}

\begingroup\footnotesize
\renewcommand{\thetable}{K.1}
\begin{longtable}{@{}>{\raggedright\arraybackslash}p{\dimexpr 0.2204\linewidth-2.2043\tabcolsep\relax}>{\raggedright\arraybackslash}p{\dimexpr 0.1154\linewidth-1.1538\tabcolsep\relax}>{\raggedright\arraybackslash}p{\dimexpr 0.1931\linewidth-1.9311\tabcolsep\relax}>{\raggedright\arraybackslash}p{\dimexpr 0.1154\linewidth-1.1538\tabcolsep\relax}>{\raggedright\arraybackslash}p{\dimexpr 0.1747\linewidth-1.7467\tabcolsep\relax}>{\raggedright\arraybackslash}p{\dimexpr 0.1810\linewidth-1.8102\tabcolsep\relax}@{}}
\caption{Development items per variant, Run 1 test items after filters, Run 2's never-sent items, Run 3's and Run 4's fresh items, and the already-unsealed Run 2 items that Run 4 reused for pool redraws. Run 5 reused items of Runs 1 and 3 and Run 6 the items of Run 4; Run 7 reused Run 6's and Run 4's items, and Run 8's fresh items are listed in Appendix U.2.}\\
\toprule
Dataset (options) & Dev & Run 1 & Run 2 & Run 3 & Run 4 \\
\midrule
\endfirsthead
\multicolumn{6}{@{}l}{\textit{Table K.1 (continued)}}\\
\toprule
Dataset (options) & Dev & Run 1 & Run 2 & Run 3 & Run 4 \\
\midrule
\endhead
\bottomrule
\endlastfoot
Banking77, 77 intents {[}\citenum{r5}{]} & 400 & 2,975 & none left & none left & none left \\
Banking77-20 (20 intents Jev confuses most) & 400 & -- & -- & -- & -- \\
CLINC150, in-scope {[}\citenum{r21}{]} & 300 & 4,443 & none left & none left & none left \\
MASSIVE en, 60 intents {[}\citenum{r57}{]} & none sent & -- & -- & 1,200 & 1,725 \\
Financial tweet topics (20) & 400 & 3,922 & 1,170 & none left & 600 of Run 2's (pool redraws) \\
CFPB complaints, 8 products & 400 & 3,067 & 1,406 & 1,024 (new rows) & 400 of Run 2's (pool redraws) \\
CFPB time split: 2025 pool, 2026 test (8) & -- & -- & -- & -- & 2,000 \\
LEDGAR, 25 clause types {[}\citenum{r6}{]} & 400 & 5,099 (1,000 sent, observed only) & -- & 1,200 & 1,500 \\
ToxicChat 0124 (2) {[}\citenum{r23}{]} & 1,200 & 4,932 & 2,244 & none left & -- \\
MetaTool, 20 tools {[}\citenum{r16}{]} & 400 & 4,384 & -- & 1,200 built, \ensuremath{\leq} 600 per arm & 2,381 \\
\end{longtable}
\endgroup

Banking77-20 was chosen from Jev's own confusions, so claims resting on it are provisional. Ten further development sets are in Appendix C.13.

\textbf{Dataset licences} (as recorded in the project's dataset builders and research notes): Banking77: CC BY 4.0; CLINC150: CC BY 3.0; MASSIVE: CC BY 4.0; Financial tweet topics: MIT; CFPB complaints: CC0 (public domain); ToxicChat: CC BY-NC 4.0; MetaTool: MIT; LEDGAR and UNFAIR-ToS (LexGLUE): CC BY 4.0; Helpdesk tickets: CC BY-NC 4.0; Enron spam: licence unstated; see the dataset card; BoolQ: CC BY-SA 3.0; When2Call: CC BY 4.0; BFCL v3: Apache-2.0; FOMC (Trillion Dollar Words): CC BY-NC 4.0; NVD severity: CC0 1.0; Financial PhraseBank: CC BY-NC-SA 3.0; NLBSE'24 issues: licence unstated (empty licence file); see the dataset card; OpenAI moderation set: MIT; PubMedQA: MIT; Synthetic tickets and severity tasks written for this study: CC BY 4.0.

\textbf{Model and calls.} \code{jev-\allowbreak{}1.\allowbreak{}13.\allowbreak{}0}; the returned model ID was checked on every response. Some screen arms were also sent to 1.14.0 (\$0.74 of spend); no result in this paper uses those screen arms. Run 3's secondary block sent four arms to 1.14.0 on 400 items per dataset, with 1.13 counterparts, for \$0.62 (Appendix I.12). Every call is logged in the spend ledger, and the request cache is keyed by the SHA-256 of the model ID, the wire request and the repetition.

\textbf{Code.} The confirmatory builders and analyses are \code{src/\allowbreak{}jev\_\allowbreak{}prompt\_\allowbreak{}lab/\allowbreak{}stages/\allowbreak{}gate.\allowbreak{}py} (Run 1), \code{gate2.\allowbreak{}py} (Run 2), \code{run3.\allowbreak{}py}, \code{run4.\allowbreak{}py}, \code{run5\_\allowbreak{}crossmodel.\allowbreak{}py} (Run 5, with \code{model\_\allowbreak{}registry.\allowbreak{}py} and its backends), \code{run6.\allowbreak{}py} (Run 6), \code{run7.\allowbreak{}py} (Run 7) and \code{run8.\allowbreak{}py} (Run 8), with item builders under \code{src/\allowbreak{}jev\_\allowbreak{}prompt\_\allowbreak{}lab/\allowbreak{}datasets/\allowbreak{}}; Runs 3 to 8 are launched by \code{scripts/\allowbreak{}run\_\allowbreak{}run3.\allowbreak{}sh} to \code{run\_\allowbreak{}run8.\allowbreak{}sh} after their freeze scripts, and the free post-unsealing analyses are \code{experiments/\allowbreak{}analysis\_\allowbreak{}r13.\allowbreak{}py} to \code{analysis\_\allowbreak{}r17.\allowbreak{}py} (read-only cache).

\textbf{Data, retrievers and local models.} Dataset sources are Hugging Face IDs (Banking77 \code{PolyAI/\allowbreak{}banking77}, CLINC150 \code{clinc/\allowbreak{}clinc\_\allowbreak{}oos}, tweets \code{zeroshot/\allowbreak{}twitter-\allowbreak{}financial-\allowbreak{}news-\allowbreak{}topic}, CFPB \code{Mouwiya/\allowbreak{}cfpb-\allowbreak{}consumer-\allowbreak{}complaints} streamed, ToxicChat \code{lmsys/\allowbreak{}toxic-\allowbreak{}chat}, LEDGAR \code{coastalcph/\allowbreak{}lex\_\allowbreak{}glue} {[}\citenum{r6}{]}; the full list is in Extended Appendix K). TF-IDF uses word 1-2-grams with sublinear tf, fitted on pool plus development texts; the dense retriever is \code{BAAI/\allowbreak{}bge-\allowbreak{}small-\allowbreak{}en-\allowbreak{}v1.\allowbreak{}5} through fastembed 0.9.1, cosine. Both show the top 20 neighbours, at most 3 per label (10 for binary), in the option descriptions. The fixed vote weights the top 10 dense neighbours by max(cos, 0), adds 10\textsuperscript{\ensuremath{-}}\textsuperscript{3} per class and prior-corrects by prior(c)/pool share(c). Local classifiers choose C from \{0.1, 1, 10, 100, 1000\} by stratified cross-validation, are temperature-scaled on out-of-fold logits and prior-corrected. The three-way product adds 0.001 to each of the bare, vote and classifier vectors and multiplies them, with no tuned exponent. Run 2's frozen cascade thresholds were 0.7122 (tweets), 0.5879 (CFPB) and 0.7779 (ToxicChat).

\textbf{Seeds, grid and software.} Option order is permuted per item by a seed derived from the item ID, identically in control and treatment. Confirmatory bootstraps use seed 20261108 (Run 1), 20261109 (Run 2), 20261110 (Run 3), 20261111 (Run 4) or 20261113 (Run 6) plus a fixed offset per test. Responses are on a 0.01 grid and sum to 1.000; top-1 ties occur in 0.08 to 0.42\% of responses. Python 3.12.10, numpy 2.5.4, pandas 3.1.1, scikit-learn 1.9.1, scipy 1.18.1, datasets 5.1.0 and fastembed 0.9.1.

\textbf{Cost.} Jev's billed rate was \$0.042 per million input tokens (other vendors' prices in Appendix H.4); every token count is a billed count. The runs cost \$4.26 (Run 1), \$1.63 (Run 2), \$2.80 and \$0.62 (Run 3's main and Jev 1.14 blocks), \$4.19 (Run 4), \$2.98 (Run 5, seven models plus Jev 1.14), \$0.95 (Run 6), \$3.57 (Run 7) and \$17.81 (Run 8, Jev, six other models and the chat model), against calibrated estimates of \$4.62, \$1.70, \$2.97, \$0.66, \$4.35, \$6.46, \$1.13, \$4.03 and a \$25 target. All experiments together cost \$64.013 by the spend ledger (\$42.628 before Run 7, \$46.199 before Run 8). Spend by phase, the promotion-rule code, the attacker-search loop and the open-weight scoring details are in Extended Appendix K.

\textbf{Confirmatory decision rule} (all four runs; margin m = 0.02 in Run 1 and in Run 3's H1 and H3, and 0.015 for Run 2's non-inferiority tests, Run 3's H5 and Run 4's X1; Runs 3 and 4 resample gold classes within each dataset, and Run 3's H1 also stimuli and paraphrases).

\begin{lstlisting}
for each test: d* = 20,000 cluster-bootstrap means of the paired accuracy difference
    superiority p = (1 + #{d* <= 0}) / (B + 1);  harm p = (1 + #{d* >= 0}) / (B + 1)
    non-inferiority p = (1 + #{d* <= -m}) / (B + 1)
    TOST p = max((1 + #{d* <= -m}) / (B + 1), (1 + #{d* >= +m}) / (B + 1))
    intersection-union tests: p = max over components
    pooled tests: each draw = equal-weight average of the per-dataset draws
Holm step-down over the run's family at family-wise alpha 0.05
\end{lstlisting}

\appsub{K.1}{Model versions}
\textbf{K.1 Model versions.} Every model whose output is used in this paper, with the exact identifier sent, the identifier the provider returned (where it returns one; for OpenRouter, the canonical slug and the serving provider), the dates of use (UTC, from the spend ledger and the response cache) and the ledger spend over the whole study: \$95.26 in all (\$64.01 for Runs 1 to 8 and the screen, \$11.72 for Run 9 including generation, \$19.53 for Run 10). The source of every row is in the repository's \code{reproduce/\allowbreak{}README.\allowbreak{}md}.

\begingroup\footnotesize
\renewcommand{\thetable}{K.2}
\begin{longtable}{@{}>{\raggedright\arraybackslash}p{\dimexpr 0.2231\linewidth-2.6771\tabcolsep\relax}>{\raggedright\arraybackslash}p{\dimexpr 0.1558\linewidth-1.8696\tabcolsep\relax}>{\raggedright\arraybackslash}p{\dimexpr 0.1558\linewidth-1.8696\tabcolsep\relax}>{\raggedright\arraybackslash}p{\dimexpr 0.1433\linewidth-1.7199\tabcolsep\relax}>{\raggedright\arraybackslash}p{\dimexpr 0.1254\linewidth-1.5050\tabcolsep\relax}>{\raggedright\arraybackslash}p{\dimexpr 0.1146\linewidth-1.3747\tabcolsep\relax}>{\raggedright\arraybackslash}p{\dimexpr 0.0820\linewidth-0.9840\tabcolsep\relax}@{}}
\caption{Decision models (typed APIs).}\\
\toprule
Model (paper name) & Identifier sent & Returned / pinned & Endpoint & Used in & Dates (UTC) & Spend \\
\midrule
\endfirsthead
\multicolumn{7}{@{}l}{\textit{Table K.2 (continued)}}\\
\toprule
Model (paper name) & Identifier sent & Returned / pinned & Endpoint & Used in & Dates (UTC) & Spend \\
\midrule
\endhead
\bottomrule
\endlastfoot
Jev 1.13 (TypeSafe) & \code{jev-\allowbreak{}1.\allowbreak{}13.\allowbreak{}0} & \code{jev-\allowbreak{}1.\allowbreak{}13.\allowbreak{}0} (checked on every response) & \code{https:\allowbreak{}/\allowbreak{}/\allowbreak{}api.\allowbreak{}typesafe.\allowbreak{}ai/\allowbreak{}v1/\allowbreak{}systemone} & development screen, Runs 1 to 4, 6 to 9 (reference model of Runs 5, 8, 10) & 2026-11-08 to 2026-11-12 & \$56.55 \\
Jev 1.14 (TypeSafe) & \code{jev-\allowbreak{}1.\allowbreak{}14.\allowbreak{}0} & \code{jev-\allowbreak{}1.\allowbreak{}14.\allowbreak{}0} (checked on every response) & \code{https:\allowbreak{}/\allowbreak{}/\allowbreak{}api.\allowbreak{}typesafe.\allowbreak{}ai/\allowbreak{}v1/\allowbreak{}systemone} & screen arms (not used in the paper), Run 3 secondary block, Run 5 & 2026-11-08 to 2026-11-10 & \$1.39 \\
Jev 1.14 (via OpenRouter, Run 10) & \code{typesafe/\allowbreak{}jev-\allowbreak{}1.\allowbreak{}14} & pinned and returned \code{typesafe/\allowbreak{}jev-\allowbreak{}1.\allowbreak{}14-\allowbreak{}20261014}, provider TypeSafe & \code{https:\allowbreak{}/\allowbreak{}/\allowbreak{}openrouter.\allowbreak{}ai/\allowbreak{}api/\allowbreak{}v1/\allowbreak{}systemone} & Run 10 (TypeSafe's API was down) & 2026-11-11 & \$4.74 \\
Perplexity Decider v1.2 & \code{pplx-\allowbreak{}decider-\allowbreak{}v1.\allowbreak{}2} & \code{pplx-\allowbreak{}decider-\allowbreak{}v1.\allowbreak{}2} (the API echoes the name sent; no dated snapshot exists, so a silent update cannot be detected) & \code{https:\allowbreak{}/\allowbreak{}/\allowbreak{}api.\allowbreak{}perplexity.\allowbreak{}ai/\allowbreak{}v1/\allowbreak{}decisions} & Runs 5, 8, 10 & 2026-11-10 to 2026-11-11 & \$2.13 \\
Perplexity Decider Flash (v1.2) & \code{pplx-\allowbreak{}decider-\allowbreak{}flash-\allowbreak{}v1.\allowbreak{}2} & \code{pplx-\allowbreak{}decider-\allowbreak{}flash-\allowbreak{}v1.\allowbreak{}2} (echoed, as above) & \code{https:\allowbreak{}/\allowbreak{}/\allowbreak{}api.\allowbreak{}perplexity.\allowbreak{}ai/\allowbreak{}v1/\allowbreak{}decisions} & Runs 5, 8, 10 & 2026-11-10 to 2026-11-11 & \$1.07 \\
OpenAI Decisions (gpt-6-luna) & \code{gpt-\allowbreak{}6-\allowbreak{}luna} on the Decisions endpoint & \code{gpt-\allowbreak{}6-\allowbreak{}luna} (the only snapshot; the response model must start with the id sent) & OpenAI Decisions API & Runs 5, 8, 10 & 2026-11-10 to 2026-11-11 & \$10.69 \\
Microsoft-Decision-1 & \code{microsoft/\allowbreak{}microsoft-\allowbreak{}decision-\allowbreak{}1} & pinned and returned \code{microsoft/\allowbreak{}microsoft-\allowbreak{}decision-\allowbreak{}1-\allowbreak{}20261009}, provider Azure & \code{https:\allowbreak{}/\allowbreak{}/\allowbreak{}openrouter.\allowbreak{}ai/\allowbreak{}api/\allowbreak{}v1/\allowbreak{}systemone} & Runs 5, 8, 10 & 2026-11-10 to 2026-11-11 & \$4.12 \\
Liquid d1-ultra & \code{liquid/\allowbreak{}d1-\allowbreak{}ultra} & pinned and returned \code{liquid/\allowbreak{}d1-\allowbreak{}ultra-\allowbreak{}20261015}, provider Liquid & \code{https:\allowbreak{}/\allowbreak{}/\allowbreak{}openrouter.\allowbreak{}ai/\allowbreak{}api/\allowbreak{}v1/\allowbreak{}systemone} & Runs 5, 8, 10 & 2026-11-10 to 2026-11-11 & \$3.74 \\
Kev-4B (open-weights Jev clone, Apache-2.0) & \code{jaredpalmer/\allowbreak{}kev-\allowbreak{}4b} & pinned and returned \code{jaredpalmer/\allowbreak{}kev-\allowbreak{}4b-\allowbreak{}20260924}; served by Phala in Runs 5 and 8 (2026-11-10) and by SiliconFlow in Run 10 (2026-11-11) & \code{https:\allowbreak{}/\allowbreak{}/\allowbreak{}openrouter.\allowbreak{}ai/\allowbreak{}api/\allowbreak{}v1/\allowbreak{}systemone} & Runs 5, 8, 10 & 2026-11-10 to 2026-11-11 & \$3.83 \\
Upstage Solar Decide (dropped) & \code{upstage/\allowbreak{}solar-\allowbreak{}decide} & pinned and returned \code{upstage/\allowbreak{}solar-\allowbreak{}decide-\allowbreak{}20260928}, provider Upstage; 3 responses cached, 17 of 20 pilot requests failed with HTTP 422 & \code{https:\allowbreak{}/\allowbreak{}/\allowbreak{}openrouter.\allowbreak{}ai/\allowbreak{}api/\allowbreak{}v1/\allowbreak{}systemone} & Run 5 pilot only; left out of Runs 8 and 10 & 2026-11-10 & \$0.0001 \\
\end{longtable}
\endgroup

Registered but never used for a result, with no response cached: Cloudflare Clef-flash (\code{cloudflare/\allowbreak{}clef-\allowbreak{}flash}, failed its probe with HTTP 400 \code{unsupported\_\allowbreak{}api}), Inception Mercury Decide (\code{inception/\allowbreak{}mercury-\allowbreak{}decide-\allowbreak{}20260930}, not in OpenRouter's decisions catalogue), Nace.AI Drex v2.0 (\code{nace-\allowbreak{}ai/\allowbreak{}drex-\allowbreak{}v2.\allowbreak{}0-\allowbreak{}20261009}, fallback not needed) and Cloudflare Clef (full bundle only).

\begingroup\footnotesize
\renewcommand{\thetable}{K.3}
\begin{longtable}{@{}>{\raggedright\arraybackslash}p{\dimexpr 0.1780\linewidth-2.1359\tabcolsep\relax}>{\raggedright\arraybackslash}p{\dimexpr 0.1496\linewidth-1.7957\tabcolsep\relax}>{\raggedright\arraybackslash}p{\dimexpr 0.1213\linewidth-1.4555\tabcolsep\relax}>{\raggedright\arraybackslash}p{\dimexpr 0.1872\linewidth-2.2465\tabcolsep\relax}>{\raggedright\arraybackslash}p{\dimexpr 0.1355\linewidth-1.6256\tabcolsep\relax}>{\raggedright\arraybackslash}p{\dimexpr 0.1496\linewidth-1.7957\tabcolsep\relax}>{\raggedright\arraybackslash}p{\dimexpr 0.0788\linewidth-0.9451\tabcolsep\relax}@{}}
\caption{The chat model read through log-probabilities.}\\
\toprule
Model & Identifier sent & Returned & Settings & Used in & Dates (UTC) & Spend \\
\midrule
\endfirsthead
\multicolumn{7}{@{}l}{\textit{Table K.3 (continued)}}\\
\toprule
Model & Identifier sent & Returned & Settings & Used in & Dates (UTC) & Spend \\
\midrule
\endhead
\bottomrule
\endlastfoot
OpenAI gpt-6-luna (chat completions) & \code{gpt-\allowbreak{}6-\allowbreak{}luna} & \code{gpt-\allowbreak{}6-\allowbreak{}luna} & options as letters; \code{max\_\allowbreak{}completion\_\allowbreak{}tokens} per Run 8 addendum 3, \code{temperature} 0, \code{seed} 0, \code{logprobs} true, \code{top\_\allowbreak{}logprobs} 5, \code{reasoning\_\allowbreak{}effort} ``none'' & Run 8 (MetaTool only) & 2026-11-10 & \$1.13 \\
\end{longtable}
\endgroup

\begingroup\footnotesize
\renewcommand{\thetable}{K.4}
\begin{longtable}{@{}>{\raggedright\arraybackslash}p{\dimexpr 0.1703\linewidth-1.7032\tabcolsep\relax}>{\raggedright\arraybackslash}p{\dimexpr 0.1669\linewidth-1.6692\tabcolsep\relax}>{\raggedright\arraybackslash}p{\dimexpr 0.1876\linewidth-1.8760\tabcolsep\relax}>{\raggedright\arraybackslash}p{\dimexpr 0.1828\linewidth-1.8285\tabcolsep\relax}>{\raggedright\arraybackslash}p{\dimexpr 0.1600\linewidth-1.6000\tabcolsep\relax}>{\raggedright\arraybackslash}p{\dimexpr 0.1323\linewidth-1.3231\tabcolsep\relax}@{}}
\caption{Language models that wrote or labelled data (Run 9's private datasets and blind definitions), all through OpenRouter chat completions with \code{provider.\allowbreak{}data\_\allowbreak{}collection =\allowbreak{} "deny"}.}\\
\toprule
Role & Identifier sent & Returned & Settings & Dates (UTC) & Spend \\
\midrule
\endfirsthead
\multicolumn{6}{@{}l}{\textit{Table K.4 (continued)}}\\
\toprule
Role & Identifier sent & Returned & Settings & Dates (UTC) & Spend \\
\midrule
\endhead
\bottomrule
\endlastfoot
Generator A (items of the four private datasets) & \code{anthropic/\allowbreak{}claude-\allowbreak{}haiku-\allowbreak{}5.\allowbreak{}5} (Claude Haiku 5.5) & \code{anthropic/\allowbreak{}claude-\allowbreak{}haiku-\allowbreak{}5.\allowbreak{}5}, canonical \code{anthropic/\allowbreak{}claude-\allowbreak{}haiku-\allowbreak{}5.\allowbreak{}5-\allowbreak{}20261007} per the OpenRouter catalogue, on all 12,346 of its items in the private item file & reasoning disabled (\code{reasoning:\allowbreak{} \{enabled:\allowbreak{} false\}}), default temperature & pilot and full generation, 2026-11-11 03:59 to 05:56 & \$0.87 \\
Generator B (items of the four private datasets) & \code{google/\allowbreak{}gemini-\allowbreak{}3.\allowbreak{}1-\allowbreak{}flash-\allowbreak{}lite} (Gemini 3.1 Flash-Lite) & \code{google/\allowbreak{}gemini-\allowbreak{}3.\allowbreak{}1-\allowbreak{}flash-\allowbreak{}lite}, canonical \code{google/\allowbreak{}gemini-\allowbreak{}3.\allowbreak{}1-\allowbreak{}flash-\allowbreak{}lite-\allowbreak{}20260507} per the catalogue, on all 12,419 of its items in the private item file & \code{temperature} 1.0, reasoning effort ``minimal'' & 2026-11-11 03:59 to 05:56 & \$1.51 \\
Validator V (blind label for every item; third family) & \code{mistralai/\allowbreak{}mistral-\allowbreak{}large-\allowbreak{}2512} (Mistral Large) & not stored per item & \code{temperature} 0.0 & 2026-11-11 03:59 to 05:56 & \$3.47 \\
Blind one-line definitions for Run 9 (CLINC150 validation split and three private sets; label names only) & \code{mistralai/\allowbreak{}mistral-\allowbreak{}large-\allowbreak{}2512} & not stored & at most 18 words, no other label's name & 2026-11-11 04:17 to 04:22 & \$0.020 \\
Tie-breaker C (only where V disagreed with the intended label) & the other generator family (A or B above) & as above & as above & as above & in the generator rows \\
\end{longtable}
\endgroup

The development screen's option components, the Banking77 blind definitions and MetaTool paraphrased definitions, the five persona and five politeness prefixes, Run 3's blind definitions and new instruction paraphrases, the screen's ``generated descriptions'', the red-team attack paragraphs, and Run 9's label sets, definitions and generation briefs were written with Claude models through Claude Code; the exact model versions were not recorded.

\begingroup\footnotesize
\renewcommand{\thetable}{K.5}
\begin{longtable}{@{}>{\raggedright\arraybackslash}p{\dimexpr 0.2290\linewidth-1.3739\tabcolsep\relax}>{\raggedright\arraybackslash}p{\dimexpr 0.2218\linewidth-1.3310\tabcolsep\relax}>{\raggedright\arraybackslash}p{\dimexpr 0.3203\linewidth-1.9217\tabcolsep\relax}>{\raggedright\arraybackslash}p{\dimexpr 0.2289\linewidth-1.3734\tabcolsep\relax}@{}}
\caption{Local models (no API).}\\
\toprule
Model & Identifier & Revision & Used for \\
\midrule
\endfirsthead
\multicolumn{4}{@{}l}{\textit{Table K.5 (continued)}}\\
\toprule
Model & Identifier & Revision & Used for \\
\midrule
\endhead
\bottomrule
\endlastfoot
bge-small embeddings & \code{BAAI/\allowbreak{}bge-\allowbreak{}small-\allowbreak{}en-\allowbreak{}v1.\allowbreak{}5} through fastembed 0.9.1 (ONNX export \code{Qdrant/\allowbreak{}bge-\allowbreak{}small-\allowbreak{}en-\allowbreak{}v1.\allowbreak{}5-\allowbreak{}onnx-\allowbreak{}Q}) & fastembed cache snapshot \code{aa8f\allowbreak{}8b06\allowbreak{}0edb\allowbreak{}00e0\allowbreak{}3bfd\allowbreak{}d088\allowbreak{}13a2\allowbreak{}9499\allowbreak{}46c8\allowbreak{}ba55\allowbreak{}}; the Hugging Face repo snapshot in the local cache is \code{5c38\allowbreak{}ec7c\allowbreak{}405e\allowbreak{}c4b4\allowbreak{}4b94\allowbreak{}cc5a\allowbreak{}9bb9\allowbreak{}6e73\allowbreak{}5b38\allowbreak{}267a\allowbreak{}} (not pinned in code) & dense retrieval (cosine, top 20), the fixed vote, the logistic regression, nearest centroid; Run 9 near-duplicate checks \\
MiniLM classifier & \code{sentence-\allowbreak{}transformers/\allowbreak{}all-\allowbreak{}MiniLM-\allowbreak{}L6-\allowbreak{}v2}, fine-tuned for at most 5 epochs (also the SetFit baseline) & local cache snapshot \code{1110\allowbreak{}a243\allowbreak{}fdf4\allowbreak{}706b\allowbreak{}3f48\allowbreak{}f1d9\allowbreak{}5db1\allowbreak{}a4f5\allowbreak{}529b\allowbreak{}4d41\allowbreak{}} (not pinned in code) & the fine-tuned local classifier; SetFit comparator \\
TF-IDF retriever, logistic regression, nearest centroid, nearest-neighbour vote & scikit-learn 1.9.1 (TF-IDF word 1-2-grams, sublinear tf; LR with C chosen from \{0.1, 1, 10, 100, 1000\}) & no weights & retrieval and the local models of Section 5 \\
\end{longtable}
\endgroup

\begingroup\footnotesize
\renewcommand{\thetable}{K.6}
\begin{longtable}{@{}>{\raggedright\arraybackslash}p{\dimexpr 0.2732\linewidth-1.0930\tabcolsep\relax}>{\raggedright\arraybackslash}p{\dimexpr 0.3052\linewidth-1.2209\tabcolsep\relax}>{\raggedright\arraybackslash}p{\dimexpr 0.4215\linewidth-1.6861\tabcolsep\relax}@{}}
\caption{Open-weight models of the scoping pilot (Appendix H; CPU, no API).}\\
\toprule
Model & Identifier & Revision \\
\midrule
\endfirsthead
\multicolumn{3}{@{}l}{\textit{Table K.6 (continued)}}\\
\toprule
Model & Identifier & Revision \\
\midrule
\endhead
\bottomrule
\endlastfoot
Qwen2.5-0.5B-Instruct & \code{Qwen/\allowbreak{}Qwen2.\allowbreak{}5-\allowbreak{}0.\allowbreak{}5B-\allowbreak{}Instruct} & local cache snapshot \code{7ae5\allowbreak{}5760\allowbreak{}4adf\allowbreak{}67be\allowbreak{}5041\allowbreak{}7f59\allowbreak{}c2c2\allowbreak{}f167\allowbreak{}def9\allowbreak{}a775\allowbreak{}} \\
Qwen2.5-1.5B-Instruct & \code{Qwen/\allowbreak{}Qwen2.\allowbreak{}5-\allowbreak{}1.\allowbreak{}5B-\allowbreak{}Instruct} & local cache snapshot \code{989a\allowbreak{}a798\allowbreak{}0e4c\allowbreak{}f806\allowbreak{}f80c\allowbreak{}7fef\allowbreak{}2b1a\allowbreak{}db7b\allowbreak{}c71a\allowbreak{}a306\allowbreak{}} \\
Kev-0.8B & \code{jaredpalmer/\allowbreak{}kev-\allowbreak{}0.\allowbreak{}8b} on base \code{Qwen/\allowbreak{}Qwen3.\allowbreak{}5-\allowbreak{}0.\allowbreak{}8B-\allowbreak{}Base} & pinned \code{bf75\allowbreak{}a6a8\allowbreak{}848e\allowbreak{}a696\allowbreak{}0ff2\allowbreak{}ed10\allowbreak{}8d9e\allowbreak{}d44c\allowbreak{}2941\allowbreak{}174f\allowbreak{}}; base pinned \code{dc7c\allowbreak{}dfe2\allowbreak{}ee41\allowbreak{}54fa\allowbreak{}7e30\allowbreak{}f5b5\allowbreak{}1ca4\allowbreak{}1bfa\allowbreak{}4017\allowbreak{}4e68\allowbreak{}} \\
Laya & \code{convaiinnovations/\allowbreak{}laya} & pinned \code{85cc\allowbreak{}74dd\allowbreak{}ae61\allowbreak{}0907\allowbreak{}7ebe\allowbreak{}2853\allowbreak{}a9b3\allowbreak{}077f\allowbreak{}b660\allowbreak{}6a78\allowbreak{}} \\
OpenJev-1.7B & \code{alanhuangya/\allowbreak{}OpenJev-\allowbreak{}1.\allowbreak{}7B} & pinned \code{7418\allowbreak{}ae8b\allowbreak{}6e3b\allowbreak{}6a65\allowbreak{}07a8\allowbreak{}a587\allowbreak{}29ae\allowbreak{}491c\allowbreak{}d1c7\allowbreak{}32d9\allowbreak{}} \\
\end{longtable}
\endgroup

\section{Sources of the main figures and tables}

Paths are relative to \code{experiments/\allowbreak{}results/\allowbreak{}} unless noted; figure scripts read result files only and make no API call. Extended Appendix L maps every other table, figure and number to its files.

\begingroup\footnotesize
\begin{longtable}{@{}>{\raggedright\arraybackslash}p{\dimexpr 0.2450\linewidth-0.4899\tabcolsep\relax}>{\raggedright\arraybackslash}p{\dimexpr 0.7550\linewidth-1.5101\tabcolsep\relax}@{}}
\toprule
Item & Source \\
\midrule
\endfirsthead
\multicolumn{2}{@{}l}{\textit{Table (continued)}}\\
\toprule
Item & Source \\
\midrule
\endhead
\bottomrule
\endlastfoot
Figure 1 & \code{experiments/\allowbreak{}figures/\allowbreak{}visuals/\allowbreak{}fig1\_\allowbreak{}api\_\allowbreak{}v2} (\code{.\allowbreak{}png}, \code{.\allowbreak{}pdf}), drawn by \code{experiments/\allowbreak{}plot\_\allowbreak{}fig1\_\allowbreak{}options.\allowbreak{}py} (it replaces v26's Figures 1 and 2, \code{fig1\_\allowbreak{}main\_\allowbreak{}v11} and \code{v6\_\allowbreak{}api\_\allowbreak{}flow}). Panel (b) quotes Run 8 on Jev 1.13, CLINC150 and LEDGAR fresh items at 100 per class (Appendix U.2): one fixed example per option, the same for every text, +1.65 over the bare call, measured directly (\code{run8\_\allowbreak{}secondary.\allowbreak{}csv}, ``baseline: fixed1 - bare''; +4.6 on CLINC150 and \ensuremath{-}1.3 on LEDGAR, \code{run8\_\allowbreak{}descriptive.\allowbreak{}csv}, rows fixed1 and bare); label-matched random examples +3.10 (``baseline: lm100 - bare''); nearest examples +4.67 (``label budget (pooled): prompt100 - bare''). Descriptions: Run 9 names experiment, a description on two-word names, +10.7 (MetaTool) and +25.0 (OpsDeck) (\code{run9\_\allowbreak{}names\_\allowbreak{}experiment.\allowbreak{}csv}). Wording: within \ensuremath{\pm}2 on Jev (Run 3) and on 4 of 7 other models (Run 5; Section 3). Risk: 30\% confusable pool noise against the clean pool, mean of six other APIs on fresh items, \ensuremath{-}12.6 (\code{r18\_\allowbreak{}random\_\allowbreak{}effects.\allowbreak{}csv}; Jev 1.13 \ensuremath{-}7.0, \code{run8\_\allowbreak{}primary.\allowbreak{}csv} N1) \\
Figure 3 & \code{experiments/\allowbreak{}figures/\allowbreak{}visuals/\allowbreak{}fig\_\allowbreak{}pool\_\allowbreak{}size\_\allowbreak{}v1} (\code{.\allowbreak{}png}, \code{.\allowbreak{}pdf}, \code{\_\allowbreak{}values.\allowbreak{}csv}), drawn by \code{experiments/\allowbreak{}plot\_\allowbreak{}pool\_\allowbreak{}size.\allowbreak{}py} from \code{run8\_\allowbreak{}secondary.\allowbreak{}csv}, rows ``curve clinc\us{}v: \ldots{}'' and ``curve ledgar: \ldots{}'' (prompt, bare \ensuremath{\times} classifier, classifier and nearest centroid at 5, 20 and 100 per class) and the bare-call accuracies in \code{run8\_\allowbreak{}descriptive.\allowbreak{}csv} \\
Figures A.1, M.1, 2, F.1, I.13 and A.2, and Figure P.1 of the extended appendix & \code{experiments/\allowbreak{}figures/\allowbreak{}visuals/\allowbreak{}} (\code{fig0\_\allowbreak{}hero\_\allowbreak{}v9}, \code{v1\_\allowbreak{}anatomy\_\allowbreak{}v9}, \code{v3\_\allowbreak{}worked\_\allowbreak{}example}, \code{v4\_\allowbreak{}pipelines\_\allowbreak{}v8}, \code{fig26\_\allowbreak{}mechanism\_\allowbreak{}fit\_\allowbreak{}v5}, \code{v5\_\allowbreak{}rules\_\allowbreak{}v8}, \code{v2\_\allowbreak{}protocol\_\allowbreak{}v5}), each with \code{\textless{}name\textgreater{}\_\allowbreak{}values.\allowbreak{}csv} listing every plotted number with its source file and row; these figures are generated from \code{visuals\_\allowbreak{}data\_\allowbreak{}v5.\allowbreak{}csv} by \code{experiments/\allowbreak{}build\_\allowbreak{}visuals\_\allowbreak{}data\_\allowbreak{}v5.\allowbreak{}py} and \code{experiments/\allowbreak{}plot\_\allowbreak{}*\_\allowbreak{}v5.\allowbreak{}py} (Figure A.1 by \code{experiments/\allowbreak{}plot\_\allowbreak{}fig0\_\allowbreak{}hero\_\allowbreak{}v9.\allowbreak{}py}, which draws the two fresh-item tests, hollow, in place of the pooled diamond; Figure F.1 (the main text's Figure 4 in v26) by \code{experiments/\allowbreak{}plot\_\allowbreak{}v4\_\allowbreak{}pipelines\_\allowbreak{}v8.\allowbreak{}py}, whose band gives the two fresh-item tests; Figure M.1 by \code{experiments/\allowbreak{}plot\_\allowbreak{}v1\_\allowbreak{}anatomy\_\allowbreak{}v9.\allowbreak{}py}, which labels the share chips ``Don't 2'', the wrong-label chip ``Don't 3'' and the description chip ``Do 2 \ensuremath{\cdot} helps on unclear names''), except Figure A.2, which \code{experiments/\allowbreak{}plot\_\allowbreak{}rules\_\allowbreak{}v8.\allowbreak{}py} draws in Table 2's numbering from \code{run8\_\allowbreak{}primary.\allowbreak{}csv} and \code{visuals\_\allowbreak{}data\_\allowbreak{}v5.\allowbreak{}csv} plus \code{run5\_\allowbreak{}transfer.\allowbreak{}csv}, \code{run5\_\allowbreak{}primary.\allowbreak{}csv}, \code{run5\_\allowbreak{}run\_\allowbreak{}meta.\allowbreak{}json}, \code{run6\_\allowbreak{}primary.\allowbreak{}csv}, \code{run6\_\allowbreak{}secondary.\allowbreak{}csv}, \code{r15\_\allowbreak{}pool100.\allowbreak{}csv}, \code{r15\_\allowbreak{}heldout\_\allowbreak{}pairs.\allowbreak{}csv} and \code{prodtransfer\_\allowbreak{}r6\_\allowbreak{}text.\allowbreak{}csv}. Figure A.1(b) reads \code{r15\_\allowbreak{}pool100.\allowbreak{}csv}, and Figures A.1(b) and F.1 read Run 8's fresh-item test from \code{run8\_\allowbreak{}primary.\allowbreak{}csv} \\
Figure I.13, Table I.13 & \code{r13\_\allowbreak{}a1\_\allowbreak{}mechanism\_\allowbreak{}fit.\allowbreak{}csv}, \code{r14\_\allowbreak{}x3\_\allowbreak{}mechanism\_\allowbreak{}prediction.\allowbreak{}csv}; \code{experiments/\allowbreak{}analysis\_\allowbreak{}r13.\allowbreak{}py}, \code{analysis\_\allowbreak{}r14.\allowbreak{}py} \\
Tables A.2 and P.1, Appendix I & \code{gate\_\allowbreak{}primary.\allowbreak{}csv}, \code{gate2\_\allowbreak{}primary.\allowbreak{}csv}, \code{run3\_\allowbreak{}primary.\allowbreak{}csv}, \code{run3\_\allowbreak{}secondary.\allowbreak{}csv}, \code{run4\_\allowbreak{}primary.\allowbreak{}csv}, \code{run4\_\allowbreak{}secondary.\allowbreak{}csv}, \code{r12\_\allowbreak{}a1\_\allowbreak{}toxicchat\_\allowbreak{}recall.\allowbreak{}csv}, \code{r14\_\allowbreak{}holm29.\allowbreak{}csv}; for Appendices O and E.9 also \code{run4\_\allowbreak{}*.\allowbreak{}csv} \\
Post-hoc re-analyses (Table A.2, Sections 3 to 5, Appendices I.13, Appendices E.9d, E.10, I.7, I.14, P.4) & \code{r15\_\allowbreak{}ts\_\allowbreak{}prior\_\allowbreak{}correction.\allowbreak{}csv}, \code{r15\_\allowbreak{}heldout\_\allowbreak{}size.\allowbreak{}csv}, \code{r15\_\allowbreak{}heldout\_\allowbreak{}pairs.\allowbreak{}csv}, \code{r15\_\allowbreak{}pool100.\allowbreak{}csv}, \code{r15\_\allowbreak{}classrob\_\allowbreak{}run1.\allowbreak{}csv}, \code{r15\_\allowbreak{}dm\_\allowbreak{}vs\_\allowbreak{}mechanism.\allowbreak{}csv}, \code{r15\_\allowbreak{}test\_\allowbreak{}sizes.\allowbreak{}csv}; \code{experiments/\allowbreak{}analysis\_\allowbreak{}r15.\allowbreak{}py} \\
Re-analysis 1 (Sections 2.3, 4 and 6: stress test, A/A size, wrong-label dose by scale and margin, normalised sensitivity, d1-ultra) & \code{r16\_\allowbreak{}e1\_\allowbreak{}stress\_\allowbreak{}null.\allowbreak{}csv}, \code{r16\_\allowbreak{}e1\_\allowbreak{}aa\_\allowbreak{}size.\allowbreak{}csv}, \code{r16\_\allowbreak{}e1\_\allowbreak{}aa\_\allowbreak{}observed.\allowbreak{}csv}, \code{r16\_\allowbreak{}e1\_\allowbreak{}aa\_\allowbreak{}inventory.\allowbreak{}csv}, \code{r16\_\allowbreak{}e3a\_\allowbreak{}dose.\allowbreak{}csv}, \code{r16\_\allowbreak{}e3a\_\allowbreak{}strata.\allowbreak{}csv}, \code{r16\_\allowbreak{}e3a\_\allowbreak{}normalised.\allowbreak{}csv}, \code{r16\_\allowbreak{}e3a\_\allowbreak{}d1ultra.\allowbreak{}csv}, \code{r16\_\allowbreak{}*\_\allowbreak{}checks.\allowbreak{}csv}; \code{experiments/\allowbreak{}analysis\_\allowbreak{}r16.\allowbreak{}py}; note \code{reports/\allowbreak{}audit\_\allowbreak{}notes/\allowbreak{}round16/\allowbreak{}free\_\allowbreak{}analyses.\allowbreak{}md} \\
Section 6, Tables 5 and H.4, Figure H.1, Appendix H.4, Extended Appendix S.1 & \code{run5\_\allowbreak{}primary.\allowbreak{}csv} / \code{.\allowbreak{}md}, \code{run5\_\allowbreak{}secondary.\allowbreak{}csv}, \code{run5\_\allowbreak{}descriptive.\allowbreak{}csv}, \code{run5\_\allowbreak{}transfer.\allowbreak{}csv}, \code{run5\_\allowbreak{}run\_\allowbreak{}meta.\allowbreak{}json}, \code{run5\_\allowbreak{}launch\_\allowbreak{}summary.\allowbreak{}json}, \code{run5\_\allowbreak{}worker\_\allowbreak{}*.\allowbreak{}log}; plan \code{reports/\allowbreak{}preregistration\_\allowbreak{}run5.\allowbreak{}md}, manifest \code{registry\_\allowbreak{}post/\allowbreak{}freeze/\allowbreak{}manifest\_\allowbreak{}run5\_\allowbreak{}20261110T010209Z.\allowbreak{}json} \\
Figure H.1 (main-text Figure 5 up to v24) & \code{experiments/\allowbreak{}figures/\allowbreak{}visuals/\allowbreak{}fig\_\allowbreak{}crossmodel\_\allowbreak{}v4} (\code{.\allowbreak{}png}, \code{.\allowbreak{}pdf}, \code{\_\allowbreak{}values.\allowbreak{}csv}), drawn from \code{run5\_\allowbreak{}primary.\allowbreak{}csv} by \code{experiments/\allowbreak{}plot\_\allowbreak{}crossmodel\_\allowbreak{}v4.\allowbreak{}py} (v3 without the definitions panel) \\
Run 9 (Section 7, Table 6's Jev 1.13 column, Figure Y.4, the names experiment in Section 4.1, Tables 2, 4 and 7, Appendix Y) & \code{run9\_\allowbreak{}effects.\allowbreak{}csv}, \code{run9\_\allowbreak{}open\_\allowbreak{}vs\_\allowbreak{}closed.\allowbreak{}csv}, \code{run9\_\allowbreak{}adjusted.\allowbreak{}csv}, \code{run9\_\allowbreak{}tools\_\allowbreak{}decomposition.\allowbreak{}csv}, \code{run9\_\allowbreak{}names\_\allowbreak{}experiment.\allowbreak{}csv}, \code{run9\_\allowbreak{}bare\_\allowbreak{}gap.\allowbreak{}csv}, \code{run9\_\allowbreak{}pairs.\allowbreak{}csv}, \code{run9\_\allowbreak{}rank.\allowbreak{}json}; Figure Y.4 is \code{experiments/\allowbreak{}figures/\allowbreak{}run9/\allowbreak{}run9\_\allowbreak{}effect\_\allowbreak{}vs\_\allowbreak{}bare.\allowbreak{}png} (alt text in \code{alt\_\allowbreak{}text.\allowbreak{}json}); made by \code{src/\allowbreak{}jev\_\allowbreak{}prompt\_\allowbreak{}lab/\allowbreak{}stages/\allowbreak{}run9.\allowbreak{}py} and \code{run9\_\allowbreak{}analysis.\allowbreak{}py}; plan \code{reports/\allowbreak{}preregistration\_\allowbreak{}run9.\allowbreak{}md}, design \code{reports/\allowbreak{}run9/\allowbreak{}design.\allowbreak{}md}, generation report \code{experiments/\allowbreak{}run9\_\allowbreak{}closed/\allowbreak{}full/\allowbreak{}qc\_\allowbreak{}report.\allowbreak{}json} (private folder) \\
Figure 4 (Figure 5 in v25 and v26, Figure 6 up to v24) & \code{experiments/\allowbreak{}figures/\allowbreak{}run10/\allowbreak{}run10\_\allowbreak{}paper} (\code{.\allowbreak{}png}, \code{.\allowbreak{}pdf}), drawn by \code{experiments/\allowbreak{}run10\_\allowbreak{}paper\_\allowbreak{}numbers.\allowbreak{}py} from \code{run10\_\allowbreak{}open\_\allowbreak{}vs\_\allowbreak{}closed.\allowbreak{}csv} and, for Jev 1.13, \code{run9\_\allowbreak{}open\_\allowbreak{}vs\_\allowbreak{}closed.\allowbreak{}csv}; the same script reads \code{run10\_\allowbreak{}bare\_\allowbreak{}gap.\allowbreak{}csv}, \code{run10\_\allowbreak{}rank.\allowbreak{}csv}, \code{run10\_\allowbreak{}effects.\allowbreak{}csv} and \code{run10\_\allowbreak{}completeness.\allowbreak{}csv} and writes \code{reports/\allowbreak{}run10/\allowbreak{}paper\_\allowbreak{}v24\_\allowbreak{}numbers.\allowbreak{}md}, the source of every Run 10 number in the paper \\
Run 10 (Section 7, Table 6, Tables 2 and 3, Appendix Y.7) & \code{run10\_\allowbreak{}effects.\allowbreak{}csv}, \code{run10\_\allowbreak{}open\_\allowbreak{}vs\_\allowbreak{}closed.\allowbreak{}csv}, \code{run10\_\allowbreak{}verdict.\allowbreak{}csv}, \code{run10\_\allowbreak{}rank.\allowbreak{}csv}, \code{run10\_\allowbreak{}bare\_\allowbreak{}gap.\allowbreak{}csv}, \code{run10\_\allowbreak{}names\_\allowbreak{}experiment.\allowbreak{}csv}, \code{run10\_\allowbreak{}completeness.\allowbreak{}csv}, \code{run10\_\allowbreak{}failures.\allowbreak{}jsonl}, \code{run10\_\allowbreak{}spend.\allowbreak{}jsonl}, \code{run10\_\allowbreak{}run.\allowbreak{}log}, \code{run10\_\allowbreak{}run\_\allowbreak{}luna.\allowbreak{}log}, \code{run10\_\allowbreak{}run\_\allowbreak{}jev114.\allowbreak{}log}; made by \code{src/\allowbreak{}jev\_\allowbreak{}prompt\_\allowbreak{}lab/\allowbreak{}stages/\allowbreak{}run10.\allowbreak{}py} and \code{run10\_\allowbreak{}analysis.\allowbreak{}py}; plan \code{reports/\allowbreak{}preregistration\_\allowbreak{}run10.\allowbreak{}md} (with addendum 1); manifest \code{registry\_\allowbreak{}post/\allowbreak{}run10/\allowbreak{}run10\_\allowbreak{}manifest.\allowbreak{}json}, earlier freezes \code{run10\_\allowbreak{}manifest\_\allowbreak{}freeze1.\allowbreak{}json} to \code{run10\_\allowbreak{}manifest\_\allowbreak{}freeze4.\allowbreak{}json}; the long-text checks of Section 7 are cache checks recorded in \code{reports/\allowbreak{}run10/\allowbreak{}paper\_\allowbreak{}v24\_\allowbreak{}numbers.\allowbreak{}md} \\
Table 4 & \code{gate\_\allowbreak{}primary.\allowbreak{}csv}, \code{run3\_\allowbreak{}primary.\allowbreak{}csv}, \code{run4\_\allowbreak{}primary.\allowbreak{}csv}, \code{run4\_\allowbreak{}secondary.\allowbreak{}csv}, \code{run6\_\allowbreak{}primary.\allowbreak{}csv}, \code{run6\_\allowbreak{}secondary.\allowbreak{}csv}, \code{run7\_\allowbreak{}primary.\allowbreak{}csv}, \code{run8\_\allowbreak{}primary.\allowbreak{}csv}, \code{run8\_\allowbreak{}secondary.\allowbreak{}csv}, \code{gate2\_\allowbreak{}primary.\allowbreak{}csv} \\
Section 4 (placement test), Table A.2 (Do 1, Don't 4), Extended Appendix S.2 & \code{run6\_\allowbreak{}primary.\allowbreak{}csv} / \code{.\allowbreak{}md}, \code{run6\_\allowbreak{}secondary.\allowbreak{}csv}, \code{run6\_\allowbreak{}descriptive.\allowbreak{}csv}, \code{run6\_\allowbreak{}run\_\allowbreak{}meta.\allowbreak{}json}, \code{run6\_\allowbreak{}run.\allowbreak{}log}; plan \code{reports/\allowbreak{}preregistration\_\allowbreak{}run6.\allowbreak{}md}, manifest \code{registry\_\allowbreak{}post/\allowbreak{}freeze/\allowbreak{}manifest\_\allowbreak{}run6\_\allowbreak{}20261110T041942Z.\allowbreak{}json} \\
Runs 7 and 8 (Sections 3 to 6, Tables 2, 4 and 5, Appendix U) & \code{run7\_\allowbreak{}primary.\allowbreak{}csv} / \code{.\allowbreak{}md}, \code{run7\_\allowbreak{}secondary.\allowbreak{}csv}, \code{run7\_\allowbreak{}descriptive.\allowbreak{}csv}, \code{run7\_\allowbreak{}run\_\allowbreak{}meta.\allowbreak{}json}, \code{run7\_\allowbreak{}run.\allowbreak{}log}; \code{run8\_\allowbreak{}primary.\allowbreak{}csv} / \code{.\allowbreak{}md}, \code{run8\_\allowbreak{}secondary.\allowbreak{}csv}, \code{run8\_\allowbreak{}transfer.\allowbreak{}csv}, \code{run8\_\allowbreak{}addendum\_\allowbreak{}transfer.\allowbreak{}csv}, \code{run8\_\allowbreak{}chat\_\allowbreak{}reading.\allowbreak{}csv}, \code{run8\_\allowbreak{}timesplit.\allowbreak{}csv}, \code{run8\_\allowbreak{}latency.\allowbreak{}csv}, \code{run8\_\allowbreak{}descriptive.\allowbreak{}csv}, \code{run8\_\allowbreak{}run\_\allowbreak{}meta.\allowbreak{}json}, \code{run8\_\allowbreak{}failures.\allowbreak{}jsonl}, \code{run8\_\allowbreak{}spend.\allowbreak{}jsonl}; plans \code{reports/\allowbreak{}preregistration\_\allowbreak{}run7.\allowbreak{}md}, \code{preregistration\_\allowbreak{}run8.\allowbreak{}md}; manifests \code{registry\_\allowbreak{}post/\allowbreak{}freeze/\allowbreak{}manifest\_\allowbreak{}run7\_\allowbreak{}20261110T113214Z.\allowbreak{}json} and, for Run 8, \code{manifest\_\allowbreak{}run8\_\allowbreak{}20261110T124304Z.\allowbreak{}json} (freeze 3), \code{\ldots{}T135530Z.\allowbreak{}json} (4), \code{\ldots{}T143103Z.\allowbreak{}json} (5) and \code{\ldots{}T175228Z.\allowbreak{}json} (6) \\
Re-analysis 2 (Sections 2.3, 3, 5 and 6, Table 5, Appendices E.10, I.7, I.11, U.3) & \code{r17\_\allowbreak{}stress\_\allowbreak{}calibration.\allowbreak{}csv}, \code{r17\_\allowbreak{}classlevel\_\allowbreak{}tests.\allowbreak{}csv}, \code{r17\_\allowbreak{}family29.\allowbreak{}csv}, \code{r17\_\allowbreak{}run5\_\allowbreak{}mixed.\allowbreak{}csv}, \code{r17\_\allowbreak{}refusal\_\allowbreak{}bound.\allowbreak{}csv}, \code{r17\_\allowbreak{}pool100\_\allowbreak{}studies.\allowbreak{}csv}, \code{r17\_\allowbreak{}pool100\_\allowbreak{}variants.\allowbreak{}csv}, \code{r17\_\allowbreak{}checks.\allowbreak{}csv}; \code{experiments/\allowbreak{}analysis\_\allowbreak{}r17.\allowbreak{}py}; note \code{reports/\allowbreak{}audit\_\allowbreak{}notes/\allowbreak{}round17/\allowbreak{}free\_\allowbreak{}analyses.\allowbreak{}md} \\
Re-analysis 3 (Sections 2.3 and 6, Figure 1, Tables 2 and 5, Appendix U.4) & \code{r18\_\allowbreak{}wcr\_\allowbreak{}tests.\allowbreak{}csv}, \code{r18\_\allowbreak{}wcr\_\allowbreak{}components.\allowbreak{}csv}, \code{r18\_\allowbreak{}random\_\allowbreak{}effects.\allowbreak{}csv}, \code{r18\_\allowbreak{}families.\allowbreak{}csv}, \code{r18\_\allowbreak{}checks.\allowbreak{}csv}; \code{experiments/\allowbreak{}analysis\_\allowbreak{}r18.\allowbreak{}py}; note \code{reports/\allowbreak{}audit\_\allowbreak{}notes/\allowbreak{}round18/\allowbreak{}free\_\allowbreak{}analyses.\allowbreak{}md} \\
Appendix M, Figure 2 & \code{paper/\allowbreak{}worked\_\allowbreak{}examples/\allowbreak{}*.\allowbreak{}json}, \code{reproduce.\allowbreak{}py}; \code{worked\_\allowbreak{}examples\_\allowbreak{}stats.\allowbreak{}csv} \\
\end{longtable}
\endgroup

\section{Worked examples}

Each example shows one approach on real confirmatory items: the exact request sent without the technique, the same request with it, and what Jev 1.13 (\code{jev-\allowbreak{}1.\allowbreak{}13.\allowbreak{}0}) answered. The files are in \code{paper/\allowbreak{}worked\_\allowbreak{}examples/\allowbreak{}}, one JSON file per example. The unit is one \textbf{request} (one item under one instruction wording). A request is \emph{fixed} when the control is wrong and the treatment right, and \emph{broken} when the reverse happens; for every approach, (fixed \ensuremath{-} broken) / requests equals the published accuracy change exactly. Each file's \code{wire\_\allowbreak{}request} was rebuilt with the lab's request builders and its hash found in the response cache, so it is byte-for-byte the body that was sent. Banking77 is CC BY 4.0 {[}\citenum{r5}{]}; MetaTool {[}\citenum{r16}{]} is MIT. Extended Appendix M has the fixed and broken counts for eight approaches (M.0), seeded random draws of fixed and broken requests for each, and four further examples: class shares, the local combination, a persona prefix and wrong labels (M.3 to M.6).

\begin{figure}[!htbp]
\centering
\includegraphics[width=\linewidth,height=0.72\textheight,keepaspectratio,alt={An annotated request. Grey box, "sent to the vendor": state "Top up on my Google Pay isn't working."; questions.q.instructions: the question, with a persona prefix (chip "Don't 1 * within +/-2") and a class-share sentence (chip "Don't 2 * hurts if shares are wrong"); questions.q.criteria: the options and their descriptions, which carry retrieved labelled examples (chip "Do 1 * helps"), a one-line description on an uninformative name such as opt 17 (chip "Do 2 * helps on unclear names") and the same examples under a wrong label (chip "Don't 3 * hurts"); questions.q2 ... q8: extra questions on the same text (chip "Don't 1 * within +/-2"; fewer tokens per answer, observed). Below the box: the returned answers.q.probabilities of the request with examples, with the correct option apple pay or google pay at 99  and top up failed at 1 . A side note: multiplying the answer by re-estimated class shares after the call (chip "Don't 2 * observed").}]{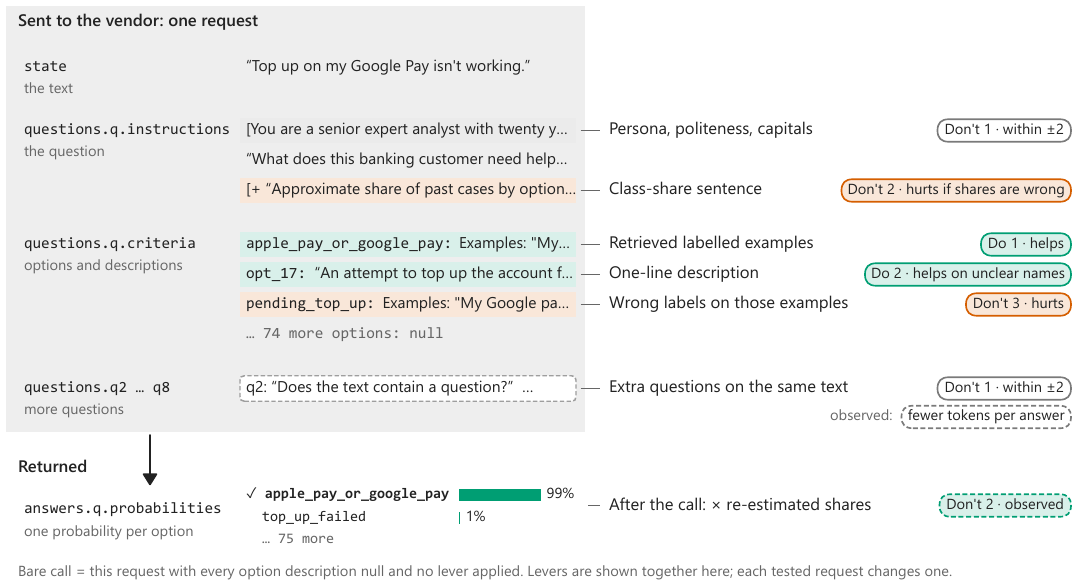}
\renewcommand{\thefigure}{M.1}
\caption{Where each lever plugs in (a main-text figure up to v22), shown on the real request for Banking77 test item 2966 (the item of Figure 2). Chips carry the rule labels of Table 2. Every lever but one edits the request; the share correction acts on the returned probabilities, on your machine. Dashed chips are observed. The full request is in Appendix M.1.}
\end{figure}

\appsub{M.1}{Retrieved labelled examples in the option descriptions}
\textbf{M.1 Retrieved labelled examples in the option descriptions} (Banking77, all 77 intents; Run 1, T1a).

\emph{What it does.} The bare request lists the 77 intents with no descriptions. The treatment retrieves the 20 pool texts most similar to the message (TF-IDF, 100 labelled training texts per intent) and writes up to three of them, under their own labels, into those options' descriptions as \code{Examples:\allowbreak{} "\ldots{}"}.

\emph{The flagship item} (Figure 2) is \code{b77full-\allowbreak{}test-\allowbreak{}2966}, paraphrase 1: \emph{``Top up on my Google Pay isn't working.''}, gold \code{apple\_\allowbreak{}pay\_\allowbreak{}or\_\allowbreak{}google\_\allowbreak{}pay}. It was drawn by a pre-declared rule from the requests present in both the T1a arm and the T4a arm (top-3 labels wrong) where the bare request was wrong, retrieval right and the relabelled request wrong: 55 of the 800 shared requests qualified, and \code{random.\allowbreak{}Random("20261109/\allowbreak{}flagship\_\allowbreak{}banking77").\allowbreak{}choice} picked index 34. The clean request of T4a is byte-identical to the T1a treatment for all 800 shared requests. Bare request (abridged; full body in \code{flagship\_\allowbreak{}banking77.\allowbreak{}json} \ensuremath{\rightarrow} \code{requests.\allowbreak{}1\_\allowbreak{}bare}):

\begin{lstlisting}
{"model": "jev-1.13.0",
 "state": "Top up on my Google Pay isn't working.",
 "questions": {"q": {"type": "choice",
   "instructions": "What does this banking customer need help with?",
   "criteria": {"top_up_by_bank_transfer_charge": null, "verify_top_up": null,
                "apple_pay_or_google_pay": null, "(*@\ldots{}@*) 74 more options": null}}}}
\end{lstlisting}

With retrieved examples, the 20 nearest pool texts all carry the gold label, and the cap of three per label leaves three shown; one option description changes:

\begin{lstlisting}
"apple_pay_or_google_pay": "Examples: \"My Google pay top up isn't working.\"; \"Google Pay isn't working. What is wrong?\"; \"My Google pay top up isn't working. Help.\""
\end{lstlisting}

With the same three examples relabelled (T4a), ranks 1 to 3 move to \code{pending\_\allowbreak{}top\_\allowbreak{}up}, \code{contactless\_\allowbreak{}not\_\allowbreak{}working} and \code{passcode\_\allowbreak{}forgotten}, and the gold option's description returns to \code{null}.

\begingroup\footnotesize
\begin{longtable}{@{}>{\raggedright\arraybackslash}p{\dimexpr 0.1079\linewidth-1.2952\tabcolsep\relax}>{\raggedright\arraybackslash}p{\dimexpr 0.2358\linewidth-2.8292\tabcolsep\relax}>{\raggedright\arraybackslash}p{\dimexpr 0.0651\linewidth-0.7810\tabcolsep\relax}>{\raggedright\arraybackslash}p{\dimexpr 0.2379\linewidth-2.8544\tabcolsep\relax}>{\raggedright\arraybackslash}p{\dimexpr 0.0651\linewidth-0.7810\tabcolsep\relax}>{\raggedright\arraybackslash}p{\dimexpr 0.2232\linewidth-2.6783\tabcolsep\relax}>{\raggedright\arraybackslash}p{\dimexpr 0.0651\linewidth-0.7810\tabcolsep\relax}@{}}
\toprule
Rank & Bare & p & + retrieved examples & p & + examples, 3 labels wrong & p \\
\midrule
\endfirsthead
\multicolumn{7}{@{}l}{\textit{Table (continued)}}\\
\toprule
Rank & Bare & p & + retrieved examples & p & + examples, 3 labels wrong & p \\
\midrule
\endhead
\bottomrule
\endlastfoot
1 & top\us{}up\us{}failed & 0.68 & \textbf{apple\us{}pay\us{}or\us{}google\us{}pay} \ding{51} & 0.99 & pending\us{}top\us{}up & 0.82 \\
2 & \textbf{apple\us{}pay\us{}or\us{}google\us{}pay} \ding{51} & 0.30 & top\us{}up\us{}failed & 0.01 & top\us{}up\us{}failed & 0.08 \\
3 & topping\us{}up\us{}by\us{}card & 0.02 & (rest \textless{} 0.005) &  & passcode\us{}forgotten & 0.07 \\
p(gold) &  & 0.30 &  & 0.99 &  & 0.01 \\
\end{longtable}
\endgroup

Billed input tokens: 1,020 \ensuremath{\rightarrow} 1,067 \ensuremath{\rightarrow} 1,080. Over the 800 shared requests (bare / kNN-20 / top-3 relabelled): right-right-right 434, right-right-wrong 222, right-wrong-right 2, right-wrong-wrong 4, wrong-right-right 38, wrong-right-wrong 55, wrong-wrong-right 7, wrong-wrong-wrong 38.

The code below re-sends the bare and the retrieval request of Figure 2 (\code{python reproduce.\allowbreak{}py \textless{}file\textgreater{}.\allowbreak{}json} does the same for any example, with \code{-\allowbreak{}-\allowbreak{}dry} and \code{-\allowbreak{}-\allowbreak{}mock} modes). Jev is not bit-for-bit deterministic, so expect small differences in the probabilities.

\begin{lstlisting}
# Re-send the bare and the retrieval request of Figure 2 (needs TYPESAFE_API_KEY).
import json, os, urllib.request

URL = "https://api.typesafe.ai/v1/systemone"
FULL = "paper/worked_examples/flagship_banking77.json"   # the exact published bodies

def decide(body):
    req = urllib.request.Request(URL, data=json.dumps(body).encode("utf-8"), method="POST",
        headers={"Authorization": "Bearer " + os.environ["TYPESAFE_API_KEY"],
                 "Content-Type": "application/json"})
    with urllib.request.urlopen(req, timeout=90) as r:
        return json.load(r)["answers"]["q"]["probabilities"]

if os.path.exists(FULL):
    reqs = json.load(open(FULL, encoding="utf-8"))["requests"]
    bare, knn = reqs["1_bare"]["wire_request"], reqs["2_retrieved_examples"]["wire_request"]
else:   # abridged fallback: 3 of the 77 intents, so the probabilities will differ from Figure 2
    bare = {"model": "jev-1.13.0",
            "state": "Top up on my Google Pay isn't working.",
            "questions": {"q": {"type": "choice",
                "instructions": "What does this banking customer need help with?",
                "criteria": {"top_up_failed": None, "apple_pay_or_google_pay": None,
                             "pending_top_up": None}}}}
    knn = json.loads(json.dumps(bare))
    knn["questions"]["q"]["criteria"]["apple_pay_or_google_pay"] = (
        'Examples: "My Google pay top up isn\'t working."; '
        '"Google Pay isn\'t working. What is wrong?"; "My Google pay top up isn\'t working. Help."')

for name, body in (("bare", bare), ("retrieval", knn)):
    print(name, sorted(decide(body).items(), key=lambda kv: -kv[1])[:3])
\end{lstlisting}

\appsub{M.2}{Definitions for cryptic option ids}
\textbf{M.2 Definitions for cryptic option ids} (MetaTool; Run 1, T7b). Without definitions the options are bare ids \code{tool\_\allowbreak{}01} \ldots{} \code{tool\_\allowbreak{}20}; with definitions each id carries MetaTool's own tool description (at most 300 characters). Over all 2,000 requests the definitions fixed 1,651 (82.5\%) and broke 10 (0.5\%); 36 were right both ways and 303 wrong both ways; accuracy 2.3\% \ensuremath{\rightarrow} 84.4\%, +82.05 points. The example request (\code{metatool\_\allowbreak{}c-\allowbreak{}test-\allowbreak{}2441}, gold \code{tool\_\allowbreak{}02} = CourseTool: p(gold) 0.03 \ensuremath{\rightarrow} 0.97) is in Extended Appendix M.2.

\section{Algorithms}

Pseudocode for three key approaches, adapted from the starter repository's \code{docs/\allowbreak{}algorithms.\allowbreak{}md} (\url{https://github.com/BjBodner/decision-levers}). Each box gives inputs, output and steps; the line under it names the starter code's function. Notation: \code{x} is the text to decide on (the state), \code{O} the ordered option list, \code{D} a labelled pool of \code{(text,\allowbreak{} label)} pairs, \code{M} the decision model, which returns a probability vector over \code{O}, and \code{sim(a,\allowbreak{} b)} cosine similarity. The research code that produced the paper's numbers is a separate code base (Appendix K). Extended Appendix N lists the other eleven algorithms, including the screen-freeze-confirm loop (N.14), class shares (N.4), the kNN vote (N.6), the tuned classifier (N.7), the null controls (N.12) and the paired cluster bootstrap (N.13).

\begin{lstlisting}
Algorithm N.2  Retrieved examples in the option descriptions (the kNN-20 prompt; Sections 4 and 5)
Input:   x, q, O, pool D, k = 20, per-label cap c (3; 10 when |O| = 2)
Output:  p over O
1. N <- the k rows of D with the highest sim(x, text), most similar first
2. S <- []; for (label, text) in N: if count(label in S) < c: append (label, text) to S
3. for each option o:
       if S has texts t1..tm labelled o: desc(o) <- 'Examples: "t1"; "t2"; ...; "tm"'   (replaces any description)
       else: desc(o) <- unchanged
4. p <- M(state = x, question = q, criteria = desc)
Note: pool texts are sent to the vendor.
\end{lstlisting}

\code{add\_\allowbreak{}examples(request,\allowbreak{} pool,\allowbreak{} k=\allowbreak{}20)}; the selection alone is \code{select\_\allowbreak{}examples}.

\begin{lstlisting}
Algorithm N.8  One bare call, combined locally (product of experts; Section 5, G5)
Input:   bare answer p (N.1), vote v (N.6), classifier f(x) (N.7), eps = 1e-3
Output:  e over O
1. e(o) <- (p(o) + eps) * (v(o) + eps) * (f(x)(o) + eps)
2. e <- e / sum(e)
Note: no pool text is sent to the vendor.
\end{lstlisting}

\code{combine(bare,\allowbreak{} vote,\allowbreak{} lr)}.

\begin{lstlisting}
Algorithm N.9  Classifier-first cascade (Section 5, G1c)
Input:   held-out items with gold y_i, local answers l_i, model answers m_i (e.g. N.2);
         target: max accuracy loss delta (default 0.005), OR min recall rho for a positive class
Output:  threshold t
1. conf_i <- max_o l_i(o)
2. grid <- the 201 quantiles of conf, plus -1 and 1 + 1e-9
3. full <- accuracy of m on the held-out items
4. for t in grid:
       local_i <- conf_i > t;   pred_i <- argmax l_i if local_i else argmax m_i
       feasible if  accuracy(pred) >= full - delta      (accuracy target)
                or  recall_positive(pred) >= rho         (recall target)
5. t <- the feasible threshold with the smallest share sent (ties: higher accuracy); none feasible -> send all
Deploy: answer locally iff conf > t, else call the model.  Freeze t before testing.
Note: with m = the kNN-20 prompt (N.2), every forwarded item is sent with its nearest pool texts.
\end{lstlisting}

\code{ClassifierFirstCas\allowbreak{}cade().\allowbreak{}fit(local,\allowbreak{} model,\allowbreak{} gold,\allowbreak{} max\_\allowbreak{}acc\_\allowbreak{}loss=\allowbreak{}.\allowbreak{}.\allowbreak{}.\allowbreak{})} or \code{fit(.\allowbreak{}.\allowbreak{}.\allowbreak{},\allowbreak{} min\_\allowbreak{}recall=\allowbreak{}.\allowbreak{}.\allowbreak{}.\allowbreak{},\allowbreak{} positive=\allowbreak{}.\allowbreak{}.\allowbreak{}.\allowbreak{})}. On rare positives choose the recall target: the accuracy target cost 7.3 points of toxic precision on ToxicChat (Appendix E.7).

Algorithm N.14 (screen, freeze, confirm; starter functions \code{compare}, \code{freeze} and \code{holm}) is in Extended Appendix N, with the other ten.

\section{Run 4: five tests on fresh items}

\textbf{Plan.} The analysis plan (\code{reports/\allowbreak{}preregistration\_\allowbreak{}run4.\allowbreak{}md}) was frozen by a manifest of 67 files, aggregate \code{34fc\allowbreak{}1153\allowbreak{}2772\allowbreak{}c44e\allowbreak{}65bf\allowbreak{}7d32\allowbreak{}bdb2\allowbreak{}145f\allowbreak{}a95a\allowbreak{}fdbb\allowbreak{}e09c\allowbreak{}2b5d\allowbreak{}cd51\allowbreak{}1850\allowbreak{}d6fa\allowbreak{}47a8\allowbreak{}}, created 2026-11-09 20:20:24 UTC (\code{registry\_\allowbreak{}post/\allowbreak{}freeze/\allowbreak{}manifest\_\allowbreak{}run4\_\allowbreak{}20261109T202024Z.\allowbreak{}json}). The manifest also covers every file that the Run 3, Run 2 and Run 1 manifests freeze. By the author's choice the freeze is internal: neither file is deposited with a third party, so, as for Runs 1 and 2, the hash shows that frozen files did not change, not when they were written. A copy can be checked like Run 3's: normalise to LF, restore the placeholder line (\code{MANIFEST\_\allowbreak{}SHA256:\allowbreak{} \textless{}filled at freeze\textgreater{}}), then hash and compare with the manifest entry (\code{scripts/\allowbreak{}freeze\_\allowbreak{}manifest\_\allowbreak{}run4.\allowbreak{}py -\allowbreak{}-\allowbreak{}check-\allowbreak{}deposit}).

\textbf{Run.} \code{scripts/\allowbreak{}run\_\allowbreak{}run4.\allowbreak{}sh} verified the four manifests (Run 4, Run 3, Run 2, Run 1) before sending and again before analysis. The never-sent scan found the observation log unchanged for MASSIVE, LEDGAR and MetaTool since the freeze, and the run recorded no validity problem (\code{run4\_\allowbreak{}run\_\allowbreak{}meta.\allowbreak{}json}). The first billed Run 4 call was at 21:09:30 UTC, 49 minutes after the freeze, and the last at 21:33:51 UTC (spend ledger). The run sent 85,592 new requests, a further 6,000 X2 requests were cache hits, and it cost \$4.19 (calibrated estimate on the unsealed items \$4.35; sealed dry run \$4.79). The ledger stood at \$34.507 before the run and \$38.699 after it (\code{run4\_\allowbreak{}run.\allowbreak{}log}). The \code{-\allowbreak{}-\allowbreak{}analyze} pass re-scored from the cache only and reproduced the primary table exactly.

\textbf{Items and design.} A Run 4 item appears in no logged request, Run 3's included, and has no exact or TF-IDF \ensuremath{\geq} 0.9 twin among texts that may have been sent; arms take the first n items of a fixed hash order, so arms are nested. \textbf{X1a} tests one bare call times a tuned, prior-corrected logistic regression on the same 5-per-class pool, and \textbf{X1b} the G5 product at 100 per class, each against the dense kNN-20 prompt from the same pool, non-inferiority at \ensuremath{-}1.5, on 1,725 MASSIVE, 1,500 LEDGAR and 2,381 MetaTool items. \textbf{X3} uses H2's display on the first 400 items per dataset; on the items that show exactly 3 gold-labelled examples (311, 380 and 366), the same first k of a seeded order of those examples (k = 1, 2, 3) are deleted (D\us{}k), relabelled to three distinct rivals (R\us{}k), or all relabelled to the first of those rivals (S\us{}k), with rank, neighbours and rivals fixed within an item. X3a tests the mean over k of acc(D\us{}k) \ensuremath{-} acc(R\us{}k) \textgreater{} 0, X3b the mean over k = 2, 3 of acc(S\us{}k) \ensuremath{-} acc(R\us{}k) \textless{} 0, and X3c (D\us{}2 \ensuremath{-} D\us{}3) \ensuremath{-} (D\us{}1 \ensuremath{-} D\us{}2) \textgreater{} 0. Gold classes are resampled within each dataset, pooled dataset-equal, 20,000 draws; Holm at \ensuremath{\alpha} = 0.05 over the five tests. Secondary estimates are in Appendix E.9; the item table and the frozen code's Table O.1 are in Extended Appendix O.

\textbf{Decisions.} X3a, X3b and X3c were confirmed (Holm p .00025 each); X1a (\ensuremath{-}0.64 {[}\ensuremath{-}1.66, +0.35{]}) and X1b (\ensuremath{-}0.73 {[}\ensuremath{-}1.63, +0.14{]}) were not shown non-inferior (Holm p .16). No planned qualifier applies: every dataset, each wording alone and each X3a and X3b component at every k lie on the planned side. With near-duplicate-group clusters both X1 intervals would lie above \ensuremath{-}1.5 ({[}\ensuremath{-}1.18, \ensuremath{-}0.09{]} and {[}\ensuremath{-}1.26, \ensuremath{-}0.19{]}), but the planned unit is the gold class. X1a had been planned at power 0.83 with a pooled SE of 0.48 points; the realised class-cluster SE was about 0.61. No deviation was recorded.

\emph{Per dataset} (accuracy points; class clusters).

\begingroup\small
\begin{longtable}{@{}lllllll@{}}
\toprule
Dataset & X1a & X1b & X3a & X3b & X3c & X3 feasible items (of 400) \\
\midrule
\endfirsthead
\multicolumn{7}{@{}l}{\textit{Table (continued)}}\\
\toprule
Dataset & X1a & X1b & X3a & X3b & X3c & X3 feasible items (of 400) \\
\midrule
\endhead
\bottomrule
\endlastfoot
MASSIVE & \ensuremath{-}0.55 & \ensuremath{-}0.75 & +14.5 & \ensuremath{-}22.4 & +12.9 & 311 \\
LEDGAR & \ensuremath{-}0.27 & \ensuremath{-}1.33 & +6.1 & \ensuremath{-}4.9 & +3.3 & 380 \\
MetaTool & \ensuremath{-}1.09 & \ensuremath{-}0.10 & +18.8 & \ensuremath{-}12.2 & +14.2 & 366 \\
\end{longtable}
\endgroup

Pooled X3 components: deletion minus distinct relabel was +2.40, +9.44 and +27.55 points at k = 1, 2, 3; one rival minus distinct rivals \ensuremath{-}6.17 at k = 2 and \ensuremath{-}20.23 at k = 3; deletion steps \ensuremath{-}1.80, \ensuremath{-}3.76 and \ensuremath{-}13.88. Against the clean display, deleting all three gold examples gave \ensuremath{-}19.43 {[}\ensuremath{-}22.75, \ensuremath{-}16.36{]} points, relabelling them to distinct rivals \ensuremath{-}46.98 and to one rival \ensuremath{-}67.20. Applied without refitting (Table I.13), the I.13 model predicts almost no difference between one rival and two at k = 2 (+0.2, against \ensuremath{-}6.2 observed), and for X3b it predicted \ensuremath{-}5.1 against \ensuremath{-}22.4 on MASSIVE and \ensuremath{-}13.2 against \ensuremath{-}4.9 on LEDGAR.

\section{Analysis plans, statistics and all 29 confirmatory tests}

Section 2.3 summarises the method in one paragraph. Runs 5 to 8 are reported in Sections 3 to 6, Appendix U and Extended Appendices S, U and V, not in the 29 counted here. Across Runs 1 to 8 there are 134 planned tests with a decision rule: 29 in Runs 1 to 4, 32 in Run 5, 3 in Run 6, 7 in Run 7 and 63 in Run 8 (moved from Section 2.3 in v25). This appendix gives the status words (P.1), the null calibration (P.2), the freeze records (P.3), and the statistics with all 29 tests (P.4). Figure P.1, a diagram of the four runs, and the longer method text are in Extended Appendix P and B.9.

\appsub{P.1}{Test identifiers and plain-language status}
\textbf{P.1 Test identifiers and plain-language status.} Test identifiers name the run: T for Run 1, G for Run 2, H for Run 3 and X for Run 4. The main text avoids them; Table P.1 and Appendices A, I and O use them. The main text, its tables and its figures use four status words, mapped onto the evidence tags of P.4 as follows; Run 5 adds four words of its own. Some extended-appendix text also uses ``descriptive'', ``not confirmed'' and ``exploratory'', which mean ``secondary'', ``not shown'' and ``observed''. Appendix tables also use \emph{item-level only} (a pass with items but not classes resampled) and \emph{reference only} (an estimate with no test); these definitions moved from Section 2.3 in v25.

\begingroup\footnotesize
\begin{longtable}{@{}>{\raggedright\arraybackslash}p{\dimexpr 0.1900\linewidth-0.7601\tabcolsep\relax}>{\raggedright\arraybackslash}p{\dimexpr 0.2031\linewidth-0.8125\tabcolsep\relax}>{\raggedright\arraybackslash}p{\dimexpr 0.6069\linewidth-2.4274\tabcolsep\relax}@{}}
\toprule
Main text & Tag & Meaning \\
\midrule
\endfirsthead
\multicolumn{3}{@{}l}{\textit{Table (continued)}}\\
\toprule
Main text & Tag & Meaning \\
\midrule
\endhead
\bottomrule
\endlastfoot
confirmed & C1, C2, C3, C4 & a planned test passed in Run 1, 2, 3 or 4, and also passed with gold-class clusters wherever a dataset has more than two classes \\
... within \ensuremath{\pm}2 points & equivalent & two one-sided tests at \ensuremath{\pm}2 passed \\
... no more than X points worse & non-inferior & a non-inferiority test at \ensuremath{-}X passed \\
observed (item-level only) & C1 or C2, item-level only & passed its planned rule with near-duplicate (item-level) clusters but not under gold-class clusters, so not counted as confirmed (Section 2.3): Run 2's G2, G4a-b and G5, and Run 1's T3b, T5a-b, T6 (via politeness on tweets) and T7b (Appendix I.7). The two-class ToxicChat passes (T3c, G1c) cannot be checked by class and stay confirmed for those two classes \\
observed (small-cluster correction) & C1, fails the wild-cluster class check & T3a and Run 8's R8-W4: passed with the pairs class bootstrap but not with the wild-cluster bootstrap that holds nominal size (re-analyses 2 and 3; Appendices U.3 and U.4) \\
secondary & secondary, descriptive & a pre-specified estimate or summary with no decision rule; ``inconclusive'' when the interval is too wide to decide \\
observed & O & exploratory: development data, or any analysis of test items chosen after labels were unsealed \\
not shown & not shown, not confirmed, not equivalent & a planned test that did not pass; never read as ``no effect'' \\
transfers & Run 5 rule & confirmed on at least 5 of the 7 other models, with no confirmed opposite effect (fixed in the plan) \\
partially transfers & Run 5 rule & confirmed on 1 to 4 of the 7, or on 5 or more while some model is opposite \\
inconclusive (Run 5 wrappers) & Run 5 R1 & neither within \ensuremath{\pm}2 nor wholly outside it \\
below the completeness rule & Run 5 protocol failure & fewer than 98\% of a cell's items complete after one re-send; reported as a protocol failure; the verdict is also given without it \\
\end{longtable}
\endgroup

\appsub{P.2}{Null calibration}
\textbf{P.2 Null calibration.}

\textbf{A promotion rule must be checked on content-free edits.} A screen arm was promoted if its 90\% bootstrap interval of \ensuremath{\Delta}logp excluded 0, its flip rate exceeded an A/A floor, both paraphrases agreed in sign and \textbar{}\ensuremath{\Delta}logp\textbar{} \ensuremath{\geq} 0.04. Without the last condition the rule promoted or flagged 12 of 53 content-free arms (23\%): 7 of 23 in-sample (four promoted, three flagged as harmful) and 5 of 30 on held-out datasets. These are negligible real effects of surface form (Extended Appendix B.2); the identical resend was never promoted. A floor tuned on the nulls is circular: 0.04 sits just above the largest nulls (below; Extended Appendix B.3).

\begingroup\footnotesize
\begin{longtable}{@{}>{\raggedright\arraybackslash}p{\dimexpr 0.2869\linewidth-1.7211\tabcolsep\relax}>{\raggedright\arraybackslash}p{\dimexpr 0.2678\linewidth-1.6070\tabcolsep\relax}>{\raggedright\arraybackslash}p{\dimexpr 0.2250\linewidth-1.3502\tabcolsep\relax}>{\raggedright\arraybackslash}p{\dimexpr 0.2203\linewidth-1.3217\tabcolsep\relax}@{}}
\toprule
Rule & In-sample, 23 & Held-out, 30 & Pooled, 53 \\
\midrule
\endfirsthead
\multicolumn{4}{@{}l}{\textit{Table (continued)}}\\
\toprule
Rule & In-sample, 23 & Held-out, 30 & Pooled, 53 \\
\midrule
\endhead
\bottomrule
\endlastfoot
Original rule as coded, both directions & 7/23 (30\%; Clopper-Pearson 13-53\%) & 5/30 (17\%; 6-35\%) & 12/53 (23\%; 12-36\%) \\
+ floor 0.03 & 1/23 & 2/30 & 3/53 \\
+ floor 0.04 (adopted for the screen) & 0/23 (upper bound 14.8\%) & 0/30 (upper bound 11.6\%) & 0/53 (upper bound 6.7\%) \\
\end{longtable}
\endgroup

The largest null effects are 0.0308 in-sample and 0.0345 held-out. Planted wrong labels among 60 fixed examples were detected 0 of 2 times at 2\% and 5\%, 1 of 2 at 10\% and 2 of 2 at 20\%.

\appsub{P.3}{Screen -> freeze -> confirm}
\textbf{P.3 Screen \ensuremath{\rightarrow} freeze \ensuremath{\rightarrow} confirm.} Each analysis plan fixed, before any test call, the hypotheses, family, endpoint (paired accuracy difference), margins, power, analysis code and the wording each outcome permits. A SHA-256 manifest hashes it with the code and a label-free item digest, and the run refuses to start if any hash differs. Labels stayed sealed until each manifest verified. For a run that was not deposited, the hashes show that frozen files did not change, not when they were written; that rests on the ledger timestamps and the author's word.

\begingroup\footnotesize
\begin{longtable}{@{}>{\raggedright\arraybackslash}p{\dimexpr 0.0492\linewidth-0.4923\tabcolsep\relax}>{\raggedright\arraybackslash}p{\dimexpr 0.2168\linewidth-2.1684\tabcolsep\relax}>{\raggedright\arraybackslash}p{\dimexpr 0.1241\linewidth-1.2408\tabcolsep\relax}>{\raggedright\arraybackslash}p{\dimexpr 0.1969\linewidth-1.9686\tabcolsep\relax}>{\raggedright\arraybackslash}p{\dimexpr 0.1773\linewidth-1.7731\tabcolsep\relax}>{\raggedright\arraybackslash}p{\dimexpr 0.2357\linewidth-2.3568\tabcolsep\relax}@{}}
\toprule
Run & Tests & Manifest & Created (UTC; 2026-11-09 unless stated) & First billed call (UTC) & Deposit \\
\midrule
\endfirsthead
\multicolumn{6}{@{}l}{\textit{Table (continued)}}\\
\toprule
Run & Tests & Manifest & Created (UTC; 2026-11-09 unless stated) & First billed call (UTC) & Deposit \\
\midrule
\endhead
\bottomrule
\endlastfoot
1 & 15, held-out items & 41 files & 00:40:59 & 00:53 & frozen internally, not deposited \\
2 & 6, never-sent items & 48 files & 06:48:02 & 06:58 & frozen internally, not deposited \\
3 & 3, fresh items of four datasets & 60 files & 15:16:30 & 15:39:59 & deposited publicly before the first call (gist created 15:17:47 UTC; URL in Statements) \\
4 & 5, fresh items of three datasets & 67 files & 20:20:24 & 21:09:30 & frozen internally, not deposited \\
5 & 4 per model, 7 models plus Jev 1.14; items Jev 1.13 answered & 72 files & 01:02:09 on 2026-11-10 & 01:07:05 on 2026-11-10 & frozen internally, not deposited; key checks and one probe per model on a synthetic text came before the freeze \\
6 & 3, Run 4's items (placement B and a drift control new) & 72 files & 04:19:42 on 2026-11-10 & 04:21:51 on 2026-11-10 & frozen internally, not deposited \\
7 & 7, Run 6's and Run 4's items & 78 files & 11:32:14 on 2026-11-10 & after the freeze (pilot, then full send) & frozen internally, not deposited \\
8 & 12 + 6 on Jev, 4 per other model, 6 on the chat model, 2 addendum tests per model; fresh items & 90 files each (freezes 3 to 6) & 12:43:04 (freeze 3), 13:55:30, 14:31:03 and 17:52:28 (freeze 6) on 2026-11-10 & after freeze 3; addendum and chat requests after their freezes & frozen internally, not deposited \\
\end{longtable}
\endgroup

The aggregate SHA-256 of each manifest:

\begin{itemize}
\item Run 1: \code{735e\allowbreak{}22ef\allowbreak{}487f\allowbreak{}7f6f\allowbreak{}9bce\allowbreak{}20cf\allowbreak{}eb92\allowbreak{}e511\allowbreak{}5550\allowbreak{}e57d\allowbreak{}4d34\allowbreak{}d275\allowbreak{}3fa3\allowbreak{}e4eb\allowbreak{}c0ef\allowbreak{}412a\allowbreak{}}, 2026-11-09 00:40:59 UTC.
\item Run 2: \code{9e68\allowbreak{}db01\allowbreak{}f0b6\allowbreak{}15db\allowbreak{}5e9c\allowbreak{}d449\allowbreak{}555d\allowbreak{}2877\allowbreak{}d8e0\allowbreak{}5965\allowbreak{}e0ed\allowbreak{}bba0\allowbreak{}f807\allowbreak{}4593\allowbreak{}98dc\allowbreak{}7ea7\allowbreak{}}, 2026-11-09 06:48:02 UTC.
\item Run 3: \code{0c77\allowbreak{}4321\allowbreak{}f162\allowbreak{}0a6f\allowbreak{}09a6\allowbreak{}2e4a\allowbreak{}98f3\allowbreak{}04bc\allowbreak{}dc21\allowbreak{}cc17\allowbreak{}3371\allowbreak{}e67f\allowbreak{}0db7\allowbreak{}f3b9\allowbreak{}57cc\allowbreak{}391f\allowbreak{}}, 2026-11-09 15:16:30 UTC.
\item Run 4: \code{34fc\allowbreak{}1153\allowbreak{}2772\allowbreak{}c44e\allowbreak{}65bf\allowbreak{}7d32\allowbreak{}bdb2\allowbreak{}145f\allowbreak{}a95a\allowbreak{}fdbb\allowbreak{}e09c\allowbreak{}2b5d\allowbreak{}cd51\allowbreak{}1850\allowbreak{}d6fa\allowbreak{}47a8\allowbreak{}}, 2026-11-09 20:20:24 UTC (first billed Run 4 call 21:09:30 UTC).
\item Run 5: \code{be2f\allowbreak{}e002\allowbreak{}8844\allowbreak{}3769\allowbreak{}0474\allowbreak{}b380\allowbreak{}7c5f\allowbreak{}cf3a\allowbreak{}8895\allowbreak{}b2d3\allowbreak{}3b71\allowbreak{}5736\allowbreak{}a214\allowbreak{}4a26\allowbreak{}b103\allowbreak{}32fc\allowbreak{}}, 2026-11-10 01:02:09 UTC (72 files; first billed Run 5 call 01:07:05 UTC).
\item Run 6: \code{0512\allowbreak{}deea\allowbreak{}c4eb\allowbreak{}b097\allowbreak{}c6ef\allowbreak{}51dc\allowbreak{}609c\allowbreak{}9fcb\allowbreak{}a45e\allowbreak{}9302\allowbreak{}27ef\allowbreak{}05b1\allowbreak{}4a93\allowbreak{}8cb1\allowbreak{}49bc\allowbreak{}4fd9\allowbreak{}}, 2026-11-10 04:19:42 UTC (72 files; fixed before any placement-B request; first billed Run 6 call 04:21:51 UTC).
\item Run 7: \code{e767\allowbreak{}541d\allowbreak{}2bd9\allowbreak{}645f\allowbreak{}9821\allowbreak{}87c2\allowbreak{}05ff\allowbreak{}b974\allowbreak{}849a\allowbreak{}f43c\allowbreak{}fbfc\allowbreak{}b26c\allowbreak{}79ac\allowbreak{}3f66\allowbreak{}6a79\allowbreak{}a8dc\allowbreak{}}, 2026-11-10 11:32:14 UTC (78 files; fixed before any Run 7 request).
\item Run 8: the original arms were sent under manifest 3, \code{e70f\allowbreak{}6de3\allowbreak{}8b5a\allowbreak{}a450\allowbreak{}9aa3\allowbreak{}e548\allowbreak{}d9b5\allowbreak{}7f92\allowbreak{}79bf\allowbreak{}7fd9\allowbreak{}0be3\allowbreak{}8777\allowbreak{}1d1d\allowbreak{}6110\allowbreak{}cd1c\allowbreak{}8408\allowbreak{}}, 2026-11-10 12:43:04 UTC (90 files). Three dated addenda were each fixed before any of their requests, the last as manifest 6, \code{adb0\allowbreak{}7972\allowbreak{}a921\allowbreak{}a198\allowbreak{}c787\allowbreak{}7cda\allowbreak{}d508\allowbreak{}0f07\allowbreak{}ba16\allowbreak{}a6f4\allowbreak{}6853\allowbreak{}0db2\allowbreak{}38d4\allowbreak{}f3a3\allowbreak{}933d\allowbreak{}de2e\allowbreak{}}, 2026-11-10 17:52:28 UTC (Appendix U.2).
\end{itemize}

Run 2 was designed after Run 1 was unsealed, and Run 4 after Run 3. Run 4's manifest also covers every file of the earlier ones, Run 5's covers everything Run 3's freezes plus its own code and request digests, Run 6's covers everything Run 4's freezes plus its own code, items and request digests, and Runs 3, 4 and 6 verified all earlier manifests before sending and again before analysis. Run 7's manifest chains to Run 6's, and Run 8's to Runs 6, 5, 4 and 3 and both Gates; both runs verified the chain before sending and after analysis. Appendices I.12 and O give the recipe for checking a deposited or released copy against its hash.

\appsub{P.4}{Statistics, evidence tags and all 29 tests}
\textbf{P.4 Statistics, evidence tags and all 29 tests.}

Tests use a cluster bootstrap of the mean paired difference (20,000 draws) {[}\citenum{r4}{]}, with Holm's procedure {[}\citenum{r15}{]} within each run at family-wise \ensuremath{\alpha} = 0.05. Runs 1 and 2 cluster mostly by near-duplicate group, which is nearly the item (``item-level''). Runs 3 and 4 cluster by gold class within each dataset and, for Finding 1, Run 3 also resamples stimuli and paraphrases (a pigeonhole bootstrap {[}\citenum{r58}{]}). Equivalence uses two one-sided tests {[}\citenum{r42}{]} at \ensuremath{\pm}2 points; non-inferiority uses \ensuremath{-}2 (Run 1, H3) or \ensuremath{-}1.5 (Run 2, H5, Run 4's X1). Multi-component tests are intersection-union tests (IUT). Intervals are 90\%. Run 3's label-budget contrasts could not reach 80\% power under class clusters, so they are secondary estimates without a decision; Run 4 tested two of them on larger samples, with planned power 0.83 (X1a) and 1.00 (X1b) at \ensuremath{\alpha}/5.

Tags C1 to C4 and O are defined in P.1; O results cannot change a decision. Each run's own Holm family is the planned method; one Holm correction over all 29 tests would no longer reject T8 and H3 (adjusted p .052 and .070) and changes no other decision; with wild-cluster class-level p, T3a fails as well (Appendix I.11).

\textbf{How to read ``25 of 29''.} 25 tests passed their planned rule. Nine of them (T3b, T5a-b, T6, T7b, G2, G4a-b, G5) passed only with near-duplicate clusters and are reported as observed (item-level only; Section 2.3), and T3a fails a small-cluster correction (re-analysis 2), so 15 are confirmed: 13 that also pass with gold-class clusters under that correction and the two two-class ToxicChat passes (T3c, G1c), which hold for those two classes only. Several passes were near-certain by design (T3a-c, T5a-b, G4's classifier component, H2 with planned power 1.00, H3's rescue part), others repeat one claim on another dataset (T1a-c, T3a-c, T4a-b, T5a-b) or share items (X3a-c), and three of the four failures (T2, X1a, X1b) bear on Finding 3. The tests used 400 (T4a) to 7,418 (T2) unique items; the passed non-inferiority tests used 1,000 to 3,424 and the failed ones 5,606.

\textbf{Why item-level passes are not counted as confirmed.} When effects vary by class, resampling items understates the uncertainty. A stress test on the cached outputs (re-analysis 1; observed) gave real paired contrasts random signs per gold class, so that the null held on average while classes differed. At a nominal two-sided 5\%, item-level intervals then rejected 38 to 46\% of the time with 8, 20 or 49 or more classes; class-cluster intervals rejected 5.5\% with 49 or more classes, 8.6 to 10.3\% with 8 to 20 and 7.0\% with 25 (re-analysis 2's re-measure: 5.85\% and 8.5 to 8.8\%, and 4.9 to 5.2\% for the wild-cluster bootstrap). So 9 of the 25 passes are reported as observed (10 after re-analysis 2 added T3a), and class-level passes near the threshold on 8 to 25 classes (the packed questions, and the definitions test H3) are the least secure confirmed ones. Two passes on ToxicChat, which has two classes, cannot be checked by class and stay confirmed for those two classes (the cascade's accuracy and the prompt against a vote); they describe the two classes tested, not a population of classes. Under identical re-sends the rule's size was near nominal (Appendix H.4).

\textbf{How many held-out items a team needs} (observed; Extended Appendix R.2). Resampling n cached items of Runs 2 to 4: at n = 100 to 200, the prompt beat bare Jev in sign on 79 to 97\% of draws, but the prompt against the local product or the cascade only on 56 to 64\% (gaps of about 0.8 points); the picked design was within 1 point of the best on 84 to 95\%. The subsamples share items with the ``truth'', so these rates are optimistic.

Margins are judgements fixed in the analysis plans: \ensuremath{\pm}2 as the smallest change a practitioner would act on, and \ensuremath{-}1.5 because replacing a prompt with a cascade or ensemble should cost less than dropping a decoration. Holm's procedure stays valid under the dependence created by tests that share control responses. The confirmatory decision rule is printed in Appendix K.

\begingroup\scriptsize
\renewcommand{\thetable}{P.1}
\begin{longtable}{@{}>{\raggedright\arraybackslash}p{\dimexpr 0.0857\linewidth-1.2000\tabcolsep\relax}>{\raggedright\arraybackslash}p{\dimexpr 0.0403\linewidth-0.5647\tabcolsep\relax}>{\raggedright\arraybackslash}p{\dimexpr 0.1197\linewidth-1.6765\tabcolsep\relax}>{\raggedright\arraybackslash}p{\dimexpr 0.0848\linewidth-1.1865\tabcolsep\relax}>{\raggedright\arraybackslash}p{\dimexpr 0.1878\linewidth-2.6294\tabcolsep\relax}>{\raggedright\arraybackslash}p{\dimexpr 0.1535\linewidth-2.1489\tabcolsep\relax}>{\raggedright\arraybackslash}p{\dimexpr 0.1268\linewidth-1.7747\tabcolsep\relax}>{\raggedright\arraybackslash}p{\dimexpr 0.2014\linewidth-2.8193\tabcolsep\relax}@{}}
\caption{All 29 confirmatory tests (25 passed as planned: 15 confirmed, 9 observed at item level only, and T3a observed after a small-cluster correction) and Run 3's secondary estimates (accuracy points; Appendix I). Holm p is within each run's own family as printed by the frozen code (Run 1 over 15 tests, Run 2 over 6, Run 3 over 3, Run 4 over 5); ``--'' marks a secondary estimate. F1, F2 and F3 are the findings of Sections 3, 4 and 5. ``Items'' counts unique items. T3, T5 and G4's classifier component were designed to be large. ``Observed, item-level only'' marks a pass that fails with gold-class clusters (Appendix I.7) and is not counted as confirmed (Section 2.3).}\\
\toprule
Finding & Run & Test; data & Items & Contrast & Estimate {[}90\% CI{]} & Holm p & Decision \\
\midrule
\endfirsthead
\multicolumn{8}{@{}l}{\textit{Table P.1 (continued)}}\\
\toprule
Finding & Run & Test; data & Items & Contrast & Estimate {[}90\% CI{]} & Holm p & Decision \\
\midrule
\endhead
\bottomrule
\endlastfoot
F1 & 1 & T6; tweets, CFPB & 3,000 & persona, capitals, politeness each vs bare (\ensuremath{\pm}2) & six intervals in {[}\ensuremath{-}1.3, +0.5{]} & 0.00075 & observed, item-level only (passed as equivalent, C1; fails with classes via politeness on tweets) \\
F1 & 1 & T8; tweets & 600 & probe + 7 unrelated questions vs probe alone, first / last (\ensuremath{\pm}2) & {[}\ensuremath{-}1.6, +0.1{]} / {[}\ensuremath{-}0.9, +0.4{]} & 0.019 & equivalent (C1); fragile: fails the 29-test correction and both stress calibrations \\
F1 & 3 & H1; MASSIVE, LEDGAR, MetaTool, CFPB & 2,400 & 5 persona, 5 politeness stimuli \ensuremath{\times} 3 paraphrases vs bare (\ensuremath{\pm}2) & persona \ensuremath{-}0.3 {[}\ensuremath{-}1.0, +0.3{]}; politeness \ensuremath{-}0.1 {[}\ensuremath{-}0.5, +0.4{]} & 0.0008 & equivalent (C3) \\
F2 & 1 & T1a-c; Banking77, CLINC150, tweets & 600; 1,400; 2,700 & TF-IDF kNN-20 vs bare & +11.2, +5.9, +8.7 & 0.00075 each & confirmed (C1) \\
F2 & 2 & G2; tweets + CFPB & 2,576 & true shares vs none & +1.9 {[}+1.1, +2.7{]} & 0.00045 & observed, item-level only (passed C2; class-cluster p .20) \\
F2 & 1 & T5a-b; tweets, CFPB & 1,800; 1,400 & true vs most distant shuffled shares & +6.8, +6.2 & 0.00075 each & observed, item-level only (passed C1) \\
F2 & 1 & T7a; MetaTool & 1,000 & names + descriptions vs names & +5.8 {[}+4.3, +7.1{]} & 0.00075 & confirmed (C1) \\
F2 & 1 & T7b; MetaTool & 1,000 & ids + descriptions vs names + descriptions (\ensuremath{-}2) & \ensuremath{-}0.5 {[}\ensuremath{-}1.5, +0.4{]} & 0.019 & observed, item-level only (passed as non-inferior, C1) \\
F2 & 3 & H3; MASSIVE, LEDGAR, CFPB & 3,424 & ids + blind definitions vs ids (\textgreater{} +10) and vs names + definitions (\ensuremath{-}2) & +73.2 {[}+63.2, +83.5{]}; \ensuremath{-}1.1 {[}\ensuremath{-}1.8, \ensuremath{-}0.5{]} (LEDGAR \ensuremath{-}2.4) & 0.014 & confirmed, not consistent across datasets (C3); fails the 29-test correction \\
F2 & 1 & T4a-b; Banking77, tweets & 400; 600 & 3 nearest shown labels wrong vs clean & \ensuremath{-}33.5, \ensuremath{-}13.4 & 0.00075 each & confirmed harm (C1) \\
F2 & 3 & H2; all four & 1,049 & relabel k = 0 to 3 of three gold examples shown: slope & \ensuremath{-}12.3 {[}\ensuremath{-}13.3, \ensuremath{-}11.3{]}; cumulative \ensuremath{-}2.9, \ensuremath{-}11.4, \ensuremath{-}38.2 (secondary) & 0.00015 & confirmed harm, monotone (C3) \\
F2 & 4 & X3a; MASSIVE, LEDGAR, MetaTool & 1,057 & delete k gold examples minus relabel them to distinct rivals, mean over k = 1 to 3 & +13.1 {[}+11.7, +14.7{]} & 0.00025 & confirmed (C4) \\
F2 & 4 & X3b; same & 1,057 & relabel to one rival minus to distinct rivals, mean over k = 2, 3 & \ensuremath{-}13.2 {[}\ensuremath{-}14.7, \ensuremath{-}11.7{]} & 0.00025 & confirmed harm (C4) \\
F2 & 4 & X3c; same & 1,057 & deletion: loss at the third gold example minus loss at the second & +10.1 {[}+7.5, +12.7{]} & 0.00025 & confirmed (C4) \\
F2 & 2 & G3; tweets & 938 & top-3 vs matched random-3 relabel & +1.2 {[}\ensuremath{-}0.4, +2.7{]} & 0.12 & not shown (power .35) \\
F3 & 2 & G4a-b; tweets, CFPB & 1,170; 1,406 & 5 per class: kNN-20 vs bare; vs tuned classifier & +3.6, +2.1; +17.3, +16.1 & 0.0003; 0.001 & observed, item-level only (passed C2) \\
F3 & 3 & H4; all four & -- & 5 per class: kNN-20 vs bare; vs tuned classifier; vs centroid & +2.1 {[}+0.0, +4.1{]}; +10.8 {[}+7.4, +15.0{]}; +12.0 {[}+8.1, +17.9{]} & -- & secondary, estimate only \\
F3 & 2 & G5; tweets + CFPB & 2,576 & bare \ensuremath{\times} vote \ensuremath{\times} classifier minus kNN-20, P = 100 (\ensuremath{-}1.5) & +1.4 {[}+0.3, +2.5{]} & 0.0003 & observed, item-level only (passed as non-inferior, C2; class-cluster p .052) \\
F3 & 3 & H5; LEDGAR, CFPB & 2,000 & the same, class clusters & +0.1 {[}\ensuremath{-}2.1, +1.5{]} & -- & secondary, inconclusive \\
F3 & 3 & H5i; MASSIVE, MetaTool & 1,200 & the same (expected to trail) & +0.6 {[}\ensuremath{-}0.7, +2.0{]} & -- & secondary, inconclusive \\
F3 & 4 & X1a; MASSIVE, LEDGAR, MetaTool & 5,606 & 5 per class: bare \ensuremath{\times} tuned classifier minus kNN-20 (\ensuremath{-}1.5) & \ensuremath{-}0.6 {[}\ensuremath{-}1.7, +0.3{]} & 0.16 & not shown non-inferior \\
F3 & 4 & X1b; same & 5,606 & 100 per class: bare \ensuremath{\times} vote \ensuremath{\times} classifier minus kNN-20 (\ensuremath{-}1.5) & \ensuremath{-}0.7 {[}\ensuremath{-}1.6, +0.1{]} & 0.16 & not shown non-inferior \\
F3 & 2 & G1c; ToxicChat & 2,244 & classifier-first cascade (8.2\% sent) minus kNN-20 (\ensuremath{-}1.5) & \ensuremath{-}0.2 {[}\ensuremath{-}0.8, +0.4{]} & 0.0003 & non-inferior on accuracy (C2; two classes, so no class-level check) \\
F3 & 1 & T2; Banking77 + CLINC150 & 7,418 & dense prompt vs fixed vote (\ensuremath{\pm}2) & +3.0 {[}+2.4, +3.7{]} & 0.99 & not equivalent \\
F3 & 1 & T3a-c; tweets, CFPB, ToxicChat & 1,300; 1,600; 2,600 & dense prompt vs fixed vote & +17.0, +9.2, +5.8 & 0.00075 each & T3c confirmed (C1, two classes); T3a observed (fails the small-cluster correction); T3b observed, item-level only \\
\end{longtable}
\endgroup

\setcounter{section}{20}
\section{Runs 7 and 8, and re-analyses 2 and 3}

Runs 7 and 8 had analysis plans fixed in advance, like Runs 4 to 6; their full tables, secondary estimates and logs are in Extended Appendices U and V. Re-analysis 2 (of Runs 1 to 6) and re-analysis 3 (of Runs 7 and 8) are post hoc and cannot change a planned decision; Extended Appendices W and X give them in full.

\appsub{U.1}{Run 7: placement controls, wrappers on hard items and stale shares}
\textbf{U.1 Run 7: placement controls, wrappers on hard items and stale shares.} The plan (\code{reports/\allowbreak{}preregistration\_\allowbreak{}run7.\allowbreak{}md}) was frozen at 2026-11-10 11:32:14 UTC by a manifest of 78 files chained to Run 6's, and all six earlier manifests were verified before sending and before analysis. A 20-request pilot passed; the run sent 52,568 new requests, all answered on the first pass, and reused 16,588 cached requests (Run 4 and Run 6 arms). It cost \$3.571 against a calibrated estimate of \$4.03 and recorded no validity problem and no deviation. Its items were reused, not fresh: for placement (R7-E6), the Run 6 items of MASSIVE (311) and LEDGAR (380); for wrappers (R7-E5), the Run 4 items of MASSIVE (331) and MetaTool (535) whose cached bare top probability was below 0.7, a label-free selection; for shares (R7-E7), 1,900 of Run 4's 2,000 CFPB 2026 complaints, with 100 held out to estimate shares. Holm ran over seven tests, with R7-E7's p-values doubled first because CFPB has 8 classes.

\begingroup\footnotesize
\renewcommand{\thetable}{U.1}
\begin{longtable}{@{}>{\raggedright\arraybackslash}p{\dimexpr 0.4029\linewidth-2.4174\tabcolsep\relax}>{\raggedright\arraybackslash}p{\dimexpr 0.3005\linewidth-1.8031\tabcolsep\relax}>{\raggedright\arraybackslash}p{\dimexpr 0.0986\linewidth-0.5913\tabcolsep\relax}>{\raggedright\arraybackslash}p{\dimexpr 0.1980\linewidth-1.1882\tabcolsep\relax}@{}}
\caption{Run 7's seven tests (accuracy points unless stated; 90\% gold-class intervals).}\\
\toprule
Test & Estimate {[}90\% CI{]} & Holm p & Decision \\
\midrule
\endfirsthead
\multicolumn{4}{@{}l}{\textit{Table U.1 (continued)}}\\
\toprule
Test & Estimate {[}90\% CI{]} & Holm p & Decision \\
\midrule
\endhead
\bottomrule
\endlastfoot
R7-E5a: persona and politeness each within \ensuremath{\pm}2 of a concurrent bare re-send, hard items (n = 866) & persona +0.59 {[}\ensuremath{-}1.80, +3.06{]}; politeness \ensuremath{-}0.35 {[}\ensuremath{-}1.18, +0.42{]} & 0.44 & not shown \\
R7-E5b: the same within \ensuremath{\pm}0.10 nats of mean log p(gold) & persona +0.01 {[}\ensuremath{-}0.02, +0.04{]}; politeness \ensuremath{-}0.02 {[}\ensuremath{-}0.04, \ensuremath{-}0.01{]} & 0.00035 & within \ensuremath{\pm}0.10 nats \\
R7-E6a: one-rival penalty inside equal-length descriptions minus a block with non-empty descriptions (n = 691) & +7.40 {[}+5.52, +9.33{]} (MASSIVE +10.9, LEDGAR +4.0) & 0.00035 & confirmed \\
R7-E6b: penalty inside descriptions minus the same text grouped in the instruction, each label named once & +8.12 {[}+5.68, +10.44{]} & 0.00035 & confirmed \\
R7-E7a: stale 2025 shares stated minus true 2026 shares stated (n = 1,900) & \ensuremath{-}11.53 {[}\ensuremath{-}20.37, \ensuremath{-}1.35{]} & 0.083 & observed (item-level only) \\
R7-E7b: shares from 100 labelled 2026 items within \ensuremath{\pm}1.5 of true shares & +0.74 {[}\ensuremath{-}0.23, +1.61{]} & 0.44 & observed (item-level only) \\
R7-E7c: label-free EM correction after the call no more than 1.5 below stating true shares & \ensuremath{-}0.68 {[}\ensuremath{-}2.34, +1.13{]} & 0.48 & not shown \\
\end{longtable}
\endgroup

\emph{Secondary.} One-rival penalties by placement were 13.7 (descriptions), 11.7 (equal-length descriptions), 5.6 (grouped block), 4.5 (plain block) and 4.3 (block with neutral descriptions). Length matching within the descriptions removed little (+1.95 {[}\ensuremath{-}0.18, +4.00{]}), and adding neutral descriptions to the block changed nothing (\ensuremath{-}0.13 {[}\ensuremath{-}0.94, +0.70{]}). With three wrong labels on one rival, the model chose that rival on 84\% of MASSIVE answers with the examples in descriptions and 38\% with them in the block (LEDGAR 28\% and 0.9\%). On the CFPB items accuracy was 77.1\% bare, 79.2\% with true shares stated, 67.7\% with stale shares stated and 80.0\% with 100-item shares; stated stale shares pushed 23\% of predictions to credit reporting, which is 3\% of the items, and applying the stale shares after the call lost less than stating them (+4.89 {[}+0.08, +8.36{]}). On the hard items, two identical bare re-sends differed by +0.03 and +0.54 points and kept the cached top option on 91 to 92\% of pairs.

\appsub{U.2}{Run 8: fresh items across vendors}
\textbf{U.2 Run 8: fresh items across vendors.}

\emph{Freeze.} The plan (\code{reports/\allowbreak{}preregistration\_\allowbreak{}run8.\allowbreak{}md}, Sections 0 to 10) gained three dated addenda: Section 11 (placement and stale shares on every model; pool noise on every model, test N1 of the per-model families, was already in freeze 3, and Section 11 only restates it as H3), Section 12 (the chat model's readout of five top log-probabilities) and Section 13 (its reply length of 16 tokens). Freezes 1 and 2 were superseded before any request (a process-listing decoding error, and a dry run that digested only part of the sealed items). The original arms were sent under freeze 3 (12:43:04 UTC); no placement or stale-share request was sent before freeze 4 (13:55:30 UTC, aggregate \code{7c01\allowbreak{}091b\allowbreak{}1059\allowbreak{}9089\allowbreak{}f857\allowbreak{}1bb3\allowbreak{}ffac\allowbreak{}e1a5\allowbreak{}d3de\allowbreak{}94e6\allowbreak{}11e5\allowbreak{}cfd7\allowbreak{}1ca4\allowbreak{}5e89\allowbreak{}18c5\allowbreak{}65a8\allowbreak{}}), freeze 5 (14:31:03 UTC, \code{4402\allowbreak{}1107\allowbreak{}f0db\allowbreak{}c7ec\allowbreak{}78fc\allowbreak{}d53b\allowbreak{}a34d\allowbreak{}f278\allowbreak{}c757\allowbreak{}c5f3\allowbreak{}f401\allowbreak{}1c6e\allowbreak{}95c7\allowbreak{}1f80\allowbreak{}a8d4\allowbreak{}2efe\allowbreak{}}) preceded any chat-model request, and no chat-model response existed before freeze 6 (17:52:28 UTC). Each manifest holds 90 files and chains to Runs 6, 5, 4 and 3 and both Gates; the chain verified after the analysis. \emph{Timing of the analysis.} The plan's Section 11 states that no Run 8 result had been analysed when it was written, and Sections 12 and 13 were prompted by pilot failures (HTTP 400, nothing billed), not by outcomes. File times agree: every Run 8 analysis output (\code{run8\_\allowbreak{}primary}, \code{\_\allowbreak{}secondary}, \code{\_\allowbreak{}transfer}, \code{\_\allowbreak{}descriptive}, \code{\_\allowbreak{}chat\_\allowbreak{}reading}, \code{\_\allowbreak{}latency}, \code{\_\allowbreak{}timesplit} and \code{run8\_\allowbreak{}addendum\_\allowbreak{}transfer}) was first created at 18:51:00 to 18:52:42 UTC, after freeze 6 (17:52:32 UTC) and the last billed call (18:34:47 UTC); no other project file written between freeze 3 and 18:51 UTC holds a Run 8 outcome, and the worker logs record only sends, spend and failures. File times cannot rule out an unrecorded look at raw responses in the cache or the terminal.

\emph{Run.} Run 8 collected 354,806 responses: 139,909 from Jev 1.13 (including 1,200 latency calls), 32,667 to 32,856 from each of the six other decision models and 17,952 from the chat model. Every pilot passed except the chat model's first two, which returned HTTP 400 and billed nothing. It cost \$17.814 against a \$25 target: Jev \$8.65, OpenAI Decisions \$3.35, Microsoft-Decision-1 \$1.29, Kev-4B \$1.22, d1-ultra \$1.18, the chat model \$1.13, Decider \$0.66 and Decider Flash \$0.33. The ledger went from \$46.199 to \$64.013.

\emph{Items.} Fresh items passed the Gate rules, had an id never logged in any request, and had no exact or TF-IDF \ensuremath{\geq} 0.9 twin among sent texts; a cache-miss check covered every model's request keys. They were the in-scope rows of the CLINC150 validation split (150 intents; 1,500 for Jev, 600 for the other models), LEDGAR (25 classes; the 555 in-scope items for Jev, 400 for the others) and MetaTool (20 tools; 399). The time split used CFPB complaints, a 2023 pool and 2,000 complaints from 2024 (8 products). The stale-share test reused Run 7's 1,000 CFPB 2026 items. Confusable noise drew each wrong label from the three classes whose centroids, in the pool's own dense embeddings, lie nearest the true class's centroid (\code{stages/\allowbreak{}run8.\allowbreak{}py}); no model's answers were used.

\emph{Analysis.} Paired item accuracy, gold-class clusters within each dataset, dataset-equal pooling and 20,000 draws; the label-budget tests also resampled pool draws. Holm ran within each family: Jev 12 tests, Jev small controls 6, each other model 4 (and one correction over all 24), the chat model 6, and the addendum 12 across models. A claim transfers if confirmed on at least 4 of 6 models with none opposite.

\begingroup\footnotesize
\renewcommand{\thetable}{U.2}
\begin{longtable}{@{}>{\raggedright\arraybackslash}p{\dimexpr 0.3546\linewidth-2.1279\tabcolsep\relax}>{\raggedright\arraybackslash}p{\dimexpr 0.2384\linewidth-1.4303\tabcolsep\relax}>{\raggedright\arraybackslash}p{\dimexpr 0.0986\linewidth-0.5913\tabcolsep\relax}>{\raggedright\arraybackslash}p{\dimexpr 0.3084\linewidth-1.8505\tabcolsep\relax}@{}}
\caption{Run 8 on Jev 1.13 (accuracy points; 90\% gold-class intervals; Holm within the family).}\\
\toprule
Test & Estimate {[}90\% CI{]} & Holm p & Decision \\
\midrule
\endfirsthead
\multicolumn{4}{@{}l}{\textit{Table U.2 (continued)}}\\
\toprule
Test & Estimate {[}90\% CI{]} & Holm p & Decision \\
\midrule
\endhead
\bottomrule
\endlastfoot
R8-N1: 30\% confusable pool labels minus clean pool (n = 2,454) & \ensuremath{-}7.02 {[}\ensuremath{-}8.16, \ensuremath{-}5.95{]} & 0.0006 & confirmed (harm) \\
R8-N2: 30\% uniform pool labels minus clean pool & \ensuremath{-}3.94 {[}\ensuremath{-}4.98, \ensuremath{-}2.92{]} & 0.0006 & confirmed (harm) \\
R8-N3: confusable minus uniform at 30\% & \ensuremath{-}3.08 {[}\ensuremath{-}3.94, \ensuremath{-}2.28{]} & 0.0006 & confirmed (harm) \\
R8-N4: 30\%-confusable pool prompt minus bare (two-sided) & +0.69 {[}\ensuremath{-}1.52, +2.87{]} & 1 & not shown (two-sided) \\
R8-N5: audit the 10\% most retrieved minus audit a random 10\% & +0.27 {[}\ensuremath{-}0.42, +0.93{]} & 1 & not shown \\
R8-T1: ids with definitions no more than 2 below names with definitions (n = 954) & \ensuremath{-}2.31 {[}\ensuremath{-}3.81, \ensuremath{-}1.00{]} & 1 & not shown \\
R8-T2: one rival minus distinct rivals, k = 2 and 3 (n = 2,213) & \ensuremath{-}13.76 {[}\ensuremath{-}15.14, \ensuremath{-}12.49{]} & 0.0006 & confirmed (harm) \\
R8-T3: examples prompt at 100 per class minus bare & +7.71 {[}+5.58, +9.89{]} & 0.0006 & confirmed \\
R8-L1: at 100 per class, local combination within \ensuremath{\pm}1 of the prompt (n = 2,055) & \ensuremath{-}0.73 {[}\ensuremath{-}1.55, +0.06{]} & 1 & not shown \\
R8-L2: at 5 per class, bare \ensuremath{\times} centroid no more than 1.5 below the prompt & \ensuremath{-}1.19 {[}\ensuremath{-}2.03, \ensuremath{-}0.45{]} & 1 & not shown \\
R8-L3: at 5 per class, prompt minus bare & +3.09 {[}+1.78, +4.61{]} & 0.0007 & confirmed \\
R8-S1: natural-share pool prompt minus bare, LEDGAR (n = 555) & +1.89 {[}\ensuremath{-}0.44, +4.90{]} & 0.64 & not shown \\
R8-W1: capitals line within \ensuremath{\pm}2 of bare (n = 799) & +0.25 {[}\ensuremath{-}0.53, +0.99{]} & 0.0003 & within \ensuremath{\pm}2 \\
R8-W2: seven packed questions within \ensuremath{\pm}2 of the probe alone, five wordings & +0.30 {[}\ensuremath{-}0.02, +0.67{]} & 0.0003 & within \ensuremath{\pm}2 \\
R8-W3: true shares stated minus bare, LEDGAR & \ensuremath{-}0.36 {[}\ensuremath{-}1.01, +0.36{]} & 0.82 & not shown \\
R8-W4: shuffled shares stated minus bare, LEDGAR & \ensuremath{-}1.08 {[}\ensuremath{-}2.00, \ensuremath{-}0.24{]} & 0.050 & confirmed (harm) as planned; observed after the wild-cluster re-check (Holm 0.093; U.4) \\
R8-W5: persona and politeness within \ensuremath{\pm}2 of a length-matched neutral prefix & +1.08 {[}\ensuremath{-}0.02, +2.22{]} & 0.19 & not shown \\
R8-W6: names with blind definitions minus names, MetaTool (n = 399) & +6.52 {[}+1.35, +12.30{]} & 0.057 & not shown \\
R8-H1 (addendum): one-rival penalty inside length-matched descriptions minus block, CLINC150 and MetaTool (n = 512) & +6.35 {[}+3.30, +9.38{]} (CLINC150 +11.2, MetaTool +1.5) & 0.00075 & confirmed \\
R8-H2 (addendum, reference): stale minus true shares stated, Run 7's cached responses (n = 1,000) & \ensuremath{-}11.40 {[}\ensuremath{-}20.57, \ensuremath{-}1.17{]} & -- & reference only \\
\end{longtable}
\endgroup

\begingroup\footnotesize
\renewcommand{\thetable}{U.3}
\begin{longtable}{@{}>{\raggedright\arraybackslash}p{\dimexpr 0.2480\linewidth-2.9759\tabcolsep\relax}>{\raggedright\arraybackslash}p{\dimexpr 0.1273\linewidth-1.5282\tabcolsep\relax}>{\raggedright\arraybackslash}p{\dimexpr 0.1153\linewidth-1.3834\tabcolsep\relax}>{\raggedright\arraybackslash}p{\dimexpr 0.1273\linewidth-1.5282\tabcolsep\relax}>{\raggedright\arraybackslash}p{\dimexpr 0.1273\linewidth-1.5282\tabcolsep\relax}>{\raggedright\arraybackslash}p{\dimexpr 0.1273\linewidth-1.5282\tabcolsep\relax}>{\raggedright\arraybackslash}p{\dimexpr 0.1273\linewidth-1.5282\tabcolsep\relax}@{}}
\caption{Run 8 on the other models (accuracy points; 90\% gold-class intervals). Bold: confirmed under one correction over the 24 tests (12 for the addendum); ``per model'': confirmed only under the per-model correction. The chat model ran on MetaTool only. Its row uses the plan's words: ``holds'' means confirmed, ``unresolved'' not shown, and ``wrong side'' opposite.}\\
\toprule
Model & R8-T1 ids vs names (\ensuremath{-}2) & R8-T2 one rival & R8-T3 retrieval & R8-N1 30\% confusable & R8-H1 placement & R8-H2 stale shares \\
\midrule
\endfirsthead
\multicolumn{7}{@{}l}{\textit{Table U.3 (continued)}}\\
\toprule
Model & R8-T1 ids vs names (\ensuremath{-}2) & R8-T2 one rival & R8-T3 retrieval & R8-N1 30\% confusable & R8-H1 placement & R8-H2 stale shares \\
\midrule
\endhead
\bottomrule
\endlastfoot
Decider v1.2 & \ensuremath{-}1.13 {[}\ensuremath{-}2.45, +0.13{]} & \textbf{\ensuremath{-}4.73} & \textbf{+3.98} & \textbf{\ensuremath{-}4.41} & \ensuremath{-}15.89 (opposite) & \ensuremath{-}9.90 \\
Decider Flash & \ensuremath{-}2.06 {[}\ensuremath{-}4.57, +0.09{]} & \textbf{\ensuremath{-}6.30} & \ensuremath{-}14.51 (opposite) & \textbf{\ensuremath{-}4.54} & \textbf{+8.99} & \ensuremath{-}4.10 \\
OpenAI Decisions & \ensuremath{-}3.57 {[}\ensuremath{-}6.15, \ensuremath{-}1.21{]} & \textbf{\ensuremath{-}7.34} & \textbf{+7.22} & \textbf{\ensuremath{-}7.00} & \ensuremath{-}4.02 (opposite) & \ensuremath{-}12.41 \\
Microsoft-Decision-1 & \ensuremath{-}1.50 {[}\ensuremath{-}2.73, \ensuremath{-}0.50{]} & \textbf{\ensuremath{-}10.09} & \textbf{+3.19} & \textbf{\ensuremath{-}10.65} & \textbf{+7.52} & \ensuremath{-}11.40 (per model) \\
Liquid d1-ultra & \textbf{+0.38} {[}\ensuremath{-}0.14, +0.91{]} & \textbf{\ensuremath{-}6.33} & +1.06 & \textbf{\ensuremath{-}28.96} & \textbf{+3.16} & \ensuremath{-}16.30 (per model) \\
Kev-4B & \ensuremath{-}2.56 {[}\ensuremath{-}5.20, \ensuremath{-}0.27{]} & \textbf{\ensuremath{-}2.80} & \textbf{+5.63} & \textbf{\ensuremath{-}20.15} & +3.92 (per model) & \ensuremath{-}3.90 \\
Verdict & partially transfers (1 of 6) & transfers (6 of 6) & partially transfers (4 of 6, one opposite) & transfers (6 of 6) & partially transfers (3 of 6, two opposite) & does not transfer (0 of 6) \\
Chat model & \ensuremath{-}0.75 {[}\ensuremath{-}3.74, +1.80{]}, unresolved & \ensuremath{-}10.20, holds & +12.16, holds & \ensuremath{-}11.65, holds & \ensuremath{-}4.81, wrong side & \ensuremath{-}12.80, unresolved \\
\end{longtable}
\endgroup

The chat model also confirmed the definitions rescue (+75.56 {[}+65.43, +84.28{]} over ids alone) and left the wrappers unresolved (+0.35 {[}\ensuremath{-}1.29, +2.05{]}).

\emph{Secondary estimates on Jev.} Against the clean pool, uniform noise at 10, 30 and 50\% cost 1.38, 3.94 and 10.22 points and confusable noise 1.82, 7.02 and 19.34; against the bare call these pools gave +8.5, +3.8 and \ensuremath{-}0.3 (uniform) and +8.1, +0.7 and \ensuremath{-}9.4 (confusable). Where the bare top probability was 0.5 to 0.8, 30\% confusable noise cost 15.5 and uniform 7.9; above 0.8, 5.3 and 2.9. Auditing 10\% of a 30\%-confusable pool gained +0.98 {[}+0.31, +1.63{]} (most retrieved first) or +0.72 {[}+0.24, +1.20{]} (random). Ids alone cost 70 to 81 points against names on every model, and definitions won back 76 to 84; the one-rival gap at three labels was \ensuremath{-}20.3 on Jev and \ensuremath{-}2.3 to \ensuremath{-}14.0 elsewhere. The prompt gained +3.62 over bare at 20 per class and +4.67 at 100 (CLINC150 and LEDGAR); bare \ensuremath{\times} tuned classifier at 5 per class was \ensuremath{-}0.98 {[}\ensuremath{-}1.80, \ensuremath{-}0.25{]} against the prompt. The examples prompt beat label-matched random examples by +1.76 {[}+0.78, +2.94{]} (random examples themselves +3.10 over bare), a fixed one-shot set by +3.21 {[}+1.53, +5.36{]}, contextual calibration by +9.96 (which itself scored \ensuremath{-}2.24 {[}\ensuremath{-}4.54, +0.04{]} against bare) and batch calibration by +7.49 (+0.22 against bare). On LEDGAR the natural-share pool scored \ensuremath{-}1.44 {[}\ensuremath{-}2.97, \ensuremath{-}0.19{]} below a balanced one, and the local combination on the natural pool beat bare by +2.43 {[}+0.44, +4.82{]}. On the 2023 to 2024 split, accuracy was 81.5\% bare, 84.1\% with the prompt, 86.7\% for a bare \ensuremath{\times} prompt product with the old shares and 86.6\% with EM-re-estimated shares; the prompt beat bare by +2.58 {[}+1.75, +3.45{]} (also with class clusters, {[}+1.69, +3.03{]}), re-estimation changed the product by \ensuremath{-}0.10 {[}\ensuremath{-}0.58, +0.39{]} although it brought the shares from a total-variation distance of 0.199 to 0.042, and it cut worst-class recall from 49.5\% to 22.2\%. On fresh items persona minus bare was +0.83 {[}\ensuremath{-}0.08, +1.80{]} and politeness +0.15 {[}\ensuremath{-}0.31, +0.61{]}. Sequential latency (n = 200 each) had p50 0.096 s and p99 0.210 s bare and 0.198 s with examples; with eight requests in flight p99 was 0.257 s and 0.277 s; the differences at p99 were \ensuremath{-}0.012 s {[}\ensuremath{-}0.19, +0.22{]} and +0.020 s {[}\ensuremath{-}0.04, +0.09{]}, and local embedding and top-20 search took 8 ms at p50 and 13 ms at p99 on CPU. On the same items, the one-rival penalty was 4.3 to 11.6 points smaller on every other model than on Jev, and the retrieval gain smaller on four of them.

\emph{Validity and anomalies.} OpenAI Decisions refused 189 requests after one re-send, so its ids-alone cells fell below 98\% complete (LEDGAR 327 of 400, MetaTool 330 of 399); they feed only secondary estimates, and T1, T3 and N1 are unchanged with refusals scored as errors. Two anomalies are unexplained. \textbf{Decider Flash on LEDGAR}: whenever correctly labelled examples were shown, its accuracy fell far below the bare call's 0.93 (examples prompt 0.39, noisy pool 0.40, clean constructed display 0.45), while relabelled displays of the same length scored 0.86 to 0.94, so wrong labels raised its LEDGAR accuracy and its one-rival penalty is reversed there; the pattern held under both wordings and over the send window, did not track prompt length (all under 2,900 tokens), and drives Flash's opposite retrieval result and its ``not consistent across datasets'' flags. \textbf{d1-ultra at two wrong labels}: it scored far lower with two wrong labels than with three (LEDGAR distinct rivals 0.31 against 0.87, one rival 0.22 against 0.86), the non-monotone curve already seen on MetaTool in Run 5, now also on LEDGAR and CLINC150; k = 2 requests were also slower in Run 5.

\emph{Deviations} (all tabulated in the run log with reason and effect): (1) and (2) freezes 1 and 2 superseded before any request, the second because the dry run digested only the first 1,982 sealed items, so LEDGAR on Jev has 555 items instead of 800 (standard error about 20\% larger); (3) workers ran one vendor at a time after database write-lock starvation (timing only); (4) the hard spending ceiling was fixed at \$76.20 after the launcher's ``ledger + \$30'' drifted (stricter than planned); (5) the budget rule was applied per launch, with OpenRouter caps at 1.25 times the estimate; (6) the placement test on CLINC150 used 200 items instead of 400, for cost (lower power); (7) the chat readout used the top five log-probabilities and a 16-token reply, because the API caps the first and rejects one-token replies (addenda 2 and 3; accuracy readout unchanged, no log-probability summary); (8) OpenAI Decisions' ids-alone cells were below 98\% complete (above); (9) MetaTool used all 399 fresh items, as planned; (10) a fill script dropped the dollar figures of one sentence of Section 11 before freeze 6 (cosmetic; the hashed text is as frozen). No arm was dropped.

\appsub{U.3}{Re-analysis 2: Runs 1 to 6}
\textbf{U.3 Re-analysis 2: Runs 1 to 6} (observed; cache only, no API call; 136 of 136 reproduction checks matched).

\begin{itemize}
\item \emph{Sizes of the class-level checks.} Under re-analysis 1's stress null, the pairs class bootstrap used by the paper's class checks rejected 8.5\% (20 to 25 classes) and 8.8\% (8 classes) at a nominal two-sided 5\%, and 5.85\% with 49 or more classes; its excess grows at small \ensuremath{\alpha}. The wild-cluster restricted bootstrap with Webb weights (WCR {[}\citenum{r49}{]}) rejected 4.9 to 5.2\% one-sided at 5\% in every bin with 8 or more classes.
\item \emph{Re-decisions.} Of the 16 tests then confirmed in Runs 1 to 4, 14 survive WCR and both stress calibrations. T3a (prompt vs fixed vote, tweets; +17.0) fails all three (WCR p .017, Holm .12) and is reported as observed. T8 (packed questions) survives WCR (Holm .022) but fails both calibrations (Holm .060 and .067) and is marked fragile; Run 8 re-tested packed questions under five wordings (within \ensuremath{\pm}2). Every confirmed test of Runs 5 and 6 survives WCR (29 tests, a different set from the 29 tests of Runs 1 to 4), and the Run 5 transfer counts are unchanged (wrappers 4, retrieval 6, wrong labels 6, definitions 6 of 7; the bin calibration alone drops d1-ultra's retrieval, leaving 5 of 7).
\item \emph{One family over the 29 tests.} 14 of 16 survive with the planned p, 13 with WCR class-level p (T3a, T8 and H3 fail), and 11 under a hierarchical split with \ensuremath{\alpha}/4 per run (T1c and T7a fail as well).
\item \emph{Run 5 as random effects} (six models; joint bootstrap over shared classes and stimuli; t on 5 df): persona \ensuremath{-}0.15 {[}\ensuremath{-}0.84, +0.54{]} and politeness \ensuremath{-}0.19 {[}\ensuremath{-}0.76, +0.38{]} (between-model SD 0.21 and 0.46; new-model prediction intervals {[}\ensuremath{-}0.96, +0.65{]} and {[}\ensuremath{-}1.28, +0.90{]}, inside \ensuremath{\pm}2); retrieval +5.64 {[}+2.02, +9.26{]} (SD 3.51; new model \ensuremath{-}2.30 to +13.58); three wrong labels \ensuremath{-}27.65 {[}\ensuremath{-}34.57, \ensuremath{-}20.73{]} (SD 6.27; Jev 1.13's \ensuremath{-}60.56 is 32.9 points below, z \ensuremath{-}4.6); definitions +78.65 {[}+71.92, +85.37{]} (SD 4.05).
\item \emph{OpenAI Decisions' refusals.} 128 of the 130 missing item-paraphrases in its definitions cells are ids-alone controls. With refused controls scored right and refused treatments wrong, the definitions gain is +58.13 {[}+52.06, +63.82{]}, still far above the +10 margin, so the verdict stays 6 of 7; the other models' gains on the refused items resemble their gains on answered ones.
\item \emph{The 100-per-class pool.} No item is shared between the 15 comparisons, but all classes are shared within a source. Sharing classes changes little, while sharing pools roughly triples the standard error (13 comparisons: 0.19 \ensuremath{\rightarrow} 0.59). Table E.10b gives the five sets.
\end{itemize}

\appsub{U.4}{Re-analysis 3: Runs 7 and 8}
\textbf{U.4 Re-analysis 3: Runs 7 and 8} (observed; cache only, no API call; 286 of 286 reproduction checks matched; \code{r18\_\allowbreak{}*.\allowbreak{}csv}).

\begin{itemize}
\item \emph{Wild-cluster re-decision.} Every Run 7 and Run 8 test was re-decided with the wild-cluster restricted bootstrap (Webb weights, 9,999 draws, one-sided p, each run's own Holm families; the 8-class doubling dropped, since WCR holds its size there). One decision changes: R8-W4 (shuffled shares on LEDGAR, 25 classes) has WCR p 0.023 and Holm 0.093 (frozen Holm 0.050), in all five Monte Carlo seeds, so it is reported as observed; its WCR interval {[}\ensuremath{-}1.99, \ensuremath{-}0.17{]} still lies below zero. Run 7 stays at 3 of 7, Run 8's Jev family at 6 of 12, and the chat model at 4 of 6. All 24 vendor decisions stand, per model and jointly, and so does every transfer count: retrieval on fresh items 4 of 6 (Microsoft-Decision-1 at joint Holm 0.0496, 0.045 to 0.049 over five seeds), pool noise and the one-rival penalty 6 of 6, placement 3 of 6 with the same two opposite (Kev-4B: +3.92, per-model Holm 0.027, joint 0.12), stale shares 0 of 6.
\item \emph{Pooled cross-vendor effects} (the six Run 8 vendors, random effects as in U.3, t on 5 df): 30\% confusable pool noise \ensuremath{-}12.62 {[}\ensuremath{-}20.91, \ensuremath{-}4.32{]} (between-model SD 9.86); one-rival penalty \ensuremath{-}6.27 {[}\ensuremath{-}8.38, \ensuremath{-}4.15{]} (SD 2.37; Jev 1.13's \ensuremath{-}13.76 lies 7.5 below the vendor mean); fresh-item retrieval +1.10 {[}\ensuremath{-}5.79, +7.99{]} (SD 7.86), or +4.22 {[}+0.94, +7.49{]} without Decider Flash. Pool noise and the one-rival penalty keep their sign when vendor families or base models are averaged first, or when Kev-4B is left out. The pooled numbers summarise these six models, not decision models in general.
\item \emph{Families} (\code{r18\_\allowbreak{}families.\allowbreak{}csv}; lineage partly from secondary sources). Decider and Decider Flash are one Perplexity family; Microsoft-Decision-1 (9B) and Kev-4B (4B) share the Qwen3.5 generation, and Kev-4B is an open Jev clone; the chat model is the same model as OpenAI Decisions, read through letter log-probabilities. With one vote per vendor family (counts, not pre-registered verdicts): pool noise 5 of 5, one-rival penalty 5 of 5, definitions 5 of 5 tested, retrieval on fresh items 4 of 5 (any member) or 3 of 5 (all members), placement 3 of 5 or 2 of 5, with the OpenAI family opposite and the Perplexity family split.
\end{itemize}

\setcounter{section}{24}
\section{Runs 9 and 10: four private synthetic datasets and the names experiment}

Run 9 asked whether the effects measured on public benchmarks are distorted by training-data leakage, and how much option names and descriptions matter (Sections 4.1 and 7). Run 9 ran on Jev 1.13; Run 10 repeated it on the six other decision models and on Jev 1.14 (Y.7). Extended Appendix Y gives the label sets, per-dataset intervals, the generation prompts in outline, Tables Y.7 and Y.8, and Figures Y.1 to Y.3; Figure Y.4, the main text's Figure 6 up to v23, is in Y.6.

\appsub{Y.1}{Plan and analysis}
\textbf{Y.1 Plan and analysis.} The plan (\code{reports/\allowbreak{}preregistration\_\allowbreak{}run9.\allowbreak{}md}) was written on 2026-11-10 before any Run 9 request to Jev. A manifest (\code{registry\_\allowbreak{}post/\allowbreak{}run9/\allowbreak{}run9\_\allowbreak{}manifest.\allowbreak{}json}) froze the code hashes, the private items, the blind definitions and every dataset's item ids, texts and request set, and every live send checked it. The plan is internal and not publicly archived. Every request was built by Run 8's code, so the arms are byte-identical to Run 8's wherever Run 8 sent them; those requests were cache hits, and the open effects reproduce Run 8 (retrieval +7.7, one rival \ensuremath{-}13.8, 30\% confusable noise \ensuremath{-}7.0). Two instruction wordings were used, except for the wrappers, which used the first wording on the first 400 items of each dataset. An item's value is the mean over wordings of {[}top option = gold{]}; effects are paired by item, with 90\% wild-cluster restricted bootstrap-t intervals (Webb weights, 9,999 draws, gold classes within each dataset), as in re-analyses 2 and 3. Pooled effects are dataset-equal means within each side (three open, four closed datasets). The primary contrast per technique is open \ensuremath{-} closed, two-sided, with Holm over the eight techniques. Secondary: the three matched pairs, the Spearman correlation of the eight pooled effects (pairs class bootstrap; also without the two largest), the bare-call accuracy gap, the names experiment and every per-dataset effect. Only items complete in both arms of a contrast enter it.

\emph{Fixed criteria.} A rule whose open effect is confirmed counts as distorted if the closed interval lies wholly on the other side of zero, or if the open \ensuremath{-} closed difference passes Holm and is at least 2 points and at least half the open effect. A null (descriptions on names, true shares, wrappers) counts as masked if it helps on closed data only and the difference passes Holm. The ranking is preserved if the lower end of the Spearman interval is at least 0.5, and distorted if its upper end is below 0.5. The bare-call gap is reported but never read alone as leakage, because the private sets are synthetic and model-validated. Absent all of these, the reading is ``no distortion large enough to detect at this size'', not ``no leakage''.

\emph{Exploratory difficulty adjustment} (added after the first results, not in the plan). Per technique, a random-effects meta-regression of the per-dataset effect on headroom (100 \ensuremath{-} bare accuracy) with an open/closed indicator: \ensuremath{\tau}\textsuperscript{2} by moments, Hartung--Knapp--Sidik--Jonkman variance and t on k \ensuremath{-} 3 degrees of freedom. The shift is open minus closed at the same headroom. It was repeated on bare-call log-odds, without ScotLease, and for the pair OpsDeck-24 and MetaTool.

\appsub{Y.2}{The private datasets}
\textbf{Y.2 The private datasets.} Every label has a public name, a public one-line definition (used only in the descriptions arm) and a private generation brief given to the generators only, so that items do not paraphrase the definitions (MetaTool's queries were generated from its descriptions {[}\citenum{r16}{]}). Each dataset has a glossary of invented people, firms, places and products; real legal and physics terms stay, because they are the jargon being tested. Items vary in persona (ten per dataset), style, length and spelling noise, and 30\% of generation batches were asked for near-misses that share vocabulary with a named confusable neighbour. The generators (Claude Haiku 5.5, Gemini 3.1 Flash-Lite) and the validator (Mistral Large) were called through OpenRouter with data collection denied; OpenAI, Qwen and Liquid models were excluded, because they back tested models. The validator labelled every item blind; when it disagreed with the intended label, the other generator labelled it blind and the majority of three decided; a three-way split or an UNCLEAR majority dropped the item. De-duplication removed exact duplicates, TF-IDF cosine \ensuremath{\geq} 0.90 and bge-small cosine \ensuremath{\geq} 0.95 within each dataset; an item with TF-IDF cosine \ensuremath{\geq} 0.90 to any text of CLINC150, Banking77, MASSIVE, LEDGAR, MetaTool, the tweets or ToxicChat was dropped as an open twin. Kept items were split per class in a seeded order that alternates the generators: the test quota first, then 100 for the pool. The items are stored only on the author's machine and in the response cache, and were sent only to the two generators, the validator and Jev 1.13, and in Run 10 to the six other decision models and to Jev 1.14 through OpenRouter.

\begingroup\footnotesize
\renewcommand{\thetable}{Y.1}
\begin{longtable}{@{}>{\raggedright\arraybackslash}p{\dimexpr 0.1676\linewidth-2.0110\tabcolsep\relax}>{\raggedright\arraybackslash}p{\dimexpr 0.2294\linewidth-2.7527\tabcolsep\relax}>{\raggedright\arraybackslash}p{\dimexpr 0.1058\linewidth-1.2692\tabcolsep\relax}>{\raggedright\arraybackslash}p{\dimexpr 0.1058\linewidth-1.2692\tabcolsep\relax}>{\raggedright\arraybackslash}p{\dimexpr 0.1181\linewidth-1.4176\tabcolsep\relax}>{\raggedright\arraybackslash}p{\dimexpr 0.1305\linewidth-1.5659\tabcolsep\relax}>{\raggedright\arraybackslash}p{\dimexpr 0.1429\linewidth-1.7143\tabcolsep\relax}@{}}
\caption{The four private datasets. Agreement: the validator's label equals the intended one; final: the majority label equals the intended one. Duplicates: share of generated items removed as near-duplicates. Max cosine: 99th percentile of each item's highest TF-IDF cosine to any open text.}\\
\toprule
Dataset & Domain; labels & Test shares & Words (median) & Generated \ensuremath{\rightarrow} kept & Agreement; final & Duplicates; open twins; max cosine \\
\midrule
\endfirsthead
\multicolumn{7}{@{}l}{\textit{Table Y.1 (continued)}}\\
\toprule
Dataset & Domain; labels & Test shares & Words (median) & Generated \ensuremath{\rightarrow} kept & Agreement; final & Duplicates; open twins; max cosine \\
\midrule
\endhead
\bottomrule
\endlastfoot
KitchenOps-80 & commercial-kitchen assistant; 80 snake\us{}case intents with kitchen jargon & balanced & 12 & 12,263 \ensuremath{\rightarrow} 11,483 & 95.6\%; 98.0\% & 5.3\%; 6; 0.53 \\
OpsDeck-24 & developer requests; 24 invented two-word tools (brand + function word) with catalogue descriptions & Zipf 0.6 & 24 & 4,661 \ensuremath{\rightarrow} 4,606 & 94.4\%; 97.5\% & 0.4\%; 0; 0.34 \\
ScotLease-25 & Scottish commercial-lease clauses; 25 jargon-heavy types & Zipf 0.8 & 88 & 4,706 \ensuremath{\rightarrow} 4,278 & 99.7\%; 99.9\% & 9.1\%; 0; 0.30 \\
BeamDesk-14 & synchrotron support tickets; 14 teams & Zipf 1.0 & 105 & 3,135 \ensuremath{\rightarrow} 3,108 & 98.5\%; 99.5\% & 0.8\%; 0; 0.19 \\
\end{longtable}
\endgroup

Each dataset keeps about 1,000 test items and 100 pool items per class. Descriptions for the descriptions arm were blind one-line definitions written from the label names by Mistral Large, following Run 3's rules, for CLINC150, KitchenOps, ScotLease and BeamDesk; OpsDeck used its catalogue descriptions and MetaTool its own; LEDGAR used Run 8's. Stated shares were the population shares of each test set.

\appsub{Y.3}{Results per dataset}
\textbf{Y.3 Results per dataset.}

\begingroup\small
\renewcommand{\thetable}{Y.2}
\begin{longtable}{@{}llllllllll@{}}
\caption{Run 9 effects per dataset on Jev 1.13 (accuracy points). D2: cryptic ids vs readable names; D3: three wrong labels vs three correct (display); D4: one rival vs distinct rivals; D1: retrieved examples vs bare call; D5: 30\% confusable pool noise vs clean pool; N1: descriptions on readable names; N2: true shares stated; N3: persona and politeness. Open: CLINC150, LEDGAR, MetaTool; private: the other four. Intervals are in Extended Appendix Y.}\\
\toprule
Dataset & D2 & D3 & D4 & D1 & D5 & N1 & N2 & N3 & Bare (\%) \\
\midrule
\endfirsthead
\multicolumn{10}{@{}l}{\textit{Table Y.2 (continued)}}\\
\toprule
Dataset & D2 & D3 & D4 & D1 & D5 & N1 & N2 & N3 & Bare (\%) \\
\midrule
\endhead
\bottomrule
\endlastfoot
CLINC150 & \ensuremath{-}91.8 & \ensuremath{-}55.3 & \ensuremath{-}23.9 & +6.4 & \ensuremath{-}8.5 & \ensuremath{-}1.0 & \ensuremath{-}1.3 & \ensuremath{-}0.1 & 92.6 \\
LEDGAR & \ensuremath{-}88.1 & \ensuremath{-}24.6 & \ensuremath{-}5.0 & +3.3 & \ensuremath{-}3.7 & \ensuremath{-}0.4 & \ensuremath{-}0.4 & +0.1 & 91.4 \\
MetaTool & \ensuremath{-}68.3 & \ensuremath{-}68.7 & \ensuremath{-}12.4 & +13.4 & \ensuremath{-}8.9 & +8.4 & +5.0 & +1.1 & 76.1 \\
KitchenOps-80 & \ensuremath{-}87.7 & \ensuremath{-}70.3 & \ensuremath{-}19.5 & +8.6 & \ensuremath{-}8.9 & +2.8 & +1.7 & +0.9 & 89.0 \\
OpsDeck-24 & \ensuremath{-}66.6 & \ensuremath{-}96.6 & \ensuremath{-}7.8 & +27.0 & \ensuremath{-}17.1 & +25.0 & +2.5 & \ensuremath{-}4.2 & 69.4 \\
ScotLease-25 & \ensuremath{-}98.1 & \ensuremath{-}31.5 & \ensuremath{-}5.1 & +0.1 & \ensuremath{-}2.3 & \ensuremath{-}1.0 & \ensuremath{-}0.1 & \ensuremath{-}0.2 & 99.8 \\
BeamDesk-14 & \ensuremath{-}90.2 & \ensuremath{-}26.1 & \ensuremath{-}17.7 & +4.7 & \ensuremath{-}3.0 & +1.6 & +2.0 & +0.5 & 95.0 \\
\end{longtable}
\endgroup

\begingroup\small
\renewcommand{\thetable}{Y.3}
\begin{longtable}{@{}llll@{}}
\caption{Matched pairs: each open dataset minus its private mirror (points, 90\% intervals with classes resampled within each dataset). BeamDesk has no open mirror in this run.}\\
\toprule
Technique & CLINC150 \ensuremath{-} KitchenOps & MetaTool \ensuremath{-} OpsDeck & LEDGAR \ensuremath{-} ScotLease \\
\midrule
\endfirsthead
\multicolumn{4}{@{}l}{\textit{Table Y.3 (continued)}}\\
\toprule
Technique & CLINC150 \ensuremath{-} KitchenOps & MetaTool \ensuremath{-} OpsDeck & LEDGAR \ensuremath{-} ScotLease \\
\midrule
\endhead
\bottomrule
\endlastfoot
D2 & \ensuremath{-}4.2 {[}\ensuremath{-}8.5, 0.0{]} & \ensuremath{-}1.7 {[}\ensuremath{-}19.3, +16.0{]} & +10.0 {[}+0.1, +19.9{]} \\
D3 & +15.0 {[}+9.9, +20.1{]} & +27.9 {[}+20.1, +36.1{]} & +6.9 {[}\ensuremath{-}2.9, +16.2{]} \\
D4 & \ensuremath{-}4.4 {[}\ensuremath{-}6.6, \ensuremath{-}2.2{]} & \ensuremath{-}4.6 {[}\ensuremath{-}7.6, \ensuremath{-}1.5{]} & +0.1 {[}\ensuremath{-}3.0, +3.1{]} \\
D1 & \ensuremath{-}2.2 {[}\ensuremath{-}5.4, +1.0{]} & \ensuremath{-}13.6 {[}\ensuremath{-}26.8, \ensuremath{-}0.5{]} & +3.2 {[}+0.5, +6.1{]} \\
D5 & +0.4 {[}\ensuremath{-}2.1, +2.9{]} & +8.2 {[}+4.9, +11.5{]} & \ensuremath{-}1.4 {[}\ensuremath{-}3.7, +0.8{]} \\
N1 & \ensuremath{-}3.8 {[}\ensuremath{-}6.2, \ensuremath{-}1.5{]} & \ensuremath{-}16.6 {[}\ensuremath{-}28.3, \ensuremath{-}4.6{]} & +0.6 {[}\ensuremath{-}1.1, +2.4{]} \\
N2 & \ensuremath{-}3.0 {[}\ensuremath{-}4.5, \ensuremath{-}1.4{]} & +2.5 {[}\ensuremath{-}3.1, +8.2{]} & \ensuremath{-}0.3 {[}\ensuremath{-}1.1, +0.5{]} \\
N3 & \ensuremath{-}1.0 {[}\ensuremath{-}2.2, +0.2{]} & +5.2 {[}+2.7, +7.7{]} & +0.3 {[}\ensuremath{-}0.4, +1.0{]} \\
Bare-call accuracy & +3.6 {[}+0.1, +7.1{]} & +6.7 {[}\ensuremath{-}6.3, +19.6{]} & \ensuremath{-}8.4 {[}\ensuremath{-}14.0, \ensuremath{-}3.0{]} \\
\end{longtable}
\endgroup

The pairs differ in both directions, and the largest differences are harms that are larger on the private sets (the display on OpsDeck, \ensuremath{-}96.6) or helps that are larger there (descriptions and retrieval on OpsDeck). Most of each tool-catalogue difference, apart from about half of the display's, is not explained by headroom (Table Y.5).

\begingroup\footnotesize
\renewcommand{\thetable}{Y.4}
\begin{longtable}{@{}>{\raggedright\arraybackslash}p{\dimexpr 0.1323\linewidth-1.3231\tabcolsep\relax}>{\raggedright\arraybackslash}p{\dimexpr 0.1977\linewidth-1.9769\tabcolsep\relax}>{\raggedright\arraybackslash}p{\dimexpr 0.1977\linewidth-1.9769\tabcolsep\relax}>{\raggedright\arraybackslash}p{\dimexpr 0.1977\linewidth-1.9769\tabcolsep\relax}>{\raggedright\arraybackslash}p{\dimexpr 0.1977\linewidth-1.9769\tabcolsep\relax}>{\raggedright\arraybackslash}p{\dimexpr 0.0769\linewidth-0.7692\tabcolsep\relax}@{}}
\caption{Exploratory: the open \ensuremath{-} closed shift at matched difficulty (points; 90\% HKSJ intervals), by covariate and with or without ScotLease, and the common slope per point of headroom (all seven datasets). D3 and D4 are very heterogeneous between datasets, so their intervals are wide.}\\
\toprule
Technique & Headroom, all & Headroom, no ScotLease & Log-odds, all & Log-odds, no ScotLease & Slope \\
\midrule
\endfirsthead
\multicolumn{6}{@{}l}{\textit{Table Y.4 (continued)}}\\
\toprule
Technique & Headroom, all & Headroom, no ScotLease & Log-odds, all & Log-odds, no ScotLease & Slope \\
\midrule
\endhead
\bottomrule
\endlastfoot
D2 & \ensuremath{-}1.0 {[}\ensuremath{-}3.7, +1.7{]} & \ensuremath{-}1.1 {[}\ensuremath{-}5.2, +3.0{]} & \ensuremath{-}3.2 {[}\ensuremath{-}17.1, +10.7{]} & \ensuremath{-}1.3 {[}\ensuremath{-}9.8, +7.2{]} & +1.05 \\
D3 & +10.0 {[}\ensuremath{-}16.7, +36.7{]} & +10.0 {[}\ensuremath{-}26.7, +46.6{]} & +18.6 {[}\ensuremath{-}19.4, +56.7{]} & +12.8 {[}\ensuremath{-}20.0, +45.7{]} & \ensuremath{-}2.16 \\
D4 & \ensuremath{-}1.5 {[}\ensuremath{-}16.3, +13.3{]} & +2.0 {[}\ensuremath{-}13.5, +17.4{]} & \ensuremath{-}0.3 {[}\ensuremath{-}15.6, +14.9{]} & +1.4 {[}\ensuremath{-}14.1, +17.0{]} & +0.11 \\
D1 & \ensuremath{-}1.5 {[}\ensuremath{-}4.8, +1.9{]} & \ensuremath{-}2.0 {[}\ensuremath{-}6.2, +2.2{]} & \ensuremath{-}3.1 {[}\ensuremath{-}11.2, +5.0{]} & \ensuremath{-}2.8 {[}\ensuremath{-}8.5, +3.0{]} & +0.73 \\
D5 & +1.3 {[}\ensuremath{-}3.3, +5.9{]} & +1.6 {[}\ensuremath{-}4.7, +7.9{]} & +3.1 {[}\ensuremath{-}3.5, +9.8{]} & +2.2 {[}\ensuremath{-}4.0, +8.3{]} & \ensuremath{-}0.43 \\
N1 & \ensuremath{-}3.3 {[}\ensuremath{-}6.0, \ensuremath{-}0.6{]} & \ensuremath{-}2.9 {[}\ensuremath{-}6.3, +0.6{]} & \ensuremath{-}4.6 {[}\ensuremath{-}13.8, +4.6{]} & \ensuremath{-}3.9 {[}\ensuremath{-}10.4, +2.7{]} & +0.55 \\
N2 & \ensuremath{-}1.9 {[}\ensuremath{-}3.8, \ensuremath{-}0.1{]} & \ensuremath{-}1.6 {[}\ensuremath{-}4.6, +1.4{]} & \ensuremath{-}2.0 {[}\ensuremath{-}5.4, +1.5{]} & \ensuremath{-}1.6 {[}\ensuremath{-}4.8, +1.5{]} & +0.20 \\
N3 & +0.7 {[}\ensuremath{-}1.6, +2.9{]} & +0.7 {[}\ensuremath{-}2.5, +3.8{]} & +0.6 {[}\ensuremath{-}2.0, +3.1{]} & +0.7 {[}\ensuremath{-}2.5, +4.0{]} & \ensuremath{-}0.03 \\
\end{longtable}
\endgroup

Without ScotLease, and on bare-call log-odds, the N1 shift's interval spans zero, so the descriptions difference depends on the ceiling set and on how difficulty is measured.

\begingroup\small
\renewcommand{\thetable}{Y.5}
\begin{longtable}{@{}llllll@{}}
\caption{OpsDeck-24 against MetaTool, the two tool catalogues (points). Headroom is 30.6 points on OpsDeck and 23.9 on MetaTool; the explained part is the common slope times the difference in headroom.}\\
\toprule
Technique & OpsDeck & MetaTool & Raw gap & Explained by headroom & Remaining {[}90\%{]} \\
\midrule
\endfirsthead
\multicolumn{6}{@{}l}{\textit{Table Y.5 (continued)}}\\
\toprule
Technique & OpsDeck & MetaTool & Raw gap & Explained by headroom & Remaining {[}90\%{]} \\
\midrule
\endhead
\bottomrule
\endlastfoot
D2 & \ensuremath{-}66.6 & \ensuremath{-}68.3 & +1.7 & +7.0 & \ensuremath{-}5.3 {[}\ensuremath{-}22.3, +11.7{]} \\
D3 & \ensuremath{-}96.6 & \ensuremath{-}68.7 & \ensuremath{-}27.9 & \ensuremath{-}14.4 & \ensuremath{-}13.5 {[}\ensuremath{-}23.8, \ensuremath{-}3.3{]} \\
D4 & \ensuremath{-}7.8 & \ensuremath{-}12.4 & +4.6 & +0.7 & +3.9 {[}\ensuremath{-}0.9, +8.6{]} \\
D1 & +27.0 & +13.4 & +13.6 & +4.9 & +8.7 {[}\ensuremath{-}3.3, +20.7{]} \\
D5 & \ensuremath{-}17.1 & \ensuremath{-}8.9 & \ensuremath{-}8.2 & \ensuremath{-}2.9 & \ensuremath{-}5.3 {[}\ensuremath{-}8.7, \ensuremath{-}2.0{]} \\
N1 & +25.0 & +8.4 & +16.6 & +3.7 & +12.9 {[}+2.0, +23.9{]} \\
N2 & +2.5 & +5.0 & \ensuremath{-}2.5 & +1.3 & \ensuremath{-}3.8 {[}\ensuremath{-}8.9, +1.3{]} \\
N3 & \ensuremath{-}4.2 & +1.1 & \ensuremath{-}5.2 & \ensuremath{-}0.2 & \ensuremath{-}5.0 {[}\ensuremath{-}7.4, \ensuremath{-}2.6{]} \\
\end{longtable}
\endgroup

\appsub{Y.4}{The names experiment}
\textbf{Y.4 The names experiment.} MetaTool's tools were renamed to one word (the first content word, or the second where the first collides) and to two spaced words (``Weather Tool''); OpsDeck's names were cut to the function word (``Queue'') or the invented brand word (``Mossgate''). The hypotheses, fixed before the run, were: one-word names lose accuracy (H-E1); descriptions add more on OpsDeck than on MetaTool (H-E2); brand-only names score near ids alone (H-E3); and on OpsDeck the gap between ids and names, both with descriptions, is smaller than on MetaTool (H-E4).

\begingroup\footnotesize
\renewcommand{\thetable}{Y.6}
\begin{longtable}{@{}>{\raggedright\arraybackslash}p{\dimexpr 0.1377\linewidth-0.8261\tabcolsep\relax}>{\raggedright\arraybackslash}p{\dimexpr 0.4810\linewidth-2.8860\tabcolsep\relax}>{\raggedright\arraybackslash}p{\dimexpr 0.2366\linewidth-1.4198\tabcolsep\relax}>{\raggedright\arraybackslash}p{\dimexpr 0.1447\linewidth-0.8681\tabcolsep\relax}@{}}
\caption{The names experiment on Jev 1.13 (accuracy points, 90\% intervals; secondary).}\\
\toprule
Dataset & Contrast & Effect {[}90\% CI{]} & Accuracy: treatment; control \\
\midrule
\endfirsthead
\multicolumn{4}{@{}l}{\textit{Table Y.6 (continued)}}\\
\toprule
Dataset & Contrast & Effect {[}90\% CI{]} & Accuracy: treatment; control \\
\midrule
\endhead
\bottomrule
\endlastfoot
MetaTool & E1: One-word vs two-word names & \ensuremath{-}10.7 {[}\ensuremath{-}17.7, \ensuremath{-}3.9{]} & 63.9\%; 74.6\% \\
MetaTool & E2: Two-word + description vs two-word & +10.7 {[}+4.0, +17.2{]} & 85.2\%; 74.6\% \\
MetaTool & E3: Two-word vs original names & \ensuremath{-}1.5 {[}\ensuremath{-}4.3, +1.2{]} & 74.6\%; 76.1\% \\
MetaTool & E4: Ids + description vs names + description & \ensuremath{-}0.8 {[}\ensuremath{-}2.6, +1.1{]} & 83.7\%; 84.5\% \\
OpsDeck-24 & E1: One-word (function word) vs two-word names & +0.4 {[}\ensuremath{-}3.4, +4.2{]} & 69.8\%; 69.4\% \\
OpsDeck-24 & E2: Two-word + description vs two-word & +25.0 {[}+14.2, +35.8{]} & 94.4\%; 69.4\% \\
OpsDeck-24 & E5: One-word (brand word) vs two-word names & \ensuremath{-}63.7 {[}\ensuremath{-}74.5, \ensuremath{-}52.9{]} & 5.8\%; 69.4\% \\
OpsDeck-24 & E4: Ids + description vs names + description & 0.0 {[}\ensuremath{-}0.4, +0.4{]} & 94.4\%; 94.4\% \\
\end{longtable}
\endgroup

H-E1 held on MetaTool but not for OpsDeck's function words, which kept the meaning of the name. H-E2 and H-E3 held: descriptions added more on OpsDeck, and brand-only names fell to 5.8\%, near the 2.8\% of ids alone on OpsDeck. H-E4 is not resolved: both gaps are within a point of zero. A larger names-over-ids advantage on MetaTool, which would have pointed to a memorised name-to-query link, did not appear.

\appsub{Y.5}{Cost and deviations}
\textbf{Y.5 Cost and deviations.} Generation cost \$5.55 on the ledger (\$3.67 as reported by the vendors), the pilot \$0.30 and the blind definitions \$0.02; Run 9's Jev requests cost \$5.85, \$1.00 of it for the private-set requests sent after the outage. Deviations, each recorded in \code{reports/\allowbreak{}run9/\allowbreak{}design.\allowbreak{}md}: (1) the plan asked for a stop after the pilot for the author's approval; the author had approved the full run in advance, so it went ahead without that stop; (2) the battery was cut to the eight techniques and the names experiment, and the rest of the original battery is costed but not run; (3) the open-twin check could not read the CFPB stream, so it covers every open dataset except CFPB (the domains do not overlap); (4) for 13 of CLINC150's 150 labels the blind definition fell back to the label name; (5) the ledger records validation at the token price, above the vendor-reported cost; (6) a TypeSafe outage interrupted the private-set requests; a relaunch under the same manifest on 2026-11-12 sent only the 24,300 missing requests and completed all four private sets, and every analysis uses only items complete in both arms of a contrast; (7) the difficulty adjustment was added after the first results and is exploratory.

\appsub{Y.6}{Run 9 summary and the difficulty adjustment (Jev 1.13; moved from Section 7 in v24)}
\textbf{Y.6 Run 9 summary and the difficulty adjustment (Jev 1.13; moved from Section 7 in v24).}

\begingroup\footnotesize
\renewcommand{\thetable}{Y.9}
\begin{longtable}{@{}>{\raggedright\arraybackslash}p{\dimexpr 0.4727\linewidth-3.7813\tabcolsep\relax}>{\raggedright\arraybackslash}p{\dimexpr 0.0746\linewidth-0.5970\tabcolsep\relax}>{\raggedright\arraybackslash}p{\dimexpr 0.0881\linewidth-0.7045\tabcolsep\relax}>{\raggedright\arraybackslash}p{\dimexpr 0.2165\linewidth-1.7323\tabcolsep\relax}>{\raggedright\arraybackslash}p{\dimexpr 0.1481\linewidth-1.1850\tabcolsep\relax}@{}}
\caption{The eight techniques on public (open) and private (closed) data, Jev 1.13 (accuracy points). Open: mean over CLINC150, LEDGAR and MetaTool; closed: mean over the four private sets; 90\% wild-cluster intervals with gold classes resampled. The last column is the exploratory open \ensuremath{-} closed shift at matched bare-call headroom, over all seven datasets and without ScotLease (Table Y.4).}\\
\toprule
Technique & Open & Closed & Open \ensuremath{-} closed {[}90\% CI{]} & Matched headroom: all; no ScotLease \\
\midrule
\endfirsthead
\multicolumn{5}{@{}l}{\textit{Table Y.9 (continued)}}\\
\toprule
Technique & Open & Closed & Open \ensuremath{-} closed {[}90\% CI{]} & Matched headroom: all; no ScotLease \\
\midrule
\endhead
\bottomrule
\endlastfoot
Cryptic ids vs readable names & \ensuremath{-}82.7 & \ensuremath{-}85.6 & +2.9 {[}\ensuremath{-}3.7, +9.6{]} & \ensuremath{-}1.0; \ensuremath{-}1.1 \\
Three wrong labels vs three correct (display) & \ensuremath{-}49.5 & \ensuremath{-}56.1 & +6.6 {[}+2.3, +11.0{]} & +10.0; +10.0 \\
One rival vs distinct rivals & \ensuremath{-}13.8 & \ensuremath{-}12.5 & \ensuremath{-}1.2 {[}\ensuremath{-}2.8, +0.3{]} & \ensuremath{-}1.5; +2.0 \\
Retrieved examples vs bare call & +7.7 & +10.1 & \ensuremath{-}2.4 {[}\ensuremath{-}6.0, +1.2{]} & \ensuremath{-}1.5; \ensuremath{-}2.0 \\
30\% confusable pool noise vs clean pool & \ensuremath{-}7.0 & \ensuremath{-}7.8 & +0.8 {[}\ensuremath{-}0.6, +2.2{]} & +1.3; +1.6 \\
Descriptions on readable names vs names & +2.3 & +7.1 & \ensuremath{-}4.8 {[}\ensuremath{-}8.0, \ensuremath{-}1.5{]} & \ensuremath{-}3.3; \ensuremath{-}2.9 \\
True class shares stated vs none & +1.1 & +1.5 & \ensuremath{-}0.4 {[}\ensuremath{-}2.2, +1.3{]} & \ensuremath{-}1.9; \ensuremath{-}1.6 \\
Persona and politeness vs bare call & +0.4 & \ensuremath{-}0.8 & +1.1 {[}+0.4, +1.9{]} & +0.7; +0.7 \\
\end{longtable}
\endgroup

The techniques ranked the same way on both sides: the rank correlation of the eight pooled effects was 1.00 {[}0.98, 1.00{]}, and 1.00 {[}0.94, 1.00{]} without the two largest (ids and the display). One small effect changed sign: persona and politeness scored +0.4 on the public sets (interval from +0.01) and \ensuremath{-}0.8 {[}\ensuremath{-}1.5, \ensuremath{-}0.1{]} on the private ones, both within about a point of zero. No open \ensuremath{-} closed difference passed Holm's correction (the smallest Holm p was 0.075, for the display, and 0.090 for the wrappers), so none of the eight rules is distorted by the plan's criteria. Nor did the public sets show the higher bare-call accuracy that memorised items would give: pooled, open minus closed was \ensuremath{-}1.6 points {[}\ensuremath{-}5.7, +2.5{]}.

\emph{Effects tracked headroom on both sides} (Figure Y.4). Where the bare call left more room, examples and descriptions helped more and the wrong-label display hurt more: retrieved examples gained about 0.7 points, and descriptions about 0.5, per point of headroom. Raw open \ensuremath{-} closed differences therefore partly reflect difficulty. At matched headroom (exploratory), descriptions on readable names differed, and stated shares narrowly: the private sets gained 3.3 points more from descriptions {[}0.6, 6.0{]}, though not clearly without ScotLease (2.9 {[}\ensuremath{-}0.6, 6.3{]}), and 1.9 more from shares {[}0.1, 3.8{]} (1.6 {[}\ensuremath{-}1.4, 4.6{]} without ScotLease). Most of the descriptions difference comes from OpsDeck, where descriptions added 25.0 points against MetaTool's 8.4, and 12.9 of that 16.6-point gap remains at matched headroom (Table Y.5). This fits the names experiment (Y.4) better than leakage: half of each OpsDeck name is an invented word, so its description carries more of the meaning. With six or seven datasets the intervals of this analysis are wide.

\begin{figure}[!htbp]
\centering
\includegraphics[width=\linewidth,height=0.72\textheight,keepaspectratio,alt={Eight panels, one per technique: x is the dataset's bare-call accuracy (70 to 100 ), y is the technique's effect in accuracy points; public (open) datasets are blue circles and private (closed) datasets orange squares, ScotLease hollow because it sits at the ceiling, each with a 90  interval; two parallel fitted lines (solid open, dashed closed) come from a meta-regression on headroom. Each panel title gives the open minus closed shift at matched headroom. In most panels the two lines nearly coincide and the shift's interval spans zero; in the descriptions panel (N1) the dashed closed line lies above the solid one, shift -3.3 [-6.0, -0.6], with OpsDeck at +25.0 well above the other points.}]{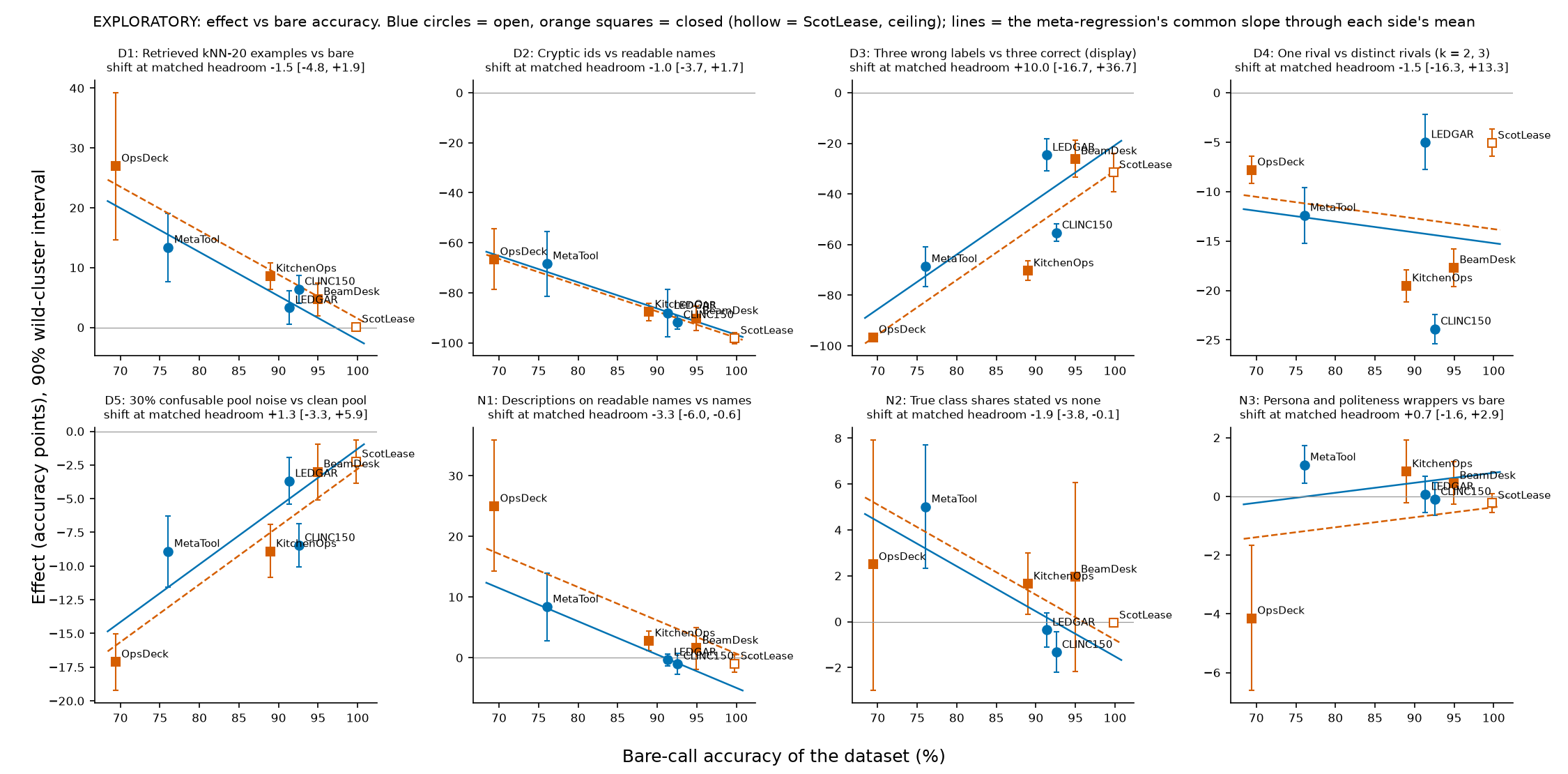}
\renewcommand{\thefigure}{Y.4}
\caption{Run 9 on Jev 1.13 (exploratory): each technique's effect against the dataset's bare-call accuracy (points, 90\% wild-cluster intervals). Blue circles: public datasets; orange squares: private ones (ScotLease hollow, at the ceiling). Lines: a meta-regression on headroom with one slope per technique and a shift between the sides; each title gives the shift at matched headroom with its 90\% interval.}
\end{figure}

\appsub{Y.7}{Run 10: the six other decision models and Jev 1.14}
\textbf{Y.7 Run 10: the six other decision models and Jev 1.14.} Run 10 asked whether Run 9's answer holds beyond Jev 1.13. Its plan (\code{reports/\allowbreak{}preregistration\_\allowbreak{}run10.\allowbreak{}md}) was written on 2026-11-11 before any Run 10 request, and a manifest (\code{registry\_\allowbreak{}post/\allowbreak{}run10/\allowbreak{}run10\_\allowbreak{}manifest.\allowbreak{}json}) froze the code hashes, the private items, the blind definitions, the item ids and the request set; every live send checked it. The plan is internal and not publicly archived. It repeats Run 9's datasets, arms, definitions, shares and name maps (Y.1 to Y.4) on Decider, Decider Flash, OpenAI Decisions, Microsoft-Decision-1, d1-ultra and Kev-4B. Items per model follow Run 8's vendor rule: CLINC150 (validation split) 600, LEDGAR 400 and MetaTool 399, the same items as Run 8's vendor cells (so those cells were cache hits), and the first 600 test items of each private set in Run 9's seeded order; the wrappers used the first 400 items under the first wording, and every other cell two wordings. About 697,000 requests to the six vendors were planned; OpenAI Decisions refused 3,830 of its requests after one re-send, mostly in the cryptic-id and brand-word cells.

\emph{Analysis.} Per model as in Y.1 (dataset-equal means per side; 90\% wild-cluster intervals, gold classes as clusters). Primary: open \ensuremath{-} closed for each technique, two-sided, with Holm over all 6 \ensuremath{\times} 8 = 48 tests jointly. A technique is \emph{leakage-sensitive} if the joint test rejects on at least two of the six models in the same direction; otherwise it shows no sign of leakage. A finding \emph{holds on closed data} on a model when the closed 90\% interval excludes zero in the open effect's direction (D1 positive, D2 to D5 negative); for N1 to N3 only estimates are reported. Secondary: the per-model Spearman correlation of the eight pooled effects (also without D2 and D3), sign flips (open and closed intervals both exclude zero, with opposite signs), the bare-call gap and the names experiment. By this rule, D4 (three rejections, all with a smaller penalty on closed data) and N1 (two, both larger on closed data) are leakage-sensitive; Section 7 explains why neither reads simply as leakage.

\emph{Addendum 1} (written before any Jev 1.14 request). Jev 1.14 was added at the author's request because TypeSafe's own API stayed unreachable and OpenRouter serves TypeSafe's Jev 1.14 (\code{typesafe/\allowbreak{}jev-\allowbreak{}1.\allowbreak{}14-\allowbreak{}20261014}, provider TypeSafe), with its own cache. It ran the same battery and joins a seven-model family (7 \ensuremath{\times} 8 = 56 tests); the six-model verdicts are reported as planned.

\begingroup\scriptsize
\renewcommand{\thetable}{Y.10}
\begin{longtable}{@{}>{\raggedright\arraybackslash}p{\dimexpr 0.2414\linewidth-3.3796\tabcolsep\relax}>{\raggedright\arraybackslash}p{\dimexpr 0.1593\linewidth-2.2308\tabcolsep\relax}>{\raggedright\arraybackslash}p{\dimexpr 0.1593\linewidth-2.2308\tabcolsep\relax}>{\raggedright\arraybackslash}p{\dimexpr 0.1084\linewidth-1.5176\tabcolsep\relax}>{\raggedright\arraybackslash}p{\dimexpr 0.0630\linewidth-0.8824\tabcolsep\relax}>{\raggedright\arraybackslash}p{\dimexpr 0.0744\linewidth-1.0412\tabcolsep\relax}>{\raggedright\arraybackslash}p{\dimexpr 0.1311\linewidth-1.8353\tabcolsep\relax}>{\raggedright\arraybackslash}p{\dimexpr 0.0630\linewidth-0.8824\tabcolsep\relax}@{}}
\caption{Run 10 verdicts on the six other models (accuracy points; 48 tests, 7 rejected by joint Holm at 0.05). Means are plain means over models of each model's dataset-equal mean per side; ``closed mean without ScotLease'' drops the ceiling set. Holds: the 90\% interval excludes zero in the expected direction.}\\
\toprule
Technique & Open mean {[}range{]} & Closed mean {[}range{]} & Closed mean without ScotLease & Holds open & Holds closed & Rejections open\textgreater{}closed / closed\textgreater{}open & Sign flips \\
\midrule
\endfirsthead
\multicolumn{8}{@{}l}{\textit{Table Y.10 (continued)}}\\
\toprule
Technique & Open mean {[}range{]} & Closed mean {[}range{]} & Closed mean without ScotLease & Holds open & Holds closed & Rejections open\textgreater{}closed / closed\textgreater{}open & Sign flips \\
\midrule
\endhead
\bottomrule
\endlastfoot
D1 Retrieved examples vs bare & +1.1 {[}\ensuremath{-}14.5, +7.2{]} & +0.3 {[}\ensuremath{-}16.0, +11.4{]} & +5.7 & 4 of 6 & 3 of 6 & 0 / 0 & 0 \\
D2 Cryptic ids vs readable names & \ensuremath{-}83.4 {[}\ensuremath{-}86.3, \ensuremath{-}74.9{]} & \ensuremath{-}81.6 {[}\ensuremath{-}88.5, \ensuremath{-}72.5{]} & \ensuremath{-}76.5 & 6 of 6 & 6 of 6 & 0 / 0 & 0 \\
D3 Three wrong labels vs three correct & \ensuremath{-}20.5 {[}\ensuremath{-}27.4, \ensuremath{-}4.9{]} & \ensuremath{-}23.6 {[}\ensuremath{-}35.3, \ensuremath{-}10.7{]} & \ensuremath{-}34.4 & 6 of 6 & 6 of 6 & 0 / 1 & 0 \\
D4 Wrong labels on one rival vs spread & \ensuremath{-}6.3 {[}\ensuremath{-}10.1, \ensuremath{-}2.8{]} & \ensuremath{-}3.3 {[}\ensuremath{-}8.9, +6.7{]} & \ensuremath{-}5.1 & 6 of 6 & 5 of 6 & 0 / 3 & 1 \\
D5 30\% wrong labels in the pool vs clean pool & \ensuremath{-}12.6 {[}\ensuremath{-}29.0, \ensuremath{-}4.4{]} & \ensuremath{-}14.1 {[}\ensuremath{-}31.0, \ensuremath{-}2.9{]} & \ensuremath{-}13.1 & 6 of 6 & 6 of 6 & 1 / 0 & 0 \\
N1 One-line descriptions on names & +1.0 {[}\ensuremath{-}2.0, +5.0{]} & +7.6 {[}+5.2, +11.2{]} & +10.7 & -- & -- & 0 / 2 & 0 \\
N2 True class shares stated & +0.5 {[}\ensuremath{-}0.2, +2.2{]} & \ensuremath{-}0.1 {[}\ensuremath{-}4.5, +2.1{]} & \ensuremath{-}0.1 & -- & -- & 0 / 0 & 0 \\
N3 Persona and politeness wrappers & +0.1 {[}\ensuremath{-}0.4, +0.8{]} & \ensuremath{-}0.5 {[}\ensuremath{-}1.3, +0.3{]} & \ensuremath{-}0.7 & -- & -- & 0 / 0 & 0 \\
\end{longtable}
\endgroup

\begingroup\scriptsize
\renewcommand{\thetable}{Y.11}
\begin{longtable}{@{}>{\raggedright\arraybackslash}p{\dimexpr 0.2414\linewidth-3.3796\tabcolsep\relax}>{\raggedright\arraybackslash}p{\dimexpr 0.1593\linewidth-2.2308\tabcolsep\relax}>{\raggedright\arraybackslash}p{\dimexpr 0.1593\linewidth-2.2308\tabcolsep\relax}>{\raggedright\arraybackslash}p{\dimexpr 0.1084\linewidth-1.5176\tabcolsep\relax}>{\raggedright\arraybackslash}p{\dimexpr 0.0630\linewidth-0.8824\tabcolsep\relax}>{\raggedright\arraybackslash}p{\dimexpr 0.0744\linewidth-1.0412\tabcolsep\relax}>{\raggedright\arraybackslash}p{\dimexpr 0.1311\linewidth-1.8353\tabcolsep\relax}>{\raggedright\arraybackslash}p{\dimexpr 0.0630\linewidth-0.8824\tabcolsep\relax}@{}}
\caption{The same with Jev 1.14 added (seven models; 56 tests, 7 rejected by joint Holm at 0.05).}\\
\toprule
Technique & Open mean {[}range{]} & Closed mean {[}range{]} & Closed mean without ScotLease & Holds open & Holds closed & Rejections open\textgreater{}closed / closed\textgreater{}open & Sign flips \\
\midrule
\endfirsthead
\multicolumn{8}{@{}l}{\textit{Table Y.11 (continued)}}\\
\toprule
Technique & Open mean {[}range{]} & Closed mean {[}range{]} & Closed mean without ScotLease & Holds open & Holds closed & Rejections open\textgreater{}closed / closed\textgreater{}open & Sign flips \\
\midrule
\endhead
\bottomrule
\endlastfoot
D1 Retrieved examples vs bare & +1.9 {[}\ensuremath{-}14.5, +7.2{]} & +1.7 {[}\ensuremath{-}16.0, +11.4{]} & +6.7 & 5 of 7 & 4 of 7 & 0 / 0 & 0 \\
D2 Cryptic ids vs readable names & \ensuremath{-}83.4 {[}\ensuremath{-}86.3, \ensuremath{-}74.9{]} & \ensuremath{-}82.2 {[}\ensuremath{-}88.5, \ensuremath{-}72.5{]} & \ensuremath{-}77.3 & 7 of 7 & 7 of 7 & 0 / 0 & 0 \\
D3 Three wrong labels vs three correct & \ensuremath{-}25.0 {[}\ensuremath{-}51.6, \ensuremath{-}4.9{]} & \ensuremath{-}28.0 {[}\ensuremath{-}54.4, \ensuremath{-}10.7{]} & \ensuremath{-}38.9 & 7 of 7 & 7 of 7 & 0 / 1 & 0 \\
D4 Wrong labels on one rival vs spread & \ensuremath{-}7.6 {[}\ensuremath{-}15.4, \ensuremath{-}2.8{]} & \ensuremath{-}4.9 {[}\ensuremath{-}14.6, +6.7{]} & \ensuremath{-}6.6 & 7 of 7 & 6 of 7 & 0 / 3 & 1 \\
D5 30\% wrong labels in the pool vs clean pool & \ensuremath{-}11.8 {[}\ensuremath{-}29.0, \ensuremath{-}4.4{]} & \ensuremath{-}13.3 {[}\ensuremath{-}31.0, \ensuremath{-}2.9{]} & \ensuremath{-}12.8 & 7 of 7 & 7 of 7 & 1 / 0 & 0 \\
N1 One-line descriptions on names & +1.1 {[}\ensuremath{-}2.0, +5.0{]} & +7.6 {[}+5.2, +11.2{]} & +10.6 & -- & -- & 0 / 2 & 0 \\
N2 True class shares stated & +0.5 {[}\ensuremath{-}0.2, +2.2{]} & +0.1 {[}\ensuremath{-}4.5, +2.1{]} & +0.1 & -- & -- & 0 / 0 & 0 \\
N3 Persona and politeness wrappers & +0.1 {[}\ensuremath{-}0.4, +0.8{]} & \ensuremath{-}0.6 {[}\ensuremath{-}1.3, +0.3{]} & \ensuremath{-}0.9 & -- & -- & 0 / 0 & 0 \\
\end{longtable}
\endgroup

\begingroup\footnotesize
\renewcommand{\thetable}{Y.12}
\begin{longtable}{@{}>{\raggedright\arraybackslash}p{\dimexpr 0.1323\linewidth-1.3231\tabcolsep\relax}>{\raggedright\arraybackslash}p{\dimexpr 0.2846\linewidth-2.8462\tabcolsep\relax}>{\raggedright\arraybackslash}p{\dimexpr 0.0769\linewidth-0.7692\tabcolsep\relax}>{\raggedright\arraybackslash}p{\dimexpr 0.1747\linewidth-1.7471\tabcolsep\relax}>{\raggedright\arraybackslash}p{\dimexpr 0.1714\linewidth-1.7144\tabcolsep\relax}>{\raggedright\arraybackslash}p{\dimexpr 0.1600\linewidth-1.6000\tabcolsep\relax}@{}}
\caption{Per model and technique, D1 to D5: open effect, closed effect with its 90\% interval, and the open \ensuremath{-} closed difference with its interval and two-sided p (accuracy points). D1: retrieved examples vs bare call; D2: cryptic ids vs readable names; D3: three wrong labels vs three correct; D4: wrong labels on one rival vs spread; D5: 30\% confusable pool noise vs clean pool. Jev 1.13 is Run 9 (complete); Jev 1.14's KitchenOps cells are about 54\% complete.}\\
\toprule
Technique & Model & Open & Closed {[}90\% CI{]} & Open \ensuremath{-} closed {[}90\% CI{]} & p (two-sided) \\
\midrule
\endfirsthead
\multicolumn{6}{@{}l}{\textit{Table Y.12 (continued)}}\\
\toprule
Technique & Model & Open & Closed {[}90\% CI{]} & Open \ensuremath{-} closed {[}90\% CI{]} & p (two-sided) \\
\midrule
\endhead
\bottomrule
\endlastfoot
D1 & Decider & +4.0 & +4.4 {[}+1.6, +7.2{]} & \ensuremath{-}0.4 {[}\ensuremath{-}4.0, +3.2{]} & 0.8720 \\
D1 & Decider Flash & \ensuremath{-}14.5 & \ensuremath{-}16.0 {[}\ensuremath{-}21.2, \ensuremath{-}10.8{]} & +1.5 {[}\ensuremath{-}4.2, +7.1{]} & 0.6772 \\
D1 & OpenAI Decisions & +7.2 & +11.4 {[}+6.6, +16.0{]} & \ensuremath{-}4.1 {[}\ensuremath{-}9.2, +1.0{]} & 0.1926 \\
D1 & Microsoft-Decision-1 & +3.2 & +6.9 {[}+3.4, +10.6{]} & \ensuremath{-}3.7 {[}\ensuremath{-}7.7, +0.2{]} & 0.1156 \\
D1 & d1-ultra & +1.1 & \ensuremath{-}8.0 {[}\ensuremath{-}14.0, \ensuremath{-}2.3{]} & +9.1 {[}+2.9, +15.3{]} & 0.0198 \\
D1 & Kev-4B & +5.6 & +3.4 {[}\ensuremath{-}2.1, +8.8{]} & +2.3 {[}\ensuremath{-}4.0, +8.5{]} & 0.5580 \\
D1 & Jev 1.14 & +6.9 & +9.7 {[}+6.7, +12.7{]} & \ensuremath{-}2.8 {[}\ensuremath{-}6.8, +1.1{]} & 0.2320 \\
D1 & Jev 1.13 (Run 9) & +7.7 & +10.1 {[}+7.0, +13.2{]} & \ensuremath{-}2.4 {[}\ensuremath{-}6.0, +1.2{]} & 0.2756 \\
D2 & Decider & \ensuremath{-}86.1 & \ensuremath{-}88.5 {[}\ensuremath{-}91.0, \ensuremath{-}85.9{]} & +2.4 {[}\ensuremath{-}1.5, +6.3{]} & 0.3224 \\
D2 & Decider Flash & \ensuremath{-}86.3 & \ensuremath{-}78.7 {[}\ensuremath{-}82.5, \ensuremath{-}74.9{]} & \ensuremath{-}7.6 {[}\ensuremath{-}12.3, \ensuremath{-}3.0{]} & 0.0070 \\
D2 & OpenAI Decisions & \ensuremath{-}82.2 & \ensuremath{-}80.8 {[}\ensuremath{-}90.3, \ensuremath{-}71.3{]} & \ensuremath{-}1.4 {[}\ensuremath{-}11.4, +8.5{]} & 0.8110 \\
D2 & Microsoft-Decision-1 & \ensuremath{-}85.8 & \ensuremath{-}83.5 {[}\ensuremath{-}87.6, \ensuremath{-}79.4{]} & \ensuremath{-}2.4 {[}\ensuremath{-}8.6, +3.8{]} & 0.5322 \\
D2 & d1-ultra & \ensuremath{-}84.9 & \ensuremath{-}85.5 {[}\ensuremath{-}88.5, \ensuremath{-}82.4{]} & +0.5 {[}\ensuremath{-}4.1, +5.1{]} & 0.8508 \\
D2 & Kev-4B & \ensuremath{-}74.9 & \ensuremath{-}72.5 {[}\ensuremath{-}75.9, \ensuremath{-}69.1{]} & \ensuremath{-}2.4 {[}\ensuremath{-}7.2, +2.3{]} & 0.4110 \\
D2 & Jev 1.14 & \ensuremath{-}83.4 & \ensuremath{-}85.9 {[}\ensuremath{-}89.3, \ensuremath{-}82.5{]} & +2.4 {[}\ensuremath{-}4.2, +9.2{]} & 0.5588 \\
D2 & Jev 1.13 (Run 9) & \ensuremath{-}82.7 & \ensuremath{-}85.6 {[}\ensuremath{-}89.0, \ensuremath{-}82.2{]} & +2.9 {[}\ensuremath{-}3.7, +9.6{]} & 0.4942 \\
D3 & Decider & \ensuremath{-}15.8 & \ensuremath{-}23.0 {[}\ensuremath{-}25.7, \ensuremath{-}20.1{]} & +7.2 {[}+2.1, +12.3{]} & 0.0268 \\
D3 & Decider Flash & \ensuremath{-}4.9 & \ensuremath{-}10.7 {[}\ensuremath{-}14.2, \ensuremath{-}7.3{]} & +5.9 {[}+0.6, +11.2{]} & 0.0644 \\
D3 & OpenAI Decisions & \ensuremath{-}21.3 & \ensuremath{-}26.4 {[}\ensuremath{-}29.7, \ensuremath{-}23.2{]} & +5.1 {[}+0.8, +9.5{]} & 0.0532 \\
D3 & Microsoft-Decision-1 & \ensuremath{-}27.4 & \ensuremath{-}35.3 {[}\ensuremath{-}37.5, \ensuremath{-}33.1{]} & +7.9 {[}+4.0, +11.8{]} & 0.0016 \\
D3 & d1-ultra & \ensuremath{-}27.3 & \ensuremath{-}12.2 {[}\ensuremath{-}15.8, \ensuremath{-}8.6{]} & \ensuremath{-}15.1 {[}\ensuremath{-}21.0, \ensuremath{-}9.3{]} & 0.0002 \\
D3 & Kev-4B & \ensuremath{-}26.6 & \ensuremath{-}34.2 {[}\ensuremath{-}39.4, \ensuremath{-}29.3{]} & +7.6 {[}+1.5, +13.8{]} & 0.0408 \\
D3 & Jev 1.14 & \ensuremath{-}51.6 & \ensuremath{-}54.4 {[}\ensuremath{-}57.1, \ensuremath{-}51.5{]} & +2.8 {[}\ensuremath{-}1.9, +7.5{]} & 0.3430 \\
D3 & Jev 1.13 (Run 9) & \ensuremath{-}49.5 & \ensuremath{-}56.1 {[}\ensuremath{-}58.8, \ensuremath{-}53.4{]} & +6.6 {[}+2.3, +11.0{]} & 0.0094 \\
D4 & Decider & \ensuremath{-}4.7 & \ensuremath{-}6.5 {[}\ensuremath{-}7.5, \ensuremath{-}5.5{]} & +1.8 {[}+0.3, +3.3{]} & 0.0502 \\
D4 & Decider Flash & \ensuremath{-}6.3 & \ensuremath{-}1.5 {[}\ensuremath{-}2.3, \ensuremath{-}0.6{]} & \ensuremath{-}4.8 {[}\ensuremath{-}6.3, \ensuremath{-}3.4{]} & 0.0002 \\
D4 & OpenAI Decisions & \ensuremath{-}7.3 & \ensuremath{-}6.7 {[}\ensuremath{-}7.6, \ensuremath{-}5.7{]} & \ensuremath{-}0.7 {[}\ensuremath{-}2.1, +0.8{]} & 0.4288 \\
D4 & Microsoft-Decision-1 & \ensuremath{-}10.1 & \ensuremath{-}3.1 {[}\ensuremath{-}3.9, \ensuremath{-}2.3{]} & \ensuremath{-}7.0 {[}\ensuremath{-}8.5, \ensuremath{-}5.4{]} & 0.0002 \\
D4 & d1-ultra & \ensuremath{-}6.3 & \ensuremath{-}8.9 {[}\ensuremath{-}9.8, \ensuremath{-}8.0{]} & +2.5 {[}+1.0, +4.1{]} & 0.0076 \\
D4 & Kev-4B & \ensuremath{-}2.8 & +6.7 {[}+5.5, +7.9{]} & \ensuremath{-}9.5 {[}\ensuremath{-}11.2, \ensuremath{-}7.8{]} & 0.0002 \\
D4 & Jev 1.14 & \ensuremath{-}15.4 & \ensuremath{-}14.6 {[}\ensuremath{-}15.6, \ensuremath{-}13.5{]} & \ensuremath{-}0.9 {[}\ensuremath{-}2.9, +1.2{]} & 0.4834 \\
D4 & Jev 1.13 (Run 9) & \ensuremath{-}13.8 & \ensuremath{-}12.5 {[}\ensuremath{-}13.2, \ensuremath{-}11.8{]} & \ensuremath{-}1.2 {[}\ensuremath{-}2.8, +0.3{]} & 0.1888 \\
D5 & Decider & \ensuremath{-}4.4 & \ensuremath{-}4.7 {[}\ensuremath{-}6.6, \ensuremath{-}2.8{]} & +0.3 {[}\ensuremath{-}2.8, +3.4{]} & 0.8976 \\
D5 & Decider Flash & \ensuremath{-}4.5 & \ensuremath{-}2.9 {[}\ensuremath{-}5.4, \ensuremath{-}0.4{]} & \ensuremath{-}1.6 {[}\ensuremath{-}4.8, +1.6{]} & 0.3916 \\
D5 & OpenAI Decisions & \ensuremath{-}7.0 & \ensuremath{-}7.5 {[}\ensuremath{-}9.2, \ensuremath{-}5.7{]} & +0.5 {[}\ensuremath{-}1.7, +2.6{]} & 0.7224 \\
D5 & Microsoft-Decision-1 & \ensuremath{-}10.7 & \ensuremath{-}12.7 {[}\ensuremath{-}14.9, \ensuremath{-}10.5{]} & +2.1 {[}\ensuremath{-}0.9, +5.0{]} & 0.2386 \\
D5 & d1-ultra & \ensuremath{-}29.0 & \ensuremath{-}26.0 {[}\ensuremath{-}31.7, \ensuremath{-}20.4{]} & \ensuremath{-}2.9 {[}\ensuremath{-}9.0, +3.2{]} & 0.4608 \\
D5 & Kev-4B & \ensuremath{-}20.1 & \ensuremath{-}31.0 {[}\ensuremath{-}35.1, \ensuremath{-}27.0{]} & +10.9 {[}+6.1, +15.6{]} & 0.0002 \\
D5 & Jev 1.14 & \ensuremath{-}7.1 & \ensuremath{-}8.5 {[}\ensuremath{-}9.6, \ensuremath{-}7.3{]} & +1.4 {[}\ensuremath{-}0.3, +3.1{]} & 0.1798 \\
D5 & Jev 1.13 (Run 9) & \ensuremath{-}7.0 & \ensuremath{-}7.8 {[}\ensuremath{-}8.7, \ensuremath{-}6.9{]} & +0.8 {[}\ensuremath{-}0.6, +2.2{]} & 0.3638 \\
\end{longtable}
\endgroup

\begingroup\footnotesize
\renewcommand{\thetable}{Y.13}
\begin{longtable}{@{}>{\raggedright\arraybackslash}p{\dimexpr 0.1323\linewidth-1.3231\tabcolsep\relax}>{\raggedright\arraybackslash}p{\dimexpr 0.2846\linewidth-2.8462\tabcolsep\relax}>{\raggedright\arraybackslash}p{\dimexpr 0.0631\linewidth-0.6308\tabcolsep\relax}>{\raggedright\arraybackslash}p{\dimexpr 0.1765\linewidth-1.7648\tabcolsep\relax}>{\raggedright\arraybackslash}p{\dimexpr 0.1835\linewidth-1.8352\tabcolsep\relax}>{\raggedright\arraybackslash}p{\dimexpr 0.1600\linewidth-1.6000\tabcolsep\relax}@{}}
\caption{As Table Y.12 for N1 (one-line descriptions on names), N2 (true class shares stated) and N3 (persona and politeness wrappers).}\\
\toprule
Technique & Model & Open & Closed {[}90\% CI{]} & Open \ensuremath{-} closed {[}90\% CI{]} & p (two-sided) \\
\midrule
\endfirsthead
\multicolumn{6}{@{}l}{\textit{Table Y.13 (continued)}}\\
\toprule
Technique & Model & Open & Closed {[}90\% CI{]} & Open \ensuremath{-} closed {[}90\% CI{]} & p (two-sided) \\
\midrule
\endhead
\bottomrule
\endlastfoot
N1 & Decider & +1.8 & +5.2 {[}+2.5, +8.0{]} & \ensuremath{-}3.4 {[}\ensuremath{-}6.9, +0.1{]} & 0.1146 \\
N1 & Decider Flash & +0.6 & +10.2 {[}+6.8, +13.6{]} & \ensuremath{-}9.5 {[}\ensuremath{-}13.6, \ensuremath{-}5.6{]} & 0.0002 \\
N1 & OpenAI Decisions & +1.9 & +6.4 {[}+2.2, +10.4{]} & \ensuremath{-}4.5 {[}\ensuremath{-}9.4, +0.1{]} & 0.1154 \\
N1 & Microsoft-Decision-1 & \ensuremath{-}1.1 & +7.4 {[}+3.9, +10.8{]} & \ensuremath{-}8.4 {[}\ensuremath{-}12.4, \ensuremath{-}4.5{]} & 0.0008 \\
N1 & d1-ultra & \ensuremath{-}2.0 & +5.2 {[}+1.1, +9.5{]} & \ensuremath{-}7.2 {[}\ensuremath{-}11.9, \ensuremath{-}2.6{]} & 0.0096 \\
N1 & Kev-4B & +5.0 & +11.2 {[}+8.0, +14.3{]} & \ensuremath{-}6.2 {[}\ensuremath{-}11.1, \ensuremath{-}1.4{]} & 0.0384 \\
N1 & Jev 1.14 & +1.5 & +7.3 {[}+4.8, +9.9{]} & \ensuremath{-}5.8 {[}\ensuremath{-}9.3, \ensuremath{-}2.3{]} & 0.0086 \\
N1 & Jev 1.13 (Run 9) & +2.3 & +7.1 {[}+4.4, +9.9{]} & \ensuremath{-}4.8 {[}\ensuremath{-}8.0, \ensuremath{-}1.5{]} & 0.0198 \\
N2 & Decider & +0.4 & +0.6 {[}\ensuremath{-}1.9, +3.1{]} & \ensuremath{-}0.1 {[}\ensuremath{-}2.8, +2.6{]} & 0.9458 \\
N2 & Decider Flash & +0.1 & \ensuremath{-}4.5 {[}\ensuremath{-}6.3, \ensuremath{-}2.7{]} & +4.5 {[}+2.1, +7.0{]} & 0.0020 \\
N2 & OpenAI Decisions & +0.0 & +1.6 {[}\ensuremath{-}1.1, +4.3{]} & \ensuremath{-}1.5 {[}\ensuremath{-}4.5, +1.4{]} & 0.4208 \\
N2 & Microsoft-Decision-1 & \ensuremath{-}0.2 & \ensuremath{-}1.9 {[}\ensuremath{-}5.2, +1.5{]} & +1.7 {[}\ensuremath{-}1.7, +5.0{]} & 0.3834 \\
N2 & d1-ultra & +0.7 & +1.6 {[}\ensuremath{-}1.4, +4.6{]} & \ensuremath{-}0.9 {[}\ensuremath{-}4.1, +2.2{]} & 0.6618 \\
N2 & Kev-4B & +2.2 & +2.1 {[}\ensuremath{-}0.9, +4.9{]} & +0.1 {[}\ensuremath{-}3.6, +3.9{]} & 0.9382 \\
N2 & Jev 1.14 & +0.3 & +1.0 {[}\ensuremath{-}0.6, +2.6{]} & \ensuremath{-}0.8 {[}\ensuremath{-}2.6, +1.0{]} & 0.4940 \\
N2 & Jev 1.13 (Run 9) & +1.1 & +1.5 {[}0.0, +3.0{]} & \ensuremath{-}0.4 {[}\ensuremath{-}2.2, +1.3{]} & 0.6948 \\
N3 & Decider & +0.6 & \ensuremath{-}0.1 {[}\ensuremath{-}0.5, +0.3{]} & +0.7 {[}+0.1, +1.3{]} & 0.0452 \\
N3 & Decider Flash & \ensuremath{-}0.2 & \ensuremath{-}1.3 {[}\ensuremath{-}1.9, \ensuremath{-}0.6{]} & +1.1 {[}+0.2, +1.9{]} & 0.0418 \\
N3 & OpenAI Decisions & \ensuremath{-}0.4 & \ensuremath{-}1.1 {[}\ensuremath{-}1.7, \ensuremath{-}0.5{]} & +0.7 {[}\ensuremath{-}0.2, +1.6{]} & 0.1782 \\
N3 & Microsoft-Decision-1 & \ensuremath{-}0.2 & \ensuremath{-}0.5 {[}\ensuremath{-}1.0, +0.0{]} & +0.3 {[}\ensuremath{-}0.2, +0.9{]} & 0.3330 \\
N3 & d1-ultra & \ensuremath{-}0.3 & +0.3 {[}\ensuremath{-}0.3, +0.8{]} & \ensuremath{-}0.5 {[}\ensuremath{-}1.2, +0.2{]} & 0.2420 \\
N3 & Kev-4B & +0.8 & \ensuremath{-}0.3 {[}\ensuremath{-}1.0, +0.4{]} & +1.1 {[}+0.1, +2.2{]} & 0.0838 \\
N3 & Jev 1.14 & +0.1 & \ensuremath{-}1.3 {[}\ensuremath{-}2.1, \ensuremath{-}0.5{]} & +1.4 {[}+0.4, +2.4{]} & 0.0224 \\
N3 & Jev 1.13 (Run 9) & +0.4 & \ensuremath{-}0.8 {[}\ensuremath{-}1.5, \ensuremath{-}0.1{]} & +1.1 {[}+0.4, +1.9{]} & 0.0128 \\
\end{longtable}
\endgroup

\begingroup\footnotesize
\renewcommand{\thetable}{Y.14}
\begin{longtable}{@{}>{\raggedright\arraybackslash}p{\dimexpr 0.2846\linewidth-2.8462\tabcolsep\relax}>{\raggedright\arraybackslash}p{\dimexpr 0.0631\linewidth-0.6308\tabcolsep\relax}>{\raggedright\arraybackslash}p{\dimexpr 0.0908\linewidth-0.9077\tabcolsep\relax}>{\raggedright\arraybackslash}p{\dimexpr 0.2259\linewidth-2.2595\tabcolsep\relax}>{\raggedright\arraybackslash}p{\dimexpr 0.2310\linewidth-2.3097\tabcolsep\relax}>{\raggedright\arraybackslash}p{\dimexpr 0.1046\linewidth-1.0462\tabcolsep\relax}@{}}
\caption{Bare-call accuracy (\%, dataset-equal means) on each side, and the rank agreement of the eight pooled effects, open vs closed (Spearman, 90\% pairs class-bootstrap interval; last column without D2 and D3). Jev 1.13 (Run 9): 1.00 {[}0.98, 1.00{]}, 1.00 without the two largest effects, bare gap \ensuremath{-}1.6 {[}\ensuremath{-}5.7, +2.5{]} (Y.6).}\\
\toprule
Model & Bare open & Bare closed & Open \ensuremath{-} closed {[}90\% CI{]} & Spearman 8 techniques {[}90\%{]} & Without D2, D3 \\
\midrule
\endfirsthead
\multicolumn{6}{@{}l}{\textit{Table Y.14 (continued)}}\\
\toprule
Model & Bare open & Bare closed & Open \ensuremath{-} closed {[}90\% CI{]} & Spearman 8 techniques {[}90\%{]} & Without D2, D3 \\
\midrule
\endhead
\bottomrule
\endlastfoot
Decider & 89.5 & 91.8 & \ensuremath{-}2.3 {[}\ensuremath{-}6.3, +1.6{]} & 0.95 {[}0.83, 0.98{]} & 0.89 \\
Decider Flash & 89.0 & 84.3 & +4.7 {[}+0.4, +9.1{]} & 0.76 {[}0.50, 0.88{]} & 0.60 \\
OpenAI Decisions & 84.8 & 86.4 & \ensuremath{-}1.6 {[}\ensuremath{-}7.5, +4.4{]} & 0.98 {[}0.90, 1.00{]} & 0.94 \\
Microsoft-Decision-1 & 90.6 & 88.2 & +2.4 {[}\ensuremath{-}2.0, +6.7{]} & 0.86 {[}0.79, 0.98{]} & 0.66 \\
d1-ultra & 89.0 & 89.7 & \ensuremath{-}0.7 {[}\ensuremath{-}5.4, +3.9{]} & 0.79 {[}0.57, 0.86{]} & 0.49 \\
Kev-4B & 77.9 & 76.7 & +1.2 {[}\ensuremath{-}3.5, +5.9{]} & 0.81 {[}0.69, 0.90{]} & 0.54 \\
Jev 1.14 & 87.4 & 88.5 & \ensuremath{-}1.2 {[}\ensuremath{-}5.2, +2.8{]} & 1.00 {[}0.93, 1.00{]} & 1.00 \\
\end{longtable}
\endgroup

\begingroup\footnotesize
\renewcommand{\thetable}{Y.15}
\begin{longtable}{@{}>{\raggedright\arraybackslash}p{\dimexpr 0.2846\linewidth-2.8462\tabcolsep\relax}>{\raggedright\arraybackslash}p{\dimexpr 0.1431\linewidth-1.4311\tabcolsep\relax}>{\raggedright\arraybackslash}p{\dimexpr 0.1431\linewidth-1.4311\tabcolsep\relax}>{\raggedright\arraybackslash}p{\dimexpr 0.1387\linewidth-1.3868\tabcolsep\relax}>{\raggedright\arraybackslash}p{\dimexpr 0.1387\linewidth-1.3868\tabcolsep\relax}>{\raggedright\arraybackslash}p{\dimexpr 0.1518\linewidth-1.5179\tabcolsep\relax}@{}}
\caption{The names experiment across models (accuracy points, 90\% intervals; secondary). E1: one-word vs two-word names; E2: two-word names plus a description vs two-word names; E5: OpsDeck names cut to the invented brand word vs two-word names; E4: ids plus description vs names plus description. Jev 1.13's values are in Table Y.6. OpenAI Decisions' E5 rests on few complete items, because it refused most brand-word requests (below).}\\
\toprule
Model & MetaTool E1 one-word & MetaTool E2 & OpsDeck E2 & OpsDeck E5 brand word & OpsDeck E4 ids+desc vs names+desc \\
\midrule
\endfirsthead
\multicolumn{6}{@{}l}{\textit{Table Y.15 (continued)}}\\
\toprule
Model & MetaTool E1 one-word & MetaTool E2 & OpsDeck E2 & OpsDeck E5 brand word & OpsDeck E4 ids+desc vs names+desc \\
\midrule
\endhead
\bottomrule
\endlastfoot
Decider & \ensuremath{-}10.7 {[}\ensuremath{-}17.6, \ensuremath{-}3.9{]} & +7.9 {[}+1.8, +13.9{]} & +16.8 {[}+8.4, +25.0{]} & \ensuremath{-}76.7 {[}\ensuremath{-}84.2, \ensuremath{-}69.7{]} & \ensuremath{-}0.2 {[}\ensuremath{-}0.8, +0.3{]} \\
Decider Flash & \ensuremath{-}9.3 {[}\ensuremath{-}17.6, \ensuremath{-}1.0{]} & +9.8 {[}+1.9, +17.6{]} & +36.6 {[}+23.5, +49.7{]} & \ensuremath{-}53.4 {[}\ensuremath{-}66.9, \ensuremath{-}40.0{]} & \ensuremath{-}0.8 {[}\ensuremath{-}3.0, +1.3{]} \\
OpenAI Decisions & \ensuremath{-}11.3 {[}\ensuremath{-}19.7, \ensuremath{-}2.9{]} & +7.8 {[}+0.6, +14.8{]} & +20.8 {[}+7.5, +34.1{]} & \ensuremath{-}63.9 {[}\ensuremath{-}82.0, \ensuremath{-}44.9{]} & +0.2 {[}\ensuremath{-}0.6, +1.0{]} \\
Microsoft-Decision-1 & \ensuremath{-}11.2 {[}\ensuremath{-}19.9, \ensuremath{-}2.4{]} & +7.1 {[}\ensuremath{-}0.8, +14.9{]} & +24.5 {[}+13.0, +35.9{]} & \ensuremath{-}69.2 {[}\ensuremath{-}82.1, \ensuremath{-}56.8{]} & +0.4 {[}\ensuremath{-}0.5, +1.4{]} \\
d1-ultra & \ensuremath{-}7.9 {[}\ensuremath{-}15.9, \ensuremath{-}0.0{]} & +5.8 {[}\ensuremath{-}1.6, +13.0{]} & +18.8 {[}+8.5, +29.1{]} & \ensuremath{-}74.9 {[}\ensuremath{-}85.8, \ensuremath{-}63.8{]} & +0.0 {[}\ensuremath{-}0.6, +0.6{]} \\
Kev-4B & \ensuremath{-}12.4 {[}\ensuremath{-}21.5, \ensuremath{-}3.3{]} & +14.7 {[}+1.9, +27.3{]} & +39.8 {[}+29.4, +50.4{]} & \ensuremath{-}40.6 {[}\ensuremath{-}51.7, \ensuremath{-}29.4{]} & \ensuremath{-}0.2 {[}\ensuremath{-}2.3, +2.0{]} \\
Jev 1.14 & \ensuremath{-}11.5 {[}\ensuremath{-}19.4, \ensuremath{-}3.8{]} & +10.8 {[}+3.6, +17.8{]} & +24.2 {[}+14.5, +33.9{]} & \ensuremath{-}65.8 {[}\ensuremath{-}75.4, \ensuremath{-}56.1{]} & +0.3 {[}\ensuremath{-}0.5, +1.1{]} \\
\end{longtable}
\endgroup

\begingroup\scriptsize
\renewcommand{\thetable}{Y.16}
\begin{longtable}{@{}>{\raggedright\arraybackslash}p{\dimexpr 0.2131\linewidth-2.9839\tabcolsep\relax}>{\raggedright\arraybackslash}p{\dimexpr 0.0887\linewidth-1.2419\tabcolsep\relax}>{\raggedright\arraybackslash}p{\dimexpr 0.1094\linewidth-1.5323\tabcolsep\relax}>{\raggedright\arraybackslash}p{\dimexpr 0.0680\linewidth-0.9516\tabcolsep\relax}>{\raggedright\arraybackslash}p{\dimexpr 0.1302\linewidth-1.8226\tabcolsep\relax}>{\raggedright\arraybackslash}p{\dimexpr 0.1198\linewidth-1.6774\tabcolsep\relax}>{\raggedright\arraybackslash}p{\dimexpr 0.1302\linewidth-1.8226\tabcolsep\relax}>{\raggedright\arraybackslash}p{\dimexpr 0.1406\linewidth-1.9677\tabcolsep\relax}@{}}
\caption{Completeness: the share of unique requests answered, per model and dataset. Every contrast uses only items complete in both arms.}\\
\toprule
Model & MetaTool & OpsDeck-24 & LEDGAR & ScotLease-25 & BeamDesk-14 & CLINC150 (validation) & KitchenOps-80 \\
\midrule
\endfirsthead
\multicolumn{8}{@{}l}{\textit{Table Y.16 (continued)}}\\
\toprule
Model & MetaTool & OpsDeck-24 & LEDGAR & ScotLease-25 & BeamDesk-14 & CLINC150 (validation) & KitchenOps-80 \\
\midrule
\endhead
\bottomrule
\endlastfoot
Decider & 100\% & 100\% & 100\% & 100\% & 100\% & 100\% & 100\% \\
Decider Flash & 100\% & 100\% & 100\% & 100\% & 100\% & 100\% & 100\% \\
OpenAI Decisions & 99\% & 91\% & 99\% & 99\% & 99\% & 98\% & 94\% \\
Microsoft-Decision-1 & 100\% & 100\% & 100\% & 100\% & 100\% & 100\% & 100\% \\
d1-ultra & 100\% & 100\% & 100\% & 100\% & 100\% & 100\% & 100\% \\
Kev-4B & 100\% & 100\% & 100\% & 100\% & 100\% & 100\% & 100\% \\
Jev 1.14 & 100\% & 100\% & 100\% & 100\% & 100\% & 100\% & 54\% \\
\end{longtable}
\endgroup

\emph{Long texts} (cache checks after the run; observed). Decider Flash on ScotLease: bare 98.9\%; with three correctly labelled examples 28.0\%; with the same examples under wrong labels 96.9\%; with nearest examples (kNN-20) 28.6\%. On BeamDesk: bare 94.8\%, nearest examples 52.8\%. d1-ultra on BeamDesk: bare 90.6\%, nearest examples 53.2\%, with 490 of 1,200 answers on ``Sample Environment'' (the gold label for 94). Decider Flash showed the same pattern on public LEDGAR in Run 8 (Appendix U.2).

\emph{Refusals.} OpenAI Decisions refused 3,830 requests after one re-send: 2,627 in cryptic-id cells (KitchenOps 1,019, OpsDeck 775, CLINC150 362, BeamDesk 177, ScotLease 110, LEDGAR 93, MetaTool 91), 939 in OpsDeck's brand-word names, and fewer than 200 in any other cell. Its completeness is in Table Y.16.

\emph{Cost.} The six vendors cost \$14.8 on the ledger (OpenAI Decisions \$6.12, Microsoft-Decision-1 \$2.39, Kev-4B \$2.23, d1-ultra \$2.18, Decider \$1.24, Decider Flash \$0.62), and Jev 1.14 \$4.74. The ledger stood at \$94.26 after Run 10, and at \$95.26 after Run 9's last private-set requests (Y.5).

\emph{Deviations.} (1) OpenAI Decisions ran without its 20-request pilot: the pilot stopped on 2 refusals, the known behaviour of the ids cells. (2) Jev 1.14 was added by addendum, through OpenRouter. (3) The manifest was refrozen four times before Jev 1.14's first request; every freeze is kept in \code{registry\_\allowbreak{}post/\allowbreak{}run10/\allowbreak{}}. (4) OpenRouter credit ran out: 108,405 of 116,202 Jev 1.14 requests were answered, and the remaining 7,797, all KitchenOps, failed with HTTP 402, so KitchenOps is about 54\% complete for Jev 1.14 and its effects there use only complete items. (5) Jev 1.13's column is Run 9's, whose private cells were completed on 2026-11-12, after Run 10 (Y.5, deviation 6).

\end{document}